\documentclass{article}

     \PassOptionsToPackage{numbers, compress}{natbib}

\usepackage[eandd, final]{neurips_2026}

\usepackage[utf8]{inputenc} 
\usepackage[T1]{fontenc}    
\usepackage{hyperref}       
\usepackage{url}            
\usepackage{booktabs}       
\usepackage{amsfonts}       
\usepackage{nicefrac}       
\usepackage{microtype}      
\usepackage{xcolor}         
\usepackage{graphicx}       
\usepackage{multirow}       
\usepackage{multicol}       
\usepackage{graphicx}
\usepackage{float}          
\usepackage{subcaption}     
\usepackage{listings}
\usepackage{etoc}            
\usepackage{tcolorbox}
\tcbuselibrary{listings,breakable}

\newtcolorbox[auto counter]{promptbox}[2][]{
  colback=black!5!white,
  colframe=black!75!white,
  breakable,
  title={Prompt~\thetcbcounter: #2},
  #1
}

\definecolor{capOriginal}{HTML}{6E69DC}
\definecolor{capFlorence}{HTML}{CD5FB9}
\definecolor{capShareGPT}{HTML}{F09650}
\definecolor{capGemini}{HTML}{3CB4BE}
\definecolor{capInternVL}{HTML}{EB698C}

\title{MONET: A \textbf{M}assive, \textbf{O}pen, \textbf{N}on-redundant and \textbf{E}nriched \textbf{T}ext-to-image dataset}

\author{%
  Benjamin Aubin\thanks{Corresponding authors: name.surname@jasper.ai} \\
  Jasper Research
  \And 
  Gonzalo Iñaki Quintana\footnotemark[1] \\
  Jasper Research
  \And
  Onur Tasar \\
  Jasper Research
  \And
  Sanjeev Sreetharan \\
  Jasper Research
  \And 
  Urszula Czerwinska \\
  Jasper Research
  \And
  Damien Henry \\
  Jasper Research
  \And
  Cl\'ement Chadebec\footnotemark[1]\\
  Jasper Research
}

\begin{document}

\newcommand{\TODO}[1]{\textcolor{red}{\bf [TODO] #1}}

\maketitle

\vspace{-1cm}
\begin{figure}[H]
  \centering
  \includegraphics[width=0.6\linewidth]{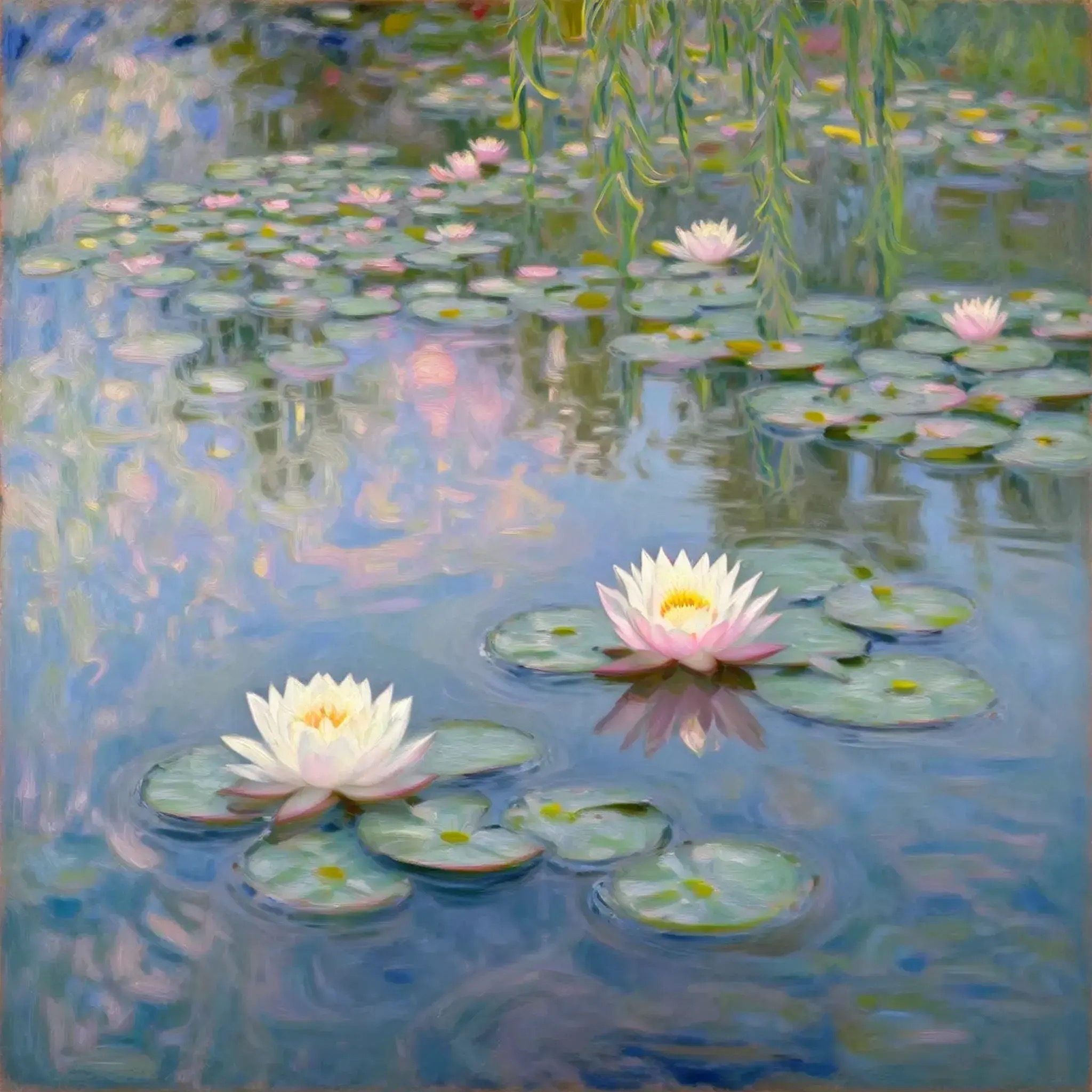}
  \caption{An impressionist water-lily painting generated at $2048\times2048$ by our 4B text-to-image model trained \emph{exclusively} on the MONET dataset, in homage to Claude Monet's \emph{Nymph\'eas} series.}
  \label{fig:monet-teaser}
\end{figure}

\begin{abstract}
    Training large text-to-image models requires high-quality, curated datasets with diverse content and detailed captions. Yet the cost and complexity of collecting, filtering, deduplicating, and re-captioning such corpora at scale hinder open and reproducible research in the field. We introduce MONET, an open \emph{Apache 2.0} dataset of $103.8$M image--text pairs collected from 2.9B raw pairs across heterogeneous open sources through successive stages of safety filtering, domain-based filtering, exact and near-duplicate removal, and re-captioning with multiple vision-language models covering short to long-form descriptions, and further augmented with synthetically generated samples. Each image is shipped with pre-computed embeddings and annotations to accelerate downstream use. To validate the effectiveness of MONET, we train a 4B-parameter latent diffusion model \emph{exclusively} on it and reach competitive GenEval and DPG scores, demonstrating that our dataset lowers the barrier to large-scale, reproducible text-to-image research.
\end{abstract}

\section{Introduction}
\label{sec:introduction}
Text-to-image (T2I) models have shown remarkable progress in generating realistic images from text descriptions \citep{ramesh2021zero,ramesh2022hierarchical,ho2022imagen,saharia2022photorealistic,rombach2022high,podell2023sdxl,kang2023scaling,chen2023pixart,chen2024pixart,esser2024scaling}. However, training such models requires large, curated corpora with strong visual diversity, high-quality images, and detailed captions. Collecting, filtering, deduplicating, and re-captioning such datasets at scale is expensive and time-consuming, restricting state-of-the-art T2I research to a handful of well-resourced players and hindering open, transparent, and reproducible work in the field. Early open initiatives, such as YFCC100M \citep{thomee2016yfcc100m}, LAION-400M / LAION-5B \citep{schuhmann2021laion,schuhmann2022laion}, and COYO-700M \citep{kakaobrain2022coyo700m}, have provided hundreds of millions to billions of web-crawled image--text pairs, but remain largely uncurated, highly redundant, and paired with short, noisy alt-text captions.

Recent works have shown that richer captions significantly boost T2I performance \citep{betker2023improving,chen2023pixart,esser2024scaling}, motivating the creation of synthetically re-captioned datasets such as ShareGPT4V \citep{chen2024sharegpt4v} (1.2M images). However, this dataset remains too small for pre-training large T2I models, and relying on a single Vision-Language Model (VLM) tends to bias the caption distribution and degrade out-of-distribution generation \citep{esser2024scaling,wu2025qwen}. To the best of our knowledge, no \textbf{openly released}, \textbf{filtered}, \textbf{deduplicated}, and \textbf{multi-VLM re-captioned} dataset is currently available for pre-training T2I models at scale.

In this paper, we bridge this gap by introducing \textbf{MONET}, a new large-scale dataset of \textbf{103.8M} image--text pairs released under the permissive \emph{Apache 2.0} license and specifically designed for training large T2I models. The dataset is available at \mbox{\url{https://huggingface.co/datasets/jasperai/monet}}. MONET is distilled from 2.9B raw pairs collected across nine heterogeneous open sources (6 \textit{real} and 3 \textit{synthetic}), using aesthetic pre-filtering, multi-classifier safety filtering, deduplication, and domain-based filtering for source governance. Each surviving image is re-captioned by multiple VLMs, ranging from short concept-level to long fine-grained descriptions, and the corpus is augmented with synthetic samples generated by \emph{Apache 2.0} T2I models. All samples are shipped with standard image embeddings (DINOv2 \citep{oquab2023dinov2}, CLIP \citep{radford2021learning}, SSCD \citep{pizzi2022self}), classifiers and detectors (YOLO \citep{jocher2023yolov8}, MediaPipe \citep{lugaresi2019mediapipe}), and pre-encoded with SANA VAE \citep{xie2025sana}. We also provide a comprehensive analysis of the dataset, including statistics, content and topic analyses, and human quality assessment, and validate its usefulness by training a 4B-parameter T2I model exclusively on MONET, which achieves competitive evaluation scores.

\section{Related work}  
\label{sec:related-works}

\paragraph{Text-to-image models}
Although early GAN-based approaches \citep{goodfellow_generative_2014,xu2018attngan,sauer2023stylegan,kang2023scaling} laid the groundwork for text-conditioned image generation, diffusion \citep{sohl2015deep,ho2020denoising,song2020score} and flow-based models \citep{liu2022flow,lipman2023flow} have become the dominant paradigms for T2I synthesis \citep{ramesh2021zero,ramesh2022hierarchical,ho2022imagen,saharia2022photorealistic,rombach2022high,razzhigaev2023kandinsky,podell2023sdxl,chen2023pixart,chen2024pixart,esser2024scaling,xie2024sana,wu2025qwen,xie2025sana}. These methods pair a powerful text encoder, whose output serves as conditioning, with a denoiser instantiated as a U-Net \citep{ronneberger2015u} or transformer \citep{vaswani2017attention,peebles2023scalable}. More recently, the success of Large Language Models (LLMs) \citep{brown2020language,hoffmann2022training,grattafiori2024llama,team2024gemini,liu2024deepseek,jiang2024mixtral} has motivated unified architectures that process all modalities in a shared space, either through autoregressive next-token prediction \citep{wang2024emu3,team2024chameleon,chen2025janus,wu2025janus,wu2025vila}, hybrid prediction--diffusion schemes \citep{zhao2024monoformer,zhou2025transfusion,ma2025janusflow,xie2025show}, or discrete diffusion \citep{swerdlow2025unified,li2026omni}.

\paragraph{Text--image datasets} Progress in VLMs and T2I models has been driven by the availability of large-scale image--text datasets. Early curated datasets such as MS-COCO \citep{lin2014microsoft}, Visual Genome \citep{krishna2017visual}, and Conceptual Captions (CC3M, CC12M) \citep{sharma2018conceptual,changpinyo2021cc12m} provide filtered image--text pairs useful for training captioning models, but their scale remains limited to several hundred thousand or several million samples, capping model scalability. Subsequent web-scale efforts \citep{hu2022scaling,jia2021scaling,desai2021redcaps} relied on noisy alt-text or social-media captions to reach hundreds of millions of pairs, culminating in YFCC100M \citep{thomee2016yfcc100m}, LAION-400M / LAION-5B \citep{schuhmann2021laion,schuhmann2022laion} and COYO-700M \citep{kakaobrain2022coyo700m}. Although these corpora enabled foundational VLMs and T2I models such as CLIP \citep{radford2021learning} and Stable Diffusion \citep{rombach2022high}, they remain highly redundant and contain misaligned, unfiltered, or unsafe content -- issues partially addressed by the safety-revised Re-LAION \citep{relaion}. To improve caption quality, several works re-caption images with VLMs \citep{betker2023improving,chen2023pixart,esser2024scaling}, most notably ShareGPT4V \citep{chen2024sharegpt4v}, which provides 1.2M GPT-4V-generated captions but is too small for large-scale pre-training and tied to a single captioner, biasing the dataset toward a single prompt distribution.

\paragraph{Dataset curation} Data quality is now widely accepted to matter more than raw quantity for training large multimodal models \citep{zhou2023lima,chen2023pixart,qin2025lumina,esser2024scaling}, since uncurated web data is noisy and often misaligned. Early curation pipelines relied on simple aesthetic scores \citep{laion_aesthetic_classifier} or CLIP-score \citep{radford2021learning} thresholding  to enforce image--text alignment. More recent efforts, such as the DataComp benchmark \citep{gadre2023datacomp}, systematically search the filter-design space, while Data Filtering Networks \citep{fang2024data} train specialized models to score and discard low-quality or misaligned samples. MONET builds upon these curation strategies by combining rigorous aesthetic, safety, and watermark filtering with pre-trained network models and careful re-captioning with captioning models of varying complexity, thereby enabling a rich and diverse prompt distribution.

\paragraph{Data deduplication} An often under-addressed issue in large multimodal corpora is the prevalence of duplicate or near-duplicate samples, which skew the data distribution and induce memorization \citep{kandpal2022deduplicating,lee2022deduplicating}, a particularly pressing concern for diffusion models \citep{somepalli2023diffusion,carlini2023extracting}. MONET addresses this issue by using a combination of deduplication methods, such as perceptual hashing \citep{venkatesan2000robust} and Self-Supervised Copy Detection (SSCD) \citep{pizzi2022self}, to remove near-duplicate images from the dataset.

\section{Dataset construction}
\label{sec:dataset-construction}

\begin{figure}[t]
  \centering
  \includegraphics[width=\textwidth]{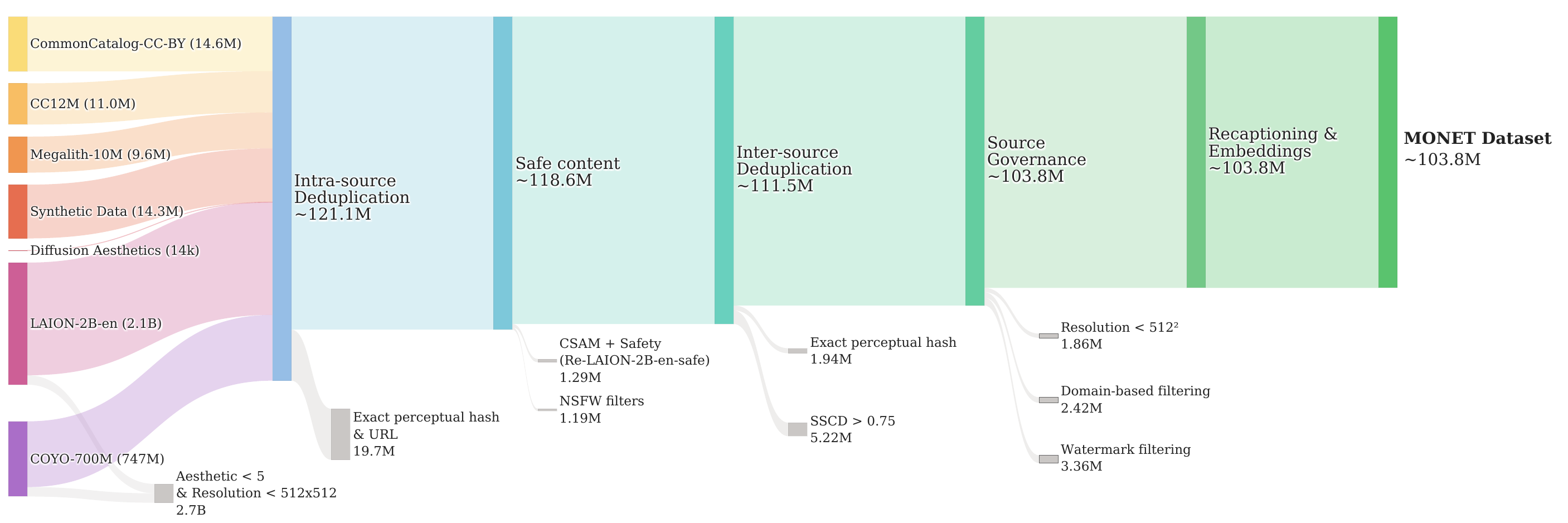}
  \caption{Curation pipeline of MONET. Each stage removes images that fail the corresponding quality, safety or source-governance checks, while the surviving pool flows to the next step.}
  \label{fig:sankey}
\end{figure}

In this section, we detail the construction of the MONET dataset. Starting from heterogeneous open sources totaling 2.9B raw image--text pairs, we apply successive stages of pre-filtering, safety filtering, deduplication, and domain-based filtering, followed by multi-VLM re-captioning and synthetic-data augmentation, to obtain a final dataset of \textbf{103.8M} high-quality and safe image--text pairs. The complete curation pipeline is illustrated in Fig.~\ref{fig:sankey}.

\subsection{Data sourcing}
MONET is built from existing open-source datasets selected via source-governance criteria, chosen to maximize diversity in content, visual style, and resolution while supporting reproducibility. As summarized in Table~\ref{tab:dataset-sources}, the bulk of the pool ($>$2.8B images, from LAION, COYO and CC12M) comes with noisy \emph{alt-text} captions, while 14.6M images are pre-captioned with VLMs such as BLIP2 \citep{li2023blip} (Common-Catalog) and 14k with GPT-4o \citep{hurst2024gpt} (Diffusion-Aesthetic-4K). Finally, 9.6M images have no captions (Megalith-10M \citep{BoerBohan2024Megalith10m}). We deliberately exclude several popular alternatives relying on Common Crawl, such as DataComp-1B \citep{gadre2023datacomp}, since they heavily overlap with LAION and COYO, as well as the non-English part of LAION-5B \citep{schuhmann2022laion}, since multilingual coverage is more reliably obtained via translation than from noisy \emph{alt-text}.

\begin{table}[ht]
  \caption{Summary of the sources used in compiling MONET, together with approximate statistics and licensing information. Top rows correspond to real image--text sources, while bottom rows report images generated synthetically (see Sec.~\ref{sec:synthetic-data}).}
\centering
\scriptsize
\setlength{\tabcolsep}{4pt}
\begin{tabular}{lcccccc}
\toprule
\textbf{Dataset Name} & \textbf{\# Images (original)} & \textbf{\# Images (final)} &  \textbf{Image source}     & \textbf{Caption source} & \textbf{License}       \\
\hline
LAION (2B-en) \citep{schuhmann2022laion}               & 2.1B       & 46.3M   & Common Crawl      & Alt-text & CC-BY-4.0     \\
COYO \citep{kakaobrain2022coyo700m}                & 747M               & 18.5M   & Common Crawl      & Alt-text & CC-BY-4.0    \\
Common-Catalog-CC-BY \citep{gokaslan2023commoncanvas}     & 14.6M                 & 11.0M   & Flickr$^*$ & BLIP2 &  CC-BY-4.0 \\
Megalith-10M \citep{BoerBohan2024Megalith10m}                                              & 9.6M              & 8.0M   & Flickr            & None & MIT (metadata)        \\
Conceptual-12M \citep{changpinyo2021cc12m}               & 11.0M                 & 6.2M   & Web               & Alt-text    & Google (Permissive)          \\
Diffusion-Aesthetic-4K \citep{zhang2025diffusion4k}       & 14k                & 12.8k   & Web               & GPT-4o  & MIT           \\
\midrule
\multicolumn{6}{l}{\emph{Synthetic data (see Sec.~\ref{sec:synthetic-data})}} \\
Z-Image \citep{zimage2025}                                & 6.2M              & 5.9M   & Synthetic         & Synthetic & Apache 2.0 \\
FLUX.2-klein-4B \citep{flux-2-2025}                       & 3.6M              & 3.5M   & Synthetic         & Synthetic & Apache 2.0 \\
FLUX.1-schnell \citep{flux2024}                           & 4.5M              & 4.4M   & Synthetic         & Synthetic & Apache 2.0 \\
\bottomrule
\tiny{$^*$Images extracted from YFCC100M \citep{thomee2016yfcc100m}}\\
\end{tabular}
\label{tab:dataset-sources}
\end{table}

\subsection{Pre-filtering}
\label{sec:prefiltering}

\begin{figure}[ht]
  \centering
  \begin{minipage}[t]{0.165\textwidth}
    \centering
    \includegraphics[height=1.95cm, width=0.9\textwidth]{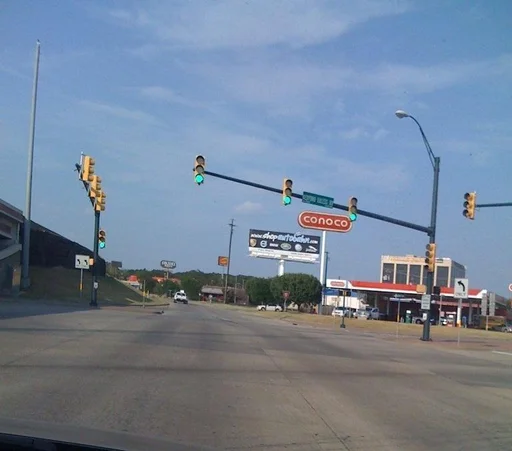}\\[2pt]
    {\small 3.76}
  \end{minipage}%
  \hfill
  \begin{minipage}[t]{0.165\textwidth}
    \centering
    \includegraphics[height=1.95cm, width=0.9\textwidth]{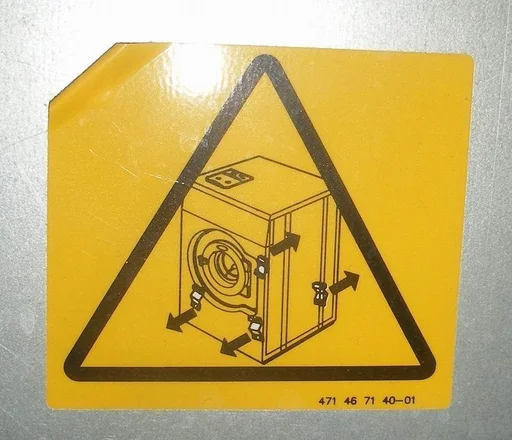}\\[2pt]
    {\small 4.09}
  \end{minipage}%
  \hfill
  \begin{minipage}[t]{0.165\textwidth}
    \centering
    \includegraphics[height=1.95cm, width=0.9\textwidth]{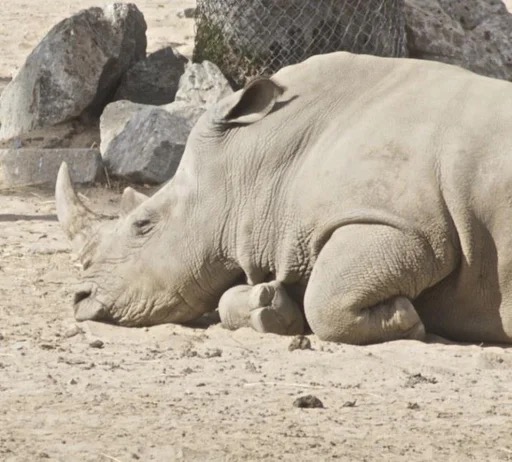}\\[2pt]
    {\small 4.71}
  \end{minipage}%
  \hfill
  \begin{minipage}[t]{0.165\textwidth}
    \centering
    \includegraphics[height=1.95cm, width=0.9\textwidth]{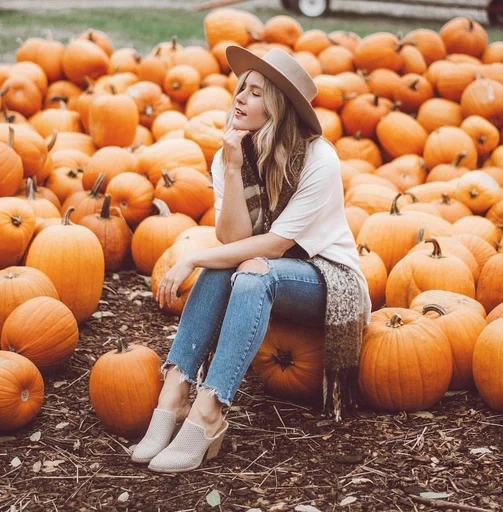}\\[2pt]
    {\small 5.34}
  \end{minipage}%
  \hfill
  \begin{minipage}[t]{0.165\textwidth}
    \centering
    \includegraphics[height=1.95cm, width=0.9\textwidth]{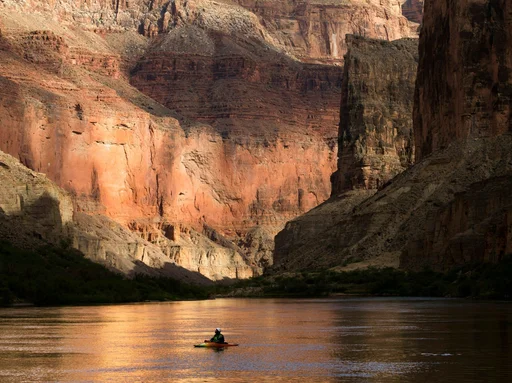}\\[2pt]
    {\small 6.26}
  \end{minipage}%
  \hfill
  \begin{minipage}[t]{0.165\textwidth}
    \centering
    \includegraphics[height=1.95cm, width=0.7\textwidth]{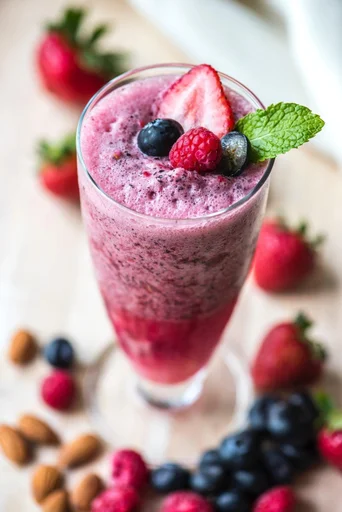}\\[2pt]
    {\small 7.51}
  \end{minipage}
  \caption{Examples of images at different aesthetic scores~\citep{laion_aesthetic_classifier}. A lower aesthetic score is correlated with lower-quality or visually unappealing images, motivating the pre-filtering stage.}
  \label{fig:aesthetic_examples}
\end{figure}

For the two largest sources, LAION and COYO, we apply two pre-filters before merging them with the smaller datasets, concentrating computational resources on images that meet our baseline quality requirements. First, we exclude images with a resolution below $512^2$ pixels, as low-resolution samples typically lack sufficient detail, reducing the effectiveness of pre-training. Second, we filter out images with an aesthetic score \citep{laion_aesthetic_classifier} below 5.0, shifting the pool toward more visually appealing samples (see Fig.~\ref{fig:aesthetic_examples}).
Combined, these two filters retain roughly 91M images from LAION and COYO. After merging with the four smaller real-image sources and the synthetic images (described in Sec.~\ref{sec:synthetic-data}), and applying intra-source URL/pHash deduplication (described in Sec.~\ref{sec:deduplication}), we obtain a \textbf{121.1M} merged pool that serves as the reference baseline for the cumulative reductions reported in the remaining stages.

\subsection{Safety filtering}
\label{sec:safety-filtering}
Since the data come mainly from the Web, we apply strict safety filters to the merged pool. First, we restrict LAION-2B-en samples to those also present in the vetted Re-LAION-2B-en-safe release \citep{relaion}, removing 1.29M images flagged during the Re-LAION safety revision. Second, we apply an ensemble of open-source \emph{Not-Safe-For-Work} (NSFW) detectors (Falcon \citep{falcon_nsfw_classifier} and Bumble \citep{bumble_nsfw_classifier}) together with an internal classifier, under a conservative union rule: an image is removed if any classifier flags it. This leverages the complementary failure modes of the detectors to minimize false negatives at the cost of some false positives, and removes an additional 1.19M images. Finally, we conduct a safety audit using DINOv2 \citep{oquab2023dinov2} embeddings by manually inspecting the 100 nearest neighbors of a small seed set of NSFW images; no additional harmful content is detected, thereby supporting the effectiveness of the previous steps. After safety filtering, the pool is reduced to \textbf{118.6M} safe images (2.1\% cumulative reduction). While no filtering pipeline can guarantee perfect coverage, this multi-layered approach substantially reduces the likelihood that harmful content will persist in the final dataset. See Appendix~\ref{sec:filtering-details} for a more in-depth discussion of the filters used.

\subsection{Deduplication}
\label{sec:deduplication}

Deduplication is crucial to ensure diversity and prevent memorization and overfitting \citep{lee2022deduplicating,kandpal2022deduplicating,somepalli2023diffusion,gu2025on}. We use a two-stage strategy combining exact and near-duplicate detection.

\begin{figure}[t]
  \centering
  \newcommand{\dedupimg}[1]{\includegraphics[height=1.5cm]{#1}}
  \newcommand{\dedupcap}[1]{{\small #1}}
  \newcommand{\deduppair}[3]{%
    \begin{tabular}[t]{@{}c@{}}
      \dedupimg{#1}\hspace{1pt}\dedupimg{#2}\\[2pt]
      \dedupcap{#3}
    \end{tabular}%
  }
  \deduppair{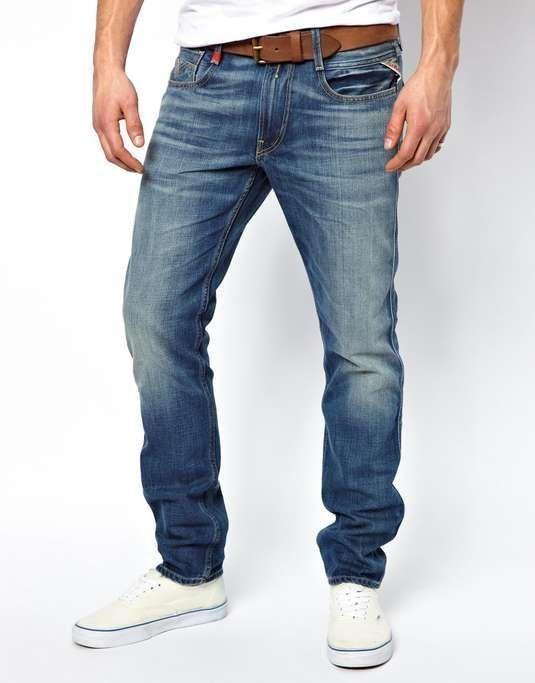}%
           {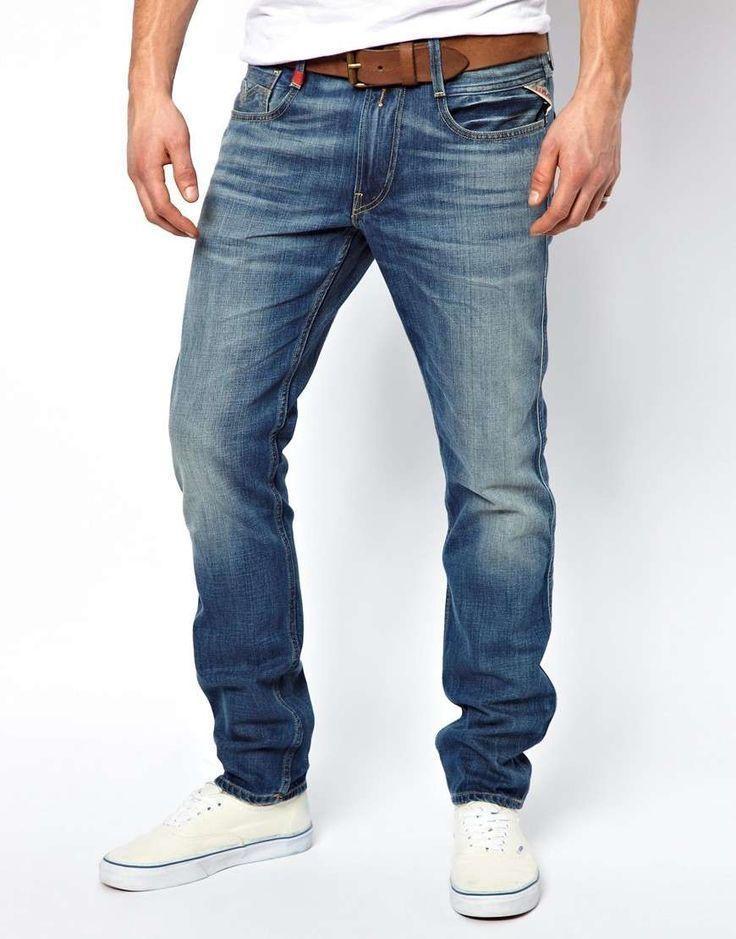}%
           {SSCD\,=\,0.92,\; $d$\,=\,2}\hfill
  \deduppair{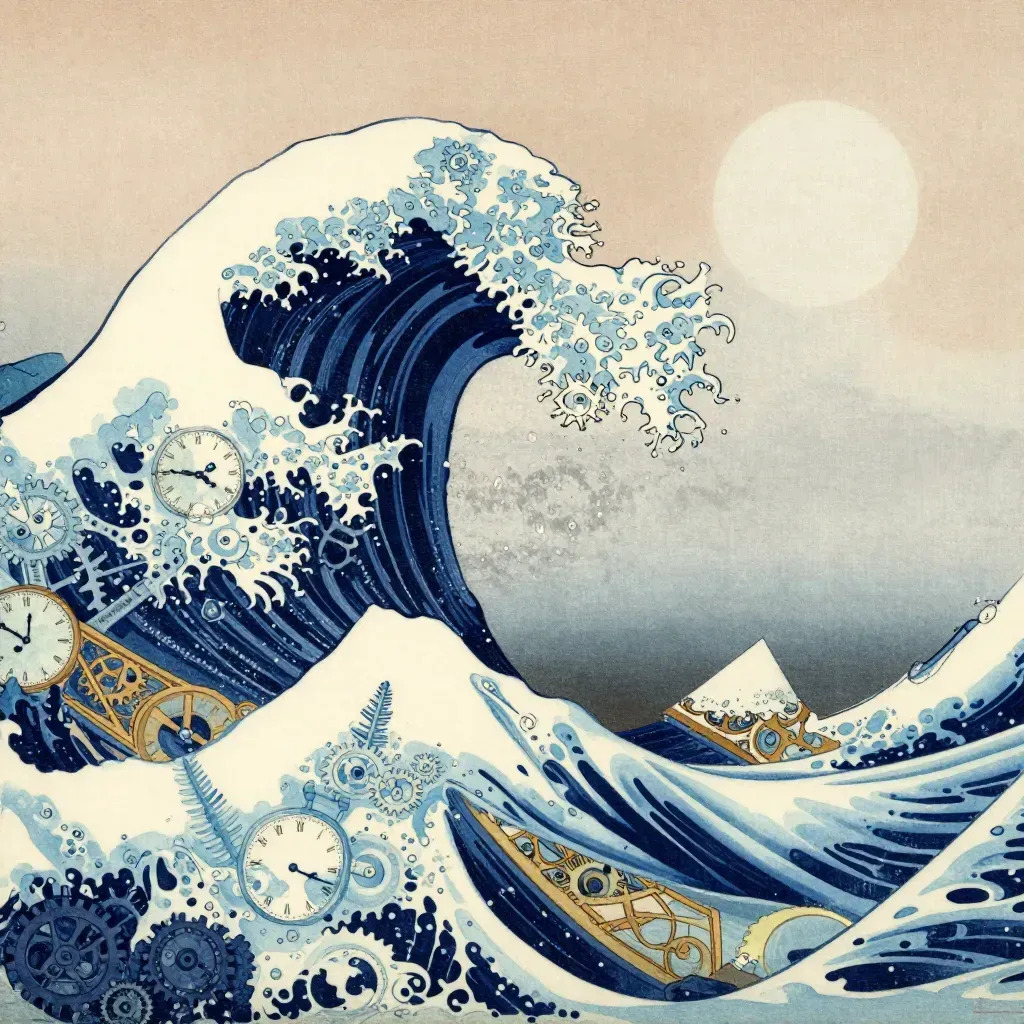}%
           {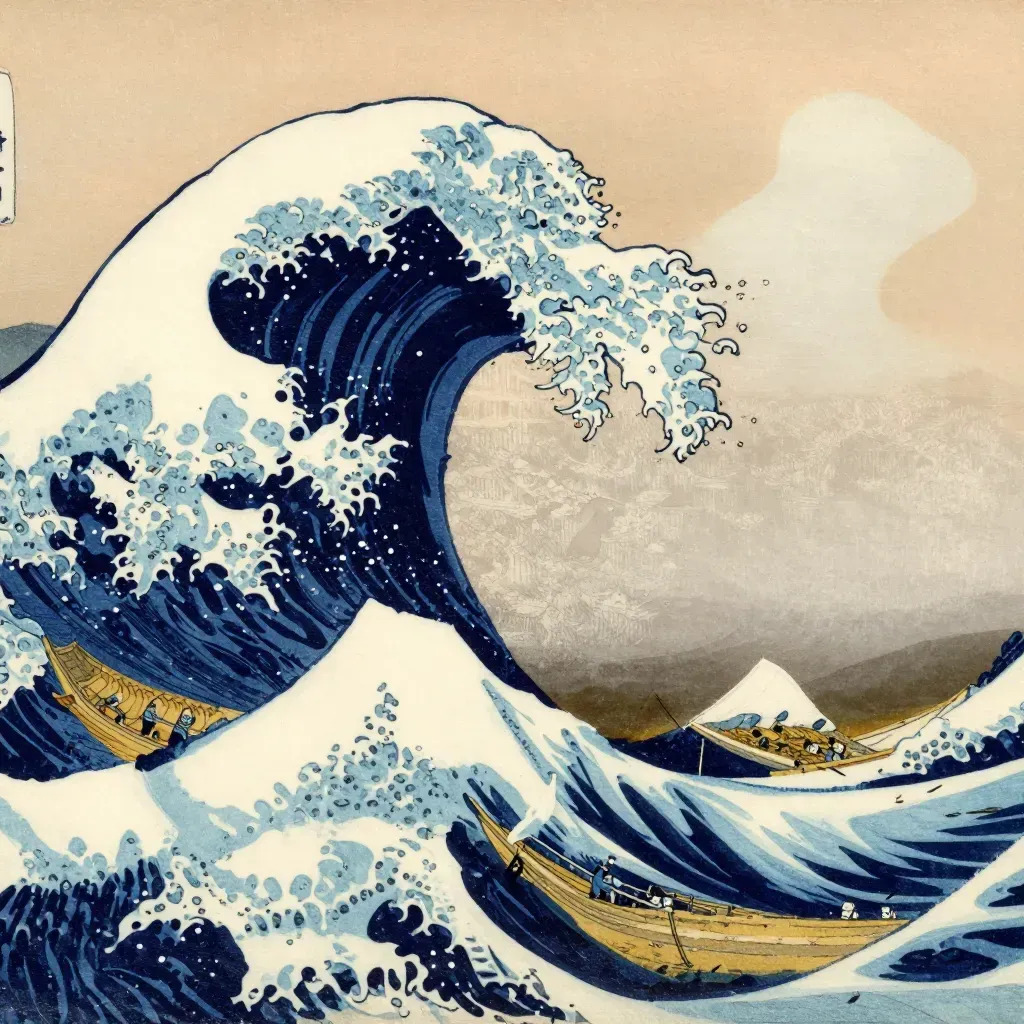}%
           {SSCD\,=\,0.90,\; $d$\,=\,4}\hfill
  \deduppair{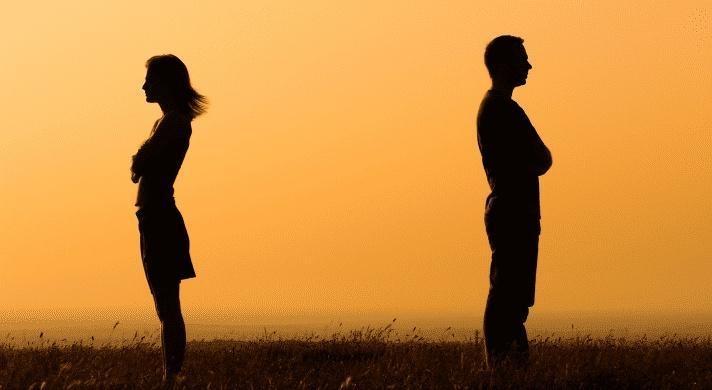}%
           {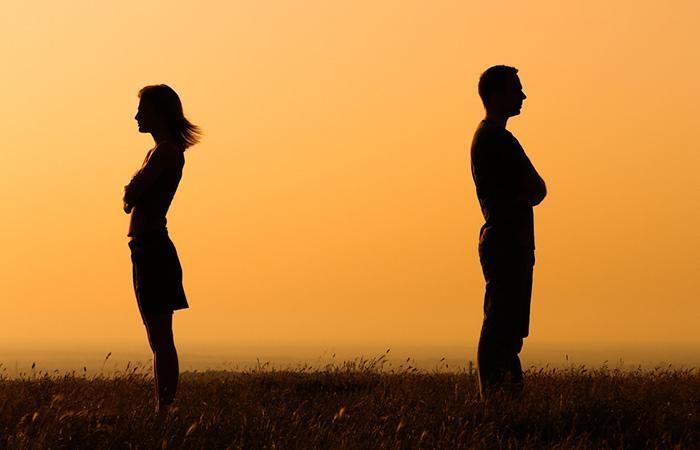}%
           {SSCD\,=\,0.86,\; $d$\,=\,10}\hfill
  \deduppair{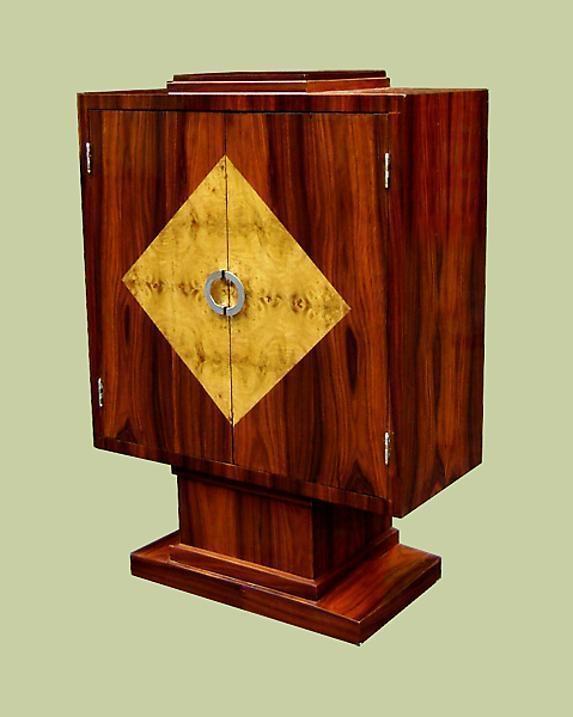}%
           {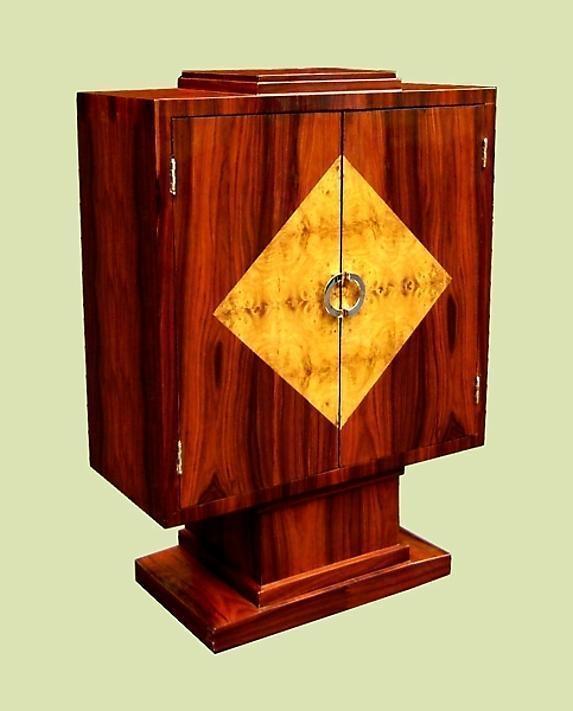}%
           {SSCD\,=\,0.76,\; $d$\,=\,26}
  \\[4pt]
  \deduppair{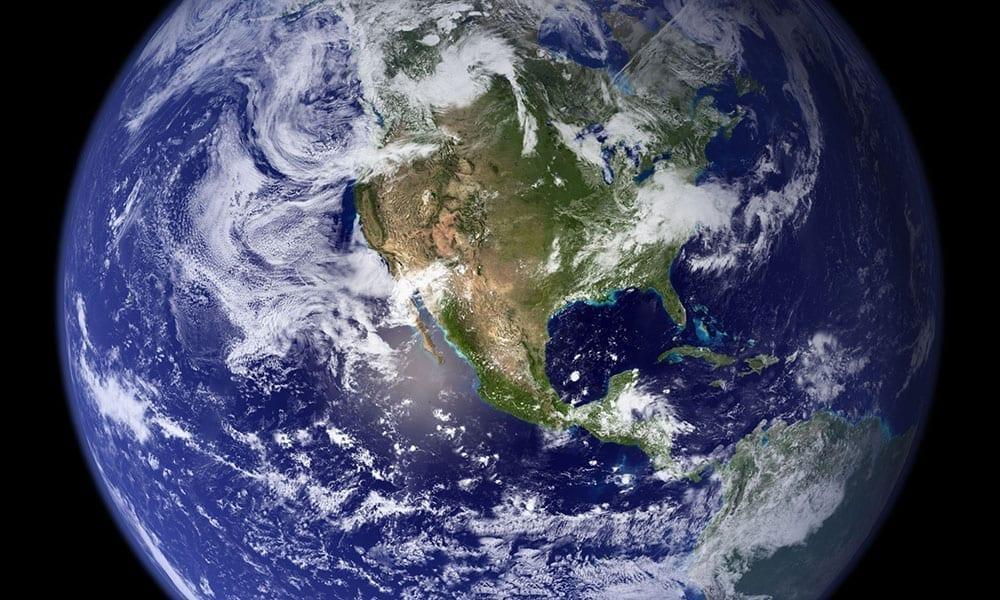}%
           {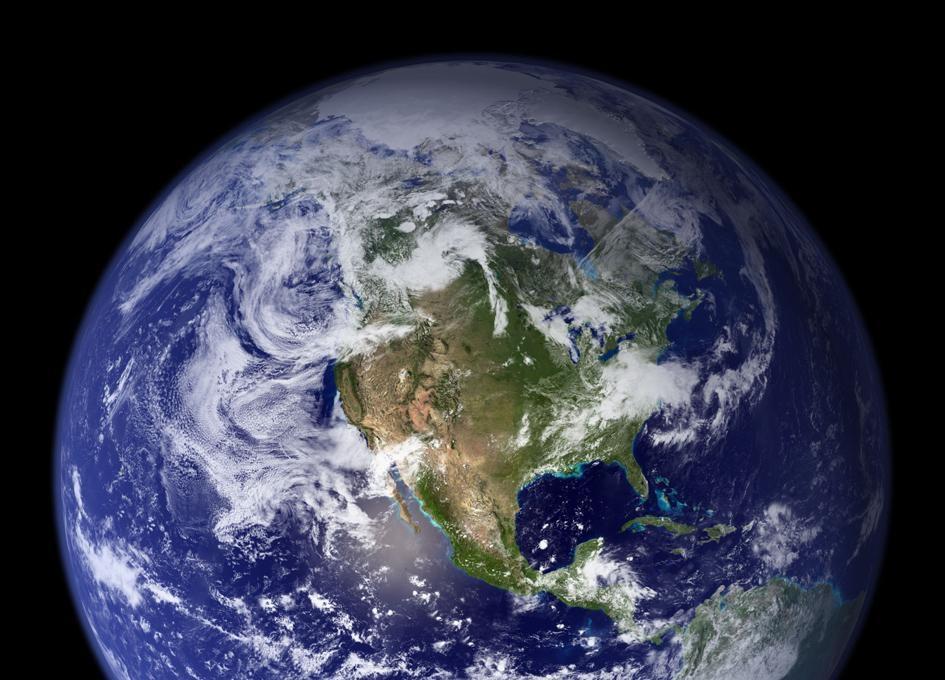}%
           {SSCD\,=\,0.71,\; $d$\,=\,32}\hfill
  \deduppair{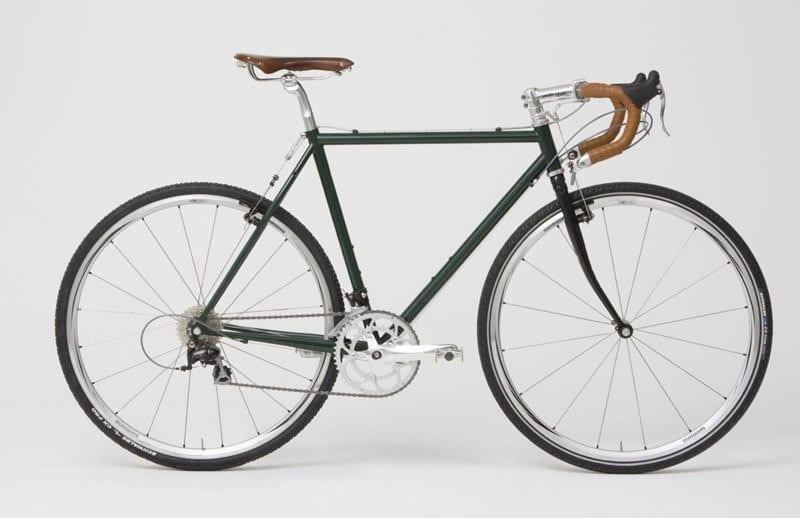}%
           {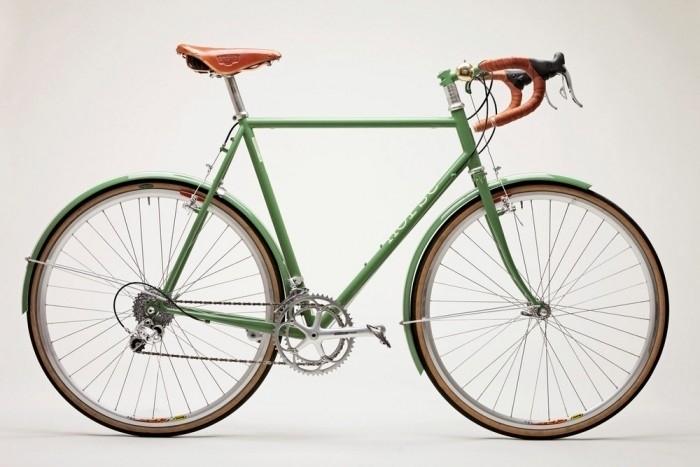}%
           {SSCD\,=\,0.65,\; $d$\,=\,20}\hfill
  \deduppair{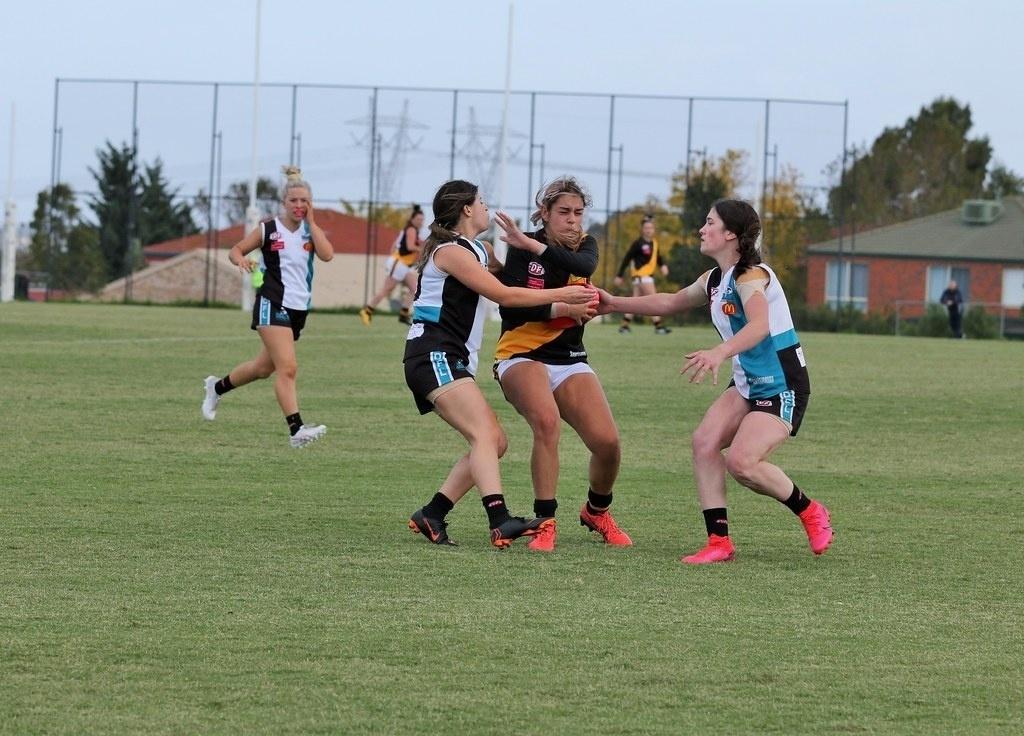}%
           {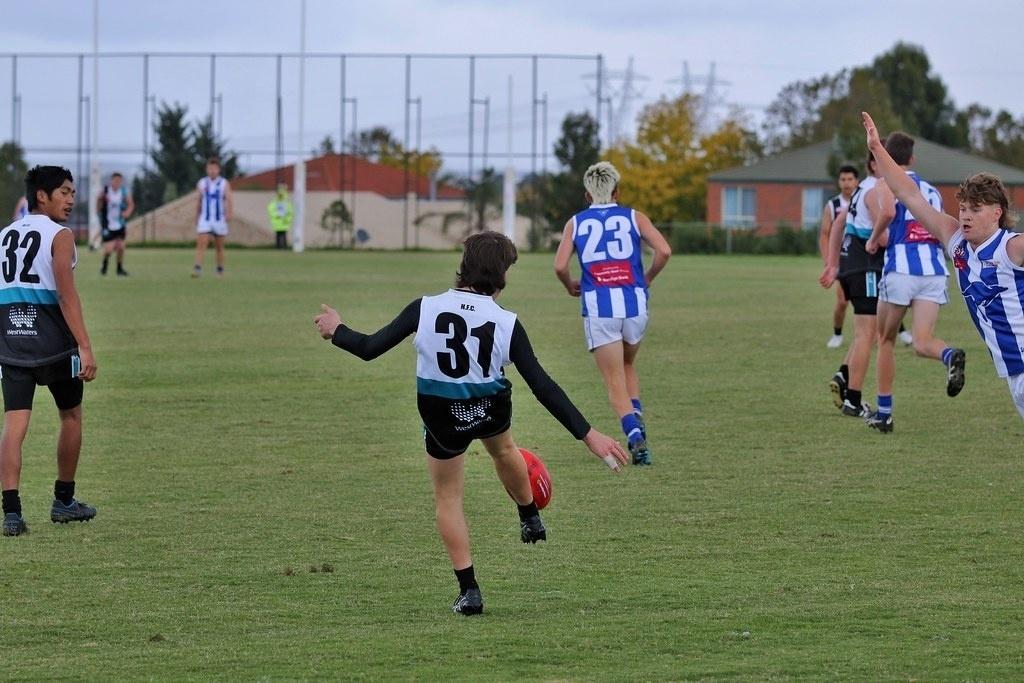}%
           {SSCD\,=\,0.51,\; $d$\,=\,20}
  \caption{SSCD nearest-neighbor pairs with cosine similarity and pHash Hamming distance~$d$. \emph{Top:} near-duplicates removed (SSCD\,$\geq$\,0.75); pHash degrades under flips, crops, or background swaps while SSCD remains high. \emph{Bottom:} semantic neighbors retained (SSCD\,$<$\,0.75).}
  \label{fig:dedup_examples}
\end{figure}

\paragraph{URL and perceptual hashing}
 We start by removing exact URL duplicates, then apply DCT-based perceptual hashing (pHash) \citep{venkatesan2000robust} to detect near-exact copies that differ only in compression or scaling. These steps are applied first to each source individually (removing ${\sim}19.7$M intra-source duplicates) and then to the merged safe pool (removing 1.94M additional inter-source duplicates).
Because pHash retains only the lowest-frequency DCT coefficients, it cannot capture geometric transforms such as flips, crops, or color shifts and is therefore unreliable for identifying near-duplicates: such pairs can reach large Hamming distances, overlapping the range of unrelated images and precluding a reliable threshold (see Fig.~\ref{fig:dedup_examples}).

\paragraph{SSCD near-duplicate detection}
To handle near-duplicates, we rely on Self-Supervised Copy Detection (SSCD) \citep{pizzi2022self}. We compute 512-d SSCD embeddings with the public \texttt{sscd\_disc\_mixup} model \citep{sscd_code} and retrieve the $k=64$ nearest neighbors per image using a FAISS index \citep{douze2024faiss}; $k=64$ trades off search speed against cluster recall, and is large enough to cover the maximum near-duplicate cluster size we observe empirically. Pairs whose cosine similarity exceeds $0.75$ are collapsed (keeping the representative with the highest resolution and aesthetic score), removing 5.22M additional images. The $0.75$ threshold corresponds to the operating point recommended by the SSCD authors at $90\%$ precision on DISC \citep{sscd_code}, and we validate it on our data by manually inspecting pair slices at $0.05$ resolution (see Fig.~\ref{fig:dedup_examples}). We find that pairs above $0.75$ are consistently near-duplicates (crops, flips, color shifts, watermarks), while pairs below $0.75$ are semantically related but visually distinct (\emph{e.g.,}\ different frames from the same series), which we retain for diversity. After deduplication, the pool contains \textbf{111.5M} unique images (8.0\% cumulative reduction). See Appendix~\ref{sec:deduplication-details} for more details and limitations of the deduplication strategy.

\subsection{Domain-based filtering and source governance}
\label{sec:compliance-filtering}

A final round of exclusion-based filters enforces resolution, source, and watermark standards. We remove images with resolution below $512^2$ pixels (1.86M images, mostly from the smaller sources, which were not pre-filtered in Sec.~\ref{sec:prefiltering}); then images originating from a blocklist of domains including known stock-photo providers such as \emph{dreamstime}, \emph{shutterstock}, \emph{freepik}, \emph{getty}, and \emph{unsplash} (2.42M images); and finally images flagged with high watermark probability (3.36M images): 2.70M by an internal detector and ${\sim}0.66$M by the LAION watermark detector \citep{laion_watermark_detector} (see Appendix~\ref{sec:filtering-details}). This leaves a final pool of \textbf{103,816,750} images (103.8M, 14.3\% cumulative reduction). These exclusion controls are not a representation of legal clearance; they are source-governance signals that reduce the prevalence of images from known restrictive providers.

\subsection{Re-captioning}
\label{sec:recaptioning}

Caption quality and diversity are both crucial for T2I models. Recent works have shown that richer captions significantly boost model performance \citep{betker2023improving,chen2023pixart,esser2024scaling,qin2025lumina}, but human-annotated captions are prohibitively expensive at the scale of hundreds of millions of images. A widely adopted alternative is to synthesize image captions using pre-trained vision-language models \citep{hurst2024gpt,chen2024sharegpt4v}. However, relying on a single captioner biases the prompt distribution and can degrade out-of-distribution generation \citep{esser2024scaling,wu2025qwen}. To mitigate this, we re-caption MONET with multiple VLMs of varying complexity. We first benchmark several candidates: BLIP2 \citep{li2023blip}, Florence-2-Large \citep{xiao2024florence}, FastVLM \citep{vasu2025fastvlm}, CogVLM1/2 \citep{wang2024cogvlm,hong2024cogvlm2}, InternVL3-8B/14B/38B \citep{zhu2025internvl3}, GPT-4V via ShareGPT4V-style captioning \citep{chen2024sharegpt4v} and Gemini 2.5 Flash-Lite \citep{comanici2025gemini}; and compare caption complexity, latency and quality on 100 diverse images. Based on these trade-offs, we retain only Florence-2-Large, InternVL3-8B, ShareGPT4V-7B, and Gemini 2.5 Flash-Lite. Florence-2-Large produces short, concept-level captions that closely match typical user prompts, while the three remaining models yield long, fine-grained descriptions. Representative outputs are shown in Appendix~\ref{sec:caption-samples-and-alignment}. To validate this selection, we correlate automatic alignment scores with Elo scores from human voting. We observe that the standard CLIP metric correlates poorly with human judgment on long captions since its 77-token context truncates most detailed outputs. We therefore report alignment with LongCLIP \citep{zhang2024long} in Fig.~\ref{fig:text_image_alignment}, which handles longer inputs and tracks human preferences more faithfully. The conclusion holds for other long-context encoders such as Jina-CLIP-v2~\citep{jinav2025jinav2} (see Appendix~\ref{sec:human-quality-assessment}). Representative re-captioning examples, CLIP/LongCLIP alignment scores, Elo correlations, and the human-voting methodology are reported in Appendix~\ref{sec:caption-samples-and-alignment}.

\subsection{Synthetic data}
\label{sec:synthetic-data}

\begin{figure}[h]
  \centering
  \setlength{\tabcolsep}{1pt}
  \renewcommand{\arraystretch}{1.1}
  \begin{subfigure}[t]{0.32\textwidth}
    \centering
    \begin{tabular}{@{}cc@{}}
      \includegraphics[width=0.49\linewidth]{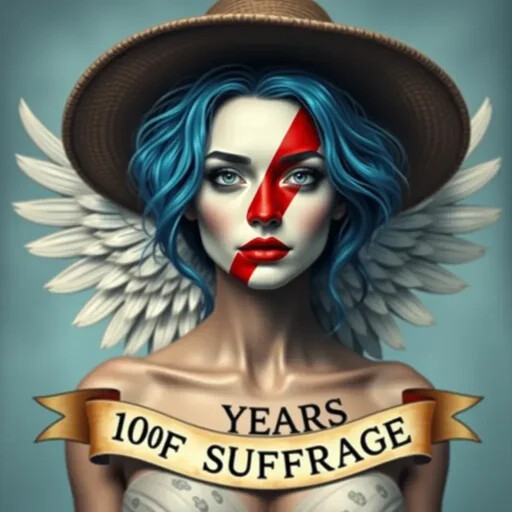} &
      \includegraphics[width=0.49\linewidth]{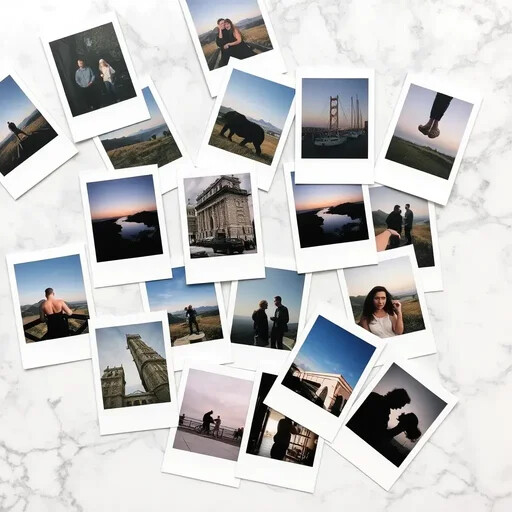} \\
    \end{tabular}
    \caption{FLUX.1-schnell}
    \label{fig:synthetic_images_schnell}
  \end{subfigure}\hfill
  \begin{subfigure}[t]{0.32\textwidth}
    \centering
    \begin{tabular}{@{}cc@{}}
      \includegraphics[width=0.49\linewidth]{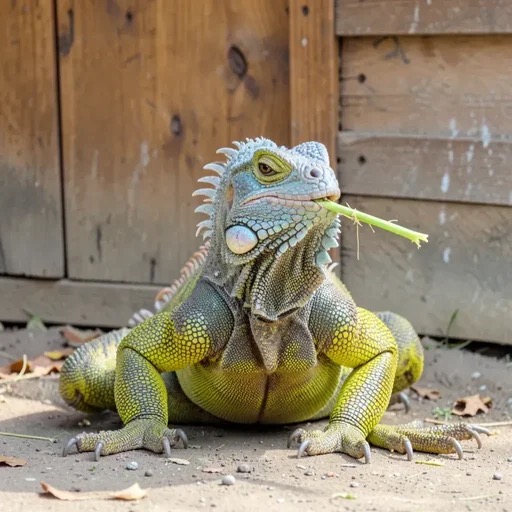} &
      \includegraphics[width=0.49\linewidth]{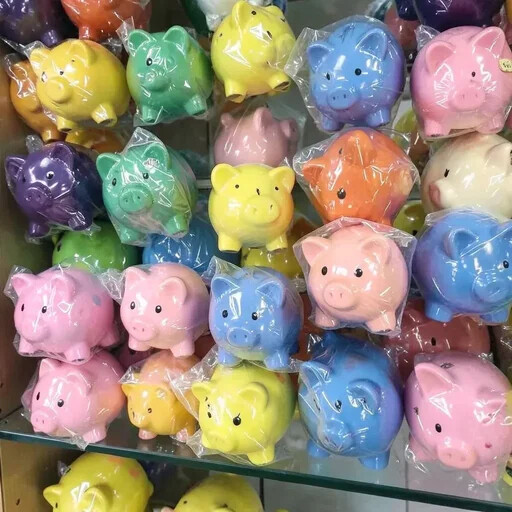} \\
    \end{tabular}
    \caption{FLUX.2-klein-4B}
    \label{fig:synthetic_images_klein}
  \end{subfigure}\hfill
  \begin{subfigure}[t]{0.32\textwidth}
    \centering
    \begin{tabular}{@{}cc@{}}
      \includegraphics[width=0.49\linewidth]{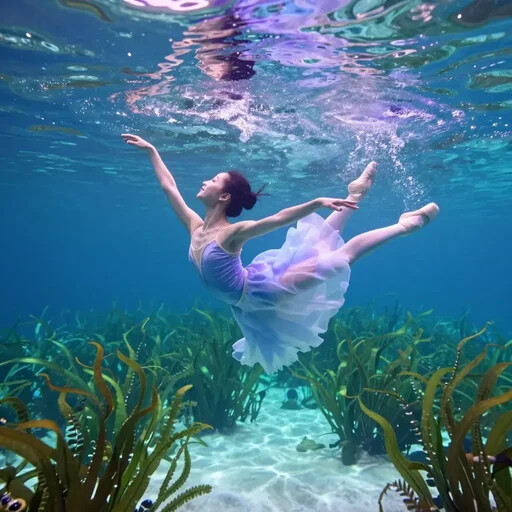} &
      \includegraphics[width=0.49\linewidth]{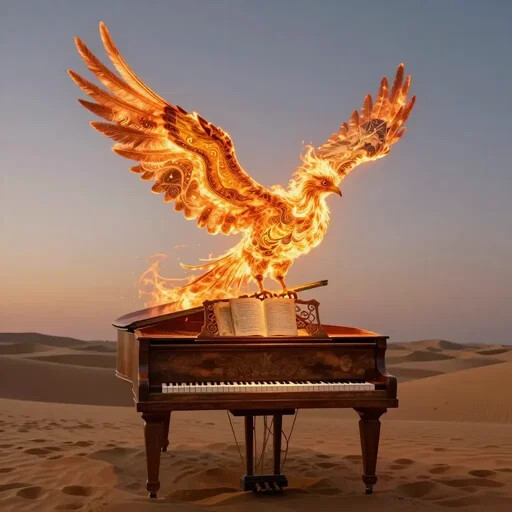} \\
    \end{tabular}
    \caption{Z-Image}
    \label{fig:synthetic_images_zimage}
  \end{subfigure}
  \caption{Examples of synthetic images generated for the MONET dataset using different models. Prompts are drawn from MONET and \cite{improved-flux-prompts}, then upsampled with Qwen3-4B \citep{yang2025qwen3}.}
  \label{fig:synthetic_images}
\end{figure}

We complement real data with synthetic images generated by FLUX.1-schnell \citep{flux2024}, FLUX.2-klein-4B \citep{flux-2-2025}, and Z-Image \citep{zimage2025}, chosen as top-performing T2I models released under the permissive \emph{Apache~2.0} license, which allows redistribution and use of their outputs for training. Prompts are drawn from re-captioning (Sec.~\ref{sec:recaptioning}) and an open-source prompt collection \cite{improved-flux-prompts}, then upsampled with Qwen3-4B \citep{yang2025qwen3} under a system prompt that removes unsafe content. The generated images are filtered with the same NSFW and watermark detectors used in Sec.~\ref{sec:safety-filtering} and the domain-based filters of Sec.~\ref{sec:compliance-filtering}. Examples are shown in Fig.~\ref{fig:synthetic_images}. As shown in Sec.~\ref{sec:impact-of-synthetic-data}, mixing in moderate amounts of synthetic data improves text--image alignment.

\subsection{Image encoding \& VAE pre-encoding}
\label{sec:image-encoding}

To accelerate downstream use, each MONET image is shipped with pre-computed embeddings, structured annotations, and latents, avoiding repeated raw-pixel processing. We store three complementary image embeddings: DINOv2-vitg14 \citep{oquab2023dinov2} for general-purpose scene representations (retrieval, classification), CLIP-vit-base-patch32 \citep{radford2021learning} for image--text alignment (cross-modal search, zero-shot classification), and SSCD \citep{pizzi2022self} for vector search and deduplication at scale. We release the FAISS indexes for all the embeddings at \url{https://huggingface.co/spaces/jasperai/monet-retrieval} and a Uniform Manifold Approximation and Projection (UMAP) visualization at \url{https://huggingface.co/spaces/jasperai/monet-umap}. We further release compact annotations from lightweight models, directly usable for filtering, balancing, and conditional generation: YOLO-v9e object detection \citep{redmon2016you,jocher2023yolov8} (80 COCO categories, for object-centric queries and layout-conditioned generation), YOLO-v8x image classification \citep{jocher2023yolov8} (distribution over 1{,}000 ImageNet-1k categories), and MediaPipe face detection \citep{lugaresi2019mediapipe} (face counts, boxes, and landmarks, for portrait filtering and privacy-aware subsampling). Finally, each image is accompanied by a pre-encoded latent from the SANA VAE \citep{xie2025sana}, enabling latent diffusion training directly on compressed representations and cutting storage, bandwidth, and encoding time.

\begin{figure}[!t]
\begin{tcolorbox}[title=Re-captioning example]
  \centering
  \begin{minipage}[c]{0.55\textwidth}
    \centering
    \includegraphics[width=\linewidth, keepaspectratio]{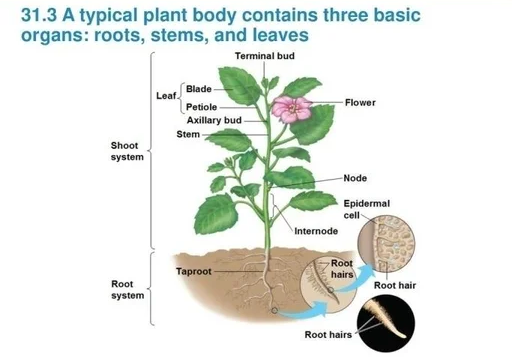}
  \end{minipage}%
  \hfill
  \begin{minipage}[c]{0.44\textwidth}
    \raggedright\footnotesize
    \textcolor{capOriginal}{\textbf{Original:}} 31.3 A typical plant body contains three basic organs: roots, stems, and leaves\par\vspace{3pt}
    \textcolor{capFlorence}{\textbf{Florence-2-Large:}} The image shows a plant with a flower, its roots, stems, and leaves clearly visible. The poster also contains text, providing further information about the plant's anatomy.\par\vspace{3pt}
    \textcolor{capShareGPT}{\textbf{ShareGPT4V-7B:}} The image presents a detailed diagram of a plant's anatomy, focusing on the root system. The plant is depicted with a pink flower at the top, surrounded by green leaves. The root system is shown in a cross-section view, revealing the taproot and root hairs. The diagram is labeled with the names of the different parts, such as the shoot system, leaf, stem, terminal bud, flower, node, intermodal cell, and root hairs. The diagram is a visual representation of the plant's structure, providing a clear understanding of its anatomy.\par\vspace{3pt}
    \textcolor{capGemini}{\textbf{Gemini 2.5 Flash-Lite:}} The image is a detailed illustration of a typical plant's basic organs: roots, stems, and leaves, accompanied by magnified views of root hairs and epidermal cells. The main object is a vibrant green plant with a pink flower blooming from its upper portion. The plant is divided into two main systems: the shoot system, which includes the leaves, stem, axillary bud, terminal bud, and flower; and the root system, which consists of a taproot and numerous root hairs extending into the soil. The scene depicts the plant partially submerged in brown soil, with the shoot system above ground and the root system below. Various parts of the plant are labeled with text, such as ``Leaf'' (further broken down into ``Blade'' and ``Petiole''), ``Stem'', ``Flower'', ``Terminal bud'', ``Axillary bud'', ``Node'', and ``Internode''. Two inset circles provide magnified views: one shows the internal structure of a root, highlighting ``Root hairs'' emerging from epidermal cells, and the other provides a close-up of these fine, hair-like structures. The style is educational and illustrative, using clear lines, vibrant colors, and informative labels to convey botanical information effectively. The caption at the top, ``31.3 A typical plant body contains three basic organs: roots, stems, and leaves,'' sets the context for the diagram.\par\vspace{3pt}
    \textcolor{capInternVL}{\textbf{InternVL3-8B:}} The image illustrates the structure of a typical plant body, emphasizing its three fundamental parts: roots, stems, and leaves. The root system is depicted beneath the soil, showcasing the taproot along with numerous root hairs that facilitate water absorption. The shoot system is illustrated vertically, highlighting nodes, internodes, and buds, including the axillary bud and terminal bud. Leaves, attached to the stem via petioles, are also detailed, showing the blade and flower structures. The image indicates that the root hair cell is not the same as the epidermal cell.
  \end{minipage}
\end{tcolorbox}
  \caption{Representative re-captioning example, comparing the \textcolor{capOriginal}{\textbf{original}} web caption with captions produced by \textcolor{capFlorence}{\textbf{Florence-2-Large}}, \textcolor{capShareGPT}{\textbf{ShareGPT4V-7B}}, \textcolor{capGemini}{\textbf{Gemini 2.5 Flash-Lite}} and \textcolor{capInternVL}{\textbf{InternVL3-8B}}. Additional examples are reported in Appendix~\ref{sec:caption-samples-and-alignment}.}
  \label{fig:caption_samples_1}
\end{figure}

\subsection{Computational cost of the dataset}
\label{sec:computational-cost-of-dataset}
Constructing MONET required ${\sim}175$k GPU-hours on a cluster of 60 L40S and 80 H200 GPUs, dominated by re-captioning (${\sim}79\%$), followed by domain-based filtering (${\sim}14\%$), and deduplication, synthetic generation, and feature / VAE pre-encoding (${\sim}2$--$3\%$ each). The end-to-end pipeline took several months of wall-clock time. By releasing MONET together with its multi-VLM captions, embeddings, annotations, and pre-computed VAE latents, we aim to substantially lower the barrier to reproducible text-to-image research at scale.
\section{Dataset analysis}
\label{sec:dataset-analysis}

\begin{figure}[t]
  \centering
  \setlength{\tabcolsep}{0pt}
  \begin{subfigure}[c]{0.4\textwidth}
    \centering\vspace{-1.5em}
    \includegraphics[width=\linewidth]{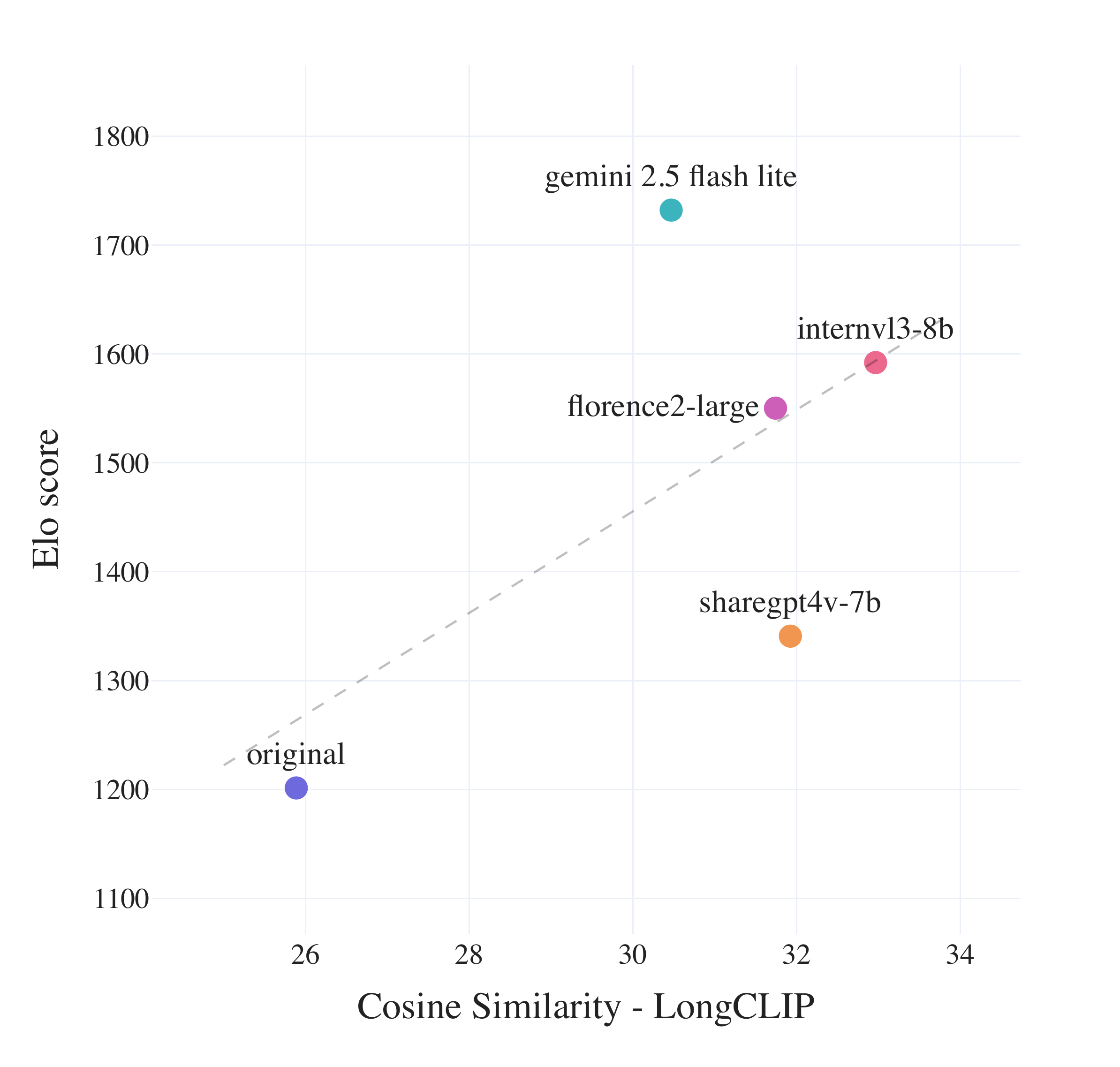}
    \caption{Text--image alignment.}
    \label{fig:text_image_alignment}
  \end{subfigure}%
  \hfill
  \begin{subfigure}[c]{0.58\textwidth}
    \centering
    \begin{subfigure}[t]{0.49\textwidth}
      \centering
      \includegraphics[width=\linewidth]{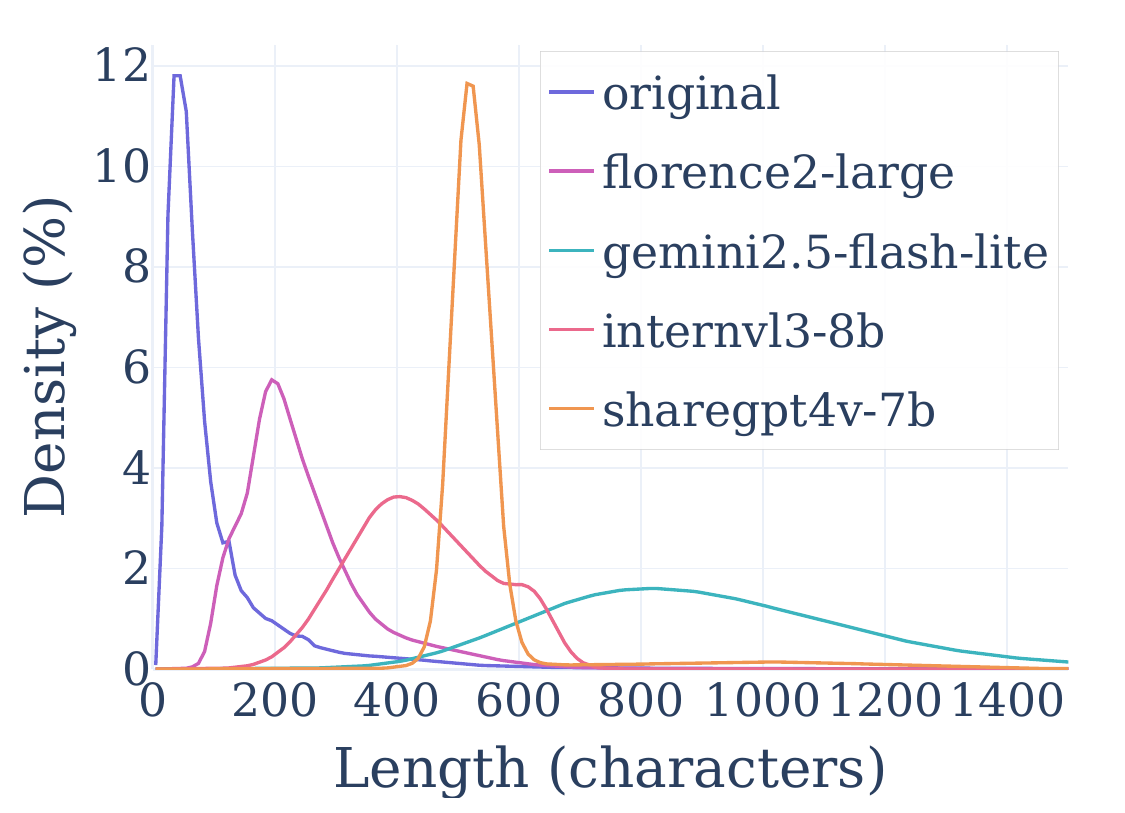}
      \caption{Caption length.}
      \label{fig:caption_len}
    \end{subfigure}%
    \begin{subfigure}[t]{0.49\textwidth}
      \centering
      \includegraphics[width=\linewidth]{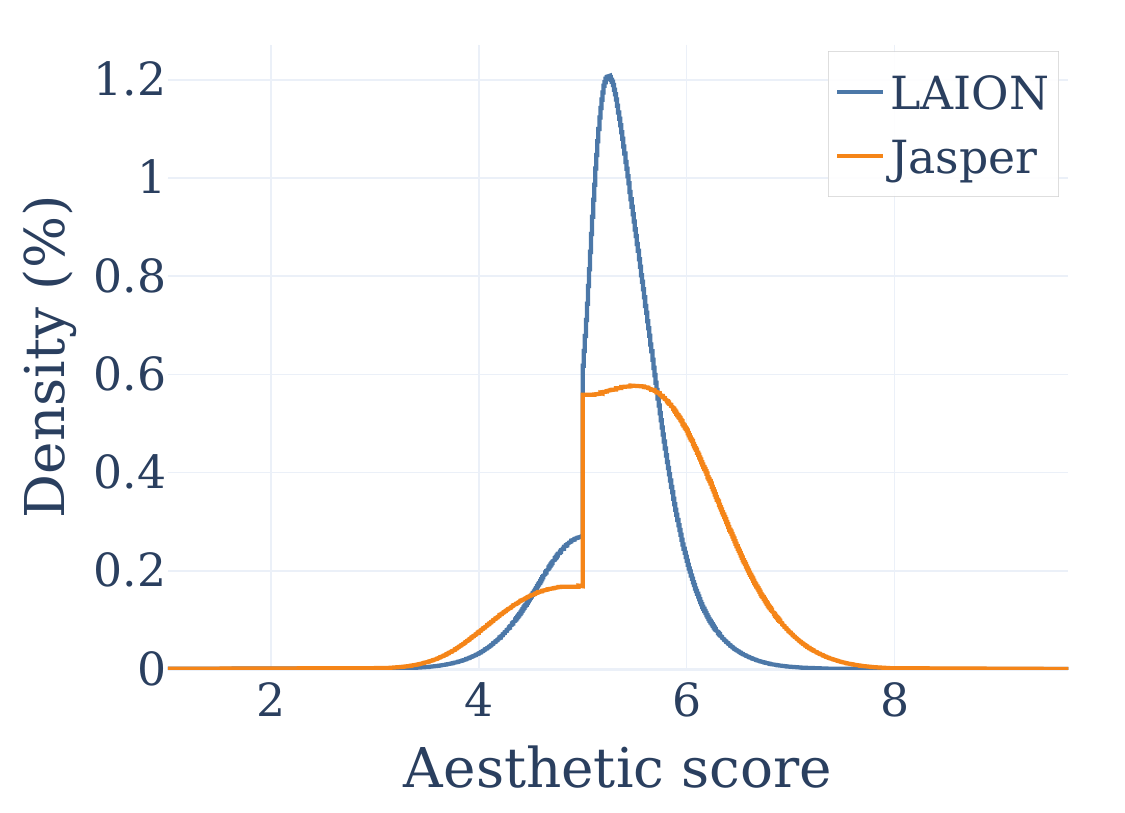}
      \caption{Aesthetic score.}
      \label{fig:aesthetic_score}
    \end{subfigure}

    \begin{subfigure}[t]{0.49\textwidth}
      \centering
      \includegraphics[width=\linewidth]{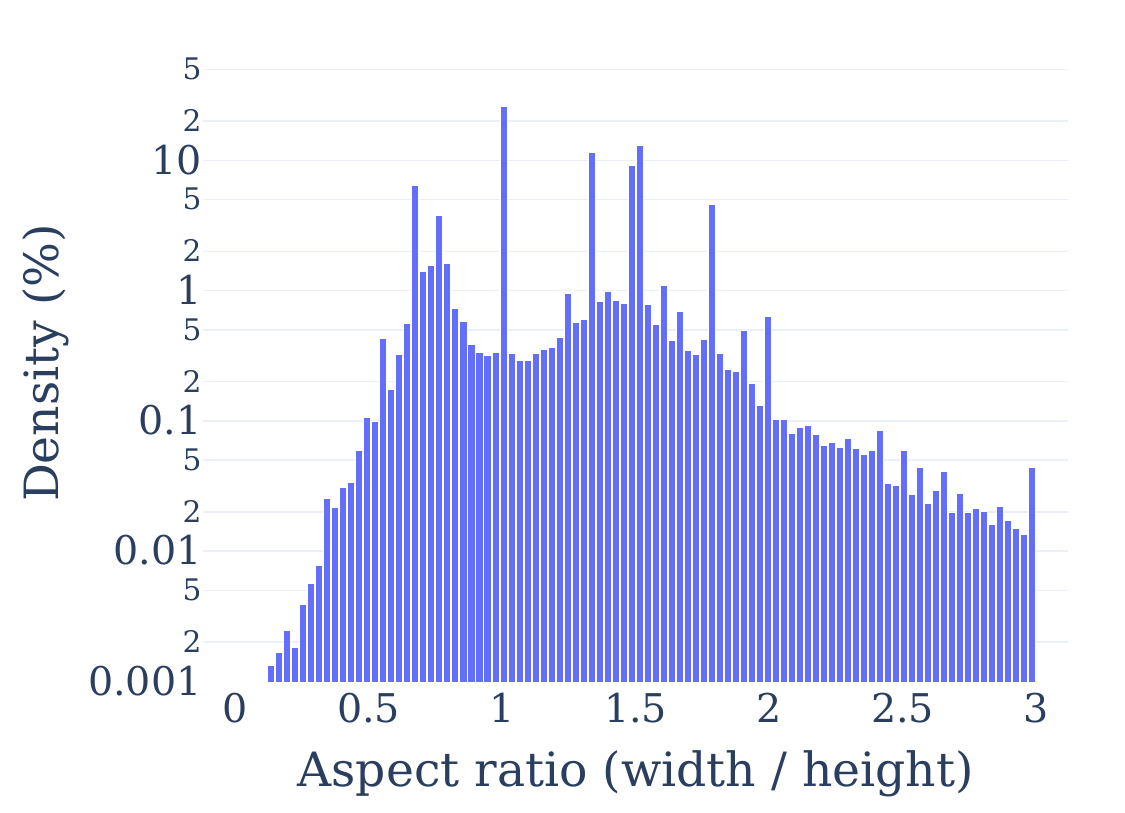}
      \caption{Image aspect ratio.}
      \label{fig:img_aspect_ratio}
    \end{subfigure}%
    \begin{subfigure}[t]{0.49\textwidth}
      \centering
      \includegraphics[width=\linewidth]{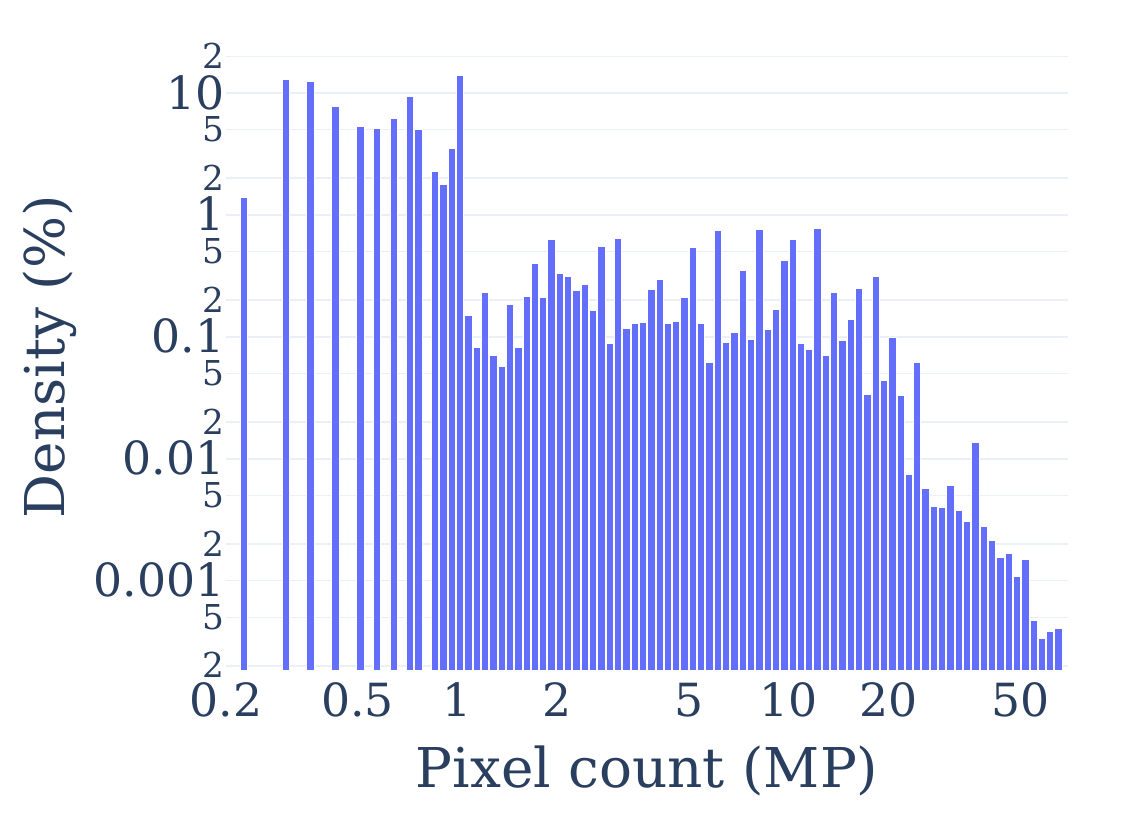}
      \caption{Image resolution.}
      \label{fig:img_res}
    \end{subfigure}
  \end{subfigure}
  \caption{Captions and image statistics of MONET. Image resolution is in Megapixels (MP).}
\end{figure}

\paragraph{Caption \& image statistics} Fig.~\ref{fig:caption_len} shows the caption length distributions for the four retained captioners and the original captions. All generated captions are substantially longer than the originals, with Gemini 2.5 Flash-Lite producing the most verbose captions, followed by ShareGPT4V-7B and InternVL3-8B, while Florence2-Large produces compact captions (see Appendix~\ref{sec:caption-samples-and-alignment}). Figs.~\ref{fig:aesthetic_score},~\ref{fig:img_aspect_ratio} and~\ref{fig:img_res} report the distributions of aesthetic score (LAION scores and scores from our internal classifier), aspect ratio and image resolution. Both aesthetic score distributions are centered in a similar interval, but our internal classifier exhibits greater spread, while LAION's is more concentrated. Notably, both distributions show a sharp jump discontinuity at a score of 5, a result of the aesthetic pre-filtering of Sec.~\ref{sec:prefiltering}. Aspect ratios are mostly within $[0.5, 3.0]$, with clear peaks at common formats such as 1:1, 3:2, 2:1, and 3:4. Finally, most images are below 20\,MP, although the distribution exhibits a long tail reaching up to 66\,MP.

\begin{figure}[t]
\captionsetup{skip=0pt}
  \centering
  \begin{subfigure}[t]{0.32\textwidth}
    \centering
    \includegraphics[width=\textwidth,trim=10 200 0 150,clip]{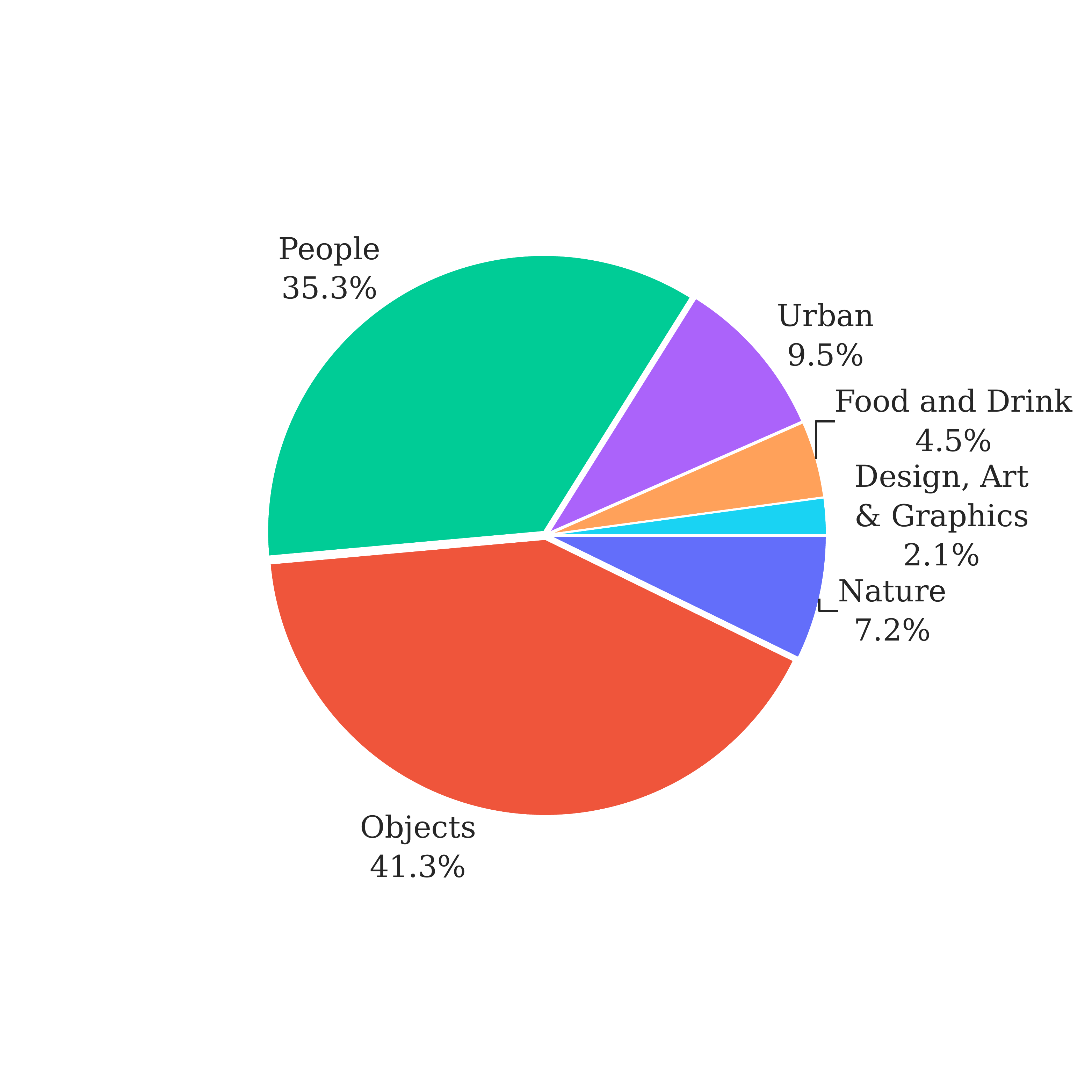}
  \end{subfigure}%
  \begin{subfigure}[t]{0.32\textwidth}
    \centering
    \includegraphics[width=\textwidth,trim=10 200 0 150,clip]{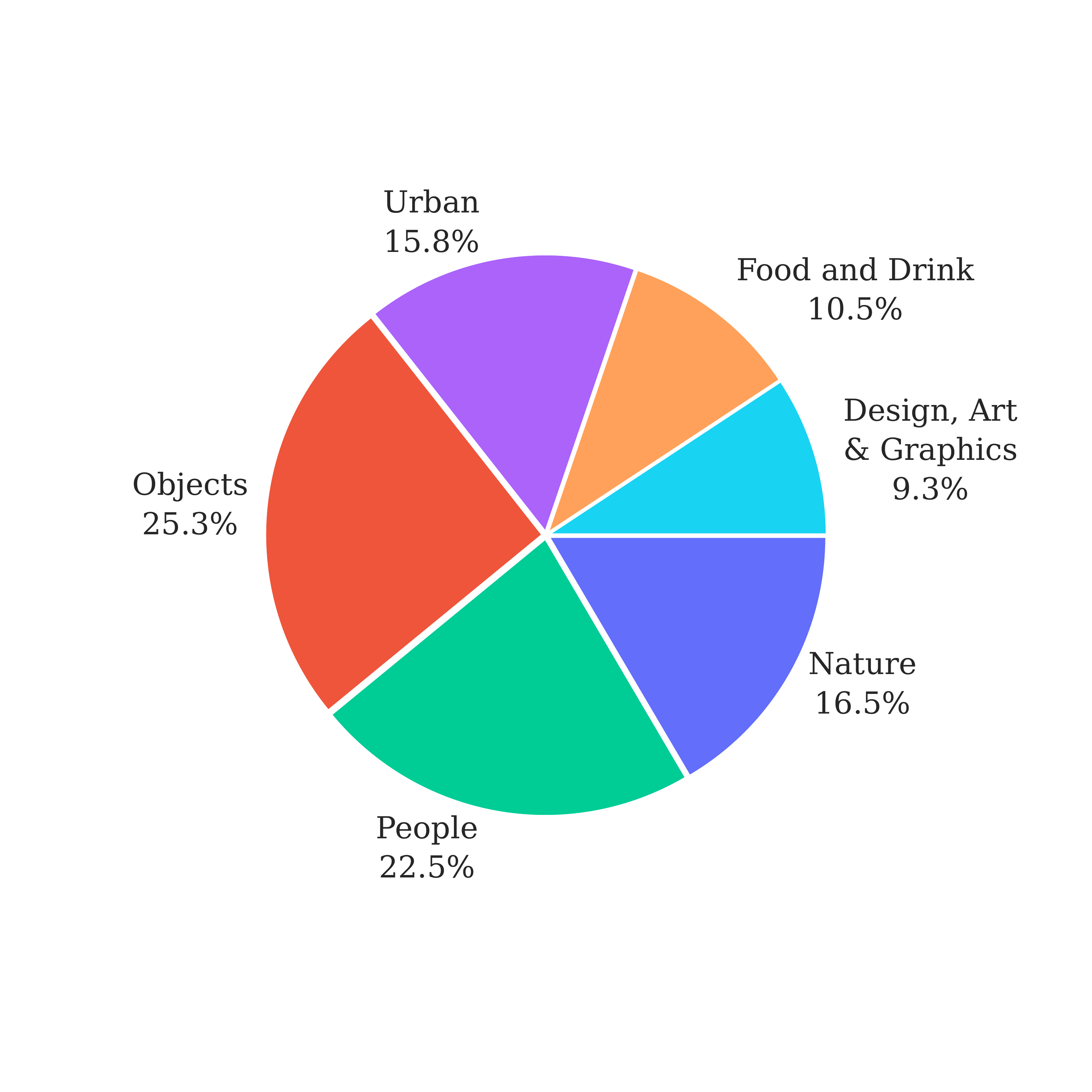}
  \end{subfigure}%
  \begin{subfigure}[t]{0.32\textwidth}
    \centering
    \includegraphics[width=\textwidth,trim=10 200 0 150,clip]{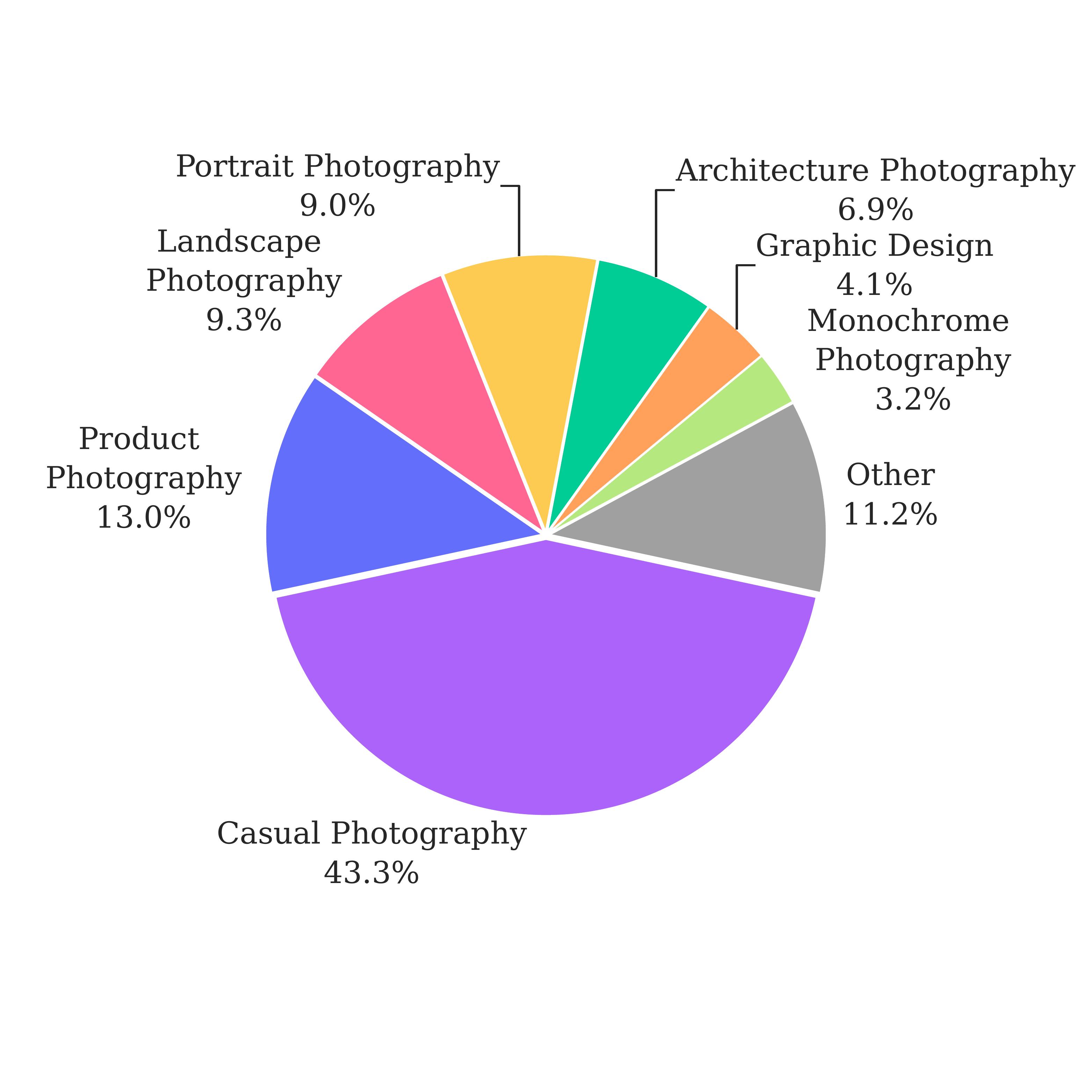}
  \end{subfigure}%
  \caption{MONET dataset distribution: (left) YOLO-based content classification, (middle) CLIP-based content classification, (right) Qwen3-VL-8B-Instruct based image style.}
  \label{fig:image-content-and-style}
\end{figure}

\paragraph{Content distribution}
We explore two approaches: (i)~top-5 YOLO object detections with COCO labels and (ii)~CLIP-based zero-shot classification. Both YOLO outputs and CLIP embeddings are pre-computed in Sec.~\ref{sec:dataset-construction} and stored in the dataset metadata. While YOLO provides structured detections, its expressiveness is limited by the COCO label space (80 object categories). For CLIP, we define ${\sim}2.7$k classes and encode them with the prompt ``a photo of a \{class\}'', where \{class\} denotes the class name. Image--class similarities are computed via cosine similarity between image and text embeddings, and the top-5 classes are retained. Both YOLO and CLIP base classes are then grouped into two hierarchical meta-levels following \citet{wu2025qwen} (see Appendix~\ref{sec:image-content-style-detailed-distribution}). Fig.~\ref{fig:image-content-and-style}~(left) shows the dataset content distribution based on YOLO detections across the highest-level classes. The distribution is dominated by \emph{objects} (41.3\%) and \emph{people} (35.3\%), with smaller shares for \emph{food and drink} (4.5\%) and \emph{design, art \& graphics} (2.1\%), reflecting the limited coverage of COCO labels (\emph{e.g.,}\ 10 food-related classes and a single relevant design class, ``book''). Fig.~\ref{fig:image-content-and-style}~(middle) reports the CLIP-based distribution, which is more balanced: 25.3\% \emph{objects}, 22.5\% \emph{people}, 16.5\% \emph{nature}, 15.8\% \emph{urban}, 10.5\% \emph{food and drink} and 9.3\% \emph{design, art \& graphics}. This improved coverage stems from the broader set of CLIP base classes, so we consider the CLIP-based estimates to be more representative of the dataset's content. Overall, MONET's distribution is consistent with comparable closed-source datasets: for example, Qwen-Image \citep{wu2025qwen} reports 12.9\% \emph{people}, 21.7\% \emph{objects}, 27.4\% \emph{design}, 7.0\% \emph{food}, 13.6\% \emph{urban} and 12.5\% \emph{nature} (Animals, Landscapes and Plants). A detailed breakdown is provided in Appendix~\ref{sec:image-content-style-detailed}.

\paragraph{Image style} We use Qwen3-VL-8B-Instruct \citep{yang2025qwen3,Qwen-VL} to classify a subset of the dataset, limited to 1.5M randomly sampled images for cost reasons, into 15 classes according to image style; the prompt and class definitions are provided in Appendix~\ref{sec:image-style-classification-prompt}. Fig.~\ref{fig:image-content-and-style}~(right) shows the resulting distribution. MONET spans a wide range of styles from graphic design and illustrations to portraits and product photography and is dominated by casual photography, a catch-all class for everyday photos. The full per-style distribution, including styles grouped under ``Other'', is reported in Fig.~\ref{fig:image-style-complete}.

\section{Downstream validation}
\label{sec:downstream-validation}

\begin{figure}
    \centering
    \includegraphics[width=0.8\linewidth]{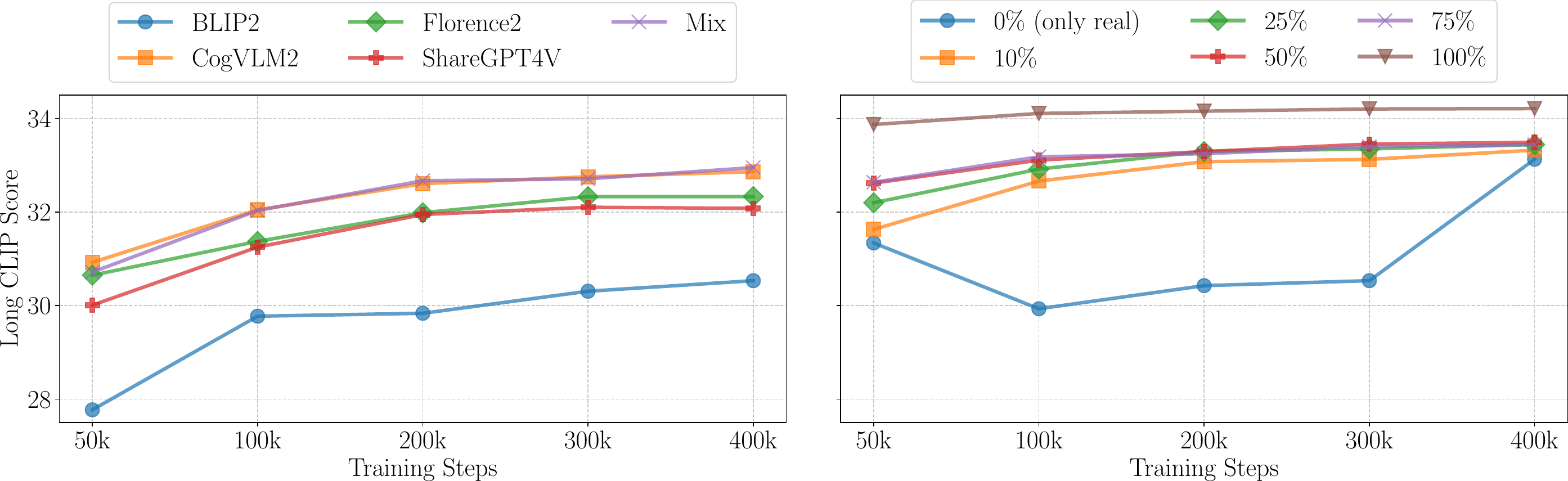}
    \caption{(Left) LongCLIP score evolution throughout training with different captioning models and (right) increasing amounts of synthetic data. }
    \label{fig:long-clip-caption-synthetic}
\end{figure}

\begin{table}[t]
    \centering
    \scriptsize
    \caption{Impact of captioners on FID (left), and impact of the synthetic data ratio on FID and DIV-LPIPS (right), on the ImageNet-512 validation set after 400k training iterations. FID is computed on 50k samples, while DIV-LPIPS is computed using 2k prompts with 10 samples per prompt.}
    \label{tab:fid-caption-synthetic}
    \begin{tabular}{lc|ccc}
        \toprule
        \multicolumn{2}{c|}{Captioner} &
        \multicolumn{3}{c}{Synthetic Data} \\
        Captioner & FID $\downarrow$ &
        Synthetic Data (\%) & FID $\downarrow$ & DIV-LPIPS $\uparrow$ \\
        \midrule
        BLIP2$^*$   & 7.5 & $0$   & 8.1  & 0.59 \\
        CogVLM2$^*$ & 7.3 & $10$  & 8.3  & 0.59 \\
        Florence2   & 7.3 & $25$  & 8.0  & 0.59 \\
        ShareGPT4V  & \textbf{5.2} & $50$  & 7.6 & \textbf{0.60} \\
        Mix         & 8.0 & $75$  & \textbf{7.2}  & 0.59 \\
        --          & --  & $100$ & 15.0 & 0.47 \\
        \bottomrule
    \end{tabular}
\end{table}

\subsection{Impact of multi-captioning}
To justify the decision to use multiple captioning models in MONET, we assess the impact of caption types on the performance of a T2I model. We re-caption the ImageNet dataset \citep{deng2009imagenet} with four captioners of different complexity: BLIP2 \citep{li2023blip}, CogVLM2\footnote{We used BLIP2 and CogVLM2 captioners even though they are not part of the final MONET dataset, as the benchmark of Sec.~\ref{sec:recaptioning} had not yet been performed at the time we launched this experiment.} , Florence2 \citep{xiao2024florence} and ShareGPT4V \citep{chen2024sharegpt4v}. We then train five T2I diffusion models, one per captioner and one with captions uniformly sampled from all four (\emph{Mix}); other training details are provided in Appendix~\ref{sec:training-details}. We report in Fig.~\ref{fig:long-clip-caption-synthetic} (left) the LongCLIP alignment score \citep{zhang2024long} and in Table~\ref{tab:fid-caption-synthetic} (left) the Fréchet Inception Distance (FID) \citep{heusel2017gans} computed on 50k samples from the ImageNet-512 validation set, where evaluation captions are uniformly sampled from all four captioners. Our findings align with prior work \citep{esser2024scaling,wu2025qwen}: multiple captioners improve robustness and generalization across prompts (higher Long CLIP Score), while verbose captioners like ShareGPT4V accelerate convergence (lower FID, less dependent on prompt differences) but hurt out-of-distribution performance when used alone, motivating MONET’s multi-captioner mix.

\subsection{Impact of synthetic data}
\label{sec:impact-of-synthetic-data}
We conduct a similar experiment to assess the impact of synthetic data, using the same diffusion-model architecture as in the previous experiment. We train six text-conditioned diffusion models on a mix of the re-captioned ImageNet training set and synthetic samples generated by FLUX.2-klein-4B~\citep{flux-2-2025}, with the proportion of synthetic data set to 0\% (real data only), 10\%, 25\%, 50\%, 75\% and 100\% (synthetic data only), and evaluate each model on the same validation set used in the previous section. As illustrated in Fig.~\ref{fig:long-clip-caption-synthetic} (right), we observe that adding synthetic data improves text--image alignment, justifying the inclusion of such samples in MONET. However, as expected, an excessive proportion of synthetic data can lead to overfitting or distribution shift, as evidenced by the markedly higher FID observed when training exclusively on synthetic data (Table~\ref{tab:fid-caption-synthetic}, right). Excessive reliance on synthetic data also reduces sample diversity, as reflected by the DIV-LPIPS \citep{ruiz2023dreambooth, voronov2023loss} measurements in Table~\ref{tab:fid-caption-synthetic} (right). Specifically, DIV-LPIPS decreases from 0.59 to 0.47 when using only synthetic data. Notably, incorporating just 25\% real data substantially improves sample diversity. In MONET, synthetic data account for approximately 13\% of the dataset, and all samples are released with source tags that enable users to easily distinguish between different data sources, specifically synthetic and real data. The impact of synthetic data on large-scale training is beyond the scope of this work.

\subsection{Text-to-image model training}
\label{sec:t2i-training}
Finally, we train a 4B-parameter text-to-image model on MONET. We rely on the latent diffusion framework \citep{rombach2022high} with a denoiser inspired by MMDiT \citep{esser2024scaling} and using a deep-compression VAE (DCVAE) \citep{chen2025deep}; text conditioning is injected using Qwen3-4B \citep{yang2025qwen3}. Table~\ref{tab:geneval-benchmark} reports the performance on the GenEval \citep{ghosh2023geneval} and DPG \citep{hu2024ella} benchmarks of our model trained \textbf{exclusively} on MONET. As reported, the model is competitive with many existing models trained on closed-source data, underlining the quality of MONET. All training details, together with qualitative samples generated at $1024^2$ and $2048^2$ resolution, are provided in Appendices~\ref{sec:training-details} and \ref{sec:app-additional-results}. These samples illustrate the diversity and quality of MONET, supporting training even beyond the standard $1024^2$ resolution.

\begin{table}
\caption{Results on the GenEval and DPG benchmarks. Our 4B model trained on the MONET dataset achieves competitive performance against models of similar size trained on closed-source data.}
\label{tab:geneval-benchmark}
\scriptsize
\setlength{\tabcolsep}{2.5pt}
\resizebox{\linewidth}{!}{%
    \begin{tabular}{l|c|ccccccc|cccccc}
        \toprule
        \multicolumn{1}{c}{}& \multicolumn{1}{c}{}&\multicolumn{7}{c}{GenEval}&\multicolumn{6}{c}{DPG}\\
        \multicolumn{1}{c}{Model} & \multicolumn{1}{c}{Num.}& \multicolumn{2}{c}{Objects}& \multicolumn{5}{c}{} \\
                 \multicolumn{1}{c}{} &\multicolumn{1}{c}{Params} (B)& Single & Two & Counting & Colors & Position & Color & \multicolumn{1}{c}{Overall $\uparrow$} &Global & Entity & Attribute & Relation & Other & Overall$\uparrow$\\
        \midrule
        SD1.5 \citep{rombach2022high}    & 0.9   & 0.97 & 0.38 & 0.35 & 0.76 & 0.04 & 0.06 & 0.43 &74.63 & 74.23 & 75.39 & 73.49 & 67.81 & 63.18 \\
        PixArt-$\alpha$  \citep{chen2023pixart}& 0.6& 0.98 & 0.50 & 0.44 & 0.80 & 0.08 & 0.07 & 0.48 & 74.97 & 79.32 & 78.60 & 82.57 & 76.96 & 71.11\\
        Emu3-Gen \citep{wang2024emu3}& 8.0 & 0.98 & 0.71 & 0.34 & 0.81 & 0.17 & 0.21 & 0.54 & 85.21 & 86.68 & 86.84 & 90.22 & 83.15 & 80.60\\
        SDXL \citep{podell2023sdxl}   &   2.6     & 0.98 & 0.74 & 0.39 & 0.85 & 0.15 & 0.23 & 0.55 & 83.27 & 82.43 & 80.91 & 86.76 & 80.41 & 74.65 \\
        SD3 Medium \citep{esser2024scaling}& 2.0 & 0.98 & 0.74 & 0.63 & 0.67 & 0.34 & 0.36 & 0.62 & 87.90 & 91.01 & 88.83 & 80.70 & 88.68 & 84.08 \\
        FLUX.1 [Dev] \citep{flux2024} &12.0 & 0.98 & 0.81 & 0.74 & 0.79 & 0.22 & 0.45 & 0.66 & 74.35 & 90.00 & 88.96 & 90.87 & 88.33 & 83.84 \\
        DALL-E 3 \citep{betker2023improving} & -- & 0.96 & 0.87 & 0.47 & 0.83 & 0.43 & 0.45 & 0.67 & 90.97 & 89.61 & 88.39 & 90.58 & 89.83 & 83.50 \\
        SANA-1.5 \citep{xie2025sana}& 4.8 & 0.99 & 0.85 & 0.77 & 0.87 & 0.34 & 0.54 & 0.72 & -- & -- & -- & -- & -- & 85.00\\
        Lumina-Image 2.0 \citep{qin2025lumina}& 2.6 & -- & 0.87 & 0.67 & -- & -- & 0.62 & 0.73 & --    & \textbf{91.97} & 90.20 & \textbf{94.85} & --    & 87.20\\
        Janus-Pro-7B \citep{chen2025janus}& 7.0 & 0.99 & 0.89 & 0.59 & 0.90 & \textbf{0.79} & 0.66 & 0.80 &86.90 & 88.90 & 89.40 & 89.32 & 89.48 & 84.19\\
        HiDream-I1-Full \citep{cai2025hidream}& 17.0 & \textbf{1.00} & \textbf{0.98} &0.79 &0.91 & 0.60 & 0.72 & 0.83 & 76.44 & 90.22 & 89.48 & 93.74 & 91.83 & 85.89 \\
        Z-Image \citep{zimage2025}& 6.0 & \textbf{1.00} & 0.94& 0.78& \textbf{0.93} &0.62& \textbf{0.77} &0.84 & \textbf{93.39} &91.22& \textbf{93.16} &92.22& 91.52 &88.14\\
        Qwen-Image \citep{wu2025qwen}& 20.0 & 0.99 &0.92 & \textbf{0.89} &0.88 &0.76 &\textbf{0.77} &\textbf{0.87} & 91.32 & 91.56 & 92.02 & 94.31 & \textbf{92.73} & \textbf{88.32}\\
        \midrule
        Ours & 4.1 & \textbf{1.00} & 0.90 & 0.73 & 0.88 & 0.35 & 0.62 & 0.74 & 84.80 & 91.76 & 89.70 & 94.16 & 79.60 &85.56 \\
        \bottomrule
        \end{tabular}%
        }
\end{table}

\subsection{Impact of data curation}

To quantify the impact of the data curation performed by MONET, we compare training on the original uncurated data pool with training on MONET. We consider two different data distributions: 1) 103.8M images randomly sampled from the uncurated original pool of data (2.9B data source pool) and 2) the MONET dataset. We then pre-train two models with the same architecture (4B model of Sec.~\ref{sec:t2i-training}) on each split, following the training details described in Appendix~\ref{sec:training-details}. Fig.~\ref{fig:impact-data-curation} shows the evolution of GenEval, CLIPScore, and Aesthetic Score throughout training, where it can be seen that the processing pipeline of MONET significantly improves the performance of the model on all three metrics. We emphasize that, beyond quality, uncurated data can introduce safety risks due to harmful or copyrighted content in the source data. MONET’s filtering stages aim to mitigate these risks.

\begin{figure}
    \centering
    \includegraphics[width=\linewidth]{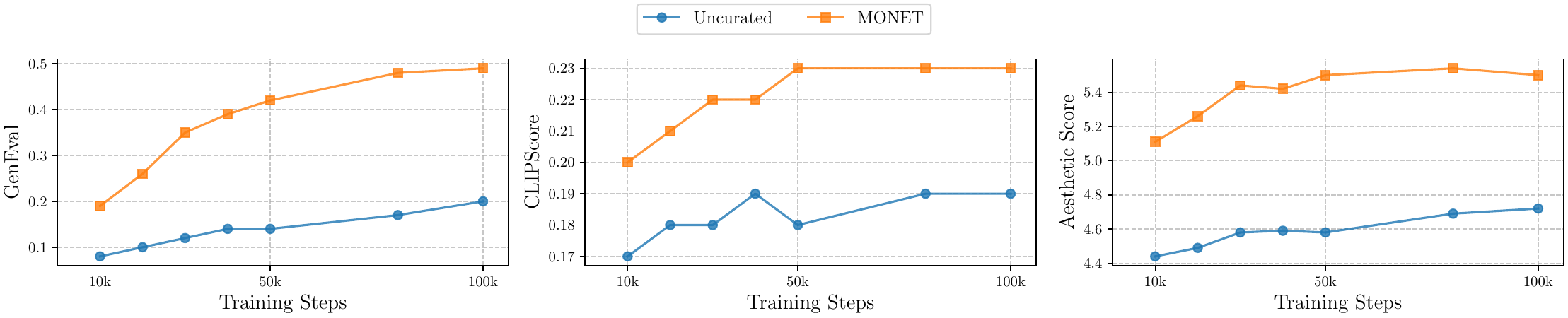}
    \caption{Comparison of training on the uncurated dataset and MONET in terms of (left) GenEval, (middle) CLIPScore, and (right) Aesthetic Score.}
    \label{fig:impact-data-curation}
\end{figure}

\section{Ethics \& responsible use}
\label{sec:ethics-licensing}

Releasing a large-scale image--text dataset carries responsibilities regarding representation, safety and downstream impact. MONET aggregates web-sourced data we do not own; we therefore focus our ethical commitments on careful curation, transparent documentation, and the release of audit statistics to the community.

\textbf{Representation audit.} We audit a random sample of $\sim$5M images using Qwen3-VL-8B-Instruct with a structured prompt that elicits concrete visual evidence before committing to a categorical label, and defaults to \emph{unknown} when evidence is insufficient (see full methodology, prompt and aggregate distributions in Appendix~\ref{sec:ethics-audit-prompt}). We focus on four demographic dimensions: \emph{cultural origin}, \emph{skin tone} (Fitzpatrick 1--6~\citep{fitzpatrick1988validity}), \emph{predominant gender} and \emph{predominant age group}. Cultural origin is dominated by European and North American contexts, consistent with documented Western biases of web-scraped corpora~\citep{schuhmann2022laion}. Skin tones concentrate around categories 3--4, with lighter (1--2) and darker (5--6) tones under-represented; gender is roughly balanced, while age skews strongly toward adults, with children, teenagers and elderly subjects less frequent. These biases are largely inherited from the upstream sources, and the released annotations should help users re-weight the dataset toward a more balanced training distribution.

\textbf{Responsible use.} Despite our curation efforts, residual risks remain. The demographic biases above may propagate to models trained on MONET; practitioners should monitor outputs for fairness and apply mitigations such as balanced sampling. Safety filters do not achieve perfect recall, so downstream deployments should add output-level safety classifiers. We encourage users to follow ethical AI guidelines and consider the societal impact of derived models.
\section{Limitations and future work}
\label{sec:limitations}
MONET inherits biases from its Common-Crawl-based sources, over-representing European and North American contexts. Due to high compute requirements (${\sim}175$k GPU-hours for the full pool), image-style and ethics-audit annotations are restricted to representative subsets and rely on a single VLM (Qwen3-VL-8B-Instruct). Extending the ethics annotations described in Sec.~\ref{sec:ethics-licensing} to the full dataset would enable downstream re-weighting and the construction of a balanced dataset; scaling the process and cross-checking with other VLMs or human review are natural next steps for future work. MONET is English-only and re-captioning targets short, medium and long descriptions without structured attributes (counts, colors, spatial relations); multilingual captions and attribute-aware prompts are natural extensions. Synthetic content may also reflect hallucinations and stylistic biases of the underlying models, only partially mitigated by our multi-model mix. Moreover, our intentionally conservative \textit{NSFW} and \textit{watermark} filtering strategy may come at the cost of discarding \textit{safe} and \textit{compliant} images. Finally, our validation focuses on a 4B-parameter T2I model trained at up to $1024^2$ resolution; scaling to larger models, higher resolutions and human preference studies is left to future work. 
\section{Conclusion}
\label{sec:conclusion}

We introduced \textbf{MONET}, an open \emph{Apache~2.0} dataset of \textbf{103.8M} curated image--text pairs built from heterogeneous open sources through successive stages of filtering, two-stage deduplication, multi-VLM re-captioning and synthetic-data augmentation, and shipped with pre-computed embeddings, annotations and latents to accelerate downstream use and deeper analysis. To the best of our knowledge, MONET is the first open, meticulously filtered, deduplicated and multi-captioned dataset for training T2I models at scale. We validated our design choices by training a 4B model that reaches competitive GenEval and DPG scores. MONET is also designed as a foundational \emph{pre-training} dataset, intended to be paired with high-quality fine-tuning subsets for task-specific applications. By releasing MONET, we aim to lower the barrier to reproducible, large-scale text-to-image research.

\begin{ack}
This work received no third-party funding or support during the 36 months preceding submission. The authors declare no competing financial interests relevant to this work during the 36 months preceding submission.
\end{ack}

\small
\bibliographystyle{plainnat}
\bibliography{references}

\newpage
\appendix
\section{Technical appendices and supplementary material}
\label{sec:appendices}

\etocsetnexttocdepth{subsection}
\localtableofcontents
\vspace{1em}
\subsection{Filtering details}
\label{sec:filtering-details}

In this section, we detail the domain-based (URL and watermark) and NSFW filters used. These act as exclusion controls and source-governance signals, not as a representation of legal clearance. 

\subsubsection{URL filtering}
URL filtering removes any image whose URL contains the name of a known stock photo provider (Dreamstime, Shutterstock, Freepik, Getty, Unsplash, Pexels, etc.). Table~\ref{tab:url-filtering-stats} reports the number of removed images and their proportion of the overall dataset, broken down by source and domain. We observed that some filtered images clearly display watermarks (such as those from Dreamstime, Shutterstock, and Getty), which validates the filtering approach. However, most images from Unsplash and Pexels, as well as some from Getty, do not include watermarks, which would make filtering at a later stage more difficult.

\begin{table}[ht]
\centering
\scriptsize
\caption{Number and proportion of removed images by URL-based filtering across stock image domains. Percentages are relative to the 121.1M merged pool.}
\begin{tabular}{l l r r}
\toprule
Source & Domain & Removed images & Percentage \\
\midrule
Dreamstime           & dreamstime.com         & 1,448,676 & 1.20\% \\
Shutterstock         & image.shutterstock.com & 225,389   & 0.19\% \\
Freepik              & freepik.com            & 176,133   & 0.15\% \\
Getty                & media.gettyimages.com  & 174,541   & 0.14\% \\
Unsplash             & unsplash.com           & 169,621   & 0.14\% \\
Pexels               & pexels.com             & 100,628   & 0.08\% \\
Envato               & envato.com             & 34,618    & 0.03\% \\
Adobe Stock / Fotolia & stock.adobe.com       & 27,120    & 0.02\% \\
Vecteezy             & vecteezy.com           & 16,092    & 0.01\% \\
123RF                & 123rf.com              & 14,135    & 0.01\% \\
Depositphotos        & depositphotos.com      & 8,990     & 0.01\% \\
Pixabay              & pixabay.com            & 8,554     & 0.01\% \\
Reuters Pictures     & reuters.com            & 6,428     & 0.01\% \\
AFP                  & afp.com                & 4,956     & 0.00\% \\
Creative Market      & creativemarket.com     & 1,727     & 0.00\% \\
iStock               & istockphoto.com        & 926       & 0.00\% \\
Alamy                & alamy.com              & 892       & 0.00\% \\
Pond5                & pond5.com              & 355       & 0.00\% \\
Bigstock             & bigstockphoto.com      & 113       & 0.00\% \\
Storyblocks          & storyblocks.com        & 77        & 0.00\% \\
AP Images            & apimages.com           & 28        & 0.00\% \\
Zuma / Newscom / Rex & zumapress.com          & 1         & 0.00\% \\
\midrule
\textbf{Total}       &                        & \textbf{2,420,000} & \textbf{2.00\%} \\
\bottomrule
\end{tabular}
\label{tab:url-filtering-stats}
\end{table}

\subsubsection{Watermark filtering}

Watermark filtering is achieved by running an internal watermark detector through the entire dataset, which produces a continuous score ranging from 0 to 1, where 1 indicates the highest likelihood of a watermark. Images are then filtered based on a manually tuned threshold; in this case, the cut-off is set to 0.34, removing 2.70M images.

We complement it with a publicly available watermark detector, the LAION watermark classifier \citep{laion_watermark_detector}. Its score is released for every image as \texttt{wk\_laion}, alongside the internal score \texttt{wk\_jasperai}. Following the conservative union rule used for NSFW filtering, an image is removed if either detector flags it; the LAION detector removes ${\sim}0.66$M additional images, bringing watermark-based removals to 3.36M.

\begin{figure}[ht]
    \centering

    \begin{subfigure}{0.9\linewidth}
        \centering
        \includegraphics[width=\linewidth]{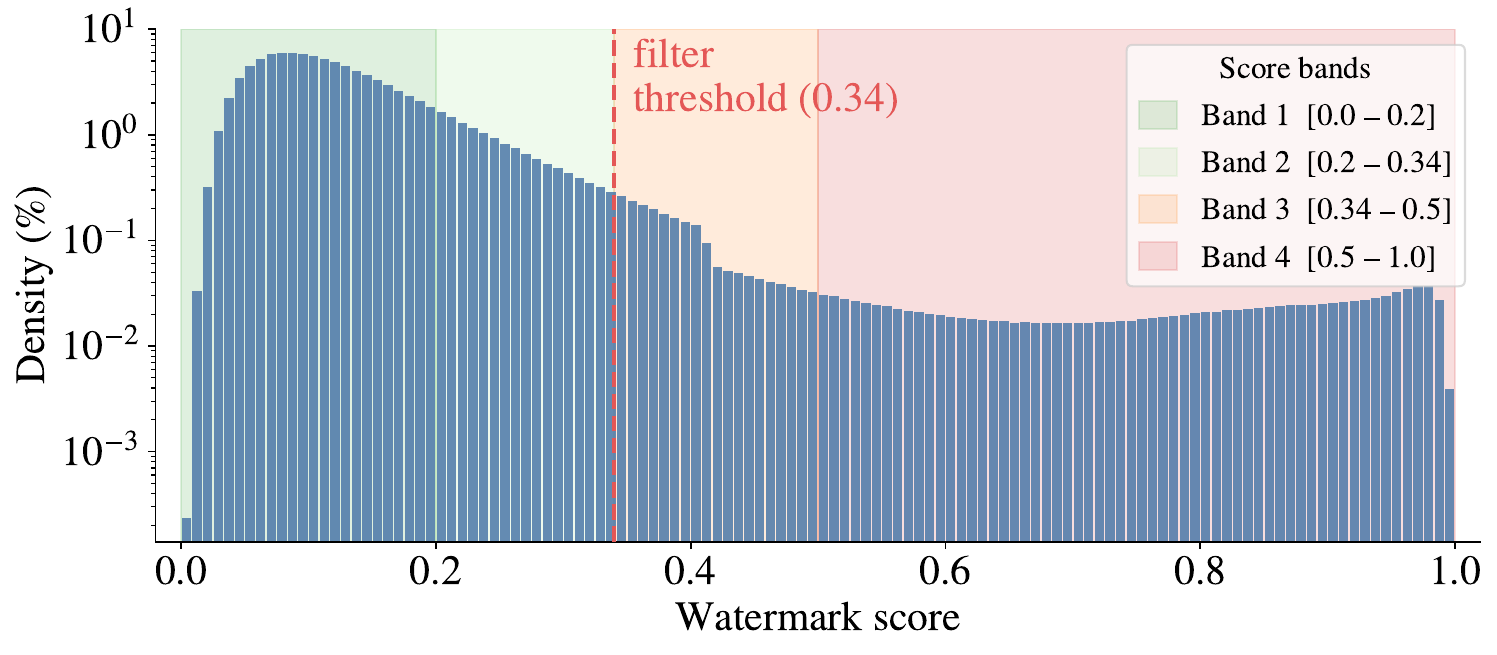}
        \caption{Histogram in logarithmic scale, with filtering threshold.}
        \label{fig:watermark-histogram}
    \end{subfigure}

    \vspace{0.5cm}

    \begin{subfigure}{0.9\linewidth}
        \centering

        \begin{subfigure}{0.22\linewidth}
            \centering
            \includegraphics[width=\linewidth,height=4cm]{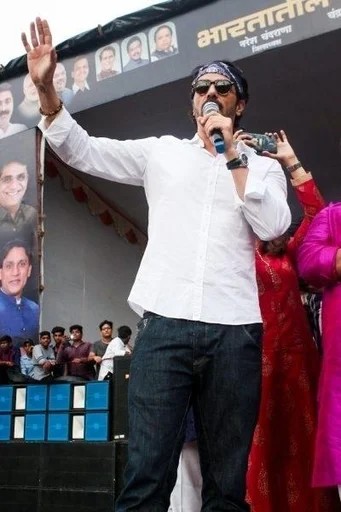}
            \caption{Band 1 (retained)}
            \label{fig:watermark-examples-band1}
        \end{subfigure}
        \hfill
        \begin{subfigure}{0.22\linewidth}
            \centering
            \includegraphics[width=\linewidth,height=4cm]{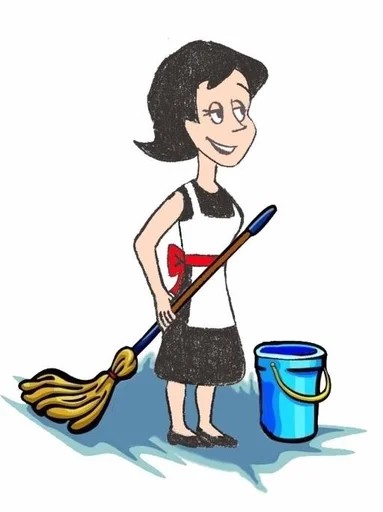}
            \caption{Band 2 (retained)}
            \label{fig:watermark-examples-band2}
        \end{subfigure}
        \hfill
        \begin{subfigure}{0.22\linewidth}
            \centering
            \includegraphics[width=\linewidth,height=4cm]{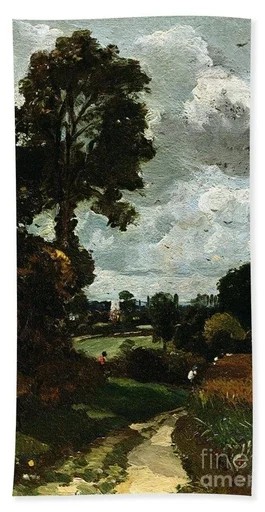}
            \caption{Band 3 (filtered)}
            \label{fig:watermark-examples-band3}
        \end{subfigure}
        \hfill
        \begin{subfigure}{0.22\linewidth}
            \centering
            \includegraphics[width=\linewidth,height=4cm]{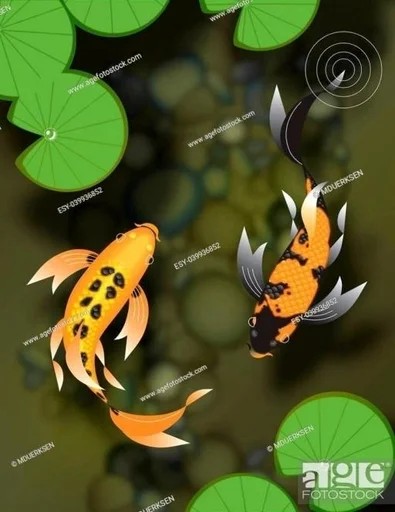}
            \caption{Band 4 (filtered)}
            \label{fig:watermark-examples-band4}
        \end{subfigure}

    \end{subfigure}

    \caption{Distribution and examples of watermark scores.}
    \label{fig:watermark-distribution-and-examples}
\end{figure}

Fig.~\ref{fig:watermark-histogram} shows the histogram of the watermark filter scores in logarithmic scale, together with the filter threshold. The scores were divided into four score bands or intervals, two of which consist of retained images and two of filtered images, and one example per score band is showcased (Figs.~\ref{fig:watermark-examples-band1}, ~\ref{fig:watermark-examples-band2}, ~\ref{fig:watermark-examples-band3}, ~\ref{fig:watermark-examples-band4}). Images in the two retained bands show no visible watermarks. We observe that band 3 typically contains more subtle watermarks, such as a single watermark in the bottom-right corner (Fig.~\ref{fig:watermark-examples-band3}), while band 4 includes images with multiple watermarks (see Fig.~\ref{fig:watermark-examples-band4}).

\subsubsection{NSFW filtering}

Three distinct methods are used to filter NSFW content: an internal (Jasper) NSFW detector and two publicly available detectors, Bumble and Falcon. Both Jasper and Bumble produce a continuous NSFW score for each image, and images with scores exceeding a specified threshold are discarded. The threshold for Jasper is manually tuned, whereas for Bumble the threshold recommended by the authors is adopted \cite{bumble_nsfw_classifier}. Falcon, in contrast, outputs a binary NSFW label, where 0 denotes safe content and 1 denotes NSFW content.

\begin{figure}[ht]
    \centering

    \begin{subfigure}{\linewidth}
        \centering
        \includegraphics[width=\linewidth]{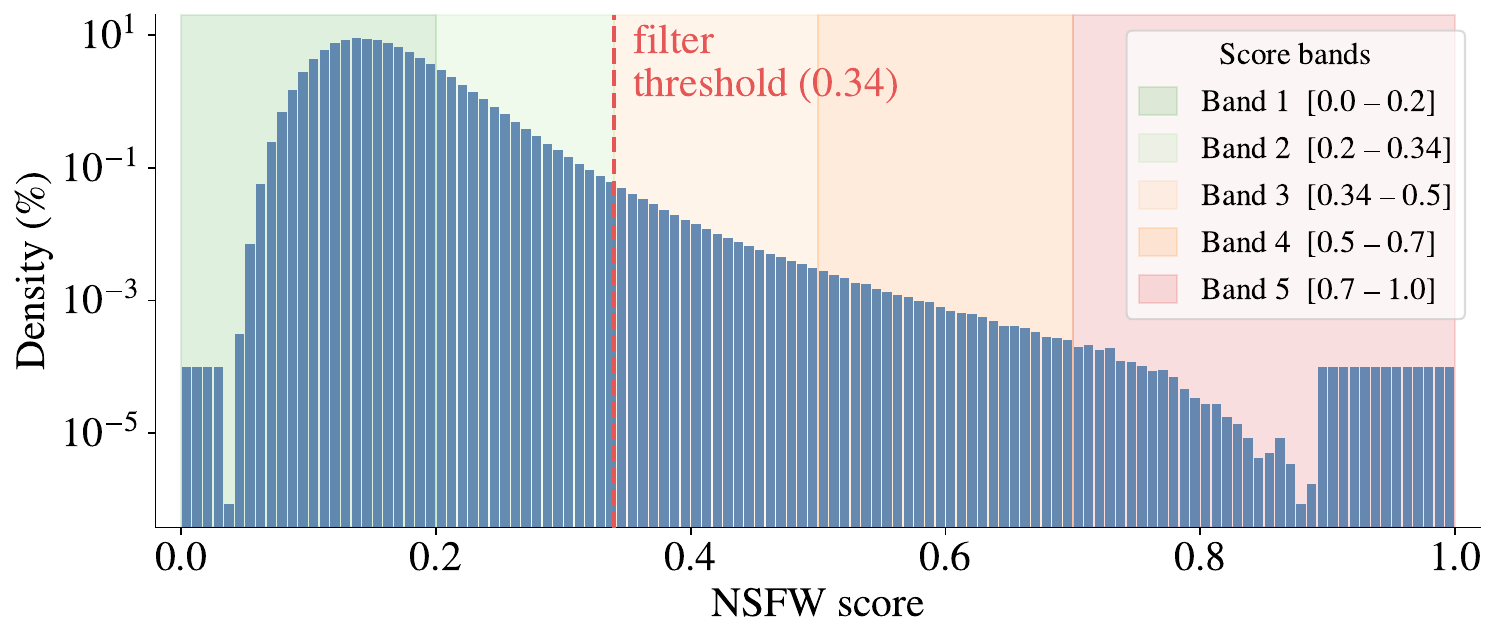}
        \caption{Histogram in logarithmic scale, with filtering threshold.}
        \label{fig:nsfw-jasper-histogram}
    \end{subfigure}

    \vspace{0.5cm}

    \begin{subfigure}{\linewidth}
        \centering

        \begin{subfigure}{0.20\linewidth}
            \centering
            \includegraphics[width=\linewidth,height=4cm]{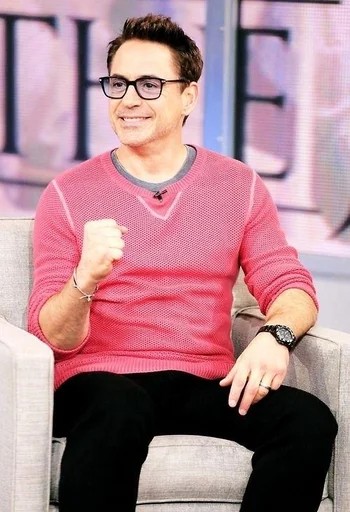}
            \caption{Band 1 (retained)}
            \label{fig:nsfw-jasper-band1}
        \end{subfigure}
        \hfill
        \begin{subfigure}{0.20\linewidth}
            \centering
            \includegraphics[width=\linewidth,height=4cm]{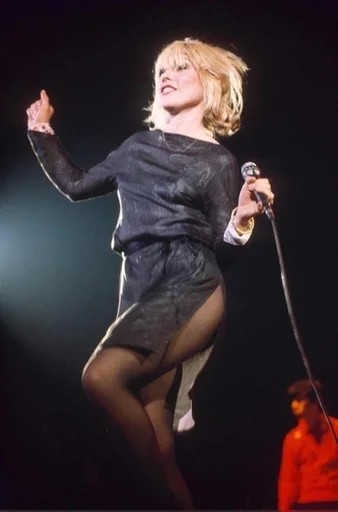}
            \caption{Band 2 (retained)}
            \label{fig:nsfw-jasper-band2}
        \end{subfigure}
        \hfill
        \begin{subfigure}{0.20\linewidth}
            \centering
            \includegraphics[width=\linewidth,height=4cm]{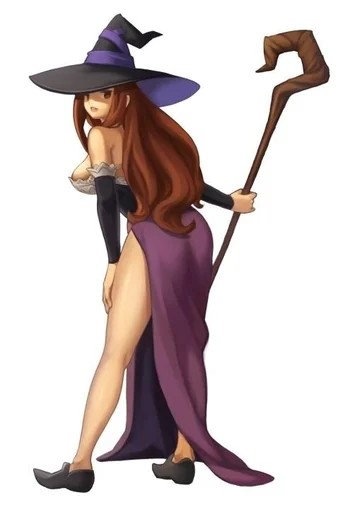}
            \caption{Band 3 (filtered)}
            \label{fig:nsfw-jasper-band3}
        \end{subfigure}
        \hfill
        \begin{subfigure}{0.20\linewidth}
            \centering
            \includegraphics[width=\linewidth,height=4cm]{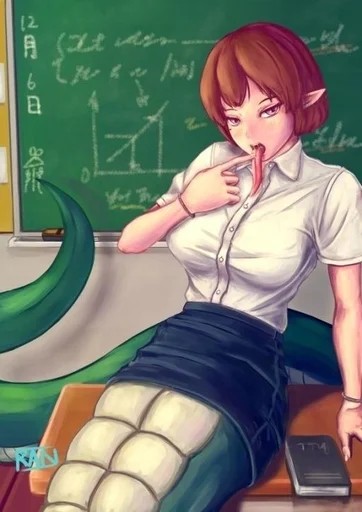}
            \caption{Band 4 (filtered)}
            \label{fig:nsfw-jasper-band4}
        \end{subfigure}

    \end{subfigure}

    \caption{Distribution and examples of Jasper NSFW score. No band 5 examples are included.}
    \label{fig:nsfw-jasper-distribution-and-examples}
\end{figure}

Fig.~\ref{fig:nsfw-jasper-distribution-and-examples} shows the histogram of the Jasper NSFW scores with some examples. The histogram is divided into 5 score bands, the first two consisting of retained images, while the latter three consist of rejected images. Examples of the first four bands are shown in Figs.~\ref{fig:nsfw-jasper-band1}–\ref{fig:nsfw-jasper-band4}, while no example from the fifth band is included due to its highly explicit NSFW content. As illustrated, the content transitions smoothly from clearly safe material to progressively more NSFW content. We note that the cut-off threshold is intentionally conservative, and that images filtered near this boundary, i.e., band 3 images, are only mildly unsafe.

\begin{figure}[ht]
    \centering

    \begin{subfigure}{\linewidth}
        \centering
        \includegraphics[width=\linewidth]{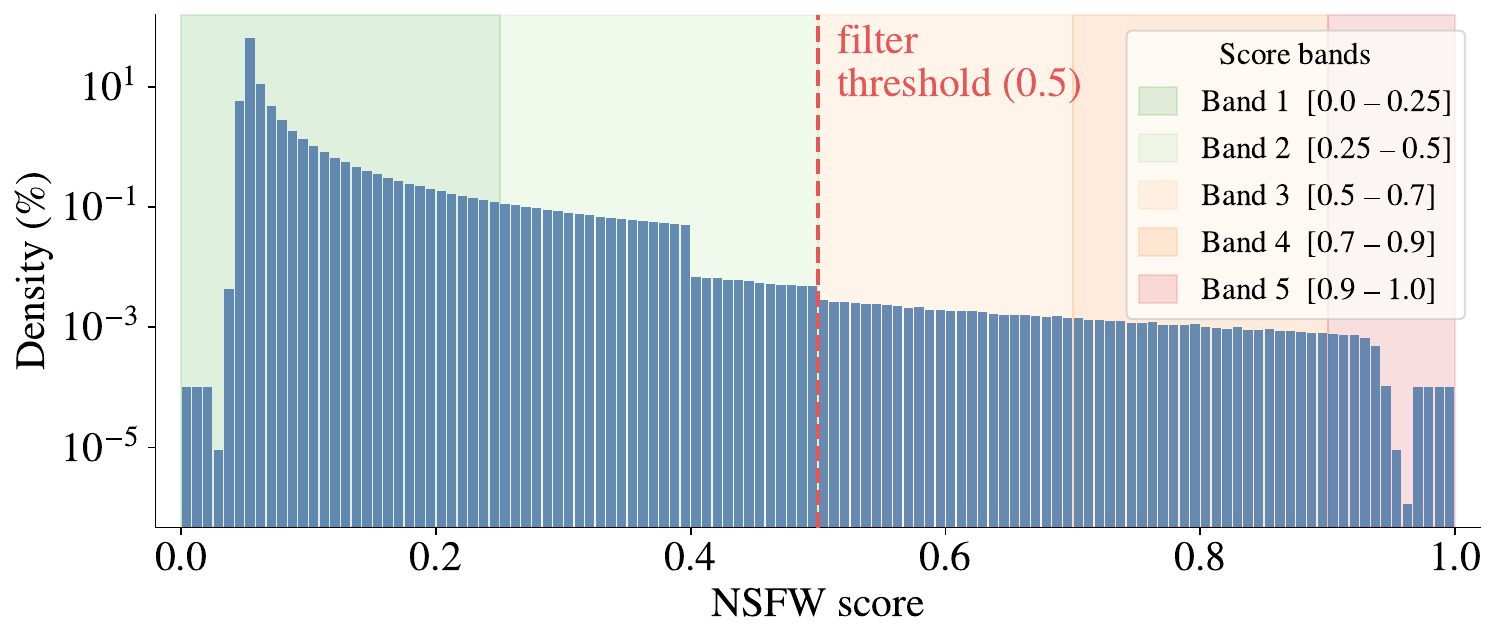}
        \caption{Histogram in logarithmic scale, with filtering threshold.}
        \label{fig:nsfw-bumble-histogram}
    \end{subfigure}

    \vspace{0.5cm}

    \begin{subfigure}{\linewidth}
        \centering

        \begin{subfigure}{0.20\linewidth}
            \centering
            \includegraphics[width=\linewidth,height=4cm]{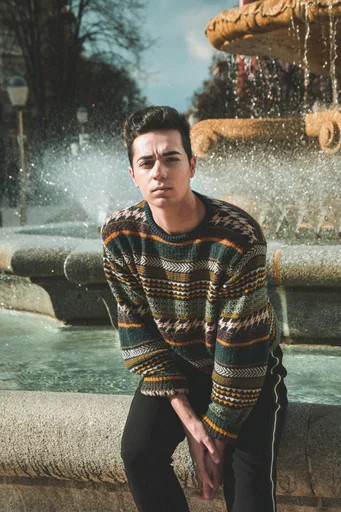}
            \caption{Band 1 (retained)}
            \label{fig:nsfw-bumble-band1}
        \end{subfigure}
        \hfill
        \begin{subfigure}{0.20\linewidth}
            \centering
            \includegraphics[width=\linewidth,height=4cm]{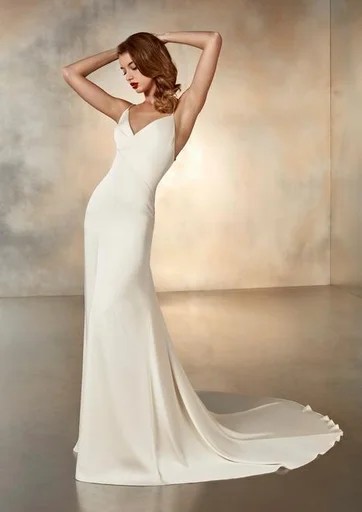}
            \caption{Band 2 (retained)}
            \label{fig:nsfw-bumble-band2}
        \end{subfigure}
        \hfill
        \begin{subfigure}{0.20\linewidth}
            \centering
            \includegraphics[width=\linewidth,height=4cm]{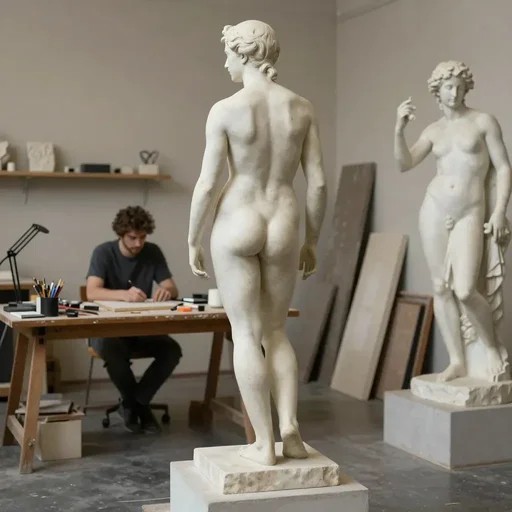}
            \caption{Band 3 (filtered)}
            \label{fig:nsfw-bumble-band3}
        \end{subfigure}
        \hfill
        \begin{subfigure}{0.20\linewidth}
            \centering
            \includegraphics[width=\linewidth,height=4cm]{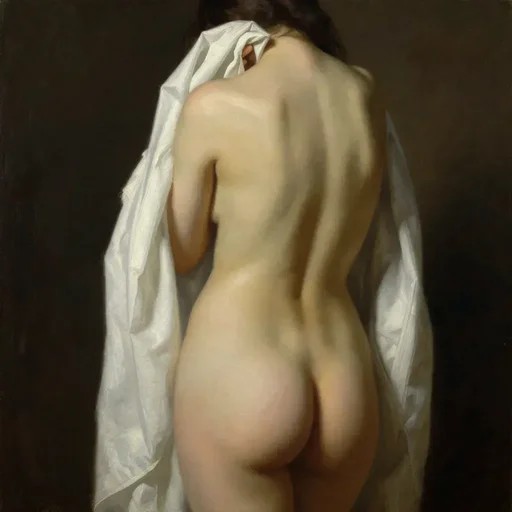}
            \caption{Band 4 (filtered)}
            \label{fig:nsfw-bumble-band4}
        \end{subfigure}
    \end{subfigure}

    \caption{Distribution and examples of Bumble NSFW score.}
    \label{fig:nsfw-bumble-distribution-and-examples}
\end{figure}

Fig.~\ref{fig:nsfw-bumble-distribution-and-examples} presents the score distribution for Bumble, divided into five score bands, along with representative examples for the first four bands. Again, no example from the last band is shown due to its explicit NSFW content. As before, the progression from safe images to NSFW content appears gradual. Band 3, illustrated in Fig.~\ref{fig:nsfw-bumble-band3}, consists mainly of mildly unsafe or even safe images, where nudity is typically associated with sculptures and paintings. The Bumble score is released for every image as \texttt{nsfw\_bumble}.

\begin{figure}[ht]
    \centering

        \centering
        \includegraphics[width=\linewidth]{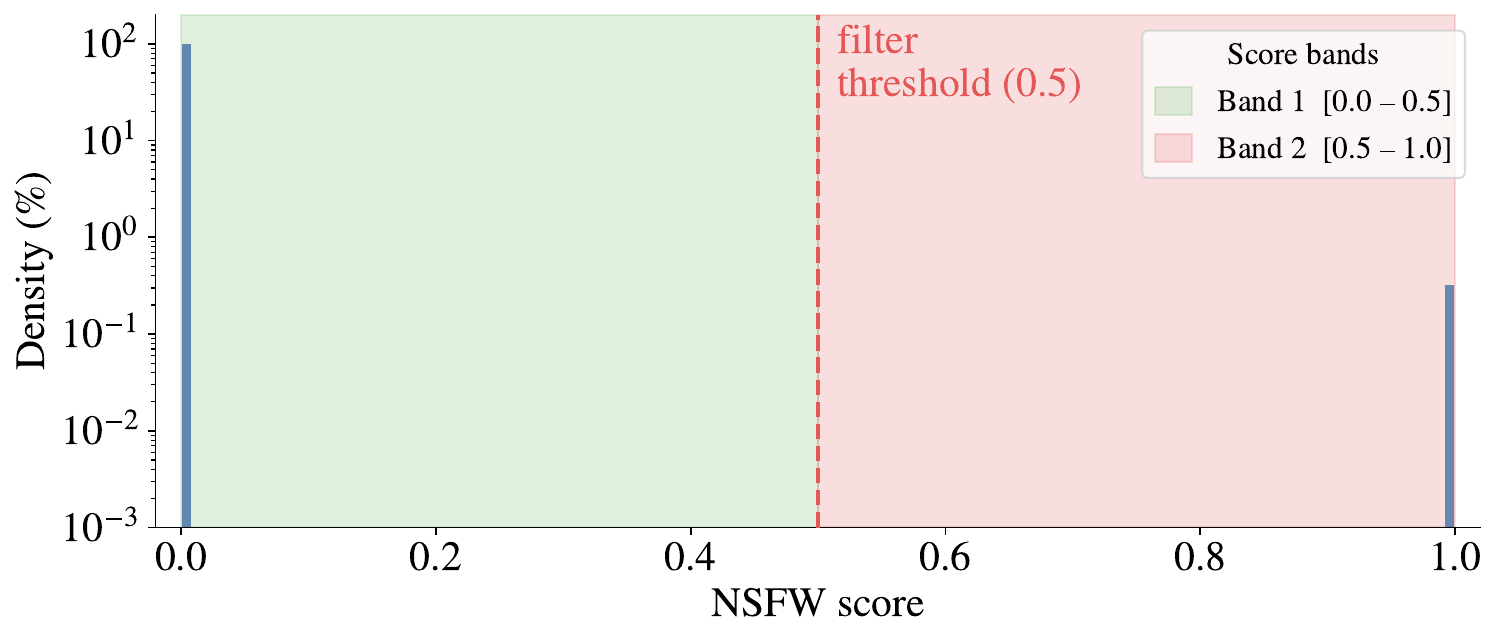}
        \caption{Histogram in logarithmic scale, with filtering threshold, of Falcon NSFW score.}
    \label{fig:nsfw-falcon-distribution-and-examples}
\end{figure}

Finally, Fig.~\ref{fig:nsfw-falcon-distribution-and-examples} shows the distribution of Falcon predictions. Since this detector produces binary predictions, we do not show individual examples, as some of the detected images contain highly explicit NSFW content. Nonetheless, this detector appears highly conservative, as many of the rejected images are only mildly unsafe or not clearly unsafe at all. 

\newpage
\subsection{Deduplication details}
\label{sec:deduplication-details}

\subsubsection{Perceptual hashing}
\label{sec:phash-technical-details}

We use the DCT-based perceptual hash (pHash) of \citet{venkatesan2000robust} as our first-pass duplicate detector. Each image is converted to grayscale, resized to $32\times 32$, and transformed by a 2D DCT; the top-left $8\times 8$ block of low-frequency coefficients is binarized against the median, yielding a 64-bit fingerprint. Pairs are compared by the Hamming distance $d$ between their hashes.

\begin{figure}[!ht]
  \centering
  \begin{minipage}[b]{0.48\linewidth}
    \vspace*{0pt}\centering
     \includegraphics[width=0.49\linewidth]{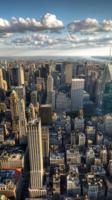}\hfill
    \includegraphics[width=0.49\linewidth]{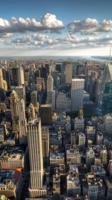}\\[2pt]
    {\small $d$\,=\,0}
  \end{minipage}%
  \hfill
  \begin{minipage}[b]{0.48\textwidth}
    \vspace*{0pt}\centering
    \includegraphics[width=0.49\linewidth]{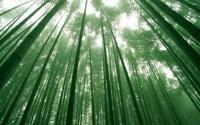}\hfill
    \includegraphics[width=0.49\linewidth]{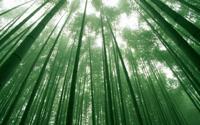}\\[2pt]
    \includegraphics[width=0.49\linewidth]{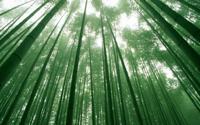}\hfill
    \includegraphics[width=0.49\linewidth]{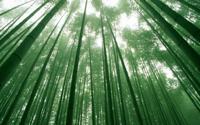}\\[2pt]
    {\small $d$\,=\,0}
  \end{minipage}
  \\[8pt]
  \begin{minipage}[t]{0.48\textwidth}
    \centering
    \includegraphics[width=0.49\linewidth]{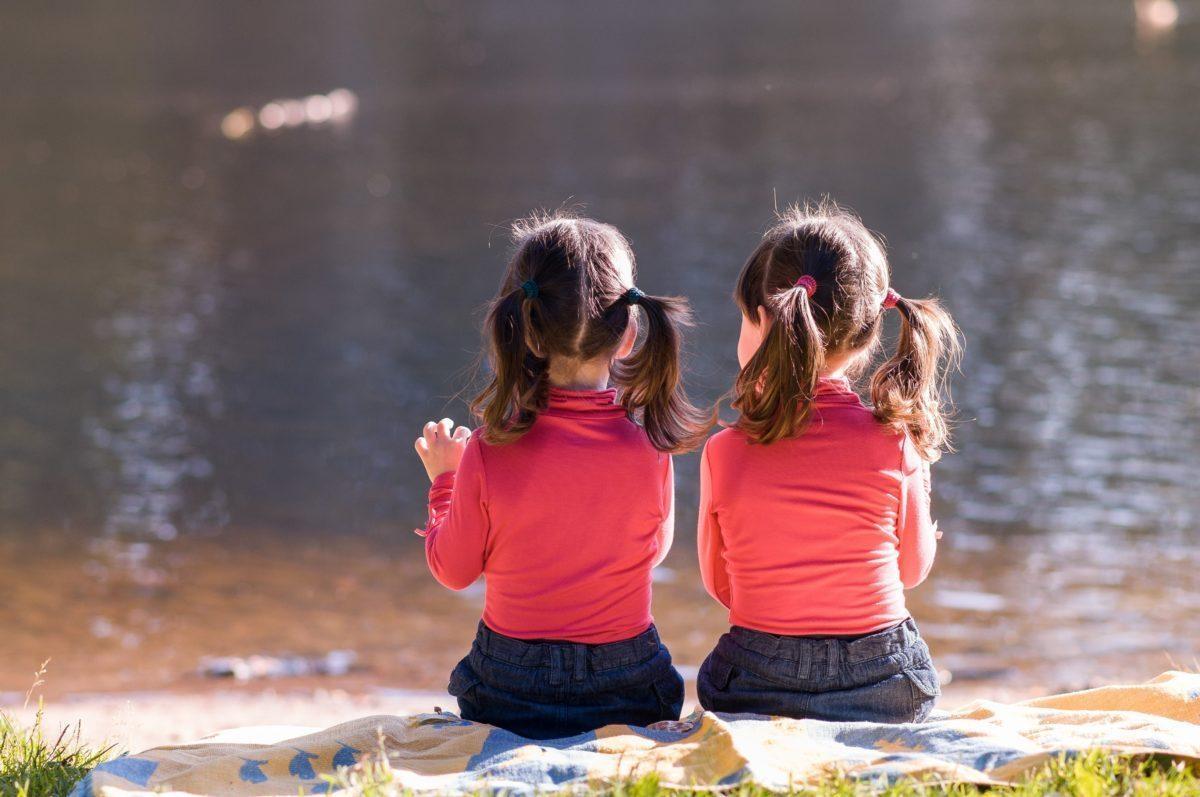}\hfill
    \includegraphics[width=0.49\linewidth]{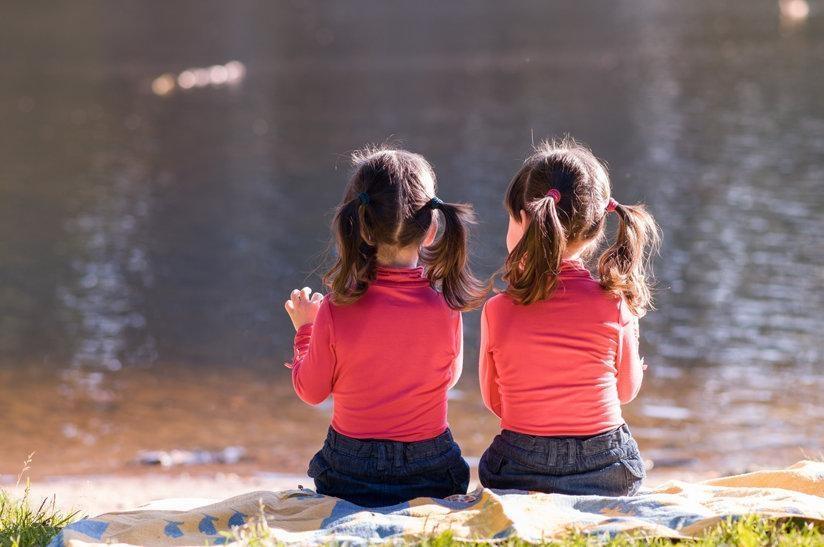}\\[2pt]
    {\small $d$\,=\,2}
  \end{minipage}%
  \hfill
  \begin{minipage}[t]{0.49\textwidth}
    \centering
    \includegraphics[width=0.49\linewidth]{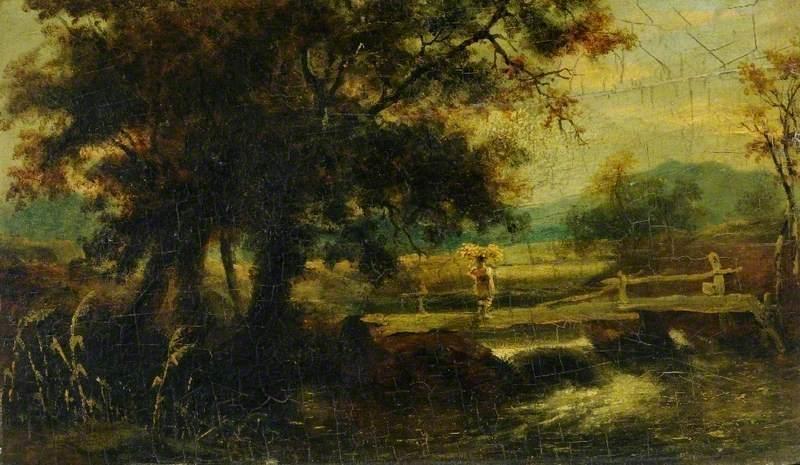}\hfill
    \includegraphics[width=0.49\linewidth]{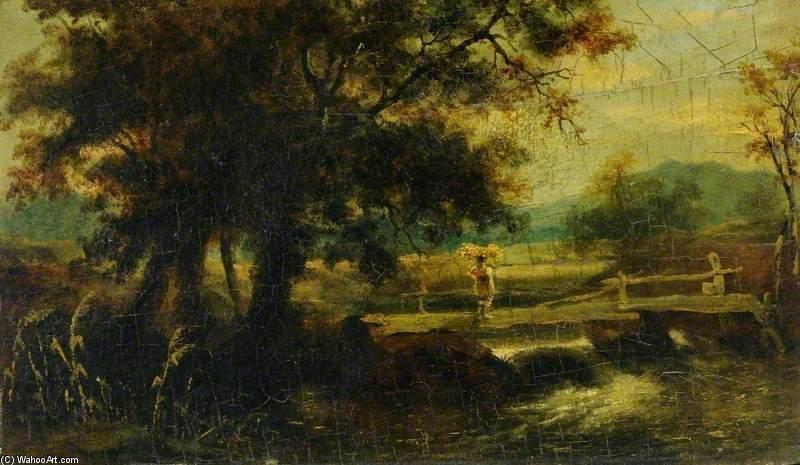}\\[2pt]
    {\small $d$\,=\,2}
  \end{minipage}
  \\[8pt]
  \begin{minipage}[t]{0.48\textwidth}
    \centering
    \includegraphics[width=0.32\linewidth]{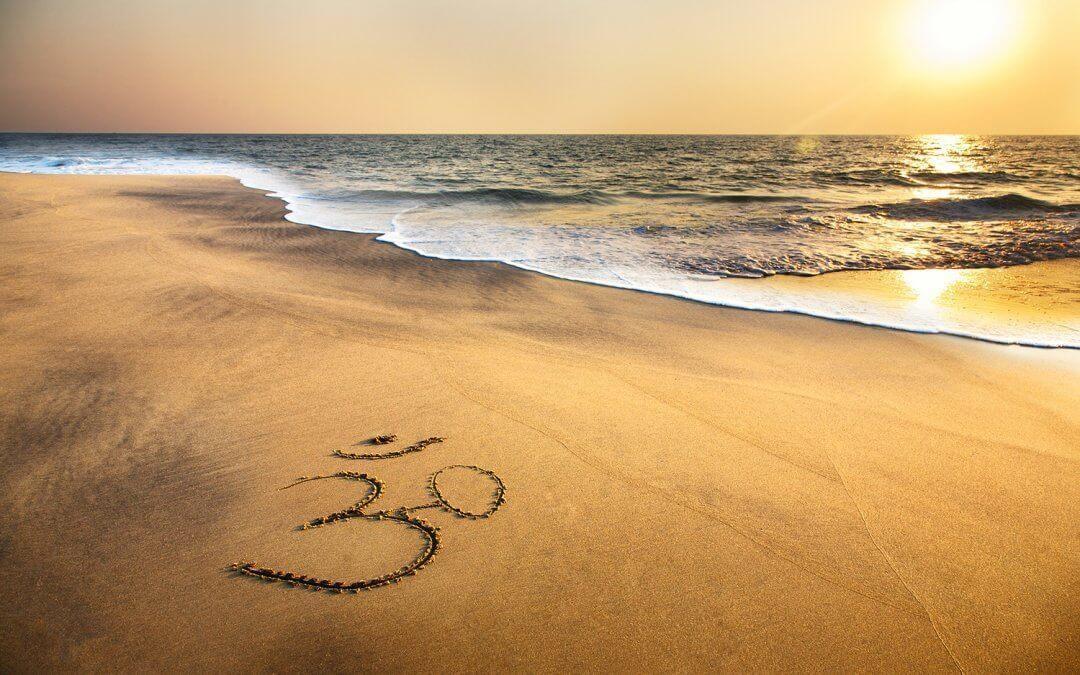}\hfill
    \includegraphics[width=0.32\linewidth]{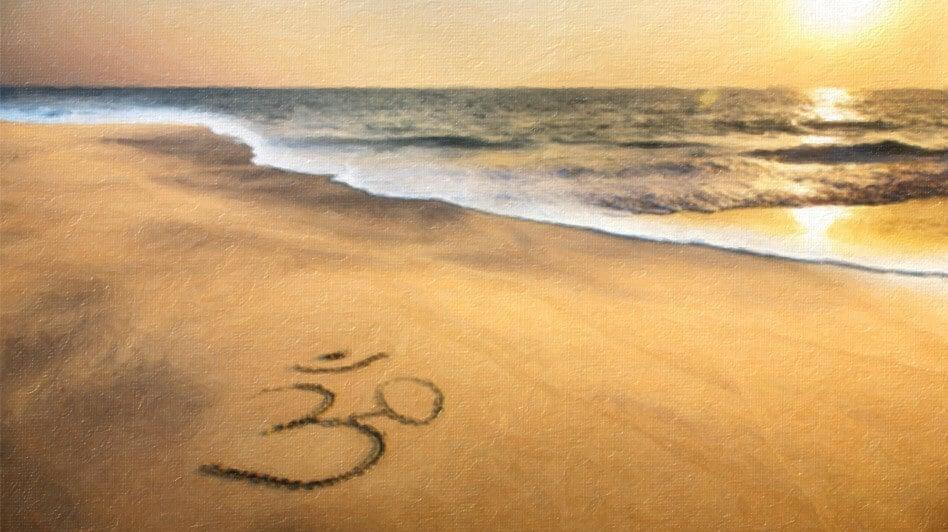}\hfill
    \includegraphics[width=0.32\linewidth]{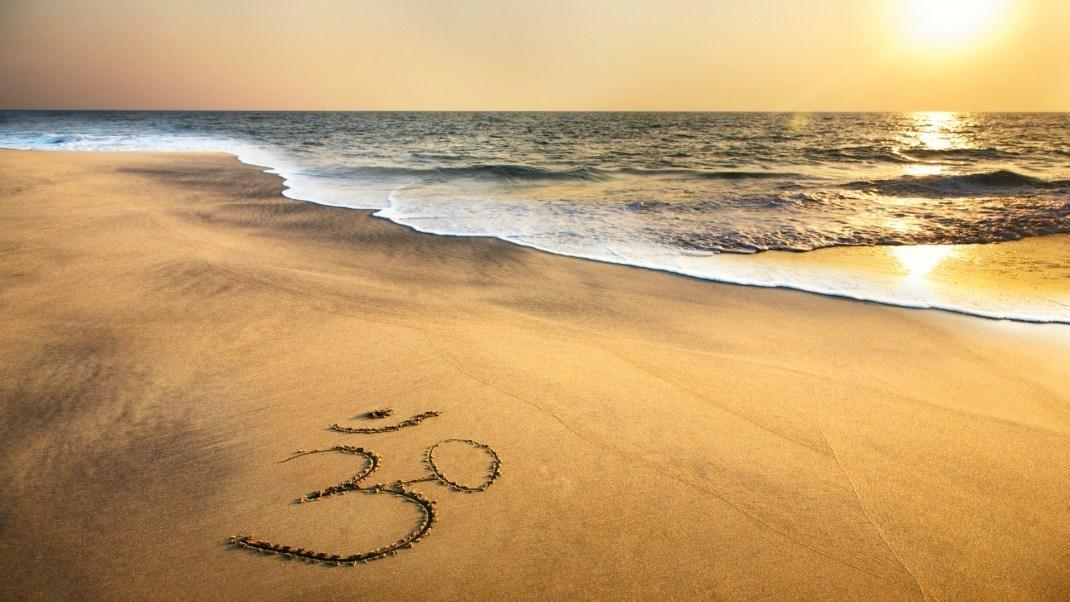}\\[2pt]
    {\small $d$\,=\,4}
  \end{minipage}%
  \hfill
  \begin{minipage}[t]{0.48\textwidth}
    \centering
    \includegraphics[width=0.32\linewidth]{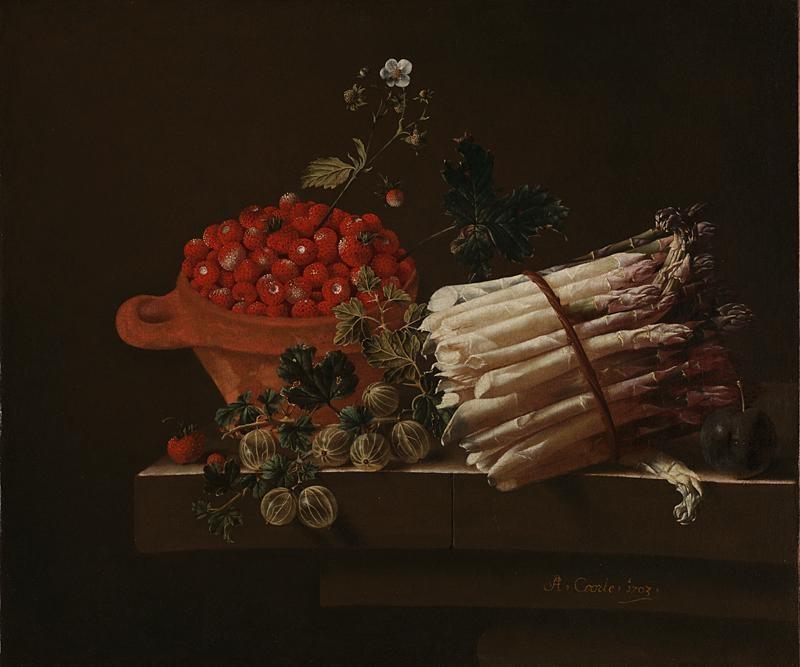}\hfill
    \includegraphics[width=0.32\linewidth]{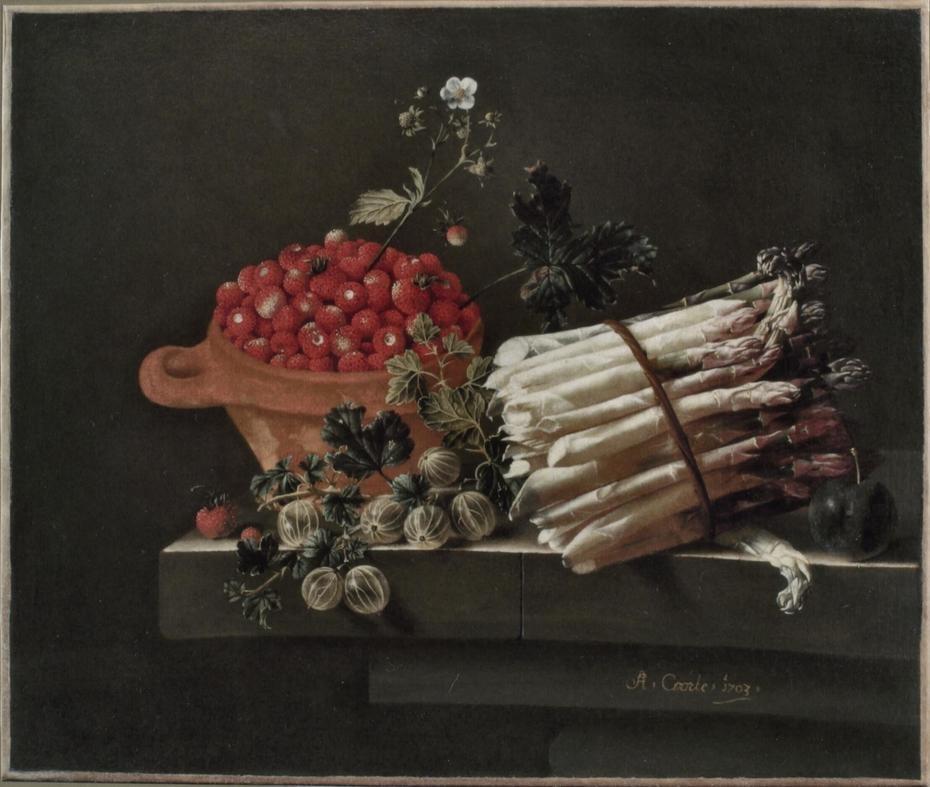}\hfill
    \includegraphics[width=0.32\linewidth]{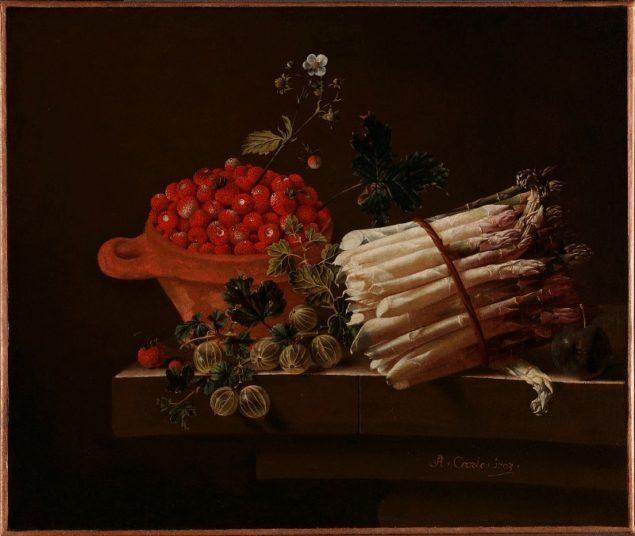}\\[2pt]
    {\small $d$\,=\,4}
  \end{minipage}
  \caption{Duplicate clusters detected by perceptual hashing at increasing Hamming distance $d$. Best viewed zoomed in.}
  \label{fig:phash_examples}
\end{figure}

In our pipeline, pHash is applied both \emph{intra-source} during merging and \emph{inter-source} after consolidation, removing approximately 19.7M and 1.94M images respectively. Because the hash discards high-frequency content, it is robust to mild JPEG re-compression, resizing, and small overlays, but sensitive to flips, crops, and color shifts. Fig.~\ref{fig:phash_examples} illustrates representative clusters at three operating points. Specifically, $d=0$ corresponds to exact-hash matches; at $d=2$, the clusters comprise visually indistinguishable copies that differ only in compression or minor pixel-level modifications; and at $d=4$, they continue to capture re-encodings of the same image with slight variations in resolution or color grading.

\begin{figure}
  \centering
  \begin{minipage}[t]{\textwidth}
    \centering
    \includegraphics[height=2.5cm]{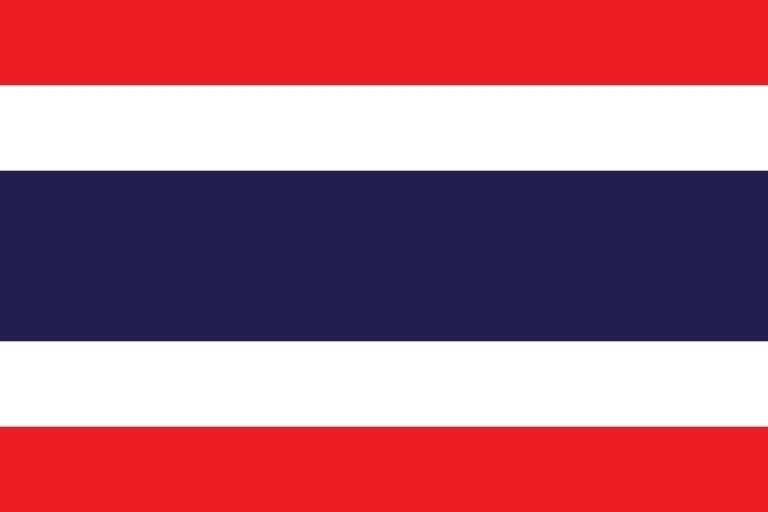}\hfill
    \includegraphics[height=2.5cm]{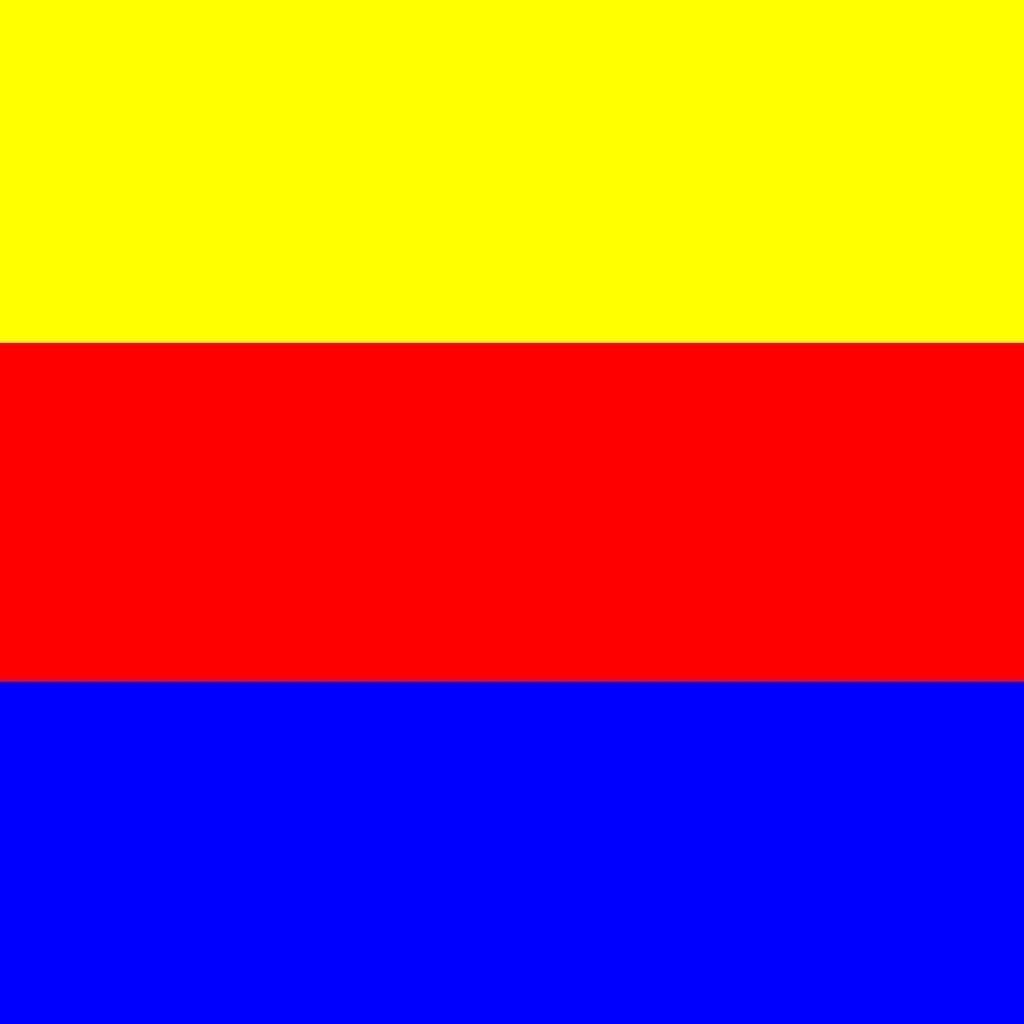}\hfill
    \includegraphics[height=2.5cm]{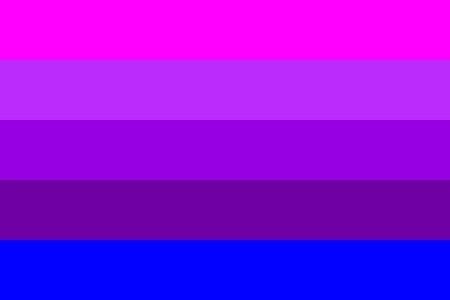}\hfill
    \includegraphics[height=2.5cm]{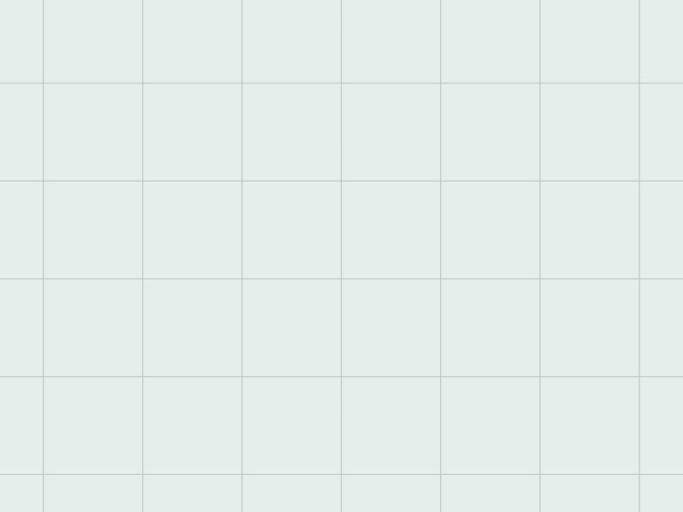}\\[2pt]
    {\small Cluster~A: horizontal stripe patterns}
  \end{minipage}
  \\[8pt]
  \begin{minipage}[t]{\textwidth}
    \centering
    \includegraphics[height=2.4cm]{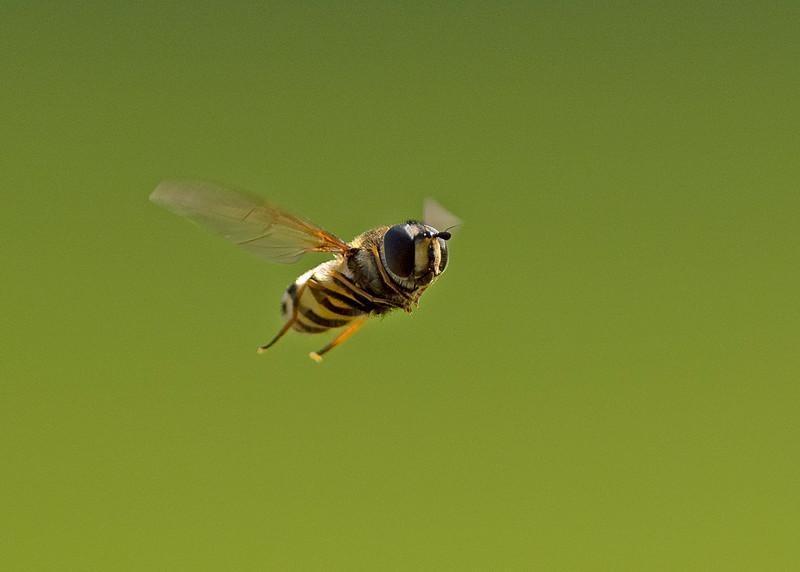}\hfill
    \includegraphics[height=2.4cm]{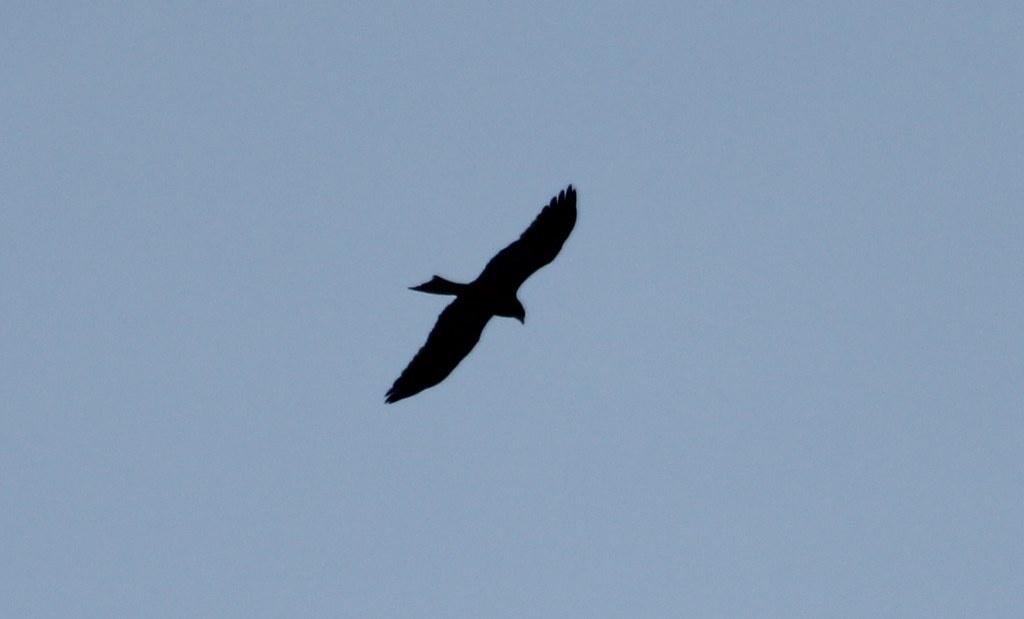}\hfill
    \includegraphics[height=2.4cm]{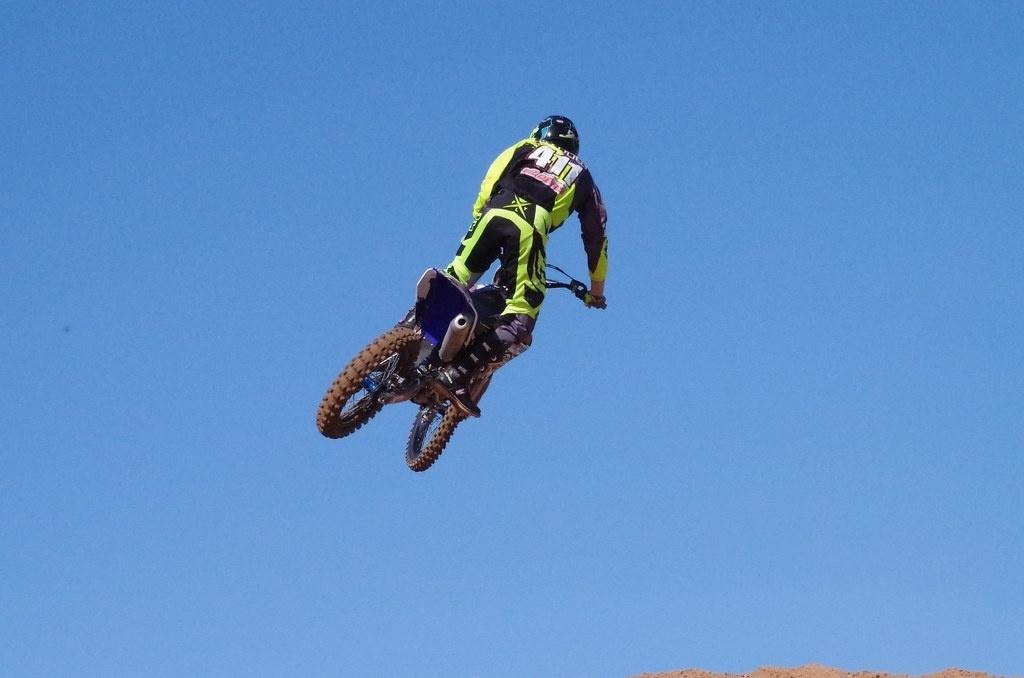}\hfill
    \includegraphics[height=2.4cm]{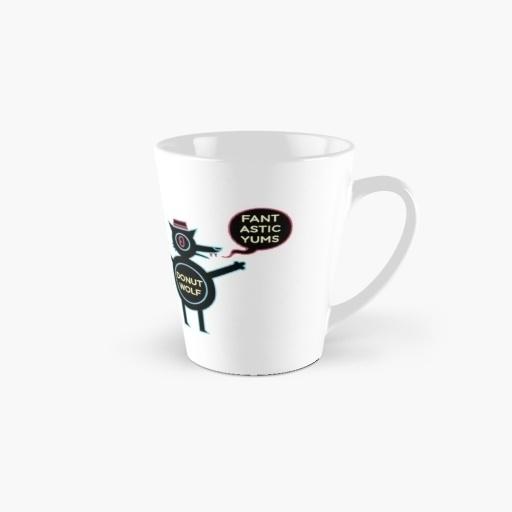}\\[2pt]
    {\small Cluster~B: small object on uniform background}
  \end{minipage}
  \\[8pt]
  \begin{minipage}[t]{\textwidth}
    \centering
    \includegraphics[height=1.3cm]{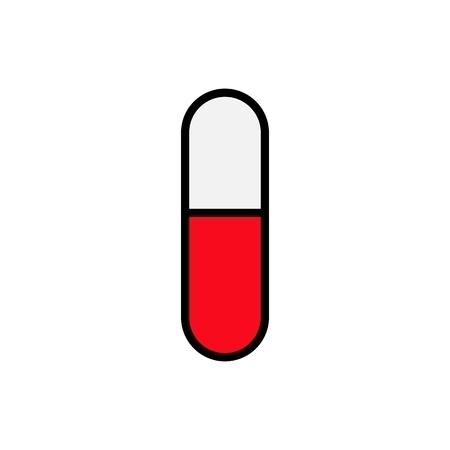}\hfill
    \includegraphics[height=1.3cm]{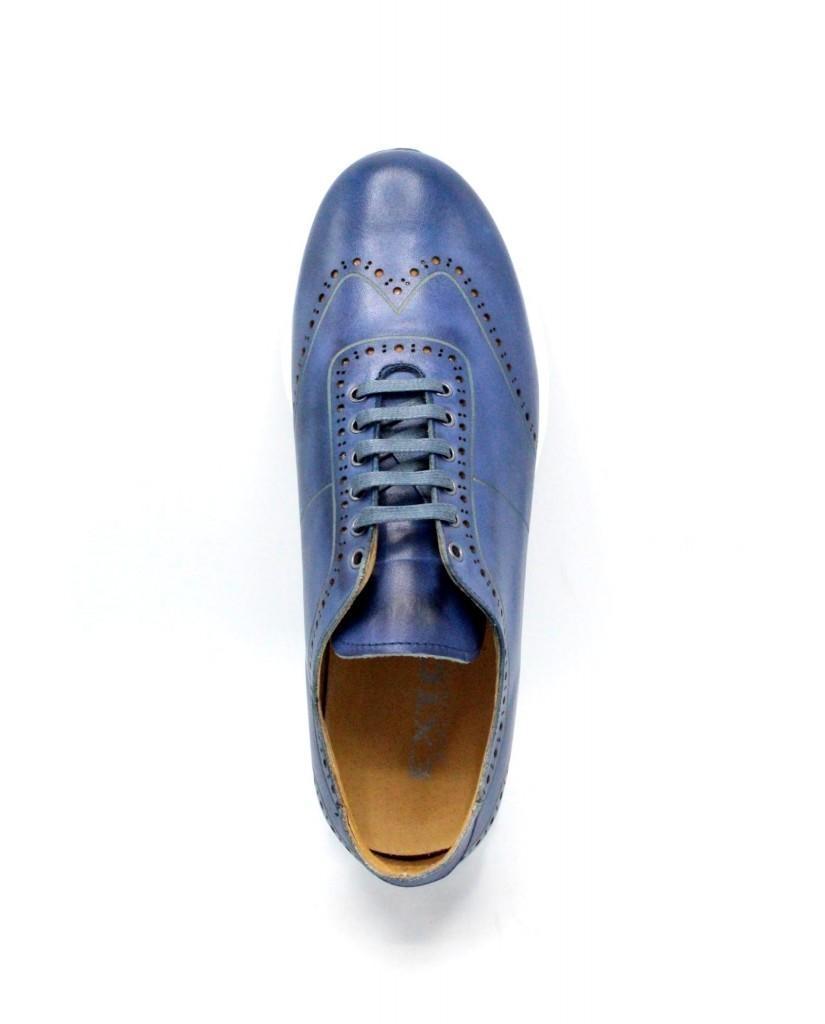}\hfill
    \includegraphics[height=1.3cm]{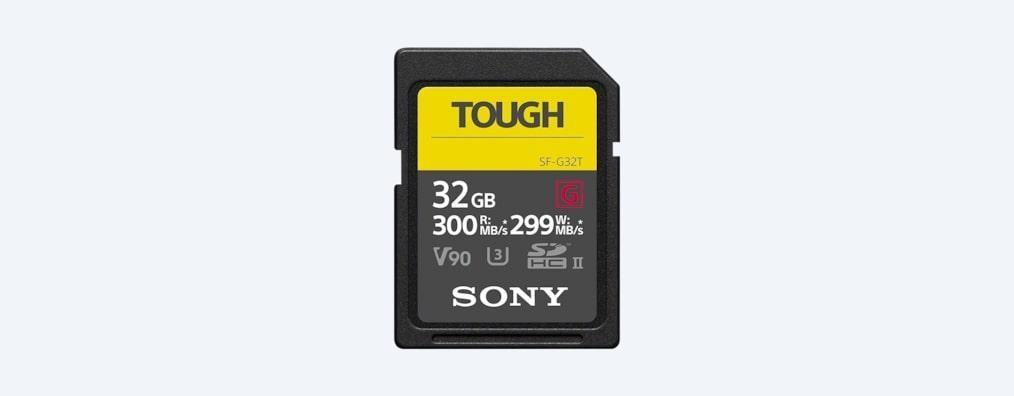}\\[2pt]
    {\small Cluster~C: centered object on white background}
  \end{minipage}
  \caption{pHash false positive clusters: each row shows images that pHash assigns a low Hamming distance $d=1$, yet the images are semantically unrelated. pHash conflates global color distributions and coarse spatial layout with perceptual identity.}
  \label{fig:dedup_limitations_phash}
\end{figure}

However, as $d$ increases, pHash starts to mix genuine duplicates with semantically unrelated images that share a similar global layout, since the 64-bit fingerprint only encodes a coarse spatial frequency representation of the image. Fig.~\ref{fig:dedup_limitations_phash} illustrates such false-positive clusters: horizontal stripe patterns, small objects on a uniform background (a bee, a bird, a motocross rider, a coffee mug), and centered products on a white background (a pill icon, a shoe, an SD card). To recover from these failure modes while still catching transformations that defeat low-frequency hashing (flips, large crops, color shifts, watermark insertion), we use a learned embedding (SSCD) for the second pass.

\clearpage
\subsubsection{SSCD}
\label{sec:sscd-technical-details}

To complement pHash and capture transformations that defeat low-frequency hashing such as horizontal flips, large crops, color and tone shifts, watermark insertion, or background substitution, we use Self-Supervised Copy Detection (SSCD) embeddings~\citep{pizzi2022self}. We use the public \texttt{sscd\_disc\_mixup} checkpoint~\citep{sscd_code}, which produces a 512-dimensional descriptor explicitly trained for copy detection. Embeddings are L2-normalized, indexed with FAISS~\citep{douze2024faiss}, and for every image we retrieve its $k=64$ nearest neighbors. Pairs with cosine similarity $\geq 0.75$ are merged into clusters via union-find, and within each cluster we keep a single representative chosen by the highest combined resolution and aesthetic score for possible ties; the remaining members are discarded. This stage removes an additional 5.22M images.

\begin{figure}[!ht]
  \centering
  \includegraphics[width=\textwidth]{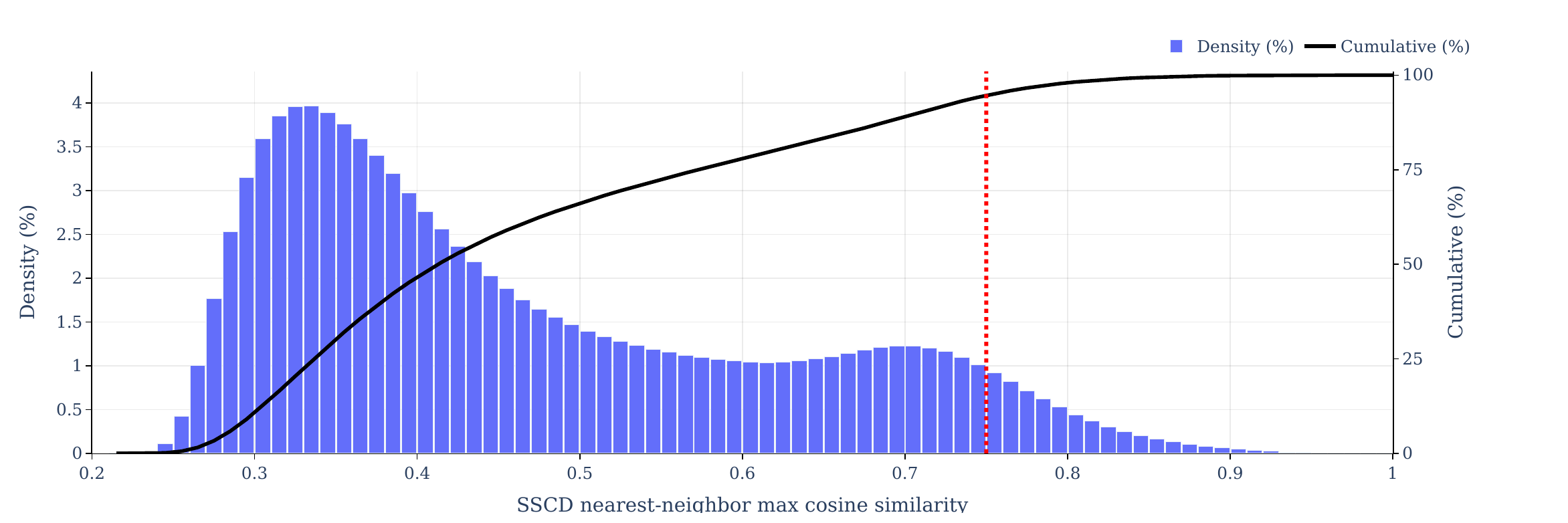}
  \caption{Distribution of SSCD nearest-neighbor maximum cosine similarities.}
  \label{fig:sscd_nn_score_distribution}
\end{figure}

\begin{figure}[!ht]
  \centering
  \newcommand{\sweepimg}[1]{\includegraphics[height=1.7cm]{#1}}
  \newcommand{\sweepcap}[1]{{\small #1}}
  \newcommand{\sweeppair}[3]{%
    \begin{tabular}[t]{@{}c@{}}
      \sweepimg{#1}\hspace{1pt}\sweepimg{#2}\\[2pt]
      \sweepcap{#3}
    \end{tabular}%
  }
  \sweeppair{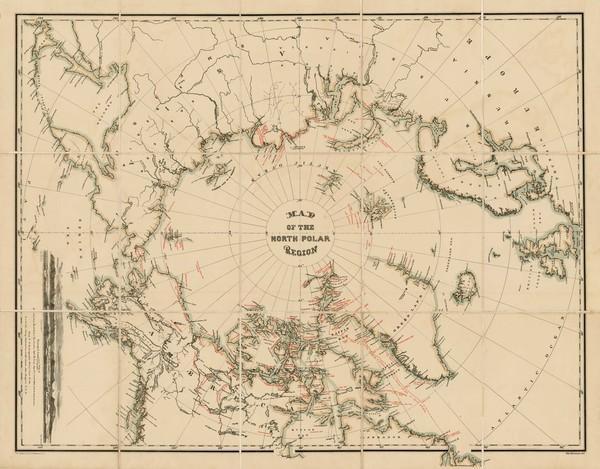}%
           {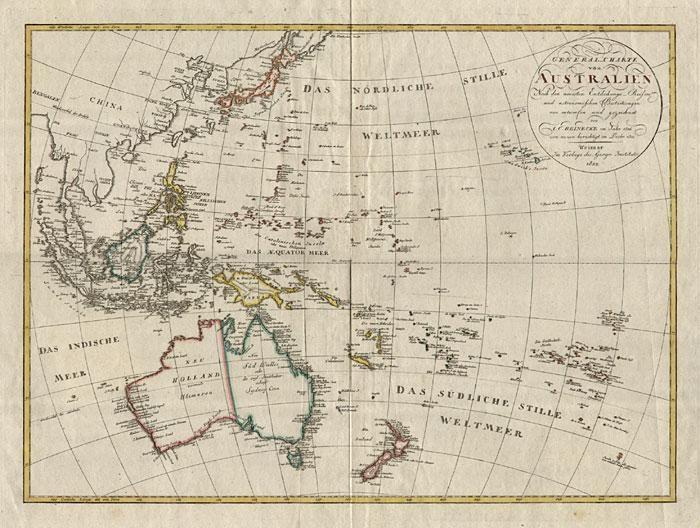}%
           {SSCD\,=\,0.51}\hfill
  \sweeppair{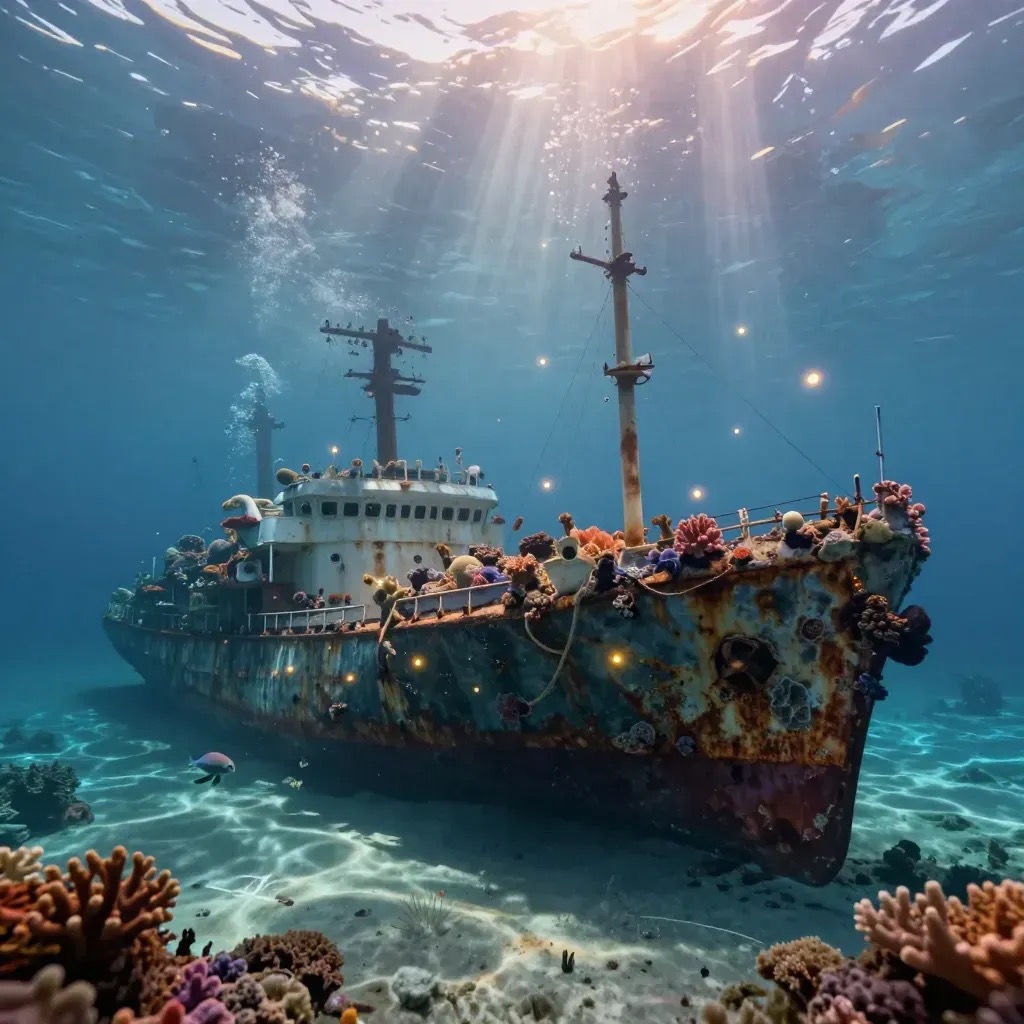}%
           {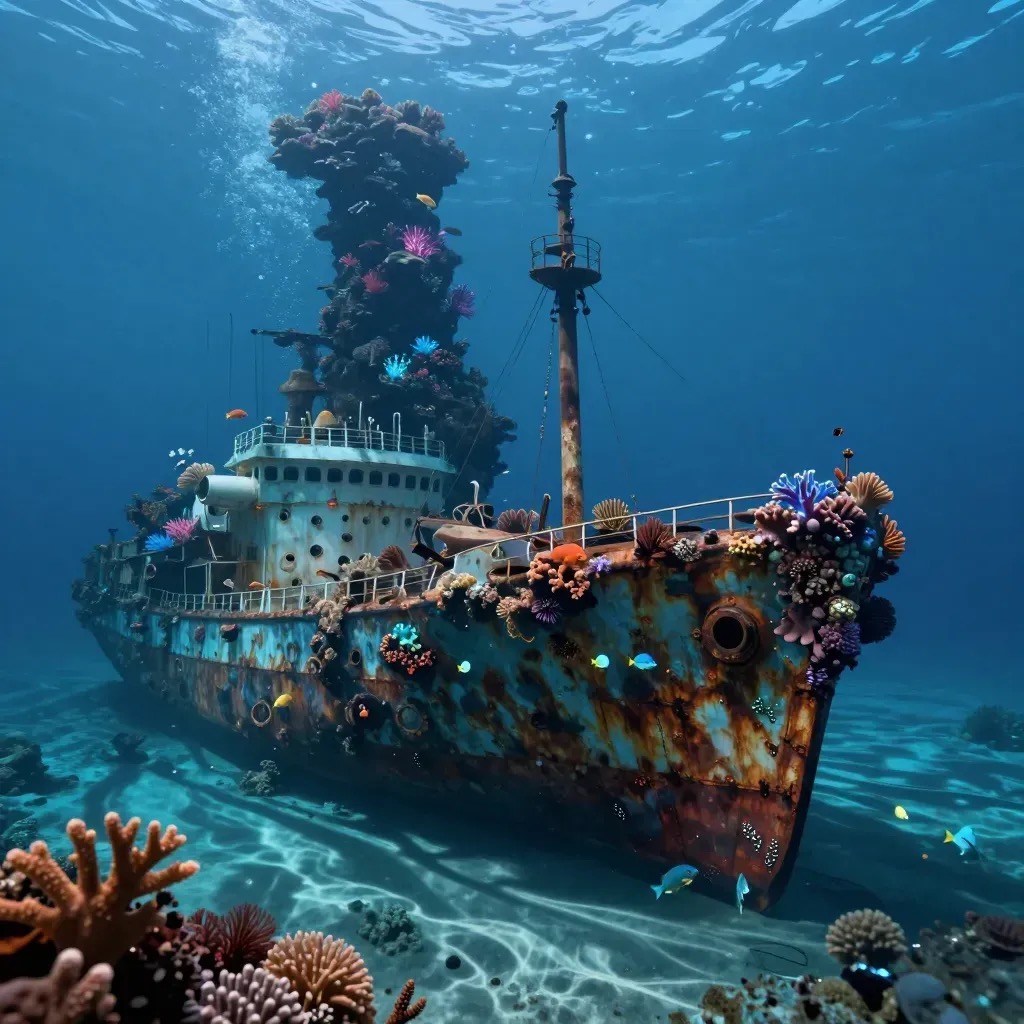}%
           {SSCD\,=\,0.58}\hfill
  \sweeppair{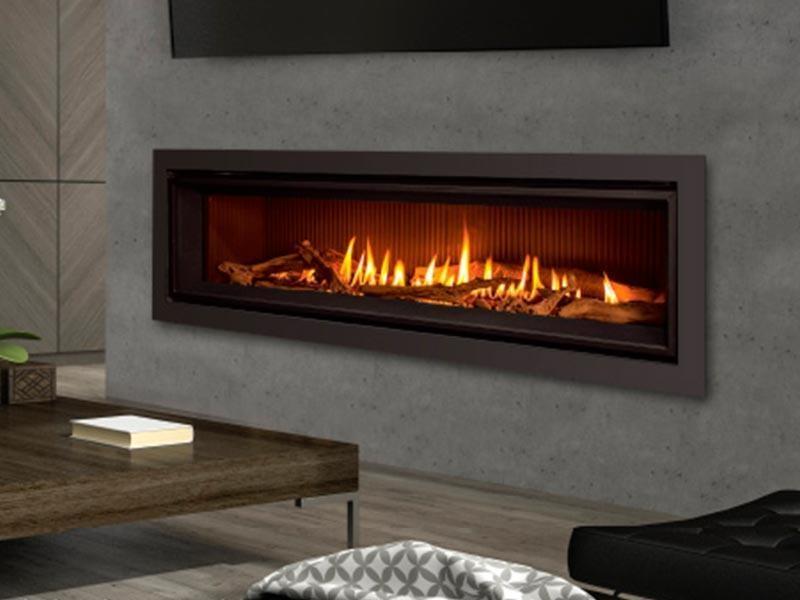}%
           {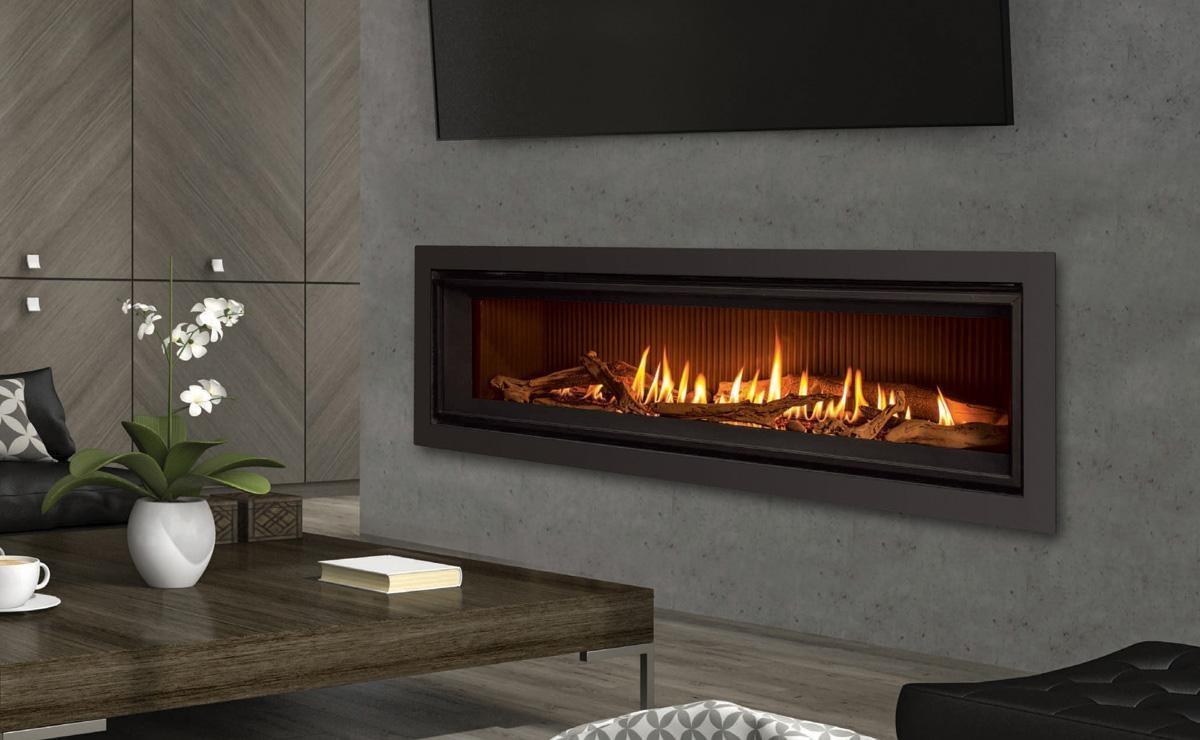}%
           {SSCD\,=\,0.65}\hfill
  \\[6pt]
  \sweeppair{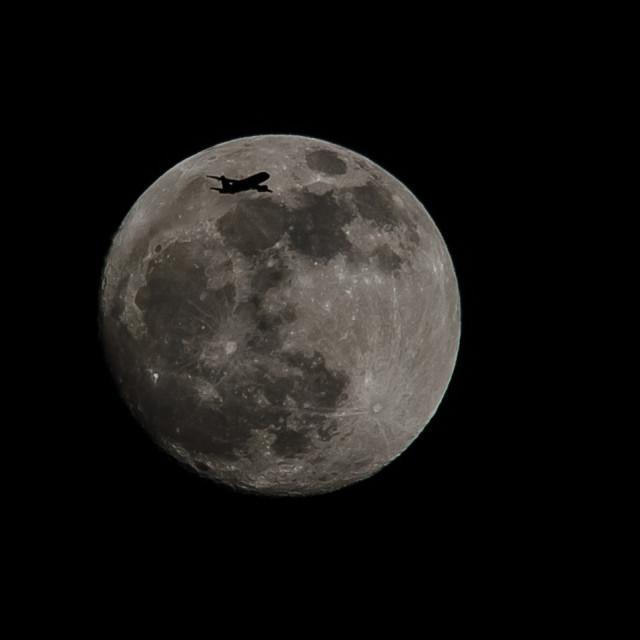}%
           {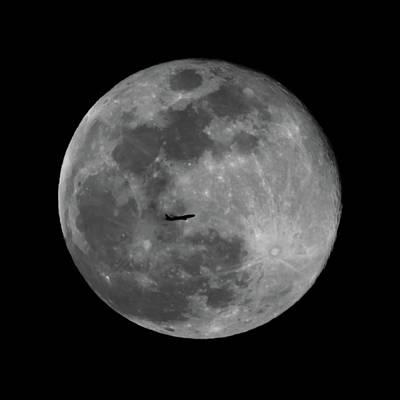}%
           {SSCD\,=\,0.71}\hfill
  \sweeppair{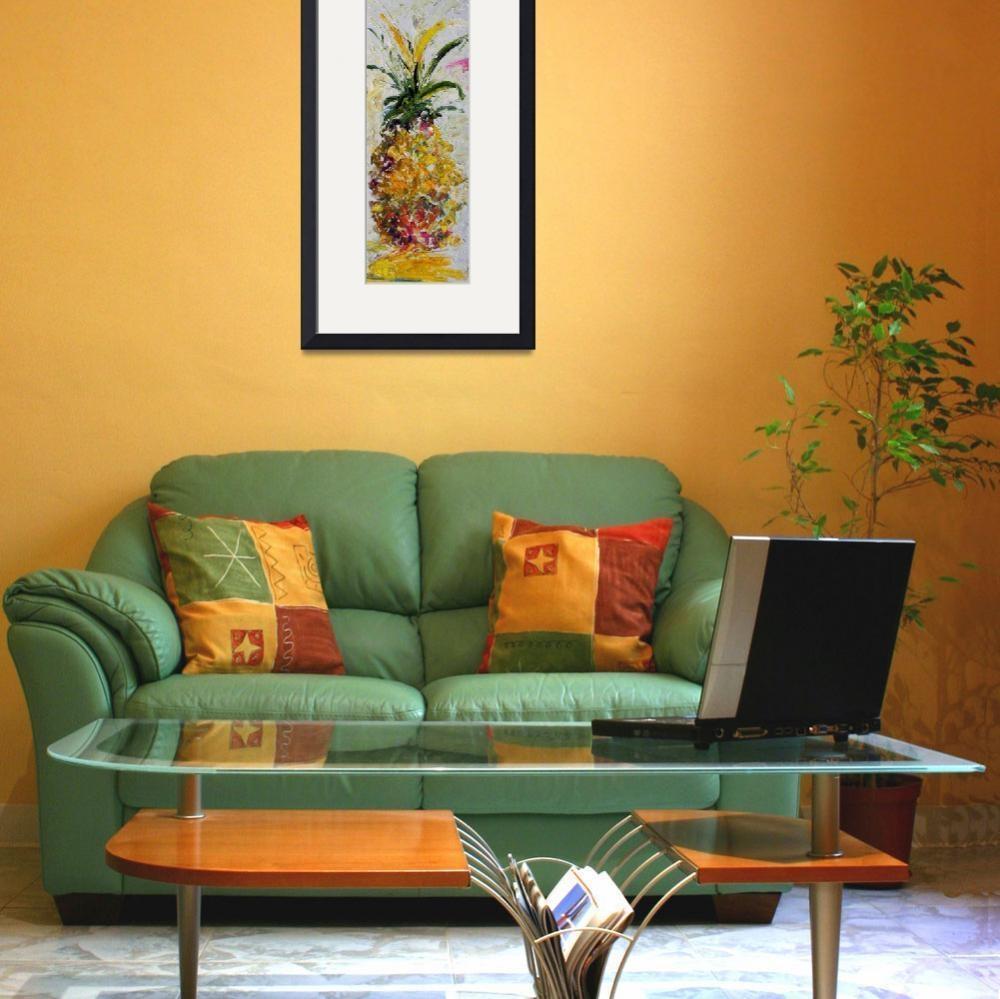}%
           {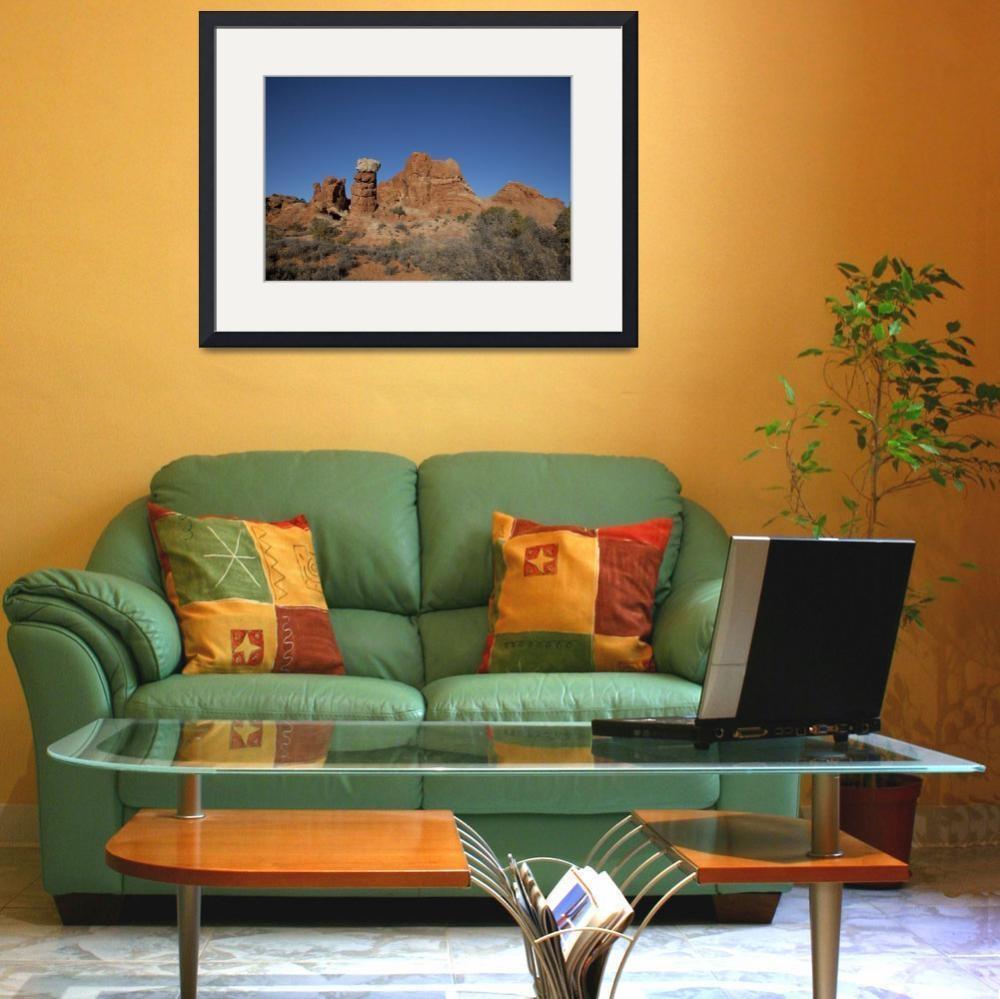}%
           {SSCD\,=\,0.77}\hfill
  \sweeppair{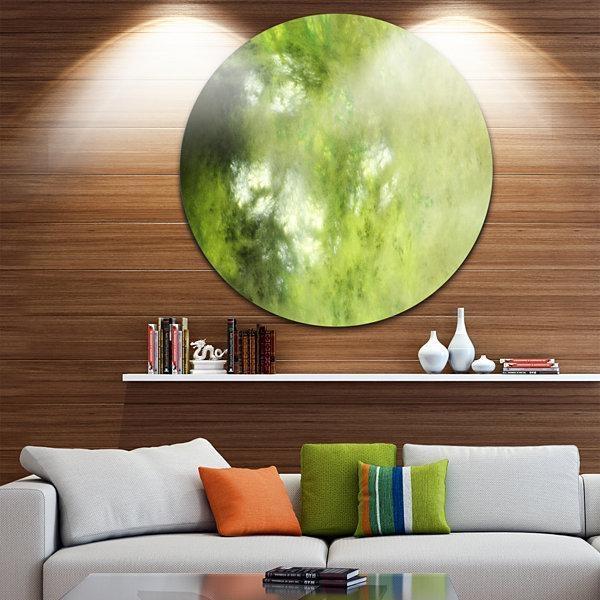}%
           {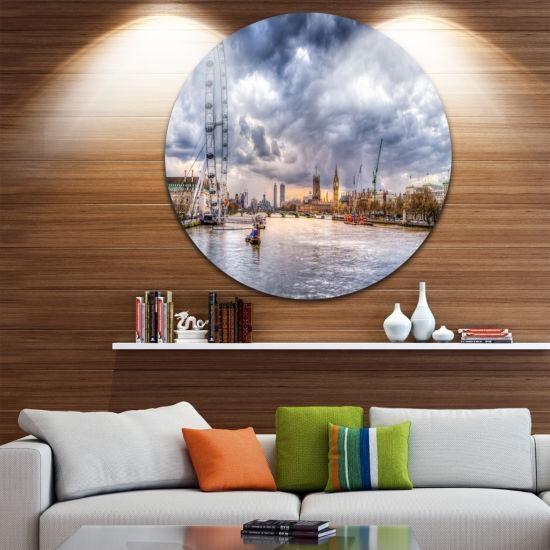}%
           {SSCD\,=\,0.82}\hfill
  \\[6pt]
  \sweeppair{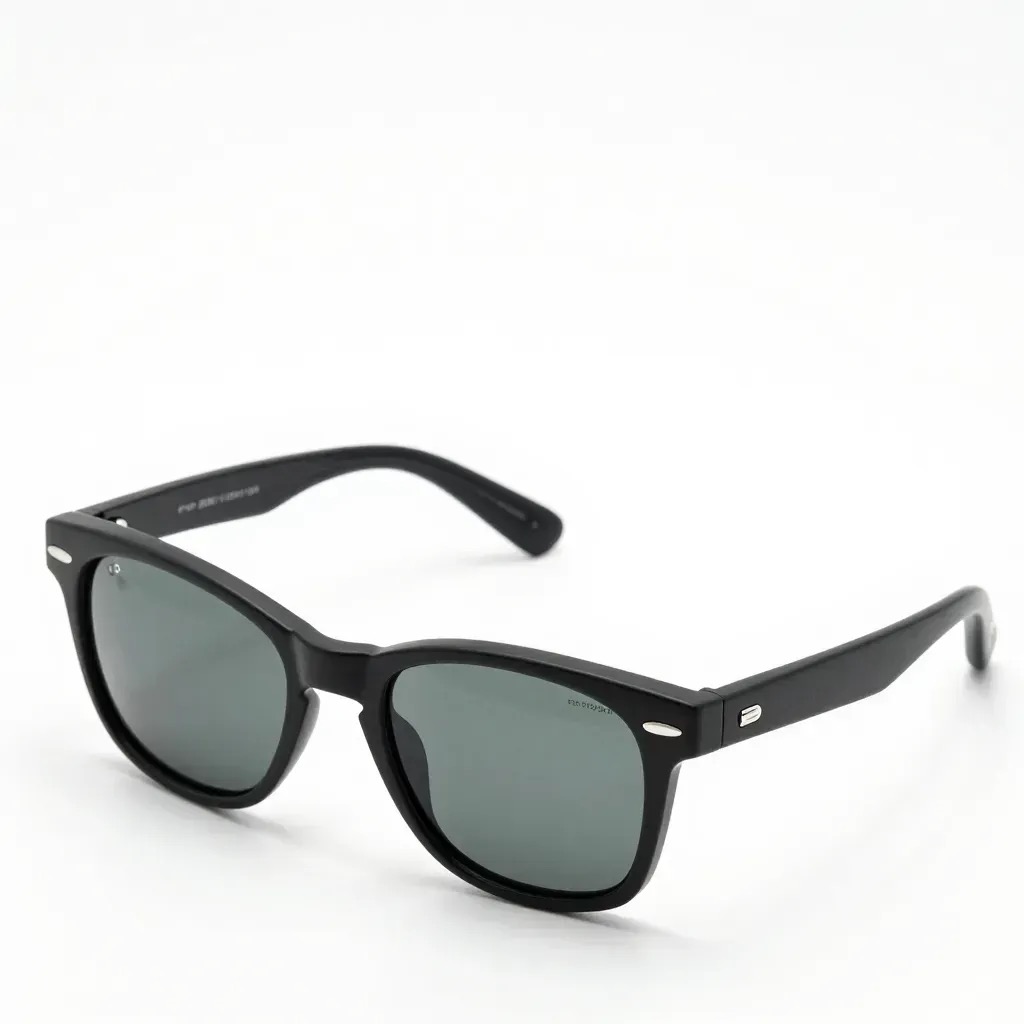}%
           {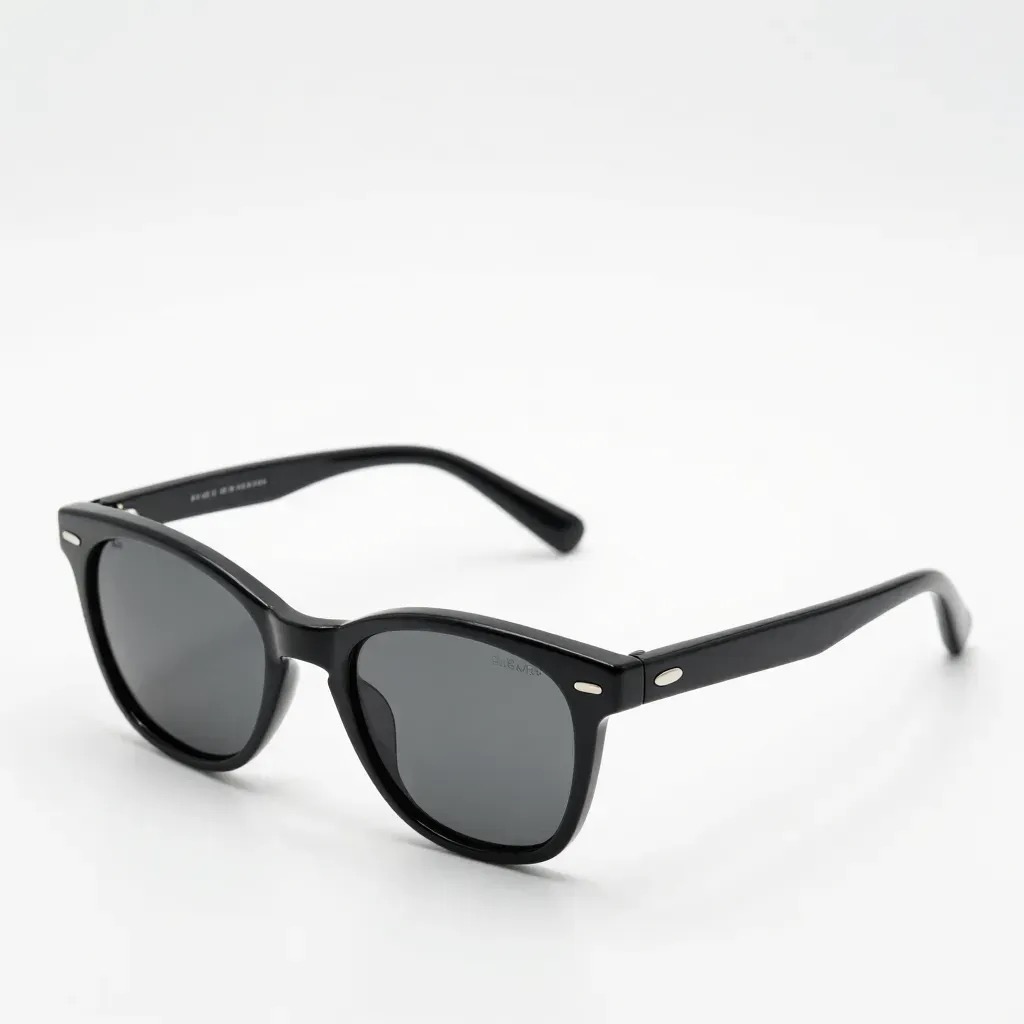}%
           {SSCD\,=\,0.89}\hfill
  \sweeppair{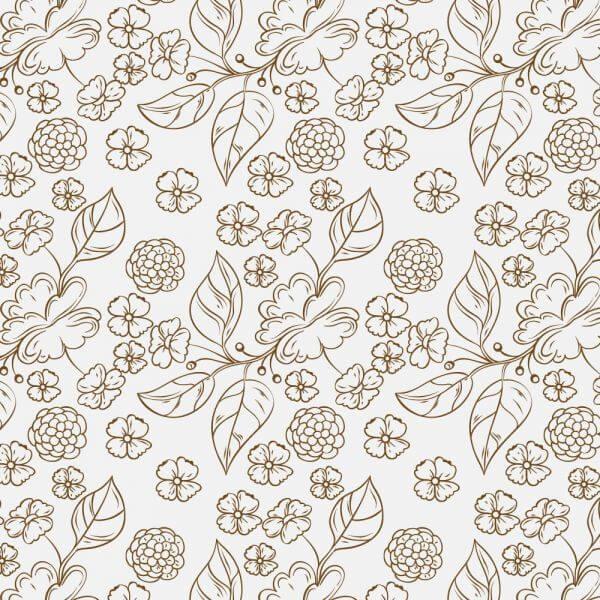}%
           {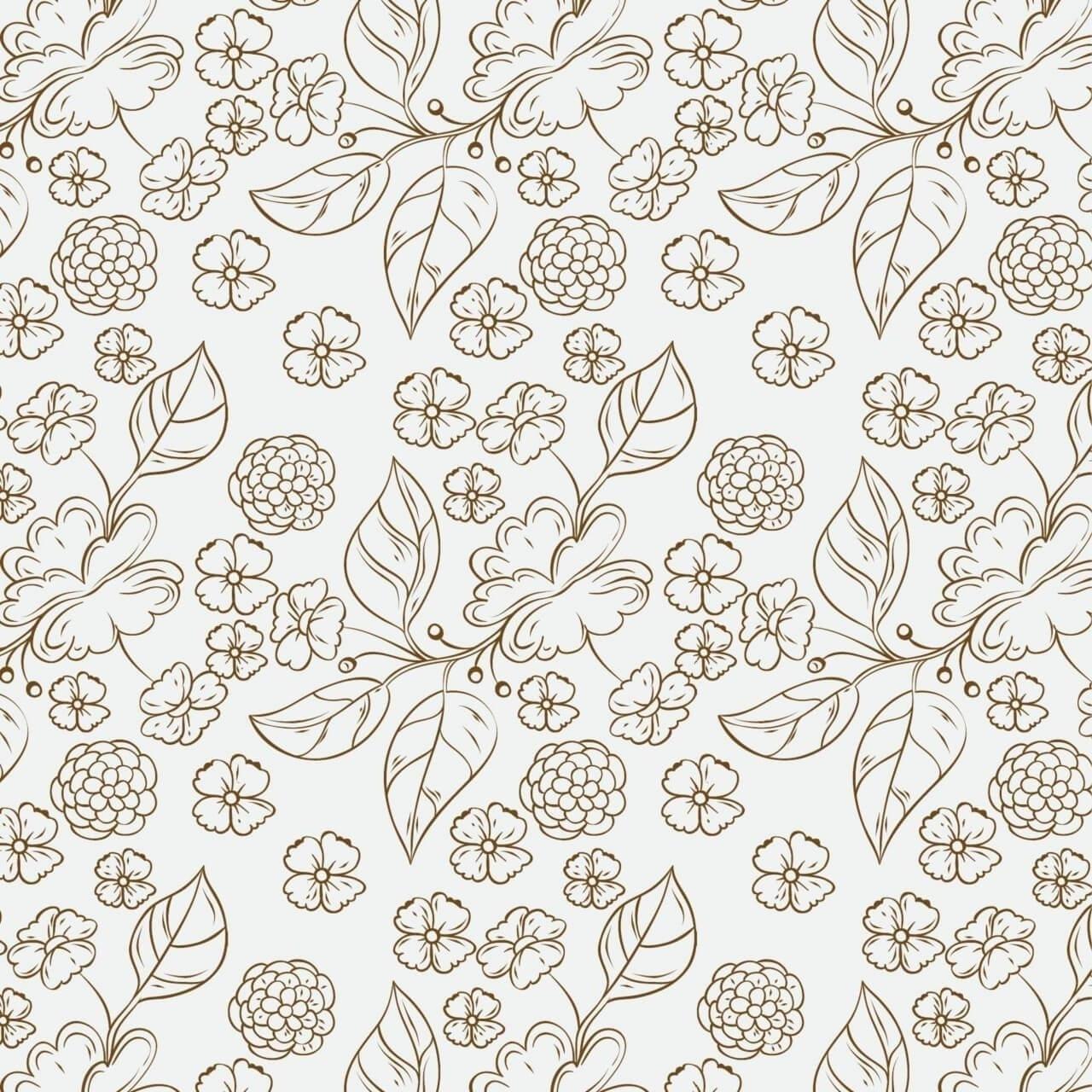}%
           {SSCD\,=\,0.92}\hfill
  \sweeppair{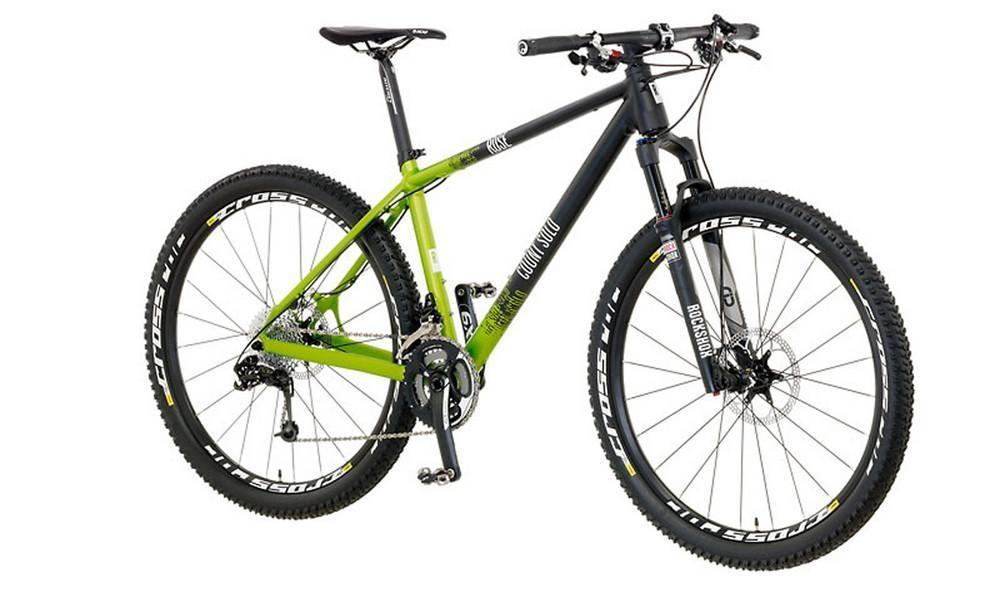}%
           {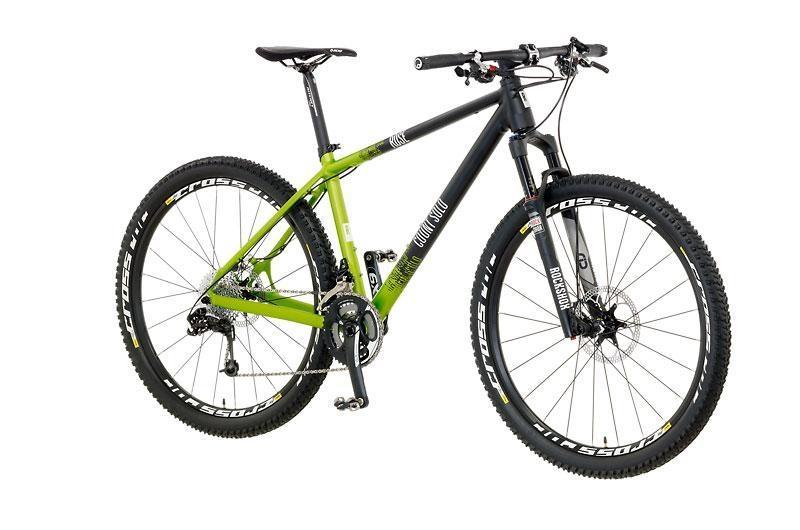}%
           {SSCD\,=\,0.96}\hfill
  \caption{SSCD threshold sweep. Nearest-neighbor pairs sampled at increasing cosine similarity. For low SSCD, pairs are merely semantically related (different photos from the same scene, object category, or visual theme) and are retained. When SSCD increases, pairs are near-duplicates}.
  \label{fig:sscd_threshold_sweep}
\end{figure}

Fig.~\ref{fig:sscd_nn_score_distribution} reports the distribution of nearest-neighbor cosine similarities over the deduplicated pool. The distribution is bimodal: a heavy mass below ${\sim}0.5$ corresponding to genuinely distinct images, and a long tail towards $1.0$ populated by near-duplicates. We use the $0.75$ operating point recommended by the SSCD authors~\citep{sscd_code} (90\% precision on DISC) and further validate it by manually inspecting random pairs from $0.05$-wide bins (cf.\ Fig.~\ref{fig:dedup_examples} of the main paper and Fig.~\ref{fig:sscd_threshold_sweep}). Above $0.75$, the neighbors are consistently near-duplicates, whereas below $0.75$, they are merely semantically related.

\begin{figure}[ht]
  \centering
  \begin{minipage}[t]{\linewidth}
    \centering
    \fcolorbox{green!60!black}{white}{\includegraphics[width=0.3\linewidth]{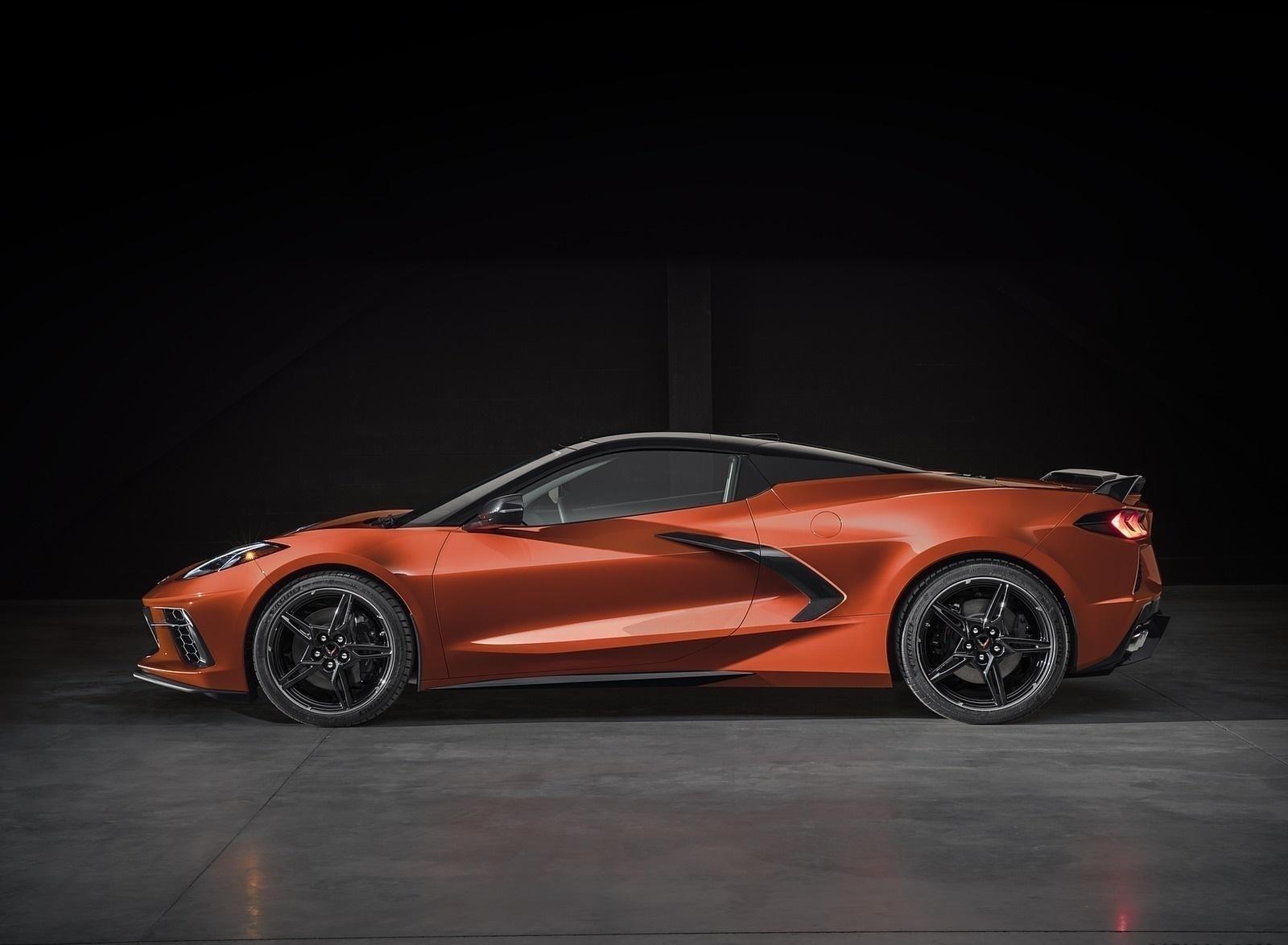}}\hfill
    \fcolorbox{white}{white}{\includegraphics[width=0.3\linewidth]{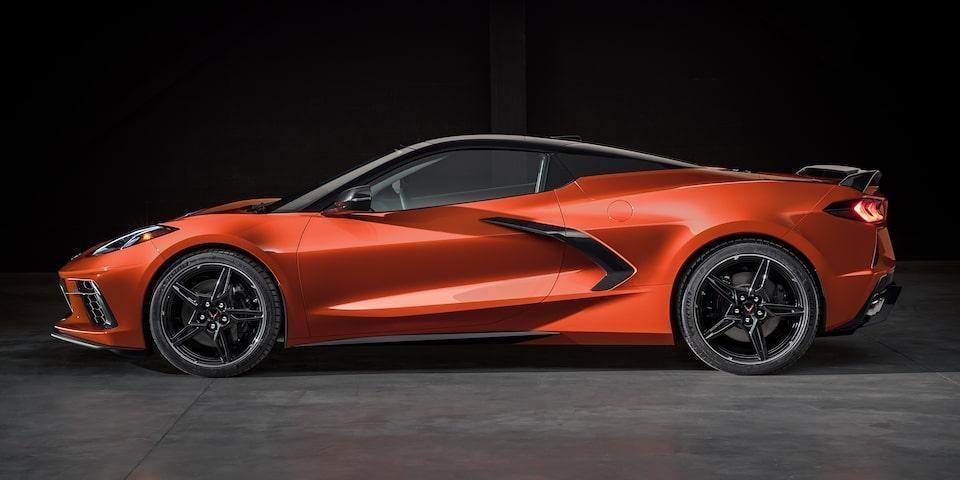}}\hfill
    \fcolorbox{white}{white}{\includegraphics[width=0.3\linewidth]{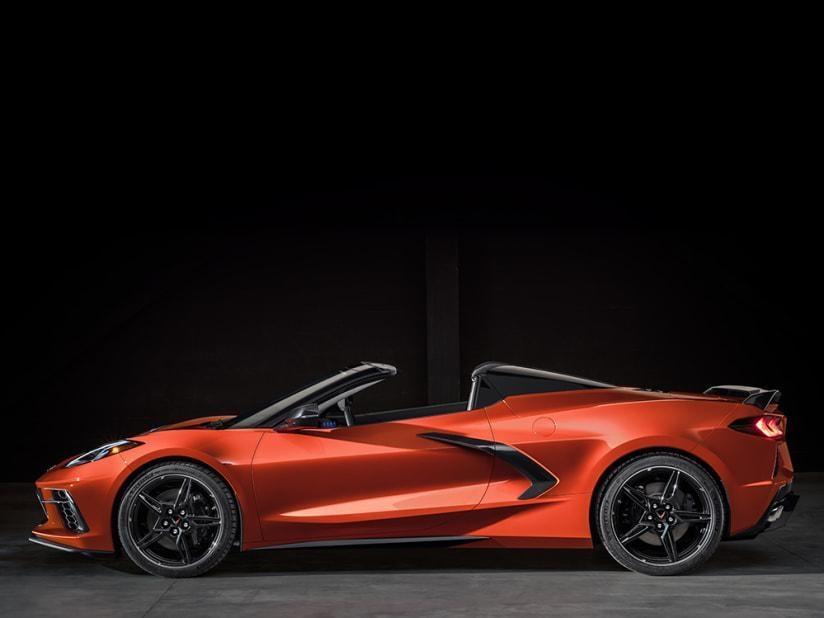}}\\[2pt]
    {\small Cluster~1}
  \end{minipage}
  \\[6pt]
  \begin{minipage}[t]{\textwidth}
    \centering
    \fcolorbox{green!60!black}{white}{\includegraphics[width=0.3\linewidth]{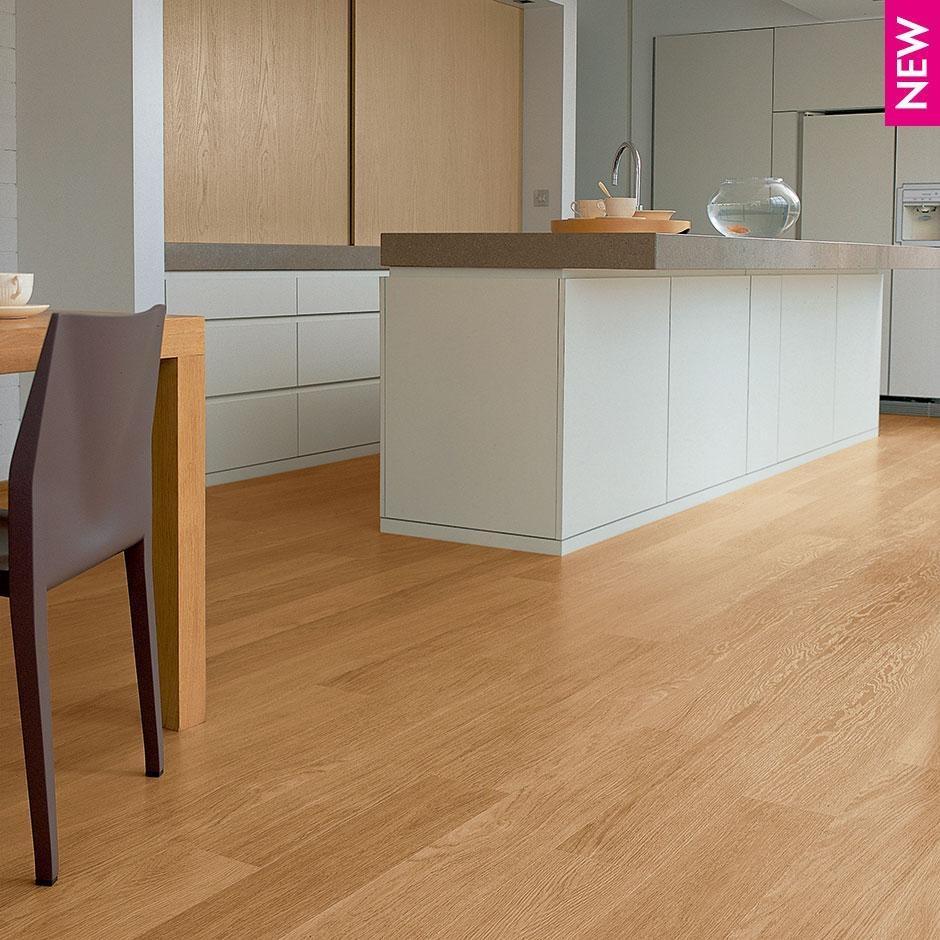}}\hfill
    \fcolorbox{white}{white}{\includegraphics[width=0.3\linewidth]{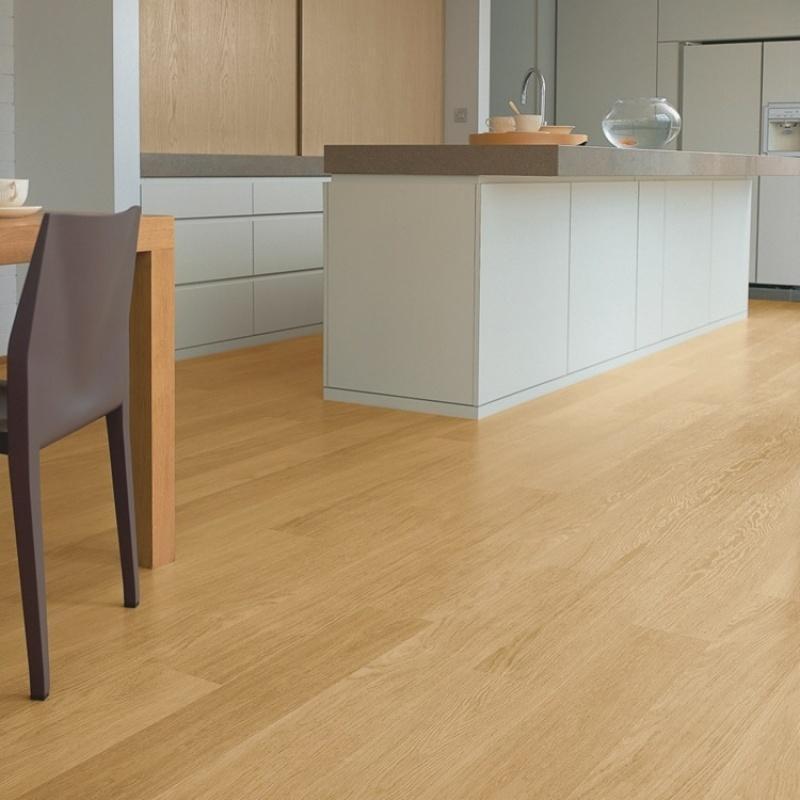}}\hfill
    \fcolorbox{white}{white}{\includegraphics[width=0.3\linewidth]{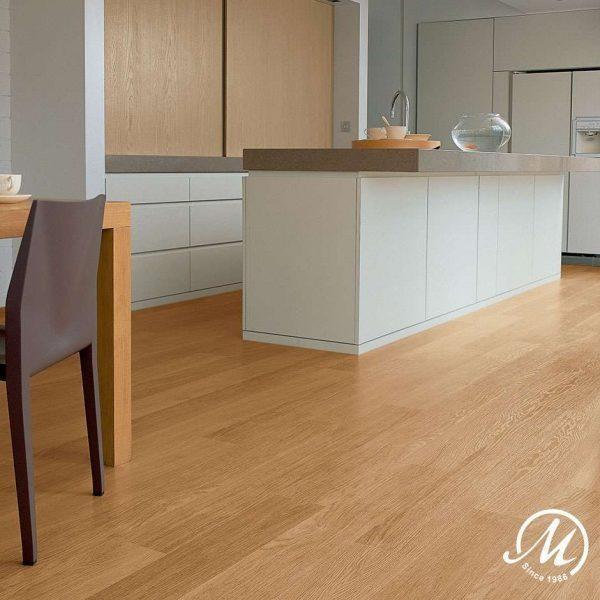}}\\[2pt]
    {\small Cluster~2}
  \end{minipage}
  \\[6pt]
  \begin{minipage}[t]{\textwidth}
    \centering
    \fcolorbox{green!60!black}{white}{\includegraphics[width=0.3\linewidth]{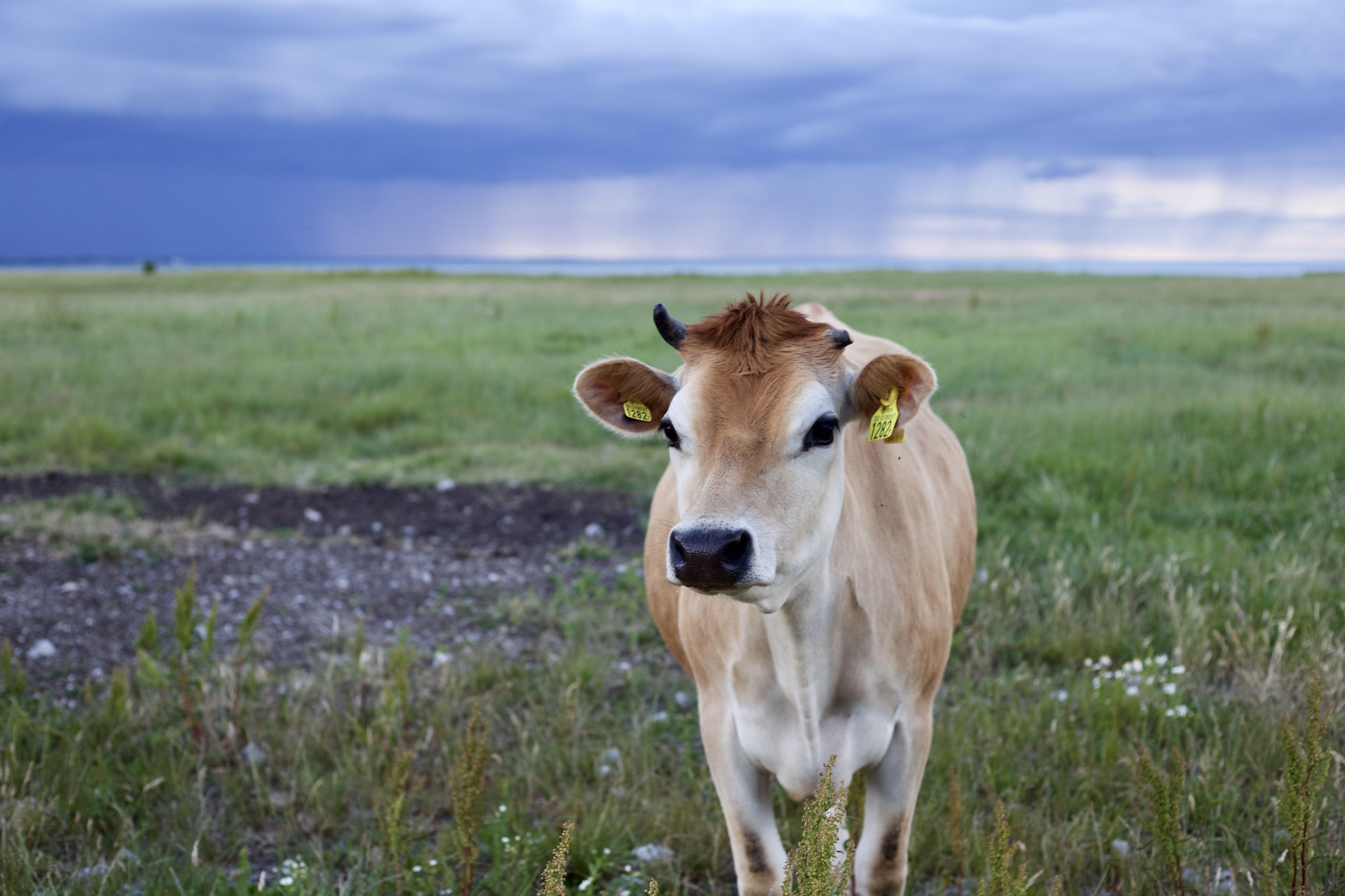}}\hfill
    \fcolorbox{white}{white}{\includegraphics[width=0.3\linewidth]{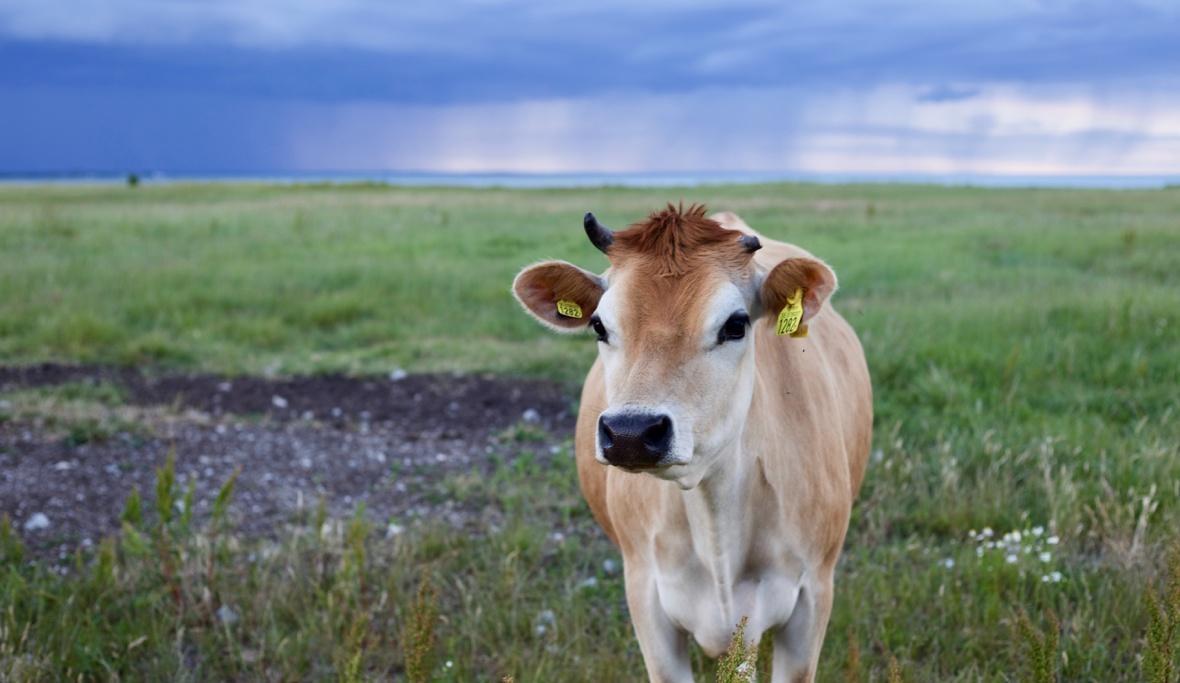}}\hfill
    \fcolorbox{white}{white}{\includegraphics[width=0.3\linewidth]{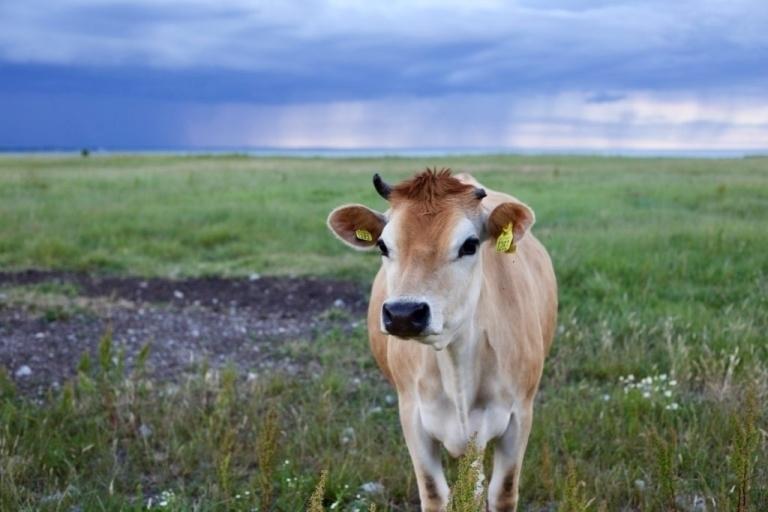}}\\[2pt]
    {\small Cluster~3}
  \end{minipage}
  \\[6pt]
  \begin{minipage}[t]{\textwidth}
    \centering
    \fcolorbox{green!60!black}{white}{\includegraphics[width=0.3\linewidth]{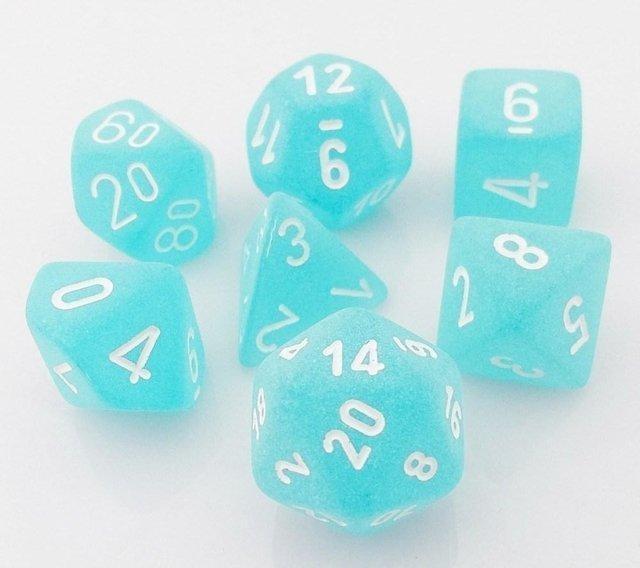}}\hfill
    \fcolorbox{white}{white}{\includegraphics[width=0.3\linewidth]{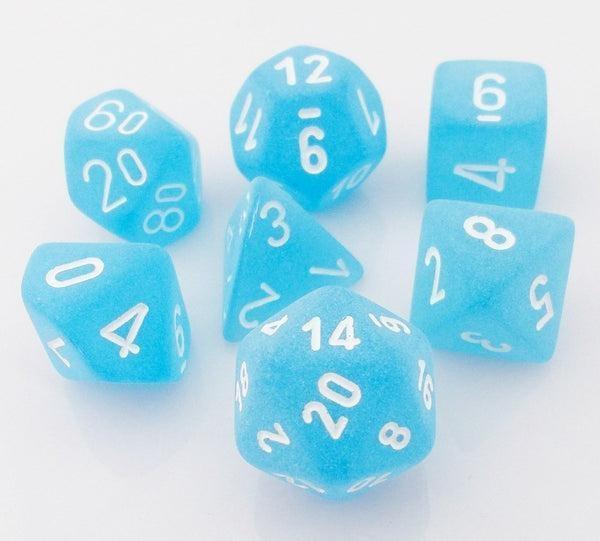}}\hfill
    \fcolorbox{white}{white}{\includegraphics[width=0.3\linewidth]{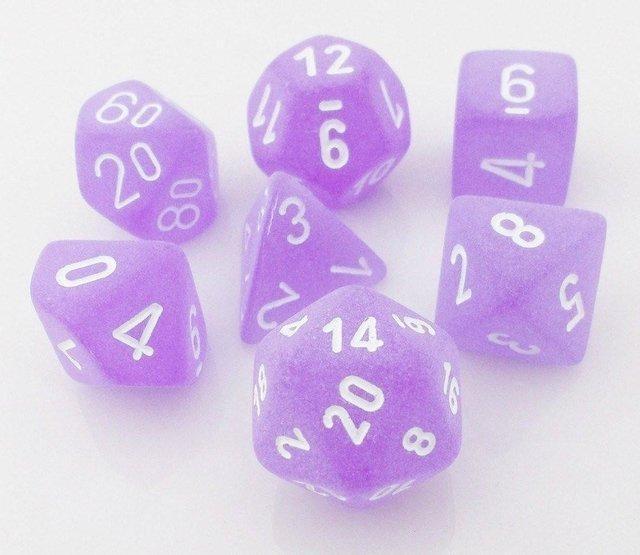}}\\[2pt]
    {\small Cluster~4}
  \end{minipage}
  \\[6pt]
  \begin{minipage}[t]{\textwidth}
    \centering
    \fcolorbox{green!60!black}{white}{\includegraphics[width=0.3\linewidth]{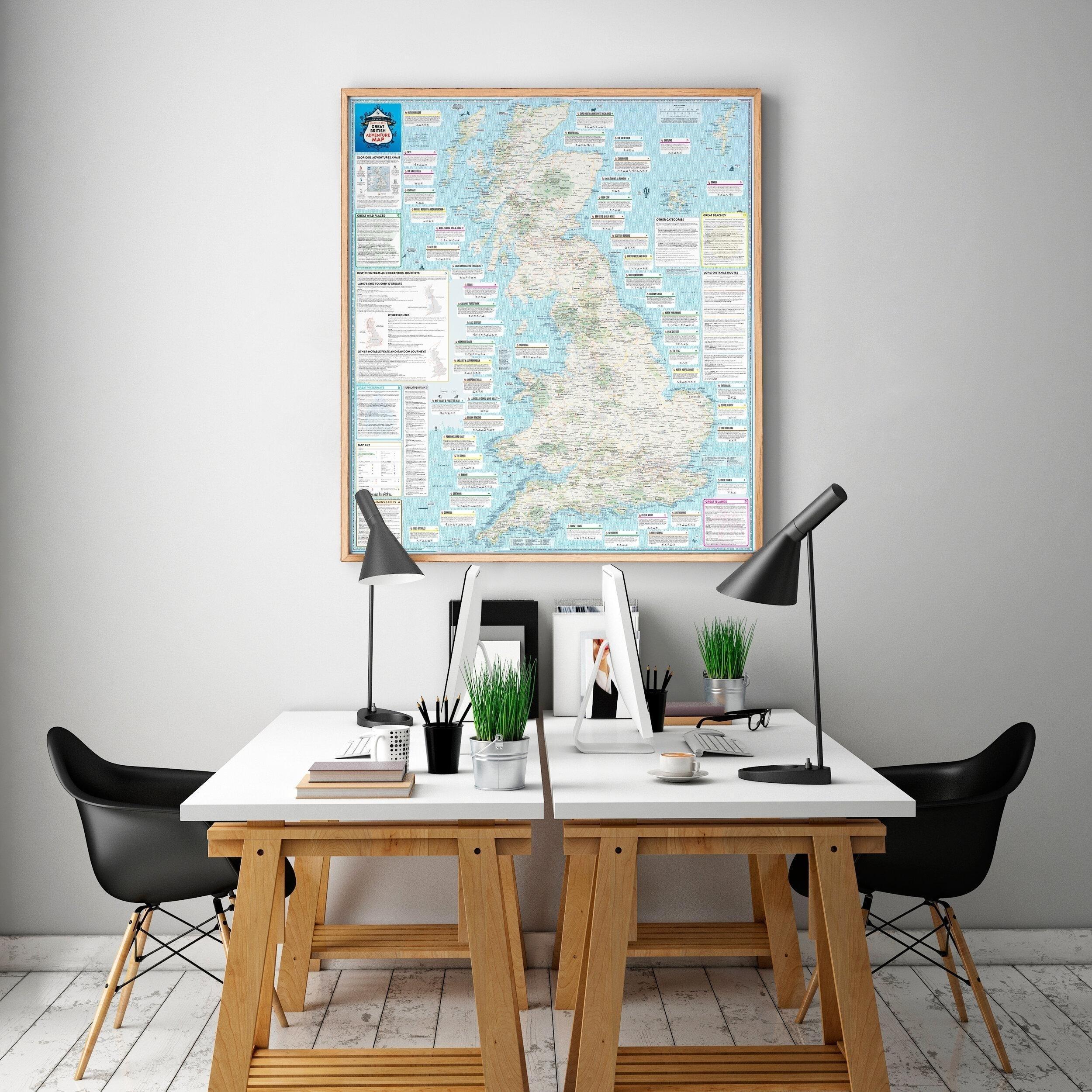}}\hfill
    \fcolorbox{white}{white}{\includegraphics[width=0.3\linewidth]{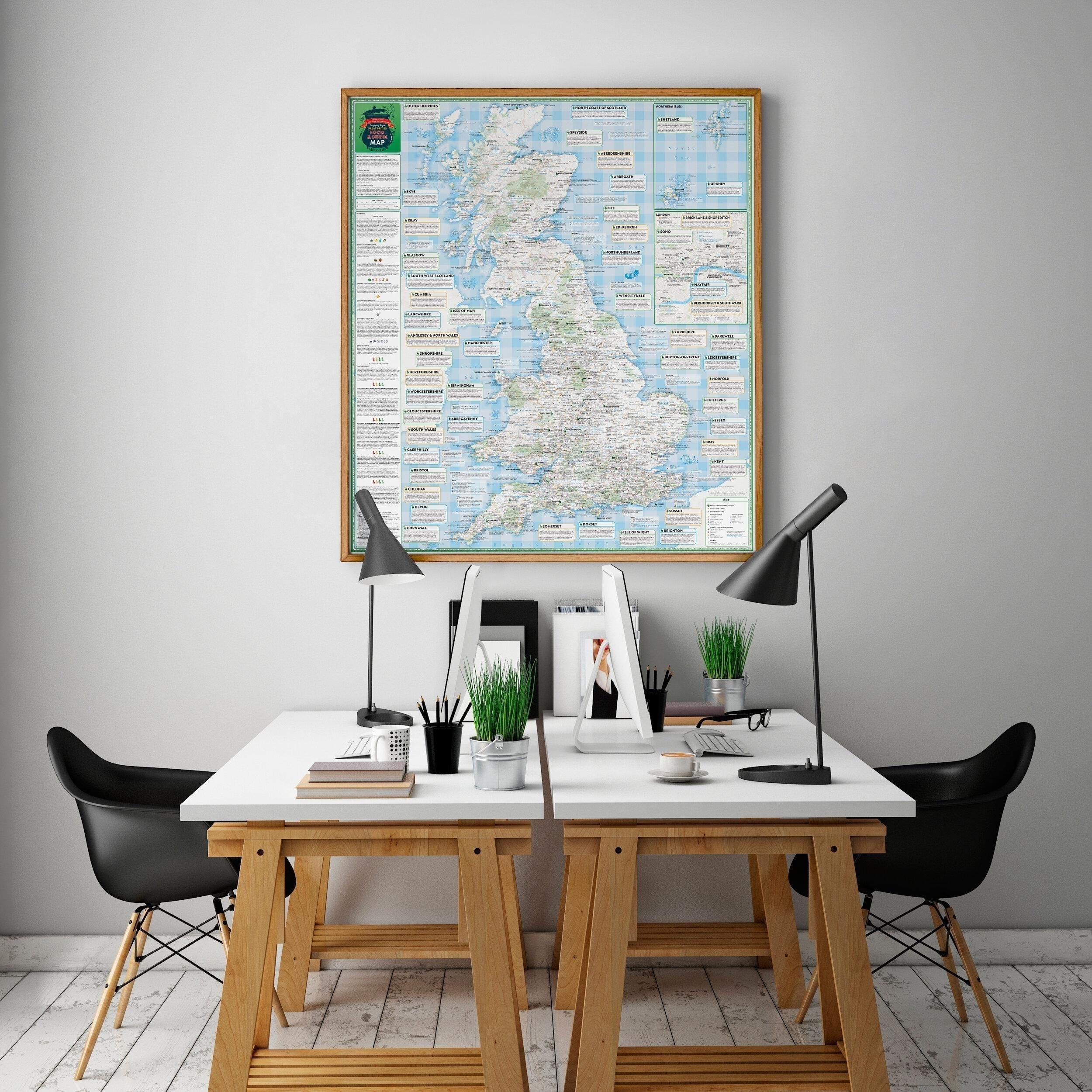}}\hfill
    \fcolorbox{white}{white}{\includegraphics[width=0.3\linewidth]{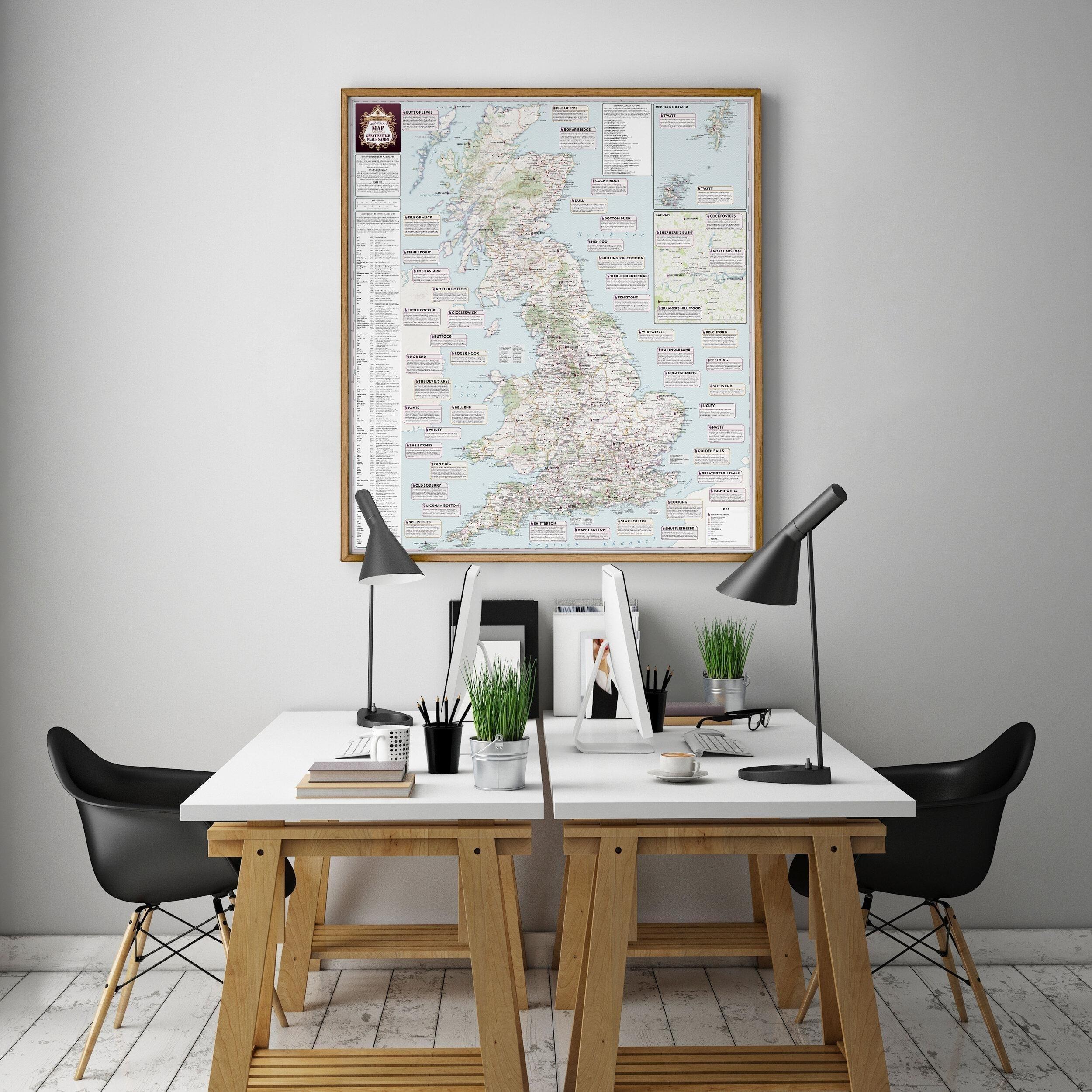}}\\[2pt]
    {\small Cluster~5}
  \end{minipage}
  \caption{Examples of near-duplicate clusters detected by SSCD. Each row shows one cluster; the image with a \textcolor{green!60!black}{green border} is the representative we keep in the dataset, while the remaining images are discarded as duplicates.}
  \label{fig:sscd_clusters}
\end{figure}

Fig.~\ref{fig:sscd_clusters} shows five representative clusters detected by SSCD that pHash misses. The image with a \textcolor{green!60!black}{green border} in each row is the representative kept in the final dataset; the others are discarded. The clusters illustrate the typical failure modes of pHash that SSCD is able to recover: cropping and re-framing of the same scene, watermark or logo overlays, color and exposure adjustments, and partial background edits. Because SSCD operates on semantic image content rather than raw spatial frequencies, it groups these variants together while remaining selective enough to leave visually distinct images untouched.

\newcommand{\keptimg}[2][0.15\linewidth]{%
  \fcolorbox{green!60!black}{white}{\includegraphics[height=#1]{#2}}%
}
\newcommand{\dupimg}[2][0.15\linewidth]{%
  \fcolorbox{white}{white}{\includegraphics[height=#1]{#2}}%
}

\begin{figure}[t]
  \centering
    \centering
    \includegraphics[width=0.48\linewidth]{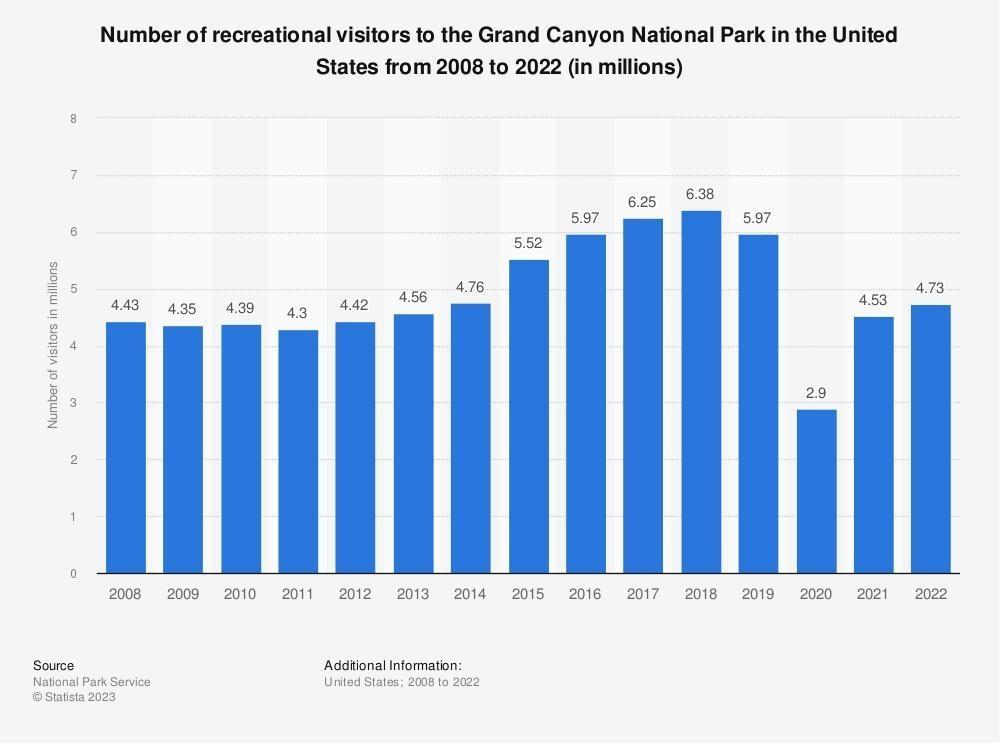}\hfill
    \includegraphics[width=0.48\linewidth]{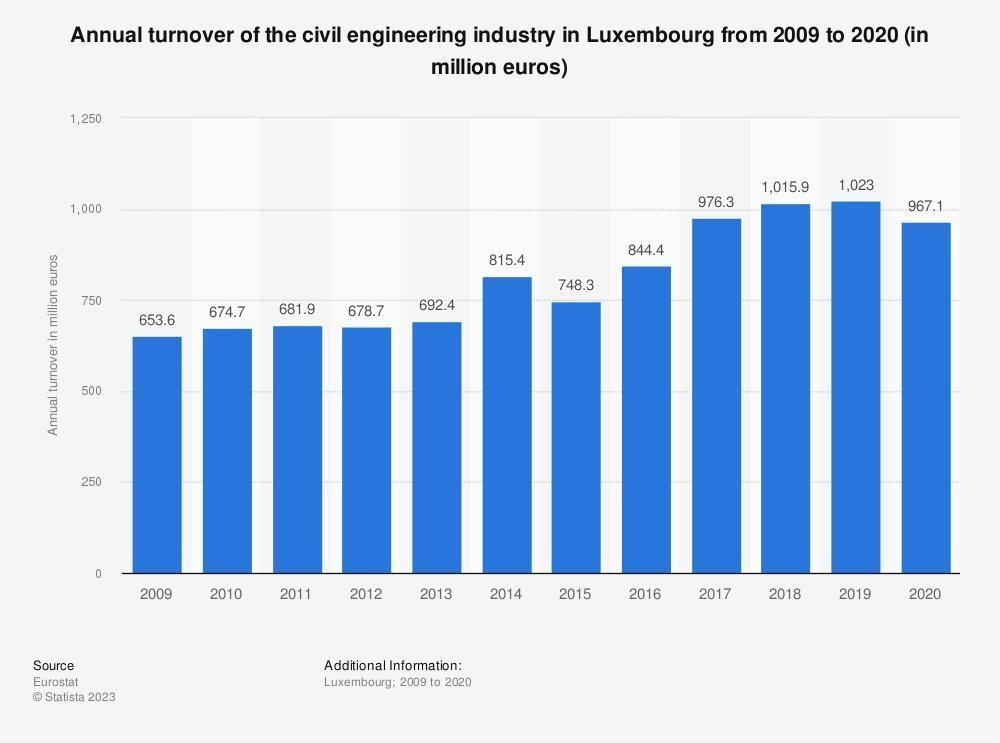}\\[2pt]
    {\small SSCD\,=\,0.92,\; $d$\,=\,20}\\
    
    \vspace{1em}
    \includegraphics[width=0.48\linewidth]{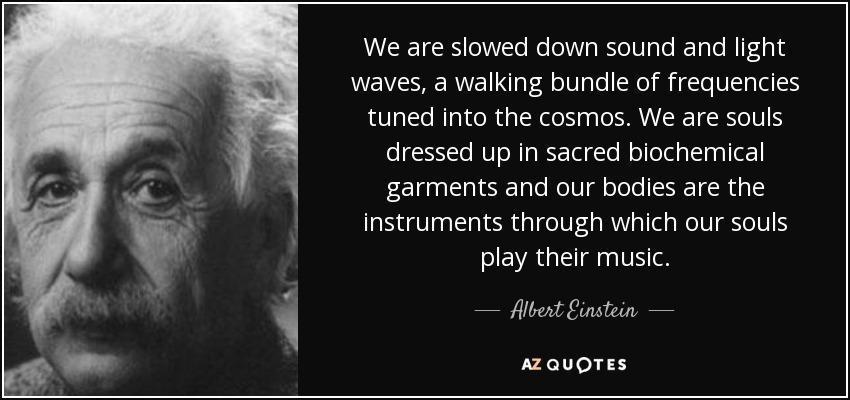}\hfill
    \includegraphics[width=0.48\linewidth]{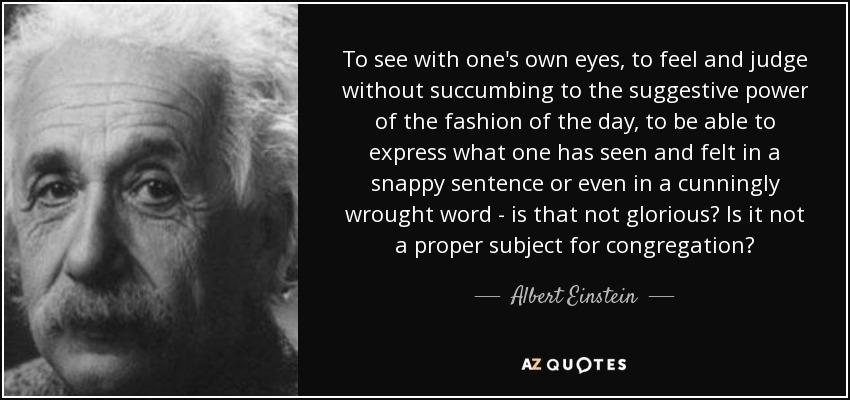}\\[2pt]
    {\small SSCD\,=\,0.91,\; $d$\,=\,4}
  \caption{SSCD false positives on template-based content. \emph{Top:} two unrelated bar charts sharing the same visual template receive SSCD\,=\,0.92 despite completely different data; pHash correctly assigns $d\!=\!20$. \emph{Bottom:} two quote images with identical portrait but different text are flagged by both SSCD (0.91) and pHash ($d\!=\!4$).}
  \label{fig:dedup_limitations_sscd}
\end{figure}

\paragraph{Limitations: false positives on template-based content}
SSCD embeddings capture mid-level visual structure but are largely insensitive to textual content rendered within images. Fig.~\ref{fig:dedup_limitations_sscd} illustrates this with two pairs, bar charts and quotes generated from identical templates but reporting entirely unrelated content. SSCD assigns them cosine similarities of 0.92/0.91, well above our 0.75 removal threshold, even though the underlying data and titles differ completely. Discarding such pairs is counterproductive, as they would otherwise help the T2I model learn to render text. Addressing this limitation, \emph{e.g.,} through OCR-aware deduplication for text-heavy images or content-aware hashing conditioned on semantic features rather than raw spatial frequencies, is an important direction for future work.

\newpage
\subsection{Re-captioning with VLMs}
\label{sec:caption-samples-and-alignment}

\subsubsection{Captioning models and prompts}
\label{sec:captioning-models-and-prompts}

VLM captioners were prompted with minimal, model-appropriate instructions to elicit each model's default captioning behavior, with no in-context examples or formatting constraints beyond what is noted below. Specifically, \textcolor{capFlorence}{\textbf{Florence-2-Large}}\footnote{\url{https://huggingface.co/microsoft/Florence-2-large}} was used in its built-in \texttt{<DETAILED\_CAPTION>} mode; \textcolor{capShareGPT}{\textbf{ShareGPT4V-7B}}\footnote{\url{https://huggingface.co/Lin-Chen/ShareGPT4V-7B}} and \textcolor{capInternVL}{\textbf{InternVL3-8B}}\footnote{\url{https://huggingface.co/OpenGVLab/InternVL3-8B-Instruct}} were prompted with \emph{``Describe this image''}; and \textcolor{capGemini}{\textbf{Gemini 2.5 Flash-Lite}}\footnote{\url{https://ai.google.dev/gemini-api/docs/models/gemini-2.5-flash-lite}} was prompted with \emph{``Describe this image in detail. Describe the main objects in the image, their relationships, the scene, the style. The caption should be in the language of the prompt without any line breaks or bullet points.''} This minimal prompt design was chosen to enable a fair comparison across captioners, since prompt phrasing can significantly affect output style, length, and content.

This section provides representative re-captioning examples from MONET with its \textcolor{capOriginal}{\textbf{original}} web caption and captions from \textcolor{capFlorence}{\textbf{Florence-2-Large}}, \textcolor{capShareGPT}{\textbf{ShareGPT4V-7B}}, \textcolor{capGemini}{\textbf{Gemini 2.5 Flash-Lite}} and \textcolor{capInternVL}{\textbf{InternVL3-8B}}, see Figs.~\ref{fig:caption_samples_1}, \ref{fig:caption_samples_2}, \ref{fig:caption_samples_3}, \ref{fig:caption_samples_4}, \ref{fig:caption_samples_5} and \ref{fig:caption_samples_6}. Qualitatively, the \textcolor{capOriginal}{\textbf{original}} web captions are often unreliable, sometimes missing the image content entirely or describing unrelated context. \textcolor{capFlorence}{\textbf{Florence-2-Large}} produces the shortest captions while remaining accurate, \textcolor{capShareGPT}{\textbf{ShareGPT4V-7B}} and \textcolor{capInternVL}{\textbf{InternVL3-8B}} bring noticeable improvements in coverage and specificity, and \textcolor{capGemini}{\textbf{Gemini 2.5 Flash-Lite}} yields substantially more detailed descriptions of objects, relationships, scene, and style.
\clearpage

\begin{tcolorbox}[title=Re-captioning example (1/5)]
\begin{figure}[H]
  \centering
  \begin{minipage}[c]{0.55\textwidth}
    \centering
    \includegraphics[width=\linewidth, keepaspectratio]{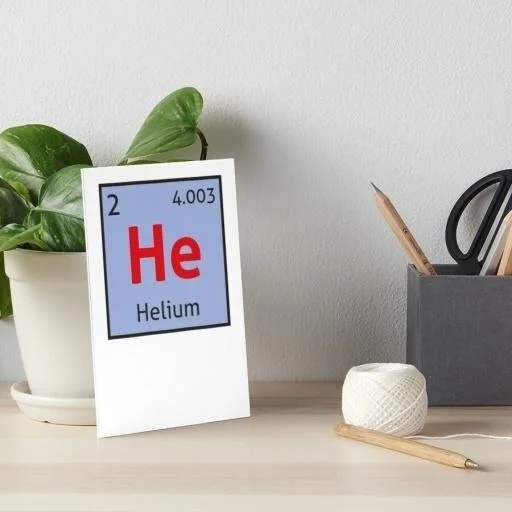}
  \end{minipage}%
  \hfill
  \begin{minipage}[c]{0.44\textwidth}
    \raggedright\footnotesize
    \textcolor{capOriginal}{\textbf{Original:}} The Periodic Table -- Helium Art Board Print\par\vspace{3pt}
    \textcolor{capFlorence}{\textbf{Florence-2-Large:}} The image shows a desk with a potted plant, a card with the element helium, a thread roll, a pencil, scissors, and a box of pencils placed on it. In the background, there is a wall.\par\vspace{3pt}
    \textcolor{capShareGPT}{\textbf{ShareGPT4V-7B:}} The image captures a serene workspace setup. Dominating the scene is a white desk, which hosts a variety of objects. On the left, a potted plant adds a touch of greenery to the setting. Adjacent to the plant, a white card stands out with its blue background and red text. The card is adorned with the atomic number 2 and the symbol ``He'' for helium, indicating a scientific theme. In the center of the desk, a gray pencil holder holds a collection of pencils, ready for use.\par\vspace{3pt}
    \textcolor{capGemini}{\textbf{Gemini 2.5 Flash-Lite:}} A white card with a blue box representing the element Helium from the periodic table stands on a wooden surface. The card shows the atomic number 2, the symbol He in red, the atomic weight 4.003, and the name Helium in black text. To the left of the card, a potted plant with large green leaves is visible. To the right of the card, a grey box holds pencils and scissors, and a spool of white twine and a wooden pencil lie on the surface. The scene is brightly lit with a soft, natural light, creating a clean and organised aesthetic with a touch of science and crafting.\par\vspace{3pt}
    \textcolor{capInternVL}{\textbf{InternVL3-8B:}} The image shows a desk with a notecard displaying the element helium from the periodic table. Next to the notecard is a potted plant, a container holding pencils and scissors, and a ball of yarn with a needle. The background is a plain wall. The notecard shows features the atomic number, atomic mass, and the chemical symbol and name of the element helium.
  \end{minipage}
  \caption{Re-captioning example (1/5)}
  \label{fig:caption_samples_2}
\end{figure}
\end{tcolorbox}

\begin{tcolorbox}[title=Re-captioning example (2/5)]
\begin{figure}[H]
  \centering
  \begin{minipage}[c]{0.55\textwidth}
    \centering
    \includegraphics[width=\linewidth, keepaspectratio]{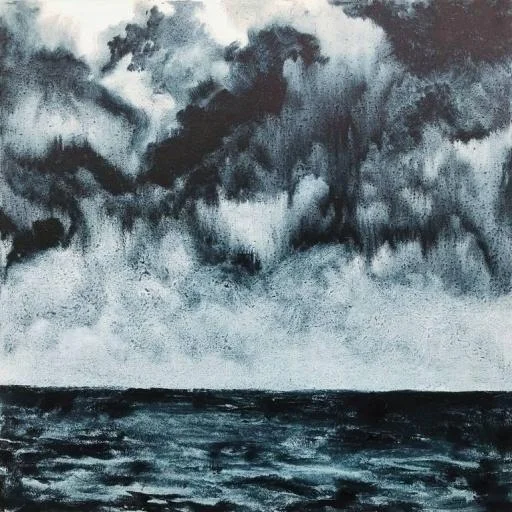}
  \end{minipage}%
  \hfill
  \begin{minipage}[c]{0.44\textwidth}
    \raggedright\footnotesize
    \textcolor{capOriginal}{\textbf{Original:}} Stormy Seas off the West Coast\par\vspace{3pt}
    \textcolor{capFlorence}{\textbf{Florence-2-Large:}} The image shows a black and white painting of a stormy sky over the ocean. The sky is filled with dark, ominous clouds, and the water below is choppy and turbulent.\par\vspace{3pt}
    \textcolor{capShareGPT}{\textbf{ShareGPT4V-7B:}} The image presents an abstract representation of a stormy sea under a dark, cloudy sky. The sea is depicted in a dark blue color, with white waves that add a sense of movement and turbulence. The sky, filled with thick, dark clouds, is rendered in a lighter shade of blue, creating a stark contrast with the sea below. The art style is abstract, characterized by loose brushstrokes that emphasize the interplay of light and shadow. The genre of the artwork is landscape, as it captures the natural elements of the sea and sky.\par\vspace{3pt}
    \textcolor{capGemini}{\textbf{Gemini 2.5 Flash-Lite:}} This is a monochromatic abstract painting depicting a turbulent sea beneath a dramatic, cloudy sky. The main objects are the sea and the sky, with the horizon line serving as their distinct separation. The sea occupies the lower third of the canvas and is rendered with dark, choppy brushstrokes, suggesting rough waves and an agitated surface. The color is a deep, almost black blue, with lighter tones highlighting the crests and troughs of the waves. Above the sea, the sky dominates the upper two-thirds of the painting. It is filled with swirling, dense clouds rendered in varying shades of dark gray, black, and white. The clouds appear heavy and ominous, with streaks and washes of paint creating a sense of movement and depth, perhaps indicating an approaching storm. The style is abstract and expressive, characterized by a loose, painterly application of pigment. The artist uses a limited palette of blues and grays, and the textured brushwork creates a sense of raw energy and emotion. The overall scene evokes a feeling of atmospheric drama and the power of nature.\par\vspace{3pt}
    \textcolor{capInternVL}{\textbf{InternVL3-8B:}} The image is a piece of art depicting a seascape under a dramatic sky. The scene shows in the painting predominantly features the ocean and the sky, with dark, heavy clouds suggesting an impending storm. The colors are predominantly shades of dark blue, black, and gray, creating a moody and turbulent atmosphere. The ocean waves are depicted with dynamic brushstrokes, enhancing the sense of movement and unrest in the scene. There is a clear division between the sea and the sky, with the horizon line situated towards the lower third of the composition, emphasizing the vastness of the sky over the sea.
  \end{minipage}
  \caption{Re-captioning example (2/5)}
  \label{fig:caption_samples_3}
\end{figure}
\end{tcolorbox}

\begin{tcolorbox}[title=Re-captioning example (3/5)]
\begin{figure}[H]
  \centering

  \begin{minipage}[c]{0.55\textwidth}
    \centering
    \includegraphics[width=\linewidth, keepaspectratio]{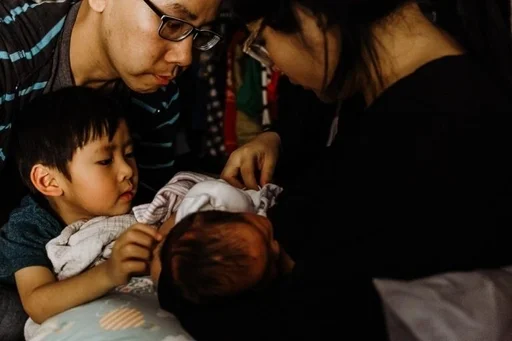}
  \end{minipage}%
  \hfill
  \begin{minipage}[c]{0.44\textwidth}
    \raggedright\footnotesize
    \textcolor{capOriginal}{\textbf{Original:}} A baby nurses while dad and brother look on.\par\vspace{3pt}
    \textcolor{capFlorence}{\textbf{Florence-2-Large:}} The image shows a man and woman holding a newborn baby in their arms, with a cloth draped over them both. In the background, there are clothes, suggesting that the family is in a room. The man is likely the father, and the woman is likely his wife, as they are both looking lovingly at the baby.\par\vspace{3pt}
    \textcolor{capShareGPT}{\textbf{ShareGPT4V-7B:}} In the image, a family of three is gathered around a baby, creating a warm and intimate atmosphere. The baby, dressed in a white onesie, is the center of attention. The family members, all wearing glasses, are looking down at the baby with expressions of love and affection. The room they are in is dimly lit, adding to the cozy ambiance. The background is filled with clothes, suggesting a home setting. The family's focus on the baby indicates a special moment being shared. The image captures a precious family moment, filled with love and warmth.\par\vspace{3pt}
    \textcolor{capGemini}{\textbf{Gemini 2.5 Flash-Lite:}} A tender, intimate scene captures a family's love as an older sibling, a young boy, gently holds a newborn baby while their parents look on with affection. The boy, with his dark hair and wide eyes, is dressed in a blue t-shirt and is cradling the infant with one hand, his fingers lightly touching the baby's face. He is nestled against a light blue and white patterned pillow. The baby, swaddled in a pink and white striped blanket and a white cloth, is lying on its back, its head turned towards the viewer. On the left, the father, wearing a striped t-shirt and glasses, leans in, his expression one of gentle observation. On the right, the mother, wearing a black top, looks down at the baby, her face mostly obscured by shadow but her posture radiating care. The overall style of the image is soft and moody, with a shallow depth of field that blurs the background and focuses attention on the family's interaction, creating a warm and emotional atmosphere.\par\vspace{3pt}
    \textcolor{capInternVL}{\textbf{InternVL3-8B:}} The image shows a family of three, including a baby, gathered closely together. One person, wearing glasses and a striped shirt, is leaning in towards the baby, who is being held by another person in a dark top. The setting appears to be indoors, with soft lighting creating a warm and intimate atmosphere. The family seems to be sharing a tender moment together.
  \end{minipage}
  \caption{Re-captioning example (3/5)}
  \label{fig:caption_samples_4}
\end{figure}
\end{tcolorbox}

\begin{tcolorbox}[title=Re-captioning example (4/5)]
\begin{figure}[H]
  \centering
  \begin{minipage}[c]{0.55\textwidth}
    \centering
    \includegraphics[width=\linewidth, keepaspectratio]{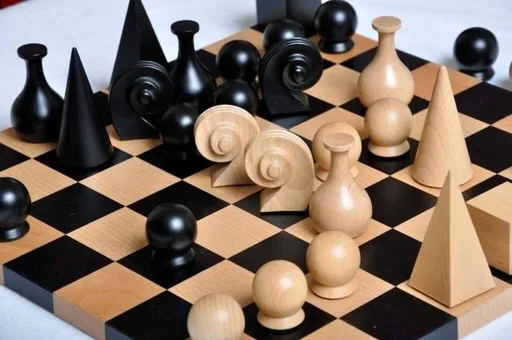}
  \end{minipage}%
  \hfill
  \begin{minipage}[c]{0.44\textwidth}
    \raggedright\footnotesize
    \textcolor{capOriginal}{\textbf{Original:}} \textlangle PERSON\textrangle{} designed this chess set in 1920 using found objects from his studio. Most of the pieces are geometric models made for still-life drawing (the king is a pyramid, the queen a cone, the rook a cube, the pawn a sphere). \textlangle PERSON\textrangle, Chess Pieces For Sale, Still Life Drawing, Geometric Form, Contemporary Art, Art Pieces, Elegant, King, Lathe\par\vspace{3pt}
    \textcolor{capFlorence}{\textbf{Florence-2-Large:}} The image shows a wooden chess board with black and white chess pieces on it, set against a white background. The board is made of wood and the pieces are arranged in a traditional chess formation, with the black pieces on one side and the white pieces on the other.\par\vspace{3pt}
    \textcolor{capShareGPT}{\textbf{ShareGPT4V-7B:}} The image presents a captivating scene of a chess game in progress. The chessboard, with its alternating black and white squares, serves as the stage for this strategic battle. On the black side of the board, there are 12 pieces, each unique in shape and color. They are arranged in a way that suggests a game in progress. The white side of the board is adorned with 11 pieces, mirroring the black side in their arrangement. The pieces on the black side are predominantly black and brown, with a few white and beige ones interspersed.\par\vspace{3pt}
    \textcolor{capGemini}{\textbf{Gemini 2.5 Flash-Lite:}} A close-up, low-angle shot showcases a modern, minimalist chess set on a wooden chessboard. The board itself is composed of alternating light wood and black squares, with the edges rendered in a natural wood tone. The chess pieces are a striking contrast of matte black and natural, unvarnished wood. The black pieces include a tall, sharp pyramid, several smooth, spherical pawns, a piece resembling a stylized vase with a flat top, and two elaborate, coiling pieces that evoke the form of a scroll or a ram's horn. The wooden pieces are also mostly spherical pawns, with one taller piece shaped like a rounded vase and another a sharp pyramid. The pieces are arranged in a seemingly random, yet strategically suggestive, pattern across the board, hinting at an ongoing game or a unique display. The style is contemporary and abstract, with a focus on geometric shapes and the interplay of light and shadow on the matte surfaces. The overall impression is one of elegant simplicity and artistic design.\par\vspace{3pt}
    \textcolor{capInternVL}{\textbf{InternVL3-8B:}} The image shows a chess set where the pieces have been replaced with various shapes and forms, yet they retain the original chess piece functions. The board itself features the traditional black and tan squares. Among the pieces, there are black pieces shaped like cones, spheres, and other geometric forms, as well as wooden pieces that resemble spirals, pyramids, and elongated ovals. The modified pieces maintain the classic chess piece roles despite their unconventional appearances.
  \end{minipage}
  \caption{Re-captioning example (4/5)}
  \label{fig:caption_samples_5}
\end{figure}
\end{tcolorbox}

\begin{tcolorbox}[title=Re-captioning example (5/5)]
\begin{figure}[H]
  \centering
  \begin{minipage}[c]{0.55\textwidth}
    \centering
    \includegraphics[width=\linewidth, keepaspectratio]{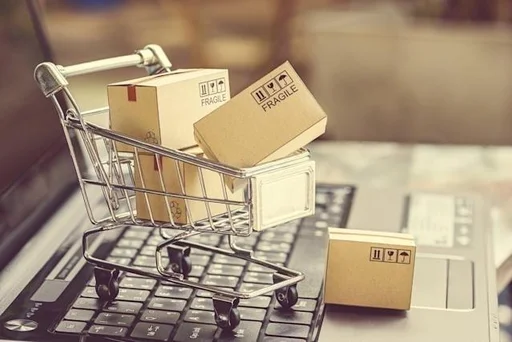}
  \end{minipage}%
  \hfill
  \begin{minipage}[c]{0.44\textwidth}
    \raggedright\footnotesize
    \textcolor{capOriginal}{\textbf{Original:}} Small shopping cart with boxes on top of a laptop keyboard\par\vspace{3pt}
    \textcolor{capFlorence}{\textbf{Florence-2-Large:}} The image shows a laptop with a shopping cart filled with cardboard boxes sitting on top of it, suggesting that the laptop is being used to help with e-commerce. The background is slightly blurred, giving the image a sense of focus on the laptop and the cart.\par\vspace{3pt}
    \textcolor{capShareGPT}{\textbf{ShareGPT4V-7B:}} The image presents a scene of online shopping. At the center of the image, a silver shopping cart is filled with three cardboard boxes. The boxes are labeled with the words ``FRAGILE'', indicating that the items inside are likely delicate and should be handled with care. The cart is placed on a black keyboard, suggesting that the shopping is being done from a computer. To the right of the cart, there's a small cardboard box with the word ``FRAGILE'' written on it, reinforcing the theme of fragile items.\par\vspace{3pt}
    \textcolor{capGemini}{\textbf{Gemini 2.5 Flash-Lite:}} A miniature shopping cart, filled with small cardboard boxes, rests on the keyboard of an open laptop. The boxes, with ``FRAGILE'' printed on them along with shipping symbols, suggest an online shopping or delivery theme. The scene is captured with a shallow depth of field, blurring the background and drawing focus to the main objects. The style is somewhat vintage and warm, possibly achieved through editing filters, evoking a sense of nostalgia or a handcrafted aesthetic.\par\vspace{3pt}
    \textcolor{capInternVL}{\textbf{InternVL3-8B:}} The image depicts a shopping cart filled with cardboard boxes labeled ``FRAGILE.'' The cart is positioned on top of a laptop keyboard, symbolizing online shopping or delivery services. Additionally, one cardboard box is placed on the laptop's touchpad area, further emphasizing the theme of e-commerce and logistics. The overall scene suggests themes of online purchases, fragility of goods, and possibly the impact on the user experience of shopping from a laptop.
  \end{minipage}
  \caption{Re-captioning example (5/5)}
  \label{fig:caption_samples_6}
\end{figure}
\end{tcolorbox}
\clearpage

\subsubsection{Human quality assessment}
\label{sec:human-quality-assessment}

To complement the automatic image--text alignment of Sec.~\ref{sec:dataset-analysis}, we ran a pairwise human study over 5000 images. Annotators were shown one image and two captions sampled at random from the five captioners (\textcolor{capOriginal}{\textbf{original}}, \textcolor{capFlorence}{\textbf{Florence-2-Large}}, \textcolor{capShareGPT}{\textbf{ShareGPT4V-7B}}, \textcolor{capGemini}{\textbf{Gemini 2.5 Flash-Lite}}, \textcolor{capInternVL}{\textbf{InternVL3-8B}}) with shuffled left/right order, and asked to vote \emph{Neither}, \emph{One is better}, or \emph{Both are good}. Votes were aggregated into a per-captioner Elo score (initialized at $1500$, draws for \emph{Both are good}, \emph{Neither} votes discarded). The full instructions provided to annotators are reproduced verbatim below.

\begin{tcolorbox}[title=Instructions provided to human annotators,breakable]
\footnotesize
In this task, you will be shown \textbf{one image alongside two captions}. Your goal is to judge which caption best describes the image. For each pair, select one of:
\begin{itemize}
    \setlength{\itemsep}{1pt}
    \item \textbf{Neither} -- neither caption accurately describes the image.
    \item \textbf{One is better} -- one caption is clearly more accurate than the other.
    \item \textbf{Both are good} -- both captions describe the image equally well.
\end{itemize}

When making your choice, pay attention to:
\begin{itemize}
    \setlength{\itemsep}{1pt}
    \item \textbf{Accuracy} -- does the caption faithfully describe what is visible in the image?
    \item \textbf{Hallucinations} -- does it mention elements not present in the image? Be especially careful with longer captions, which may sound detailed but include fabricated content.
    \item \textbf{Completeness} -- does it cover the key elements (subject, action, setting)?
    \item \textbf{Specificity} -- does it use precise rather than vague or generic language?
    \item \textbf{Objectivity} -- does it stick to what is visually observable, without subjective interpretation?
\end{itemize}
\end{tcolorbox}

\paragraph{Compensation.} Annotators were compensated at \$6/hour, above the average hourly wage of $\sim$\$1.5 in the Philippines, the country where the study was conducted. The total budget was capped at \$300 ($\sim$50 hours), so the annotators were instructed to stop once their time allowance was reached rather than voting on the full pool. Sign-in was used only to deduplicate votes and was discarded after Elo aggregation.

\begin{figure}[H]
  \centering
  \includegraphics[width=\linewidth]{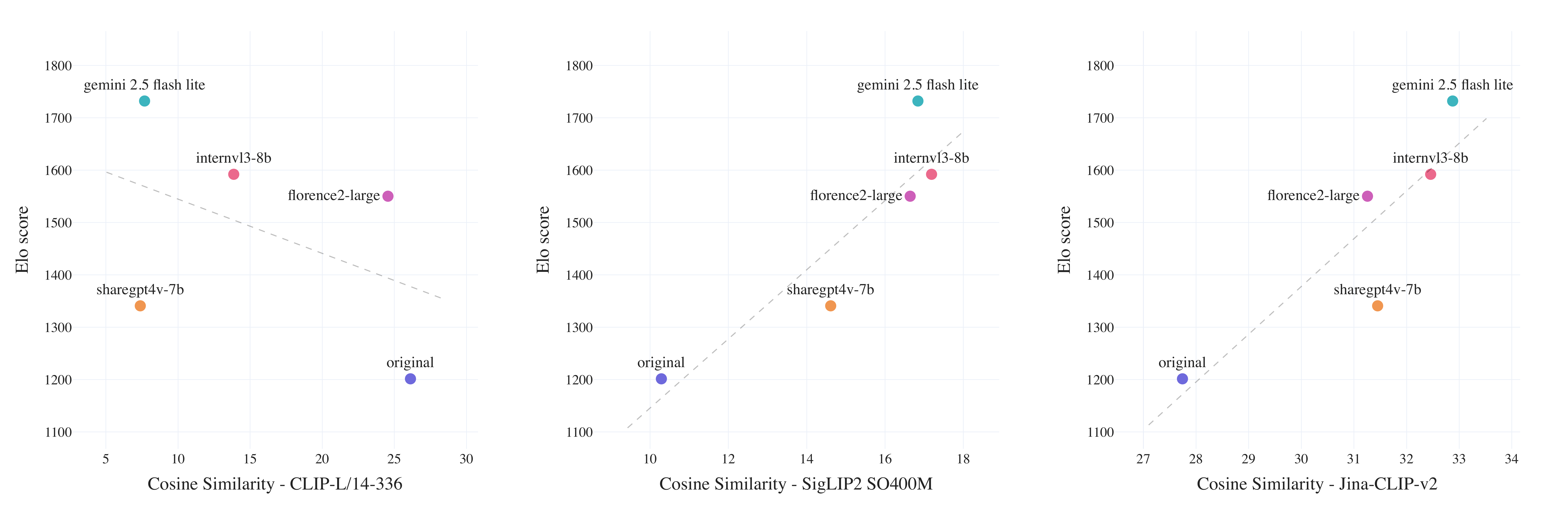}
  \caption{Human Elo scores aggregated from the pairwise voting study, plotted against the cosine similarity computed by CLIP-L/14-336 (left), SigLIP 2 (middle) and Jina-CLIP-v2 (right). The Elo ranking is consistent with LongCLIP, SigLIP 2 and Jina-CLIP-v2 cosine similarities, which support long-form captions, but is not well captured by CLIP-L/14-336 due to its $77$-token context window, which truncates the long captions produced by Gemini~2.5 Flash-Lite and ShareGPT4V-7B.}
  \label{fig:caption_image_alignment_all}
\end{figure}

\paragraph{Cross-encoder alignment with human preferences.} Fig.~\ref{fig:caption_image_alignment_all} reports the human Elo scores against the cosine similarity of three image--text encoders (\textbf{CLIP-L/14-336}~\citep{radford2021learning}, \textbf{SigLIP 2}~\citep{li2025siglip} and \textbf{Jina-CLIP-v2}~\citep{jinav2025jinav2}) complementing the \textbf{LongCLIP}~\citep{zhang2024long} alignment shown in Fig.~\ref{fig:text_image_alignment}. \textbf{LongCLIP}, \textbf{SigLIP 2} and \textbf{Jina-CLIP-v2} are consistent with the human ranking: the longer captions from \textcolor{capGemini}{\textbf{Gemini}} and \textcolor{capInternVL}{\textbf{InternVL3-8B}} obtain both higher Elo and higher cosine similarity than the \textcolor{capOriginal}{\textbf{original}} and \textcolor{capFlorence}{\textbf{Florence-2-Large}} captions, in line with their extended $248$-token text context window. We do not draw conclusions from \textbf{CLIP-L/14-336}: its $77$-token context truncates the long Gemini and ShareGPT4V-7B captions, mechanically capping their similarity and making the metric unreliable for long-form captioners. We therefore recommend long-context encoders such as LongCLIP or Jina-CLIP-v2 for evaluating long-form re-captioning pipelines.

\newpage
\subsection{Details on image content and style classifications}
\label{sec:image-content-style-detailed}

\subsubsection{Image content distribution}
\label{sec:image-content-style-detailed-distribution}

\begin{figure}[ht]
    \centering
    \includegraphics[width=0.7\linewidth]{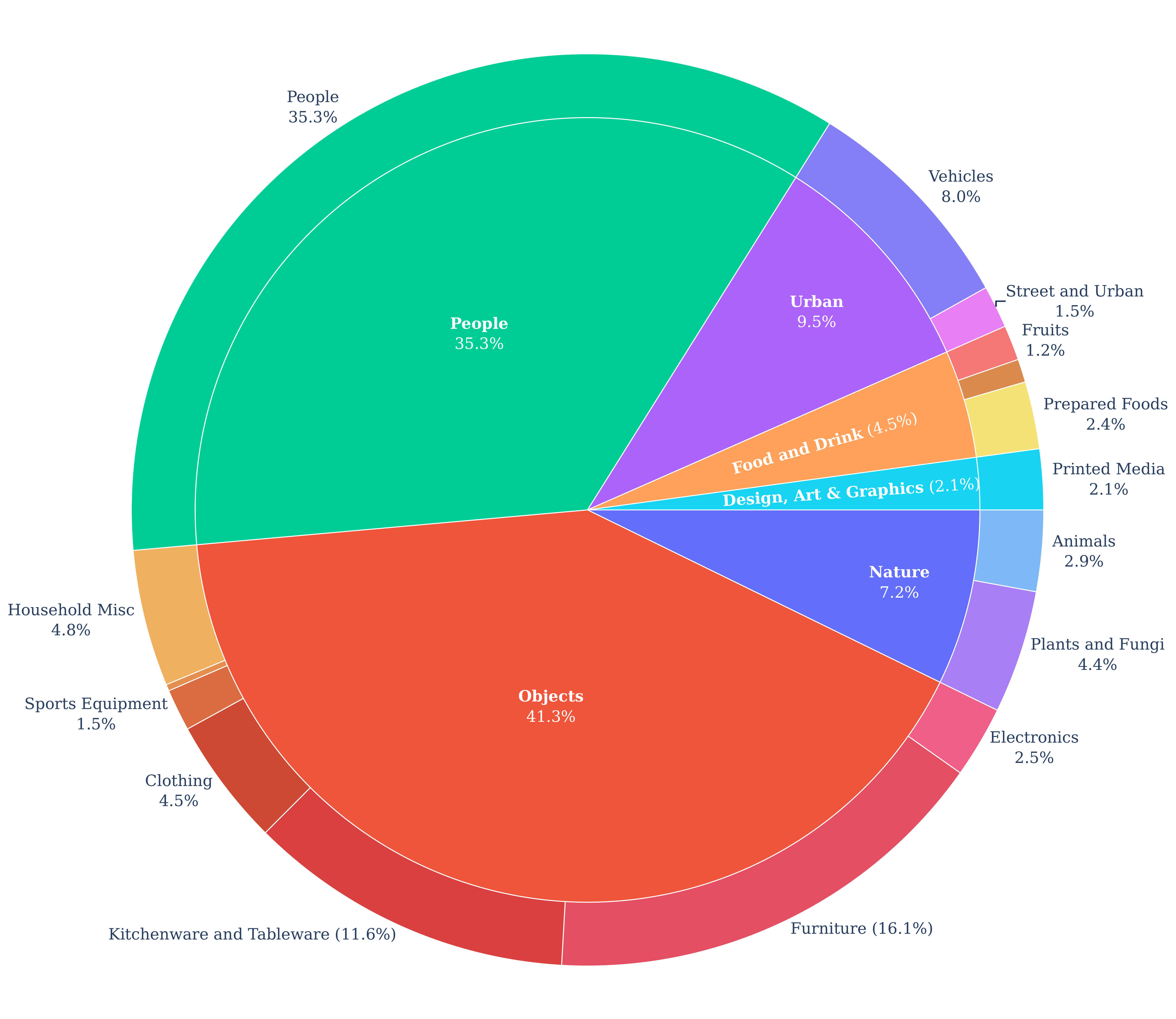}
    \caption{Hierarchical image content distribution (YOLO).}
    \label{fig:image-content-yolo-complete}
\end{figure}

\begin{figure}[ht]
    \centering
    \includegraphics[width=0.85\linewidth]{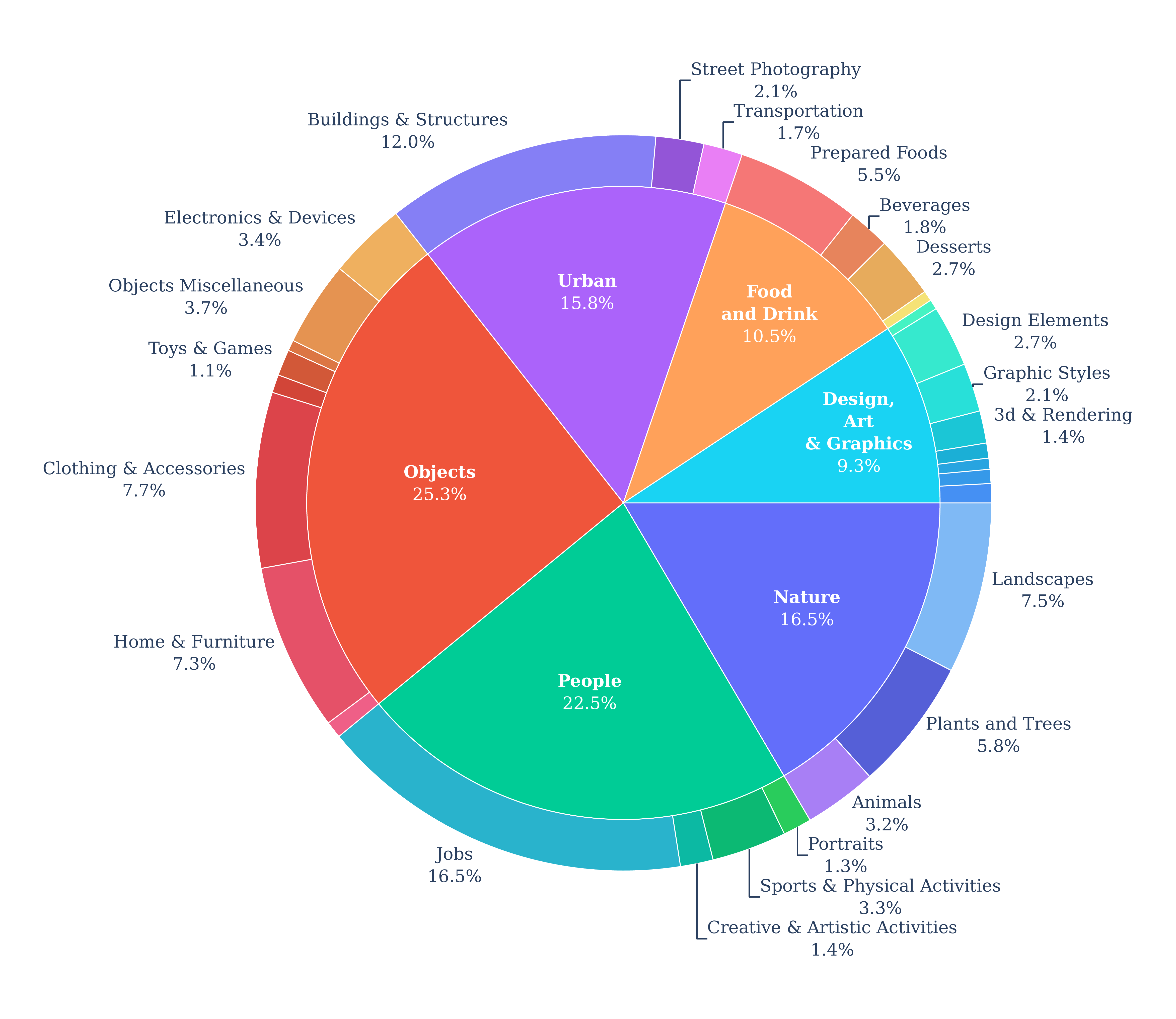}
    \caption{Hierarchical image content distribution (CLIP).}
    \label{fig:image-content-clip-complete}
\end{figure}

Figs.~\ref{fig:image-content-yolo-complete} and \ref{fig:image-content-clip-complete} show the detailed distribution of MONET's image content across the three hierarchical levels, using YOLO detection labels and CLIP-based classification, respectively. They show how individual classes are grouped into two higher hierarchy classes to produce the plots of Fig.~\ref{fig:image-content-and-style}.

\begin{figure}
    \centering
    \includegraphics[width=\linewidth]{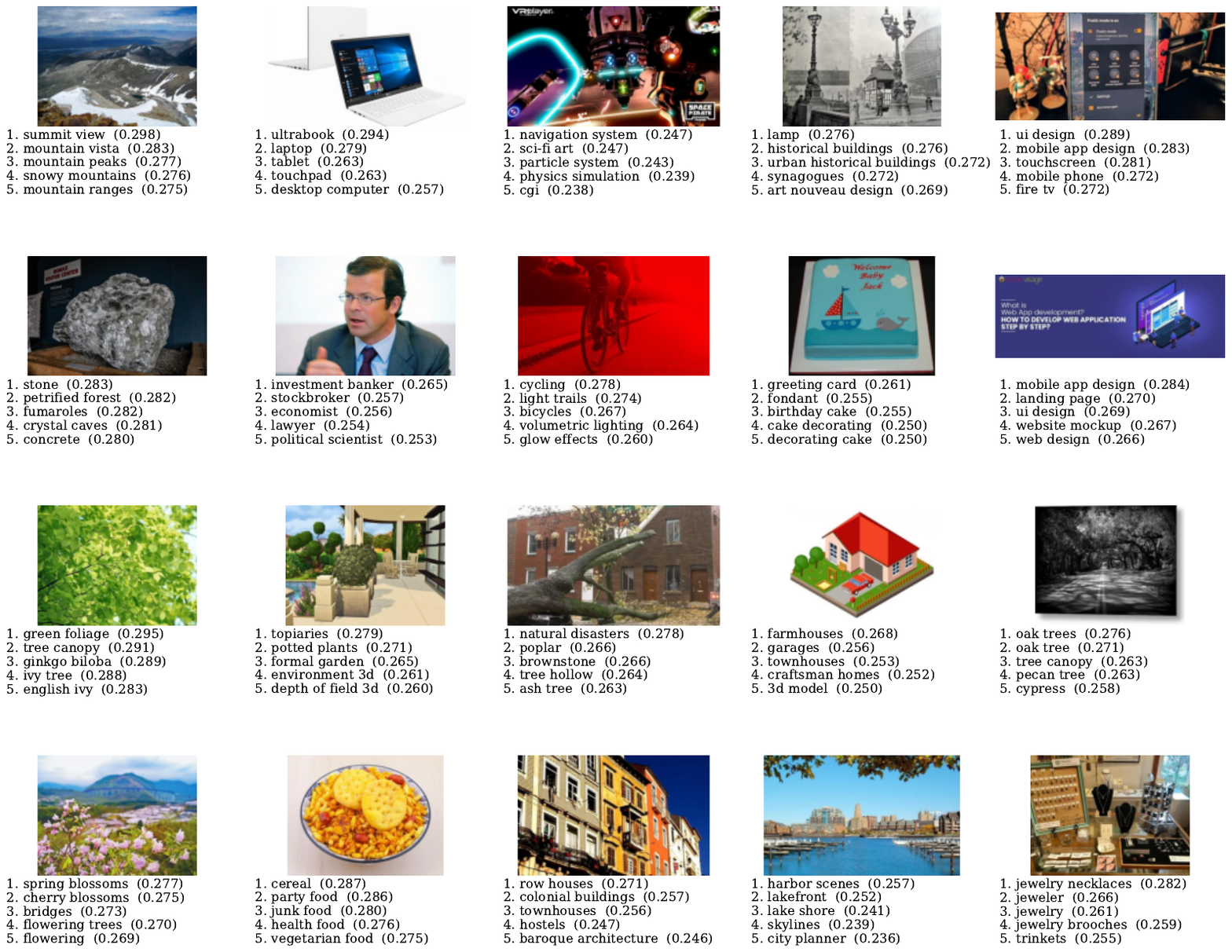}
    \caption{Examples of top-5 image content classification using CLIP with their similarity scores.}
    \label{fig:clip-classification-success}
\end{figure}

\begin{figure}
    \centering
    \includegraphics[width=\linewidth]{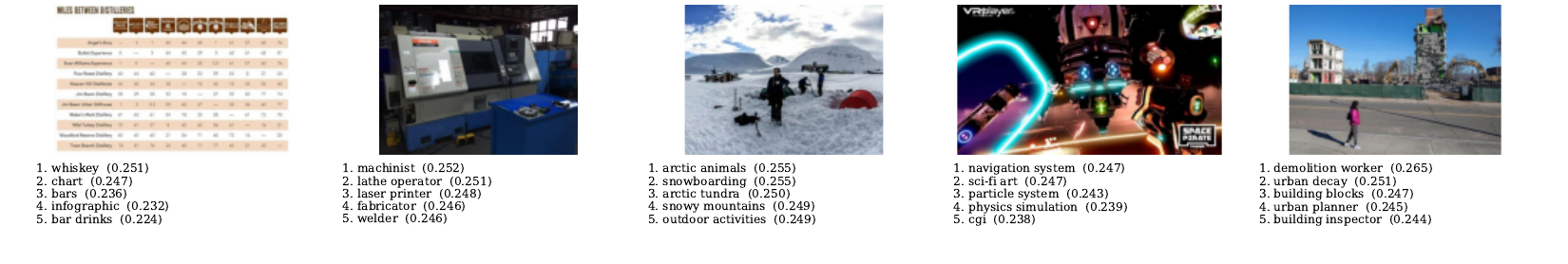}
    \caption{Examples of top-5 image content classifications using CLIP, including some incorrect or misleading classes.}
    \label{fig:clip-classification-fail}
\end{figure}

Fig.~\ref{fig:clip-classification-success} illustrates successful top-5 CLIP-based classifications on MONET images, demonstrating the model's ability to retrieve a wide variety of complex concepts using CLIP embeddings. Conversely, Fig.~\ref{fig:clip-classification-fail} showcases examples where not all top-5 labels are relevant. For instance, in the second image a ``machinist'' is predicted, while there is no human in the image, and in the third image ``arctic animals'' are predicted, while there are no visible animals in the picture. On the other hand, the first image depicts a limitation of the hierarchical classification: while predicting ``whiskey'' is correct, as the infographic is related to whiskeys, the ``whiskey'' class is subsequently associated with ``beverages'' and then with ``food and drink'', although the image should instead belong to ``design, art \& graphics''. In these failure cases, the top-1 result is frequently a false positive, which justifies our approach of evaluating the top-5 classes rather than relying solely on the top-1 result.

\clearpage

\subsubsection{Image style audit prompt and JSON schema}
\label{sec:image-style-classification-prompt}

To complement the content distribution above, we annotate every image with
a single image style label that captures \emph{how} the image was
produced rather than what it depicts. We initially explored CLIP-based zero-shot classification for this task, and found that CLIP embeddings were not reliable for distinguishing production-oriented style categories. In particular, they tend to mix visual content with how an image was produced, and they do not consistently capture fine-grained stylistic or medium-related cues. As a result, we process each image using Qwen3-VL-8B-Instruct~\citep{yang2025qwen3} with the prompt reproduced
below, which defines a 15-way taxonomy organized in three families
(photography, traditional / digital art, and utility / design) plus an
\texttt{other} escape hatch. The taxonomy is deliberately
production-oriented: photography labels are assigned from visible cues
(sensor grain, bokeh, lens geometry, lighting setup) rather than subject
matter, so that, for example, a photograph of a painting and the painting
itself receive different labels.
Fig.~\ref{fig:image-style-examples} illustrates the resulting categories
with two representative thumbnails per label sampled from the audit
preview, while Fig.~\ref{fig:image-style-complete} reports
the full distribution over MONET.

\begin{promptbox}[label={prompt:vlm-image-style}]{Image style audit prompt.}
\textbf{Picture-style audit prompt and JSON schema.}

A 15-way classification over \emph{how an image was produced}, independent
of subject matter. Content classification is handled separately.

The taxonomy is organised in three families: photography, traditional /
digital art, and utility / design, with one escape hatch
(\texttt{other}).

\medskip
\textbf{Photography (8)}\\
Always chosen based on \emph{visible production cues} (sensor grain,
bokeh, lens geometry, framing, lighting setup), never from the subject
alone.

\begin{itemize}
\item \texttt{portrait\_photography} — deliberately posed photographic
portrait of a person (studio, editorial, on-location, fashion, beauty,
headshot). The person is the intended subject and the shot is directed.

\item \texttt{product\_photography} — commercial still-life of an object
or small grouping on a clean or styled background with controlled
lighting; e-commerce, packshot, hero shot, food or beauty advertising.

\item \texttt{monochrome\_photography} — any black-and-white, greyscale,
or sepia-toned photograph. Monochrome treatment is the defining cue.

\item \texttt{landscape\_photography} — wide-framed color photograph
where the subject is the environment itself: landscape, seascape,
cityscape, mountains, forests, deserts, aerial views, astrophotography.
People may appear only as scale references.

\item \texttt{street\_photography} — documentary or photojournalistic
images of people in public spaces; unposed, observational scenes such as
events, protests, or markets.

\item \texttt{architecture\_photography} — deliberate photograph of a
building or interior as the primary subject, with strong compositional
cues (symmetry, perspective control, wide-angle geometry).

\item \texttt{wildlife\_macro\_photography} — photograph of a wild
animal, insect, or fine natural detail, with strong telephoto or macro
cues (compression, shallow depth of field).

\item \texttt{casual\_photography} — default catch-all for everyday
color photographs: snapshots, selfies, family, travel, food, interiors,
social media images.
\end{itemize}

\medskip
\textbf{Traditional / digital art (4)}

\begin{itemize}
\item \texttt{traditional\_art} — paintings and classical sculptures,
including photographic reproductions of the work itself.

\item \texttt{anime} — anime or manga-style imagery with cel shading and
stylised features.

\item \texttt{illustration} — non-anime illustration such as concept art,
vector graphics, cartoons, comics, pixel art, or digital painting.

\item \texttt{sketch} — line-based drawings such as pencil, ink, or
charcoal sketches, technical drawings, engravings, or woodcuts.
\end{itemize}

\medskip
\textbf{Utility / design (2)}

\begin{itemize}
\item \texttt{3d\_render} — computer-generated imagery (photoreal or
stylised), including product visualisation, animation stills, and game
renders.

\item \texttt{graphic\_design} — posters, advertisements, layouts,
infographics, typography-heavy designs, UI screenshots, or documents.
\end{itemize}

\medskip
\textbf{Escape hatch}

\begin{itemize}
\item \texttt{other} — anything that does not fit the above categories.
\end{itemize}
\end{promptbox}

\subsubsection{Image style distribution}
\label{sec:image-style-distribution}

\begin{figure}[ht]
    \centering
    \includegraphics[width=\linewidth,trim=0 200 0 0,clip]{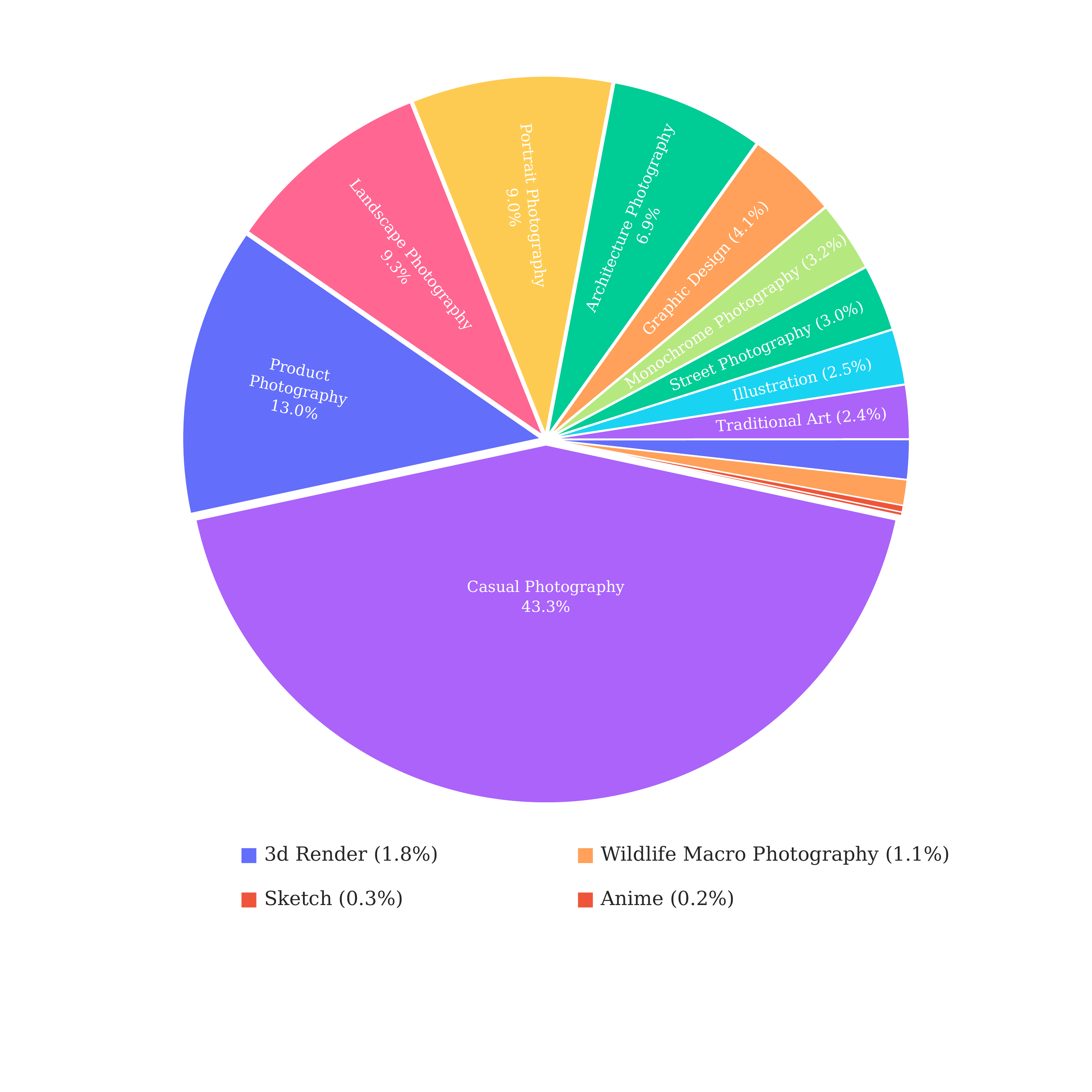}
    \caption{Detailed image style distribution.}
    \label{fig:image-style-complete}
\end{figure}

\begin{figure}[ht]
    \centering
    \newcommand{\stylethumb}[1]{%
        \includegraphics[width=\linewidth,height=\linewidth,
            keepaspectratio=false]{assets/style_audit/#1}%
    }
    \newcommand{\stylepair}[4]{%
        \begin{subfigure}[t]{\linewidth}
            \centering
            \begin{minipage}[t]{#3}\stylethumb{#1_1.jpg}\end{minipage}\hfill
            \begin{minipage}[t]{#4}\stylethumb{#1_2.jpg}\end{minipage}
            \caption{#2}
        \end{subfigure}%
    }
    \newcommand{\minthumb}[3]{%
            \begin{minipage}[t]
            {0.32\linewidth}
            {#1}
            \end{minipage}\hfill
            \begin{minipage}[t]
            {0.32\linewidth}
            {#2}\end{minipage}\hfill
            \begin{minipage}[t]
            {0.32\linewidth}
            {#3}\end{minipage}\hfill
            \vspace{0.5em}
    }
        \newcommand{\minthumbalone}[3]{%
            \begin{minipage}[t]
            {0.32\linewidth}
            {#1}
            \end{minipage}
    }
\newcommand{\minthumbtwo}[2]{%
    \begin{center}
        \begin{minipage}[t]{0.32\linewidth}
            #1
        \end{minipage}\hspace{0.04\linewidth}
        \begin{minipage}[t]{0.32\linewidth}
            #2
        \end{minipage}
    \end{center}
    \vspace{0.5em}
}
    \minthumb{\stylepair{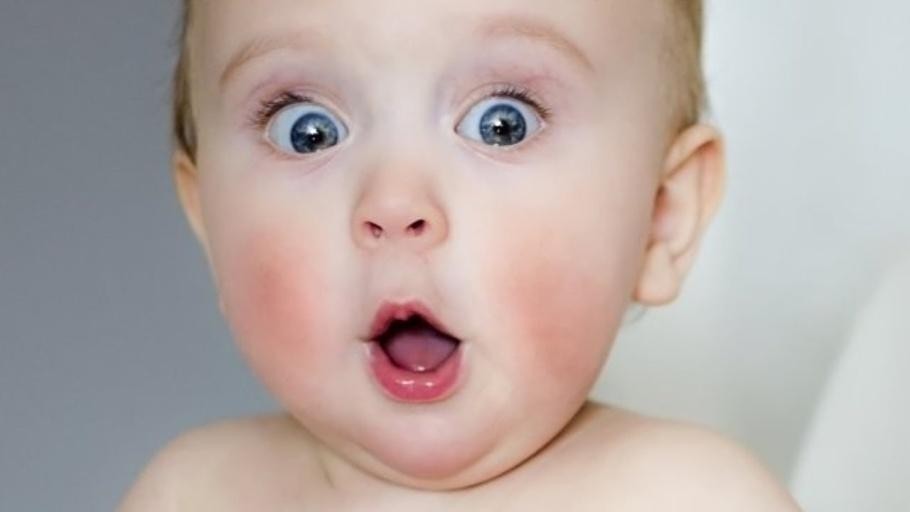}{Portrait}{0.495\linewidth}{0.495\linewidth}}{\stylepair{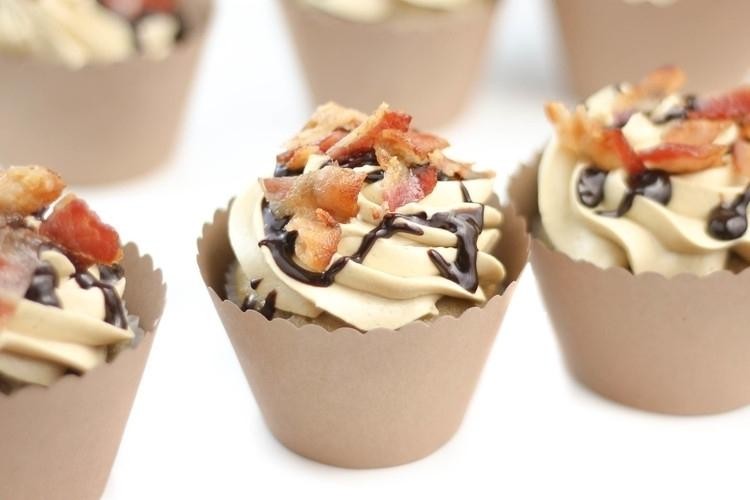}{Product}{0.495\linewidth}{0.495\linewidth}}{\stylepair{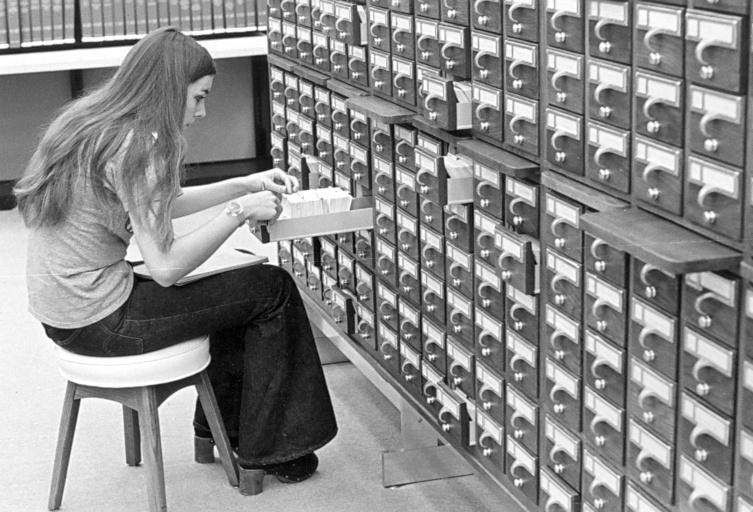}{Monochrome}{0.495\linewidth}{0.495\linewidth}}
    \minthumb{\stylepair{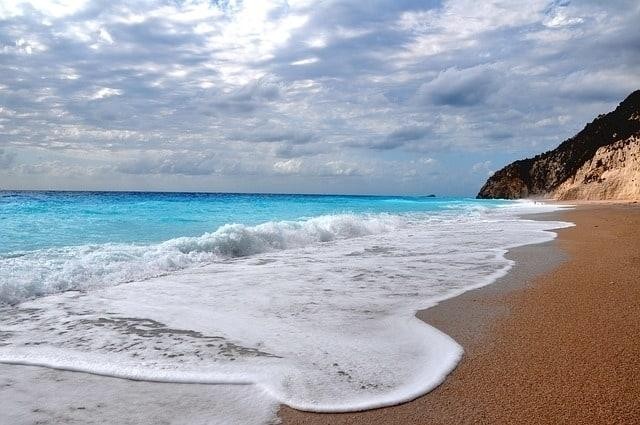}{Landscape}{0.495\linewidth}{0.495\linewidth}}{\stylepair{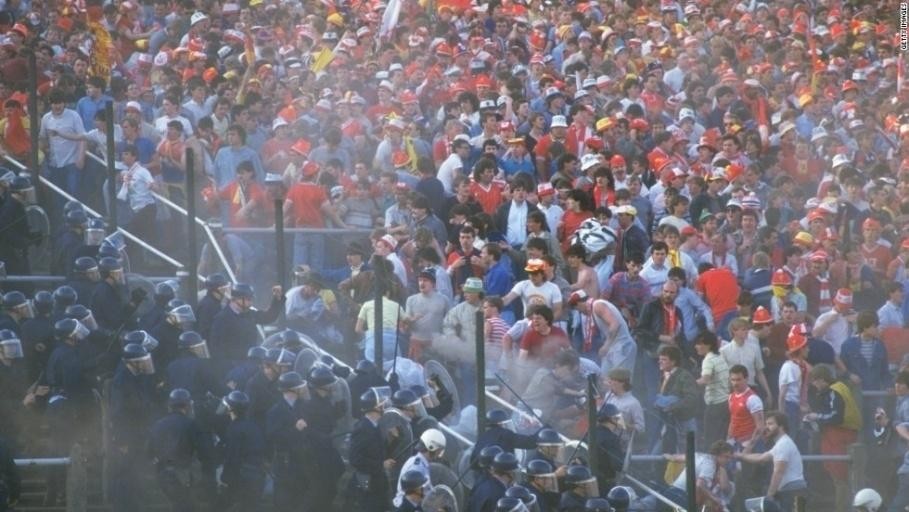}{Street}{0.495\linewidth}{0.495\linewidth}}{\stylepair{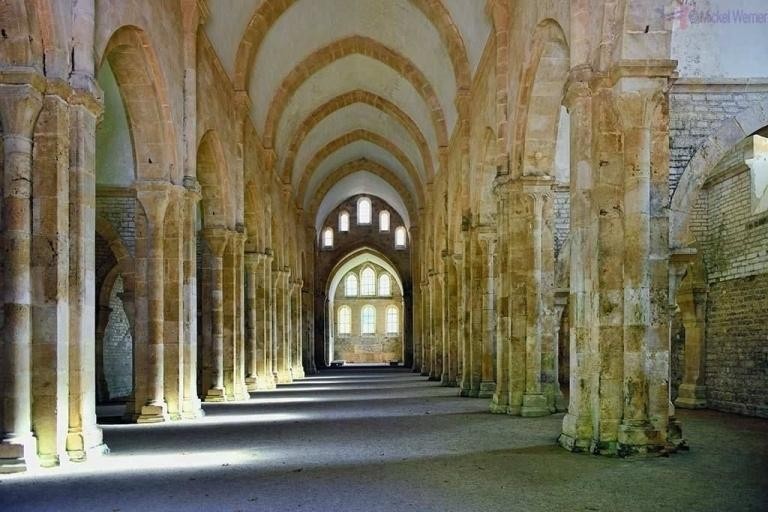}{Architecture}{0.495\linewidth}{0.495\linewidth}}
    \minthumb{\stylepair{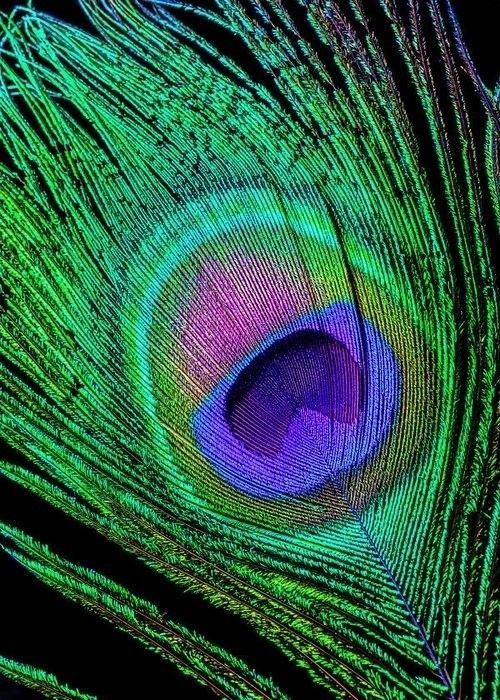}{Wildlife / macro}{0.495\linewidth}{0.495\linewidth}}{\stylepair{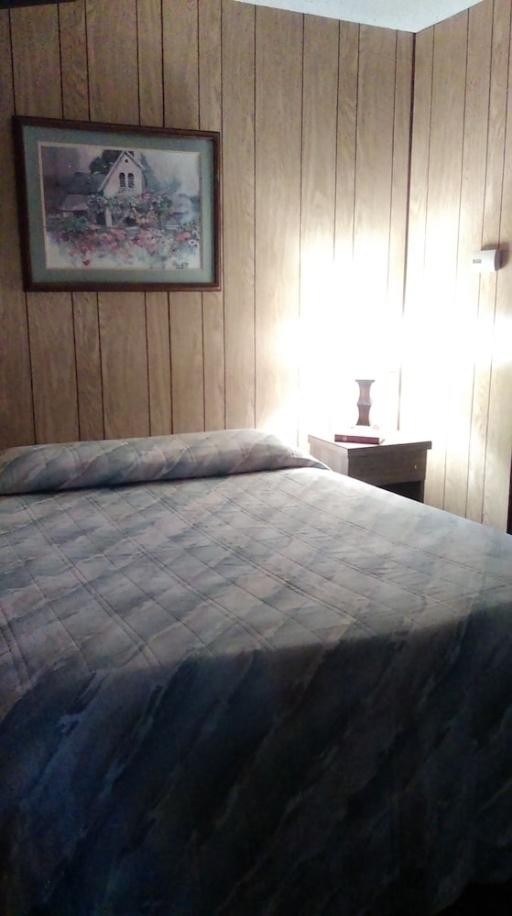}{Casual}{0.495\linewidth}{0.495\linewidth}}{\stylepair{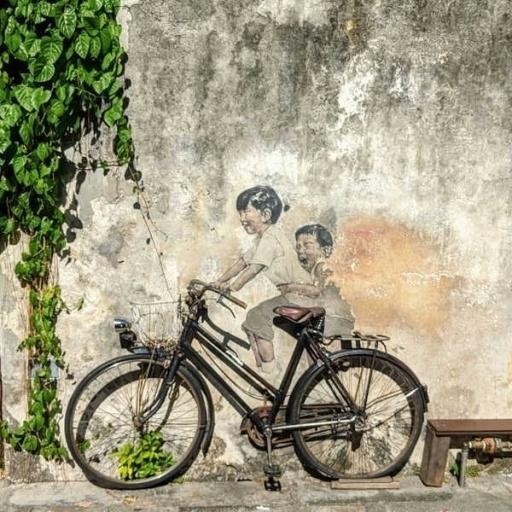}{Traditional art}{0.495\linewidth}{0.495\linewidth}}
    \minthumb{\stylepair{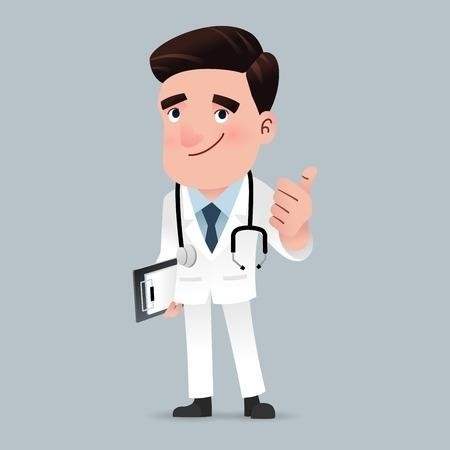}{Illustration}{0.495\linewidth}{0.495\linewidth}}{\stylepair{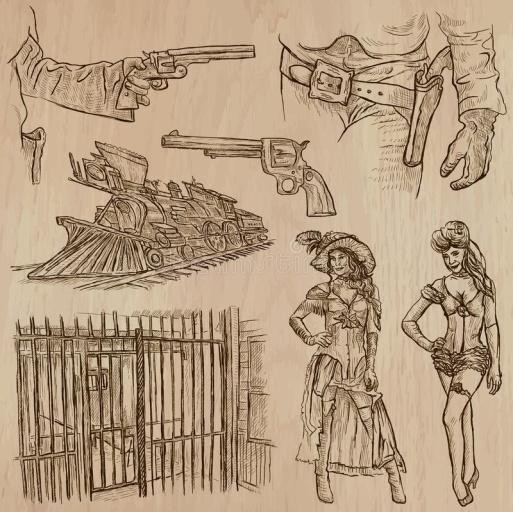}{Sketch}{0.495\linewidth}{0.495\linewidth}}{\stylepair{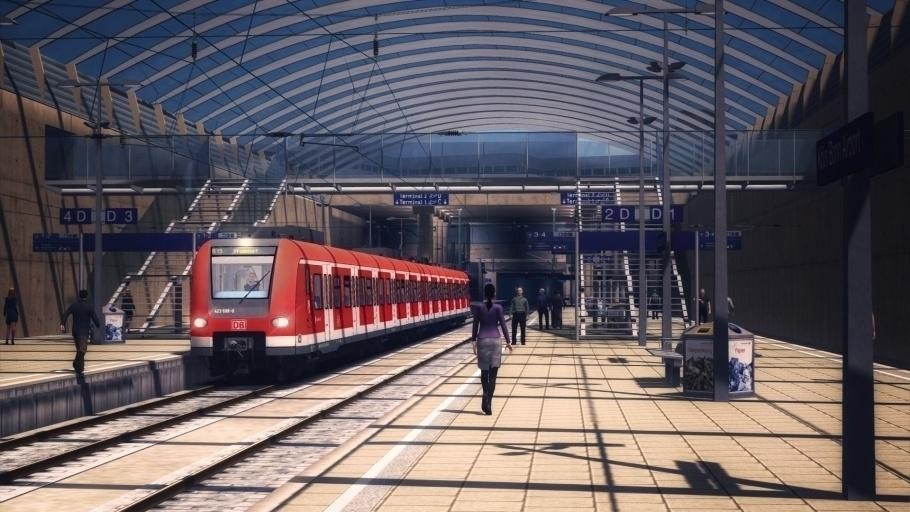}{3D render}{0.495\linewidth}{0.495\linewidth}}
    \minthumbtwo{\stylepair{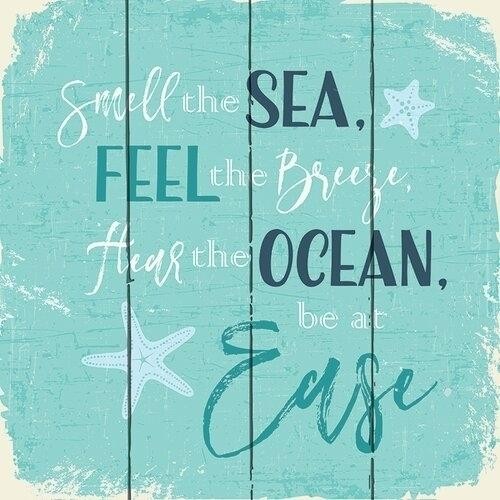}{Graphic design}{0.495\linewidth}{0.495\linewidth}}{\stylepair{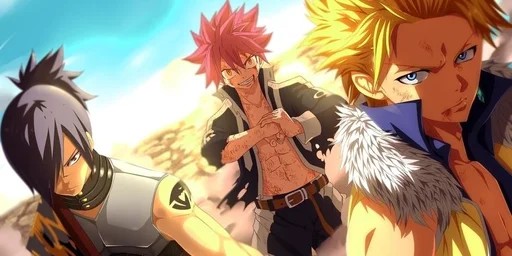}{Anime}{0.495\linewidth}{0.495\linewidth}}
    \caption{Example images per picture-style label
    sampled from the audit preview.}
    \label{fig:image-style-examples}
\end{figure}

Fig.~\ref{fig:image-style-complete} shows the detailed distribution of image styles. We highlight the high variability of image styles in the dataset. Fig.~\ref{fig:image-style-examples} shows two examples per image style category, illustrating that the VLM-based image style classification (e.g., sketch, illustration, etc.) is consistent and reliable.

\newpage
\subsection{Training details}
\label{sec:training-details}

\paragraph{Captioning models and synthetic data ablations}
For these ablations, we train the models on images of resolution $512\times512$ for 400k iterations on 2 H200 GPUs, using 16 dual-stream MMDiT blocks \citep{esser2024scaling} with 24 attention heads of size 128. The text conditioning is passed either through the Qwen3-4B pre-trained Large Language Model (LLM) \cite{yang2025qwen3} (for the synthetic data ablation) or T5 \citep{raffel2020exploring} (for the captioner ablation). The output of the antepenultimate layer (Qwen3-4B) or the last layer (T5) serves as conditioning for the denoiser. While training the model, as is standard practice, we also replace the text conditioning with an empty prompt \emph{""} 10\% of the time, enabling classifier-free guidance \citep{ho2021classifier} at inference time. We rely on the Latent Diffusion Model framework \citep{rombach2022high} to train our models using the SANA-1.5 VAE model \citep{xie2025sana}, which spatially compresses an input image by a factor of 32 and use the flow matching approach \cite{lipman2023flow,liu2022flow}. Metrics are computed using samples generated with 50 denoising steps and a guidance scale of 5. All training images are resized to $512\times512$ and we train the model with a global batch size of 512, using a learning rate of $1e^{-4}$ together with the AdamW optimizer \citep{kingma2014adam}.

\paragraph{T2I model training details}
When training our 4-billion-parameter T2I model, we rely on a denoiser combining 32 MMDiT blocks \citep{esser2024scaling} mixing single (16) and dual-stream blocks (16) all with 20 attention heads of size 128. The text conditioning is passed through the Qwen3-4B pre-trained Large Language Model (LLM). The output of the antepenultimate layer of the text encoder serves as conditioning for the denoiser. While training the model, as is standard practice, we also replace the text conditioning with an empty prompt \emph{""} 10\% of the time, enabling classifier-free guidance at inference time. We rely on the pre-trained deep-compression VAE from the SANA-1.5 model. We employ a multi-stage approach for training the model using the findings presented in the previous sections. We directly start training the model on $512 \times 512$ images using 75\% of synthetic data and the most verbose captioners, namely \emph{Gemini 2.5 Flash-Lite} (50\%) and \emph{InternVL3-8B} (50\%). We then progressively reduce the amount of synthetic data to 50\% and 30\%. We then increase the resolution of the images to $1024\times1024$ and progressively include the other captioners such that they are equally represented in the final training dataset as described in Table~\ref{tab:training-details}. The model is trained relying on the flow matching approach \cite{lipman2023flow,liu2022flow} and optimized using the AdamW optimizer \citep{kingma2014adam}.

\begin{table}[h!]
  \centering
  \scriptsize
  \caption{4B-parameter text-to-image model training details used for the benchmarks.}
  \begin{tabular}{lccccccc}
  \toprule
               & \multirow{2}{*}{Res.} & Batch& Learning & H200 & \multirow{2}{*}{Steps} \\
               & &   size & rate & GPUS & \\
  \midrule
  Phase 1      & 512    &  256 & $1e^{-4}$ & 16 & 350,000 \\
  \midrule
  Phase 2      & 512    &   1536 & $3e^{-5}$ & 24 & 420,000 \\

  \midrule
  Phase 3      & 512    &    3072 & $1e^{-5}$ & 16 & 45,000 \\
  \midrule
  Phase 4      & 1024    &   256 & $1e^{-5}$ & 16 & 35,000 \\
  \bottomrule
  \end{tabular}
  \label{tab:training-details}
  \end{table}

\clearpage

\subsection{Additional results}
\label{sec:app-additional-results}

\subsubsection{Quantitative results}
\label{sec:app-quantitative-results}

We provide in Table~\ref{tab:app-geneval-benchmark} and Table~\ref{tab:app-dpg-benchmark}
additional comparisons on the GenEval and DPG benchmarks~\citep{ghosh2023geneval,hu2024ella}, which assess short and dense prompt-following capabilities across multiple semantic categories. The tables report quantitative results with our model trained on the fully open MONET dataset against \emph{state-of-the-art} text-to-image models trained on closed-source data. Despite relying solely on open data, our 4B model achieves a competitive overall score of 0.74 on GenEval and 85.56 on DPG, outperforming several strong baselines, including FLUX.1 [Dev], SD3 Medium, and Janus-Pro-7B. It nonetheless underperforms on the \emph{other} category of DPG, which assesses among others the capacity of the model to write text on an image. Since MONET does not contain many text--image pairs of this type, the model is unable to render text faithfully. Existing datasets specifically designed for such use cases can be used for further finetuning and enriching our MONET dataset, which was mainly designed for pre-training purposes.
\begin{table}[ht]
  \scriptsize
    \centering
    \caption{GenEval benchmark. Our 4B model trained specifically on the fully open MONET dataset is able to compete with many existing text-to-image models which were trained on closed-source data.}
    \begin{tabular}{lcccccccc}
    \toprule
    \multirow{2}{*}{Model} & \multirow{2}{*}{Params. (B)} &\multicolumn{2}{c}{Objects} &\multirow{2}{*}{Counting} & \multirow{2}{*}{Colors} & \multirow{2}{*}{Position} & \multirow{2}{*}{Color} & \multirow{2}{*}{Overall $\uparrow$} \\
              & & Single & Two &  \\
    \midrule
    SD v1.5 \citep{rombach2022high}      & 0.9 & 0.97 & 0.38 & 0.35 & 0.76 & 0.04 & 0.06 & 0.43 \\
    PixArt-$\alpha$  \citep{chen2023pixart} &0.6 & 0.98 & 0.50 & 0.44 & 0.80 & 0.08 & 0.07 & 0.48 \\
    SD v2.1 \citep{rombach2022high}        &0.9 & 0.98 & 0.51 & 0.44 & 0.85 & 0.07 & 0.17 & 0.50 \\
    DALL-E 2  \citep{ramesh2022hierarchical}  & 3.5    & 0.94 & 0.66 & 0.49 & 0.77 & 0.10 & 0.19 & 0.52 \\
    Show-o \citep{xie2025show}       & 1.3 & 0.95 & 0.52 & 0.49 & 0.82 & 0.11 & 0.28 & 0.53 \\
    Emu3-Gen \citep{wang2024emu3} & 8.0&0.98 & 0.71 & 0.34 & 0.81 & 0.17 & 0.21 & 0.54 \\
    SDXL \citep{podell2023sdxl}      &  2.6   & 0.98 & 0.74 & 0.39 & 0.85 & 0.15 & 0.23 & 0.55 \\
    SDXL Turbo \citep{sauer2023adversarial}   &2.6  & 1.00 & 0.72 & 0.49 & 0.80 & 0.10 & 0.18 & 0.55 \\
    SD3 Medium \citep{esser2024scaling} & 2.0& 0.98 & 0.74 & 0.63 & 0.67 & 0.34 & 0.36 & 0.62 \\
    JanusFlow \citep{ma2025janusflow} & 1.3 & 0.97 & 0.59 & 0.45 & 0.83 & 0.53 & 0.42 & 0.63 \\
    FLUX.1 [Dev] \citep{flux2024}& 12.0 & 0.98 & 0.81 & 0.74 & 0.79 & 0.22 & 0.45 & 0.66 \\
    DALL-E 3 \citep{betker2023improving} &-      & 0.96 & 0.87 & 0.47 & 0.83 & 0.43 & 0.45 & 0.67 \\
    SD3.5 Large \citep{esser2024scaling} & 8.0& 0.98 & 0.89 & 0.73 & 0.83 & 0.34 & 0.47 & 0.71 \\
    SANA-1.5 \citep{xie2025sana} & 4.8 & 0.99 & 0.85 & 0.77 & 0.87 & 0.34 & 0.54 & 0.72\\
    Lumina-Image 2.0 \citep{qin2025lumina} &2.6 & - & 0.87 & 0.67 & - & - & 0.62 & 0.73 \\
    Janus-Pro-7B \citep{chen2025janus} & 7.0& 0.99 & 0.89 & 0.59 & 0.90 & 0.79 & 0.66 & 0.80 \\
    HiDream-I1-Full \citep{cai2025hidream} &17.0  & 1.00 & 0.98 &0.79 &0.91 & 0.60 & 0.72 & 0.83 \\
    GPT Image 1 \citep{openaigptimage} &-& 0.99 & 0.92 & 0.85 & 0.92 & 0.75 & 0.61  & 0.84\\
    Seedream 3.0 \citep{gao2025seedream}& - & 0.99 & 0.96 & 0.91 & 0.93 & 0.47 & 0.80 & 0.84\\
    Z-Image \citep{zimage2025} &6.0 &1.00 & 0.94& 0.78& 0.93 &0.62& 0.77 &0.84\\
    Qwen-Image \citep{wu2025qwen} & 20.0& 0.99 &0.92 & 0.89 &0.88 & 0.76 &0.77 &0.87\\
    \midrule
    Ours & 4.1 &1.00 & 0.90 & 0.73 & 0.88 & 0.35 & 0.62 & 0.74 \\

    \bottomrule
    \end{tabular}
    \label{tab:app-geneval-benchmark}
    \end{table}

\begin{table}[ht]
  \scriptsize
    \centering
    \caption{Quantitative evaluation results on DPG. Our 4B model trained on the fully open MONET dataset achieves competitive performance against models trained on closed-source data.}
    \begin{tabular}{lccccccc}
    \toprule
    Model & Params. (B)& Global & Entity & Attribute & Relation & Other & Overall$\uparrow$ \\
    \midrule
    SD v1.5 \citep{rombach2022high}            & 0.9  & 74.63 & 74.23 & 75.39 & 73.49 & 67.81 & 63.18 \\
    PixArt-$\alpha$ \citep{chen2023pixart}     & 0.6  & 74.97 & 79.32 & 78.60 & 82.57 & 76.96 & 71.11 \\
    Lumina-Next \citep{zhuo2024lumina}         & 2.0  & 82.82 & 88.65 & 86.44 & 80.53 & 81.82 & 74.63 \\
    SDXL \citep{podell2023sdxl}                & 2.6  & 83.27 & 82.43 & 80.91 & 86.76 & 80.41 & 74.65 \\
    Playground v2.5 \citep{li2024playground}   & 2.6  & 83.06 & 82.59 & 81.20 & 84.08 & 83.50 & 75.47 \\
    Hunyuan-DiT \citep{li2024hunyuan}          & 1.5  & 84.59 & 80.59 & 88.01 & 74.36 & 86.41 & 78.87 \\
    Janus \citep{wu2025janus}                  & 1.3  & 82.33 & 87.38 & 87.70 & 85.46 & 86.41 & 79.68 \\
    PixArt-$\Sigma$ \citep{chen2024pixart}     & 0.6  & 86.89 & 82.89 & 88.94 & 86.59 & 87.68 & 80.54 \\
    Emu3-Gen \citep{wang2024emu3}              & 8.0  & 85.21 & 86.68 & 86.84 & 90.22 & 83.15 & 80.60 \\
    Janus-Pro-1B \citep{chen2025janus}         & 1.0  & 87.58 & 88.63 & 88.17 & 88.98 & 88.30 & 82.63 \\
    DALL-E 3 \citep{betker2023improving}       & -  & 90.97 & 89.61 & 88.39 & 90.58 & 89.83 & 83.50 \\
    FLUX.1 [Dev] \citep{flux2024}              & 12.0  & 74.35 & 90.00 & 88.96 & 90.87 & 88.33 & 83.84 \\
    SD3 Medium \citep{esser2024scaling}        & 2.0  & 87.90 & 91.01 & 88.83 & 80.70 & 88.68 & 84.08 \\
    Janus-Pro-7B \citep{chen2025janus}         & 7.0  & 86.90 & 88.90 & 89.40 & 89.32 & 89.48 & 84.19 \\
    GPT Image 1 \citep{openaigptimage}         & -  & 88.89 & 88.94 & 89.84 & 92.63 & 90.96 & 85.15 \\
    HiDream-I1-Full \citep{cai2025hidream}     & 17.0  & 76.44 & 90.22 & 89.48 & 93.74 & 91.83 & 85.89 \\
    Lumina-Image 2.0 \citep{qin2025lumina}     & 2.6  & -     & 91.97 & 90.20 & 94.85 & -     & 87.20 \\
    Z-Image \citep{zimage2025}                 &6.0 &93.39&91.22& 93.16 & 92.22 &91.52 &88.14\\
        Seedream 3.0 \citep{gao2025seedream}       & -  & 94.31 & 92.65 & 91.36 & 92.78 & 88.24 & 88.27 \\
    Qwen-Image \citep{wu2025qwen}              & 20.0  & 91.32 & 91.56 & 92.02 & 94.31 & 92.73 & 88.32 \\
    \midrule
    Ours - 4B                                  & 4.1  & 84.80 & 91.76 & 89.70 & 94.16 & 79.60 & 85.56 \\
    \bottomrule
    \end{tabular}
    \label{tab:app-dpg-benchmark}
\end{table}

\clearpage

\subsubsection{Generation examples}
\label{sec:app-generation-samples}

\begin{figure*}[p]
  \centering
  \captionsetup[subfigure]{position=below, labelformat = empty}
  \subfloat[\emph{A raccoon trapped inside a glass jar full of colorful candies, the background is steamy with vivid colors}]{\includegraphics[width=2.5in]{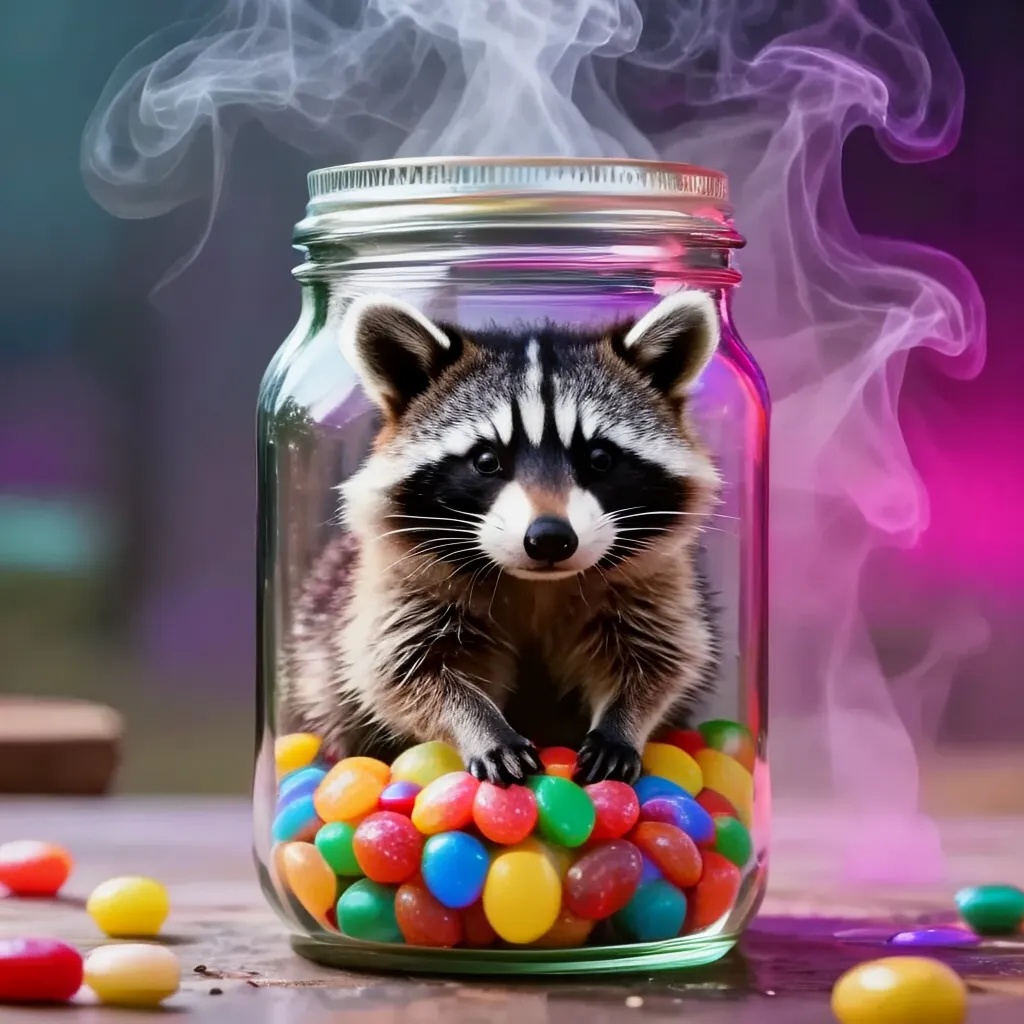}}\hspace{1em}
  \subfloat[\emph{A portrait of an old man with his eyes closed and a blue jacket. He is wearing a yellow cap with a purple square on it and red glasses.}]{\includegraphics[width=2.5in]{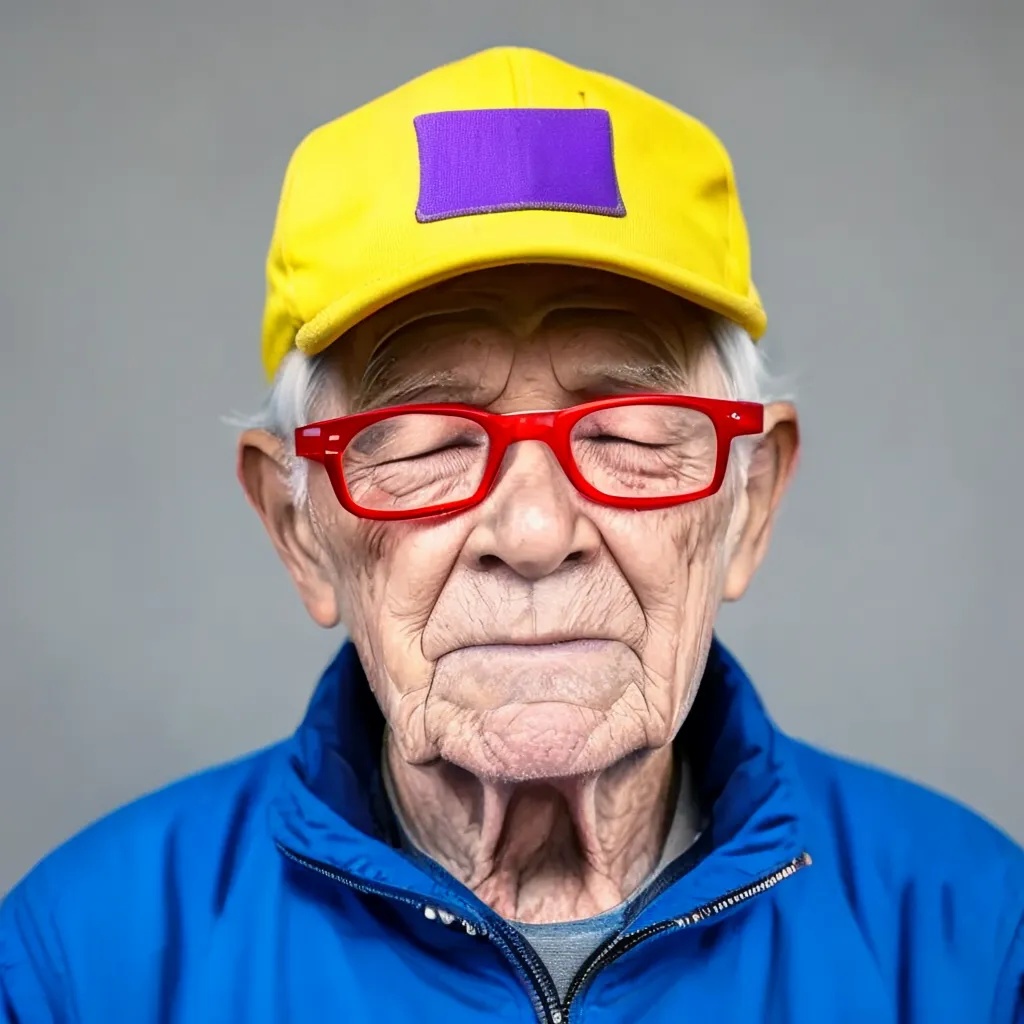}}\\\vspace{1em}
  \subfloat[\emph{A flat-lay photograph of a white desk. Four objects are placed in the four corners: Top-left: a black coffee cup. Top-right: a green succulent in a purple pot. Bottom-left: a red apple. Bottom-right: a gold pen. In the exact center of the desk is a clean, empty white notebook.}]{\includegraphics[width=2.5in]{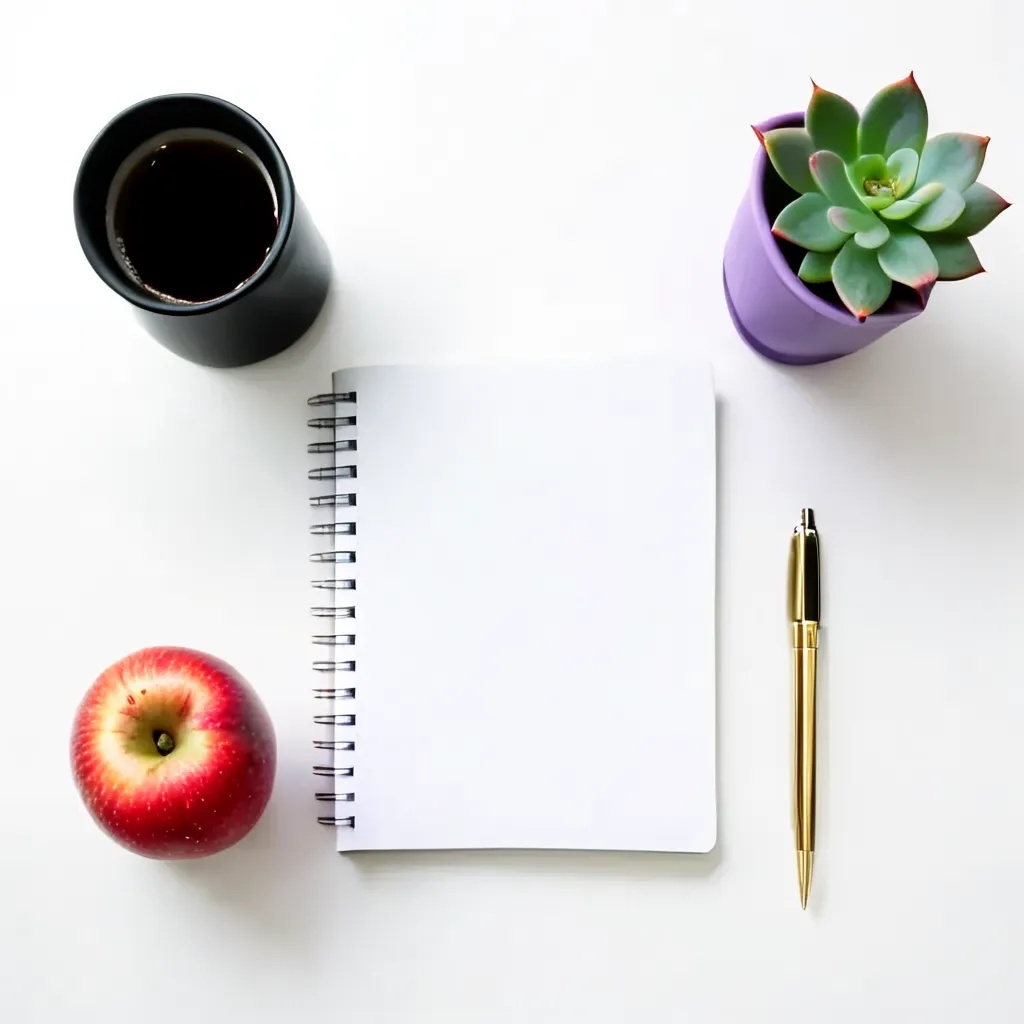}}\hspace{1em}
  \subfloat[\emph{A transparent glass teapot filled with a miniature stormy ocean, tiny lightning bolts visible inside the glass.}]{\includegraphics[width=2.5in]{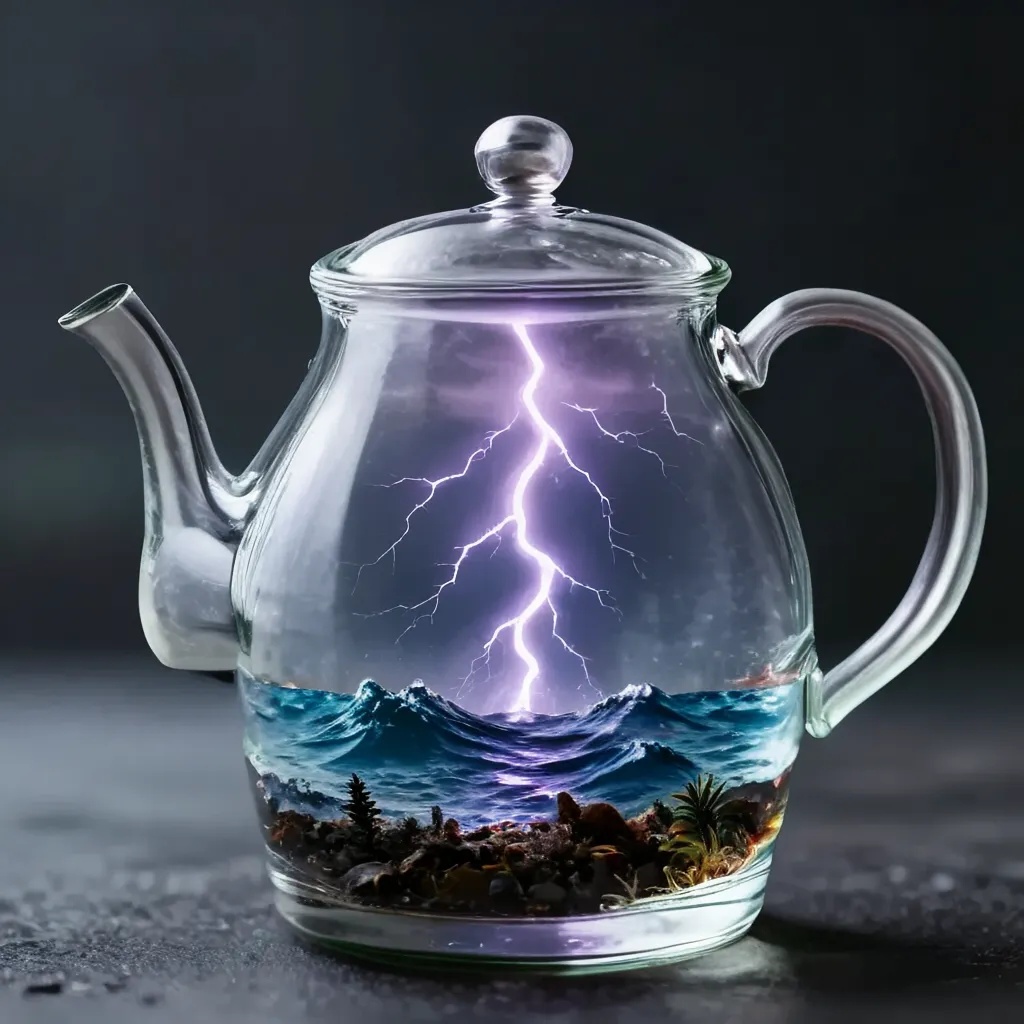}}
  \caption{Generation from our 4B model ($1024\times1024$) showcasing its ability to learn complex concepts thanks to the MONET Dataset.}
  \label{fig:gen_1024_1}
\end{figure*}

\begin{figure*}[p]
  \centering
  \captionsetup[subfigure]{position=below, labelformat = empty}
  \subfloat[\emph{A moody black and white film portrait of an older artisan. The lighting is extremely dramatic high-key chiaroscuro, illuminating only one half of the face in sharp relief, casting the rest into deep, absolute shadow. The texture of their wrinkles and beard is hyper-detailed. Grainy analog film aesthetic.}]{\includegraphics[width=2.5in]{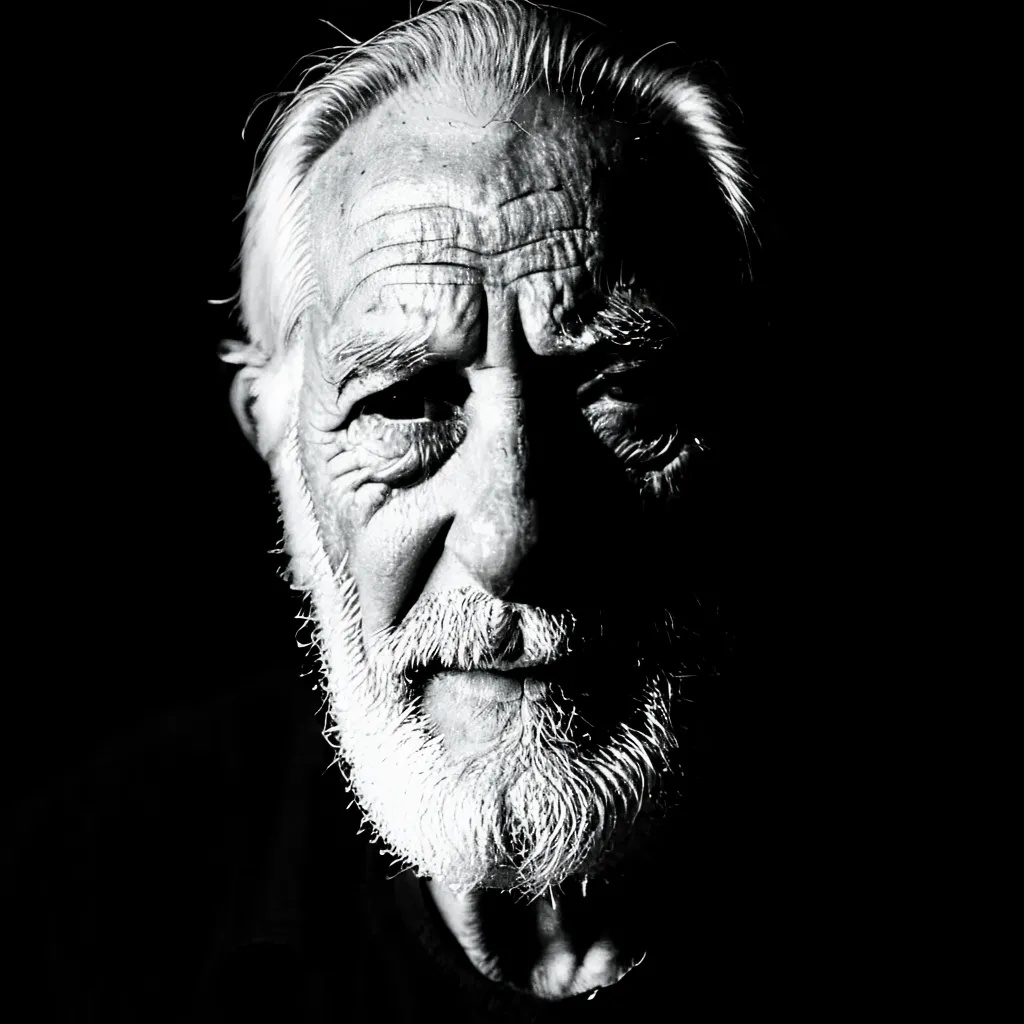}}\hspace{1em}
  \subfloat[\emph{A split-level shot (half underwater, half above) of a tropical turquoise wave breaking. Above water is misty spray; below water shows sunbeams (God rays) refracting through the moving surface, illuminating bubbles and sand.}]{\includegraphics[width=2.5in]{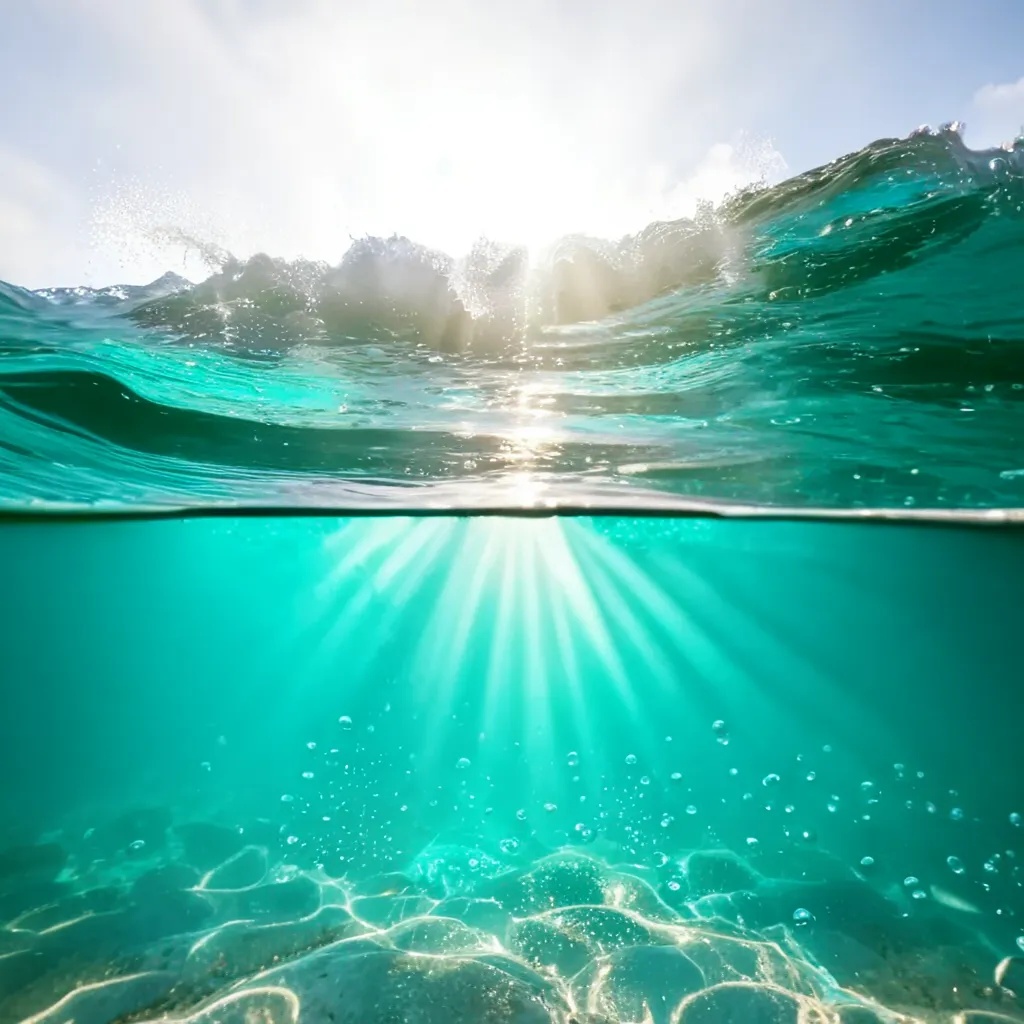}}\\\vspace{1em}
  \subfloat[\emph{A whimsical purple and pink dragon made of clay, showing tiny fingerprint marks and handcrafted texture, soft studio lighting, stop-motion aesthetic.}]{\includegraphics[width=2.5in]{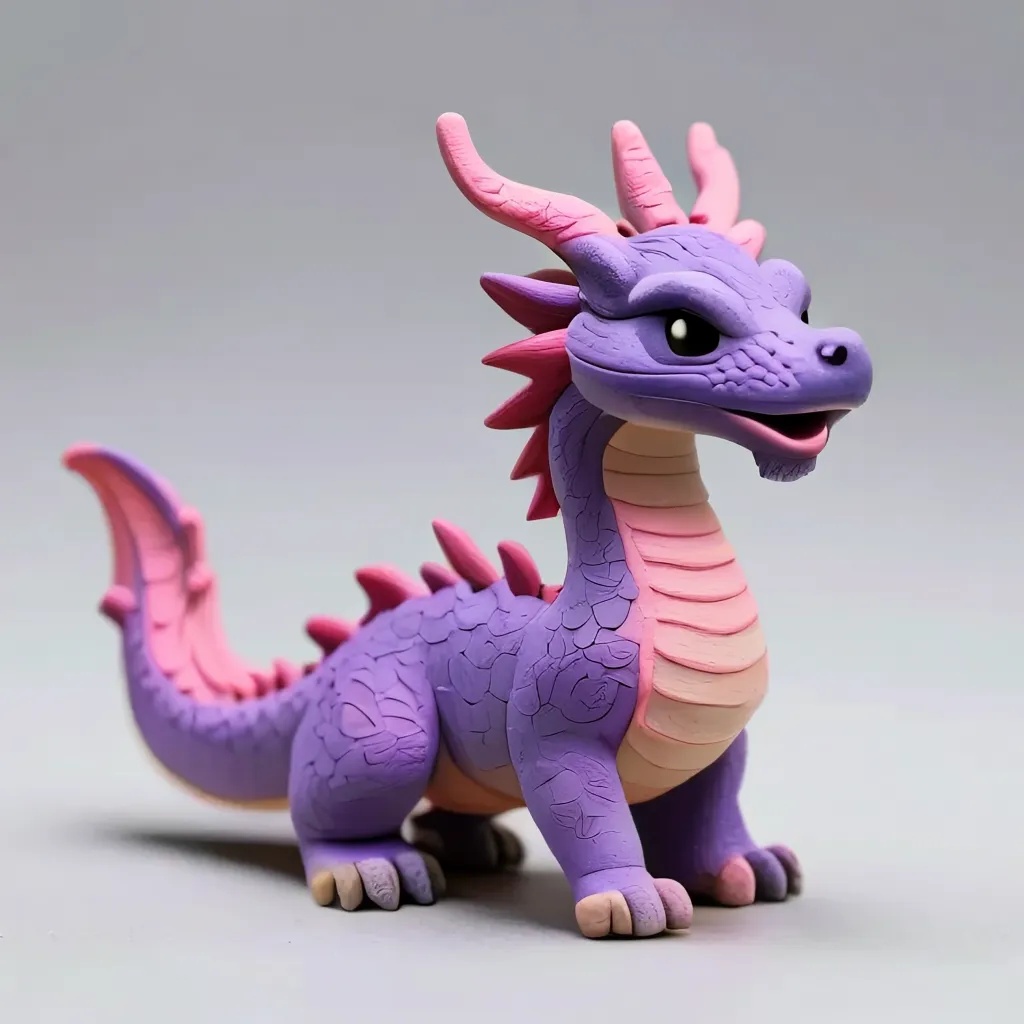}}\hspace{1em}
  \subfloat[\emph{Professional product photography of a glass skincare bottle on a floating stone plinth. Surround the bottle with clean, splash-frozen water droplets. Pure white background, studio lighting, sharp focus on the glass texture.}]{\includegraphics[width=2.5in]{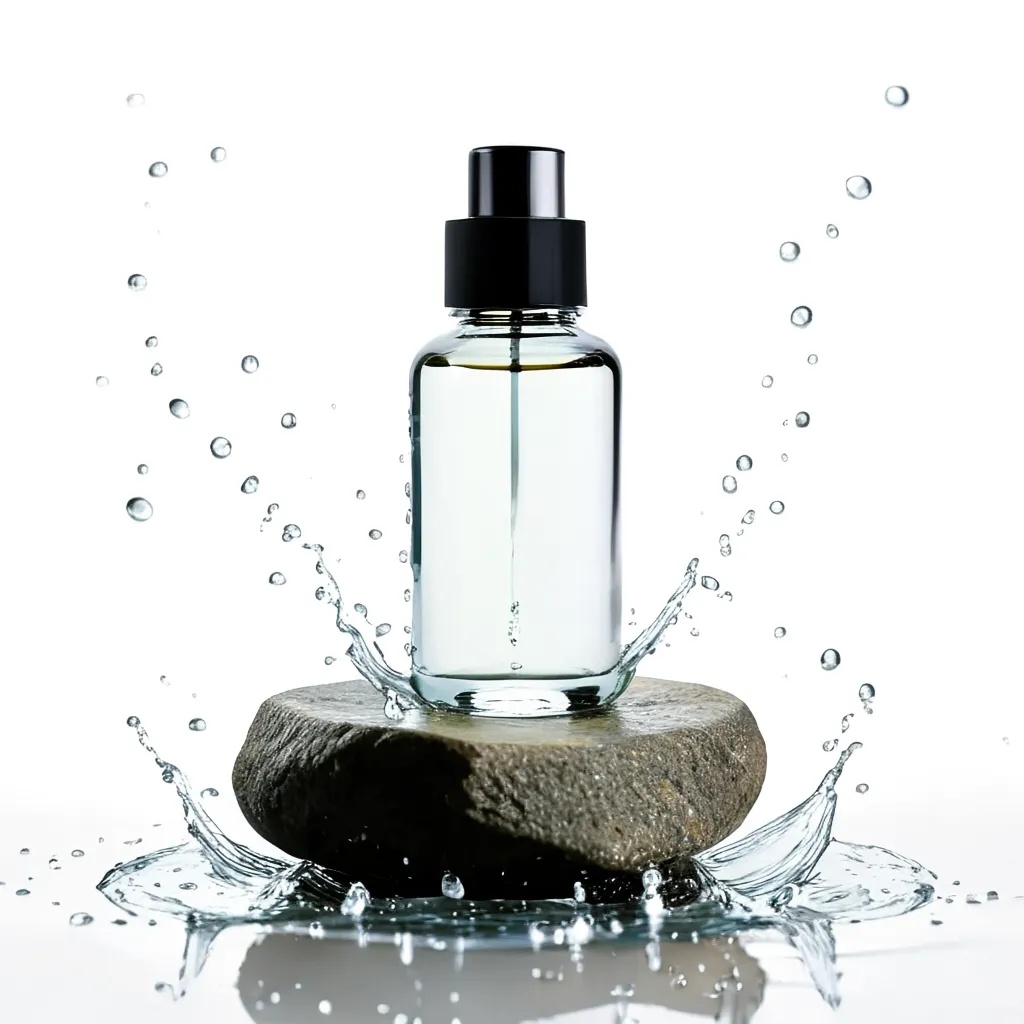}}
  \caption{Generation from our 4B model ($1024\times1024$) showcasing its ability to generate high resolution images thanks to the MONET Dataset.}
  \label{fig:gen_1024_2}
\end{figure*}

\begin{figure*}[p]
  \centering
  \captionsetup[subfigure]{position=below, labelformat = empty}
  \subfloat[\emph{A portrait of a samurai rendered in the style of a traditional 18th-century Japanese woodblock print, weathered paper texture.}]{\includegraphics[width=2.5in]{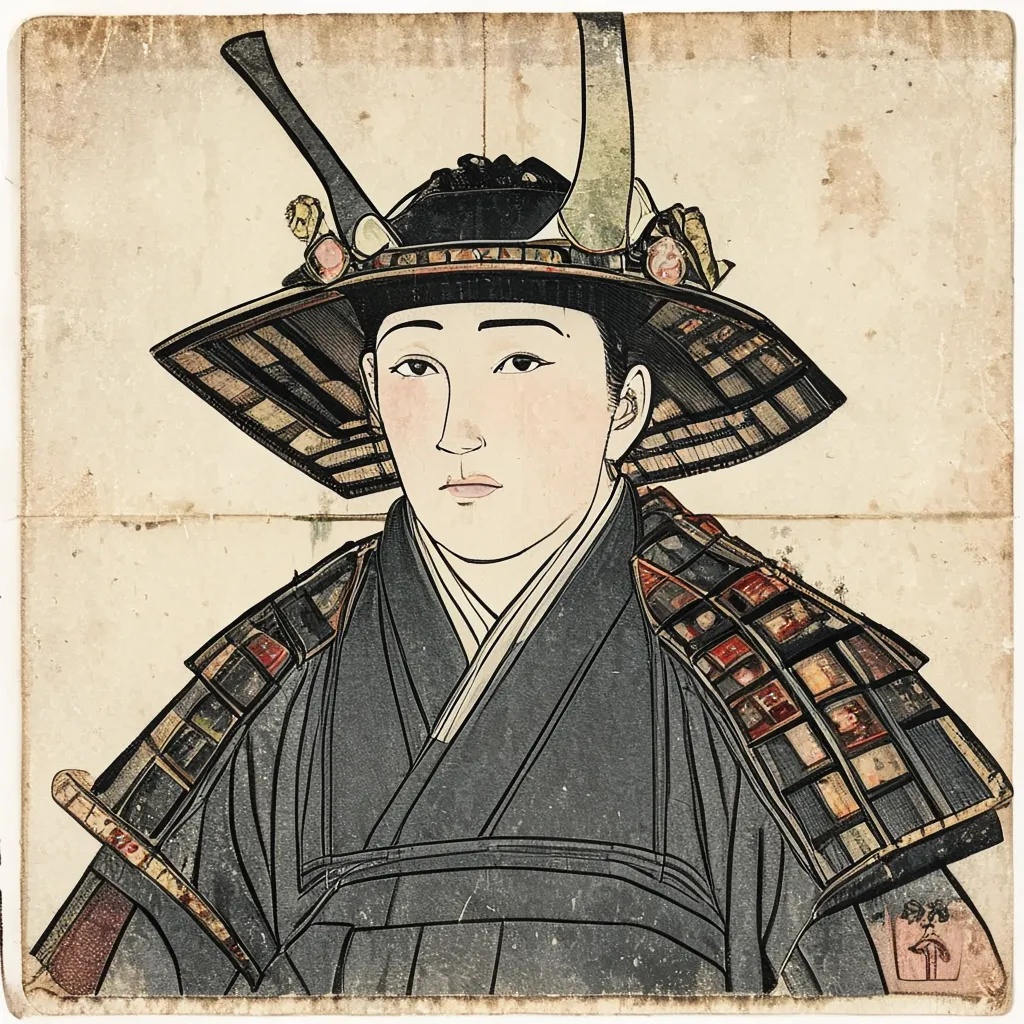}}\hspace{1em}
  \subfloat[\emph{A dynamic street scene of Tokyo reimagined entirely as an abstract expressionist oil painting. The image must have violent, thick, impasto brushstrokes of primary colors (red, yellow, blue, black) applied messily and dynamically with a palette knife on raw canvas. Show the extreme texture of the paint.}]{\includegraphics[width=2.5in]{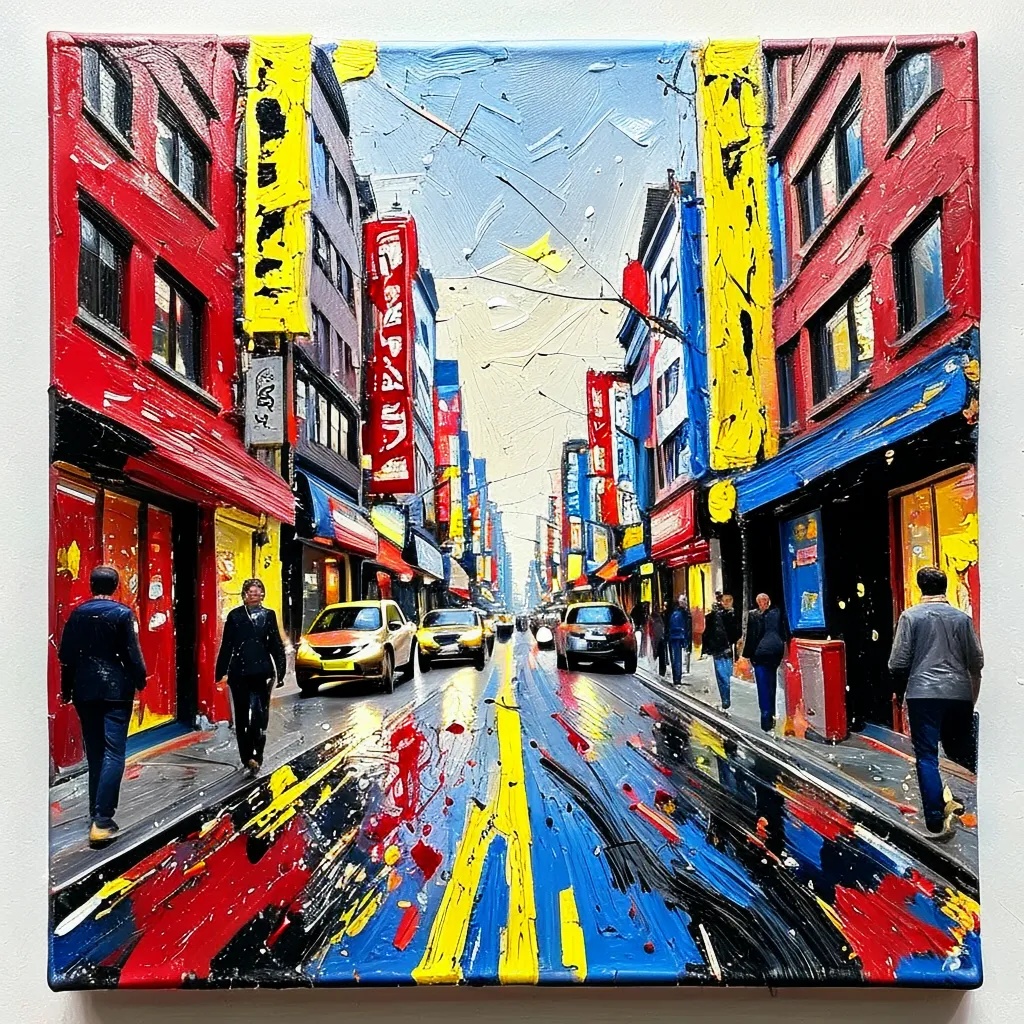}}\\\vspace{1em}
\subfloat[\emph{A vibrant pointillist painting of a garden in Paris. The entire scene is composed of millions of tiny, precise, distinct dots of pure, unmixed color that blend optically.}]{\includegraphics[width=2.5in]{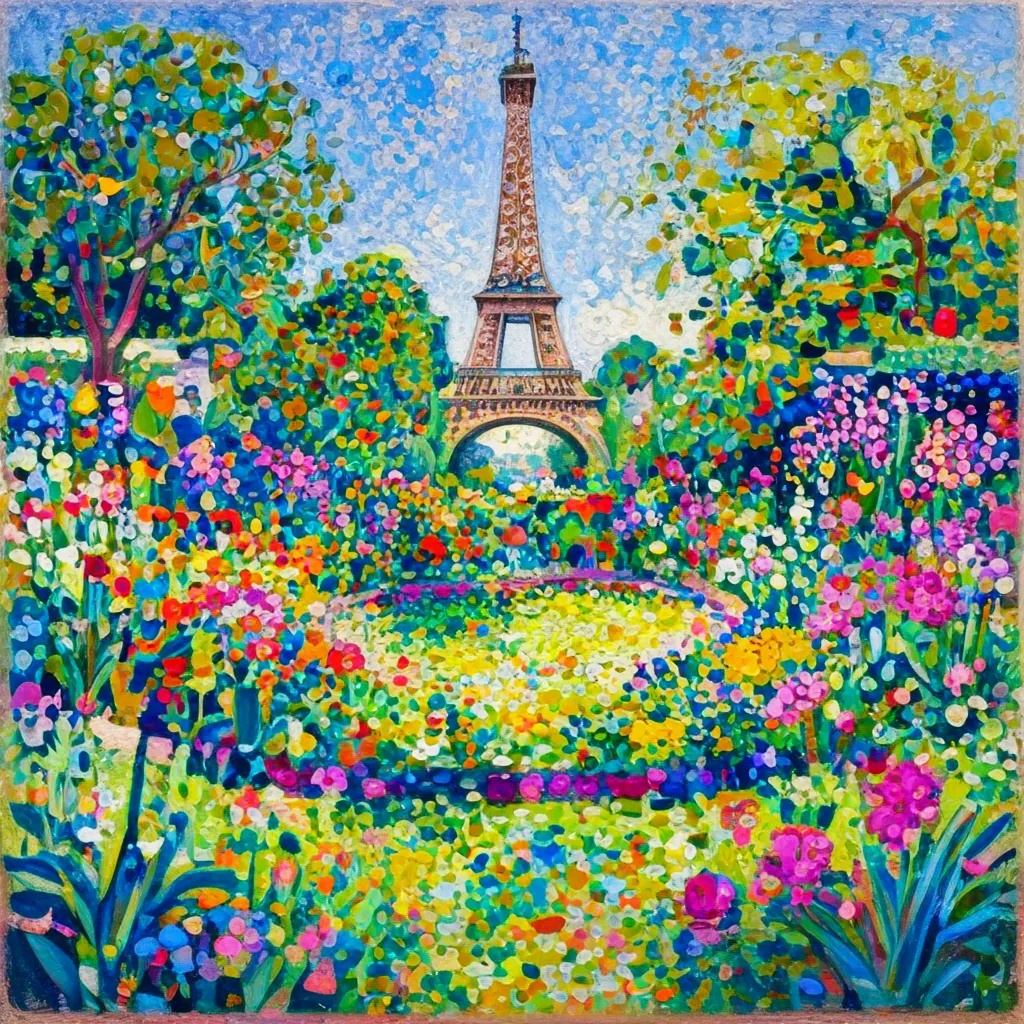}}\hspace{1em}
  \subfloat[\emph{A luminous aquarelle painting of a misty Venice canal. The style emphasizes translucent washes of paint, soft blending edges, and visible bleeding. The texture of the textured watercolor paper is prominent, and the light passes through the paint.}]{\includegraphics[width=2.5in]{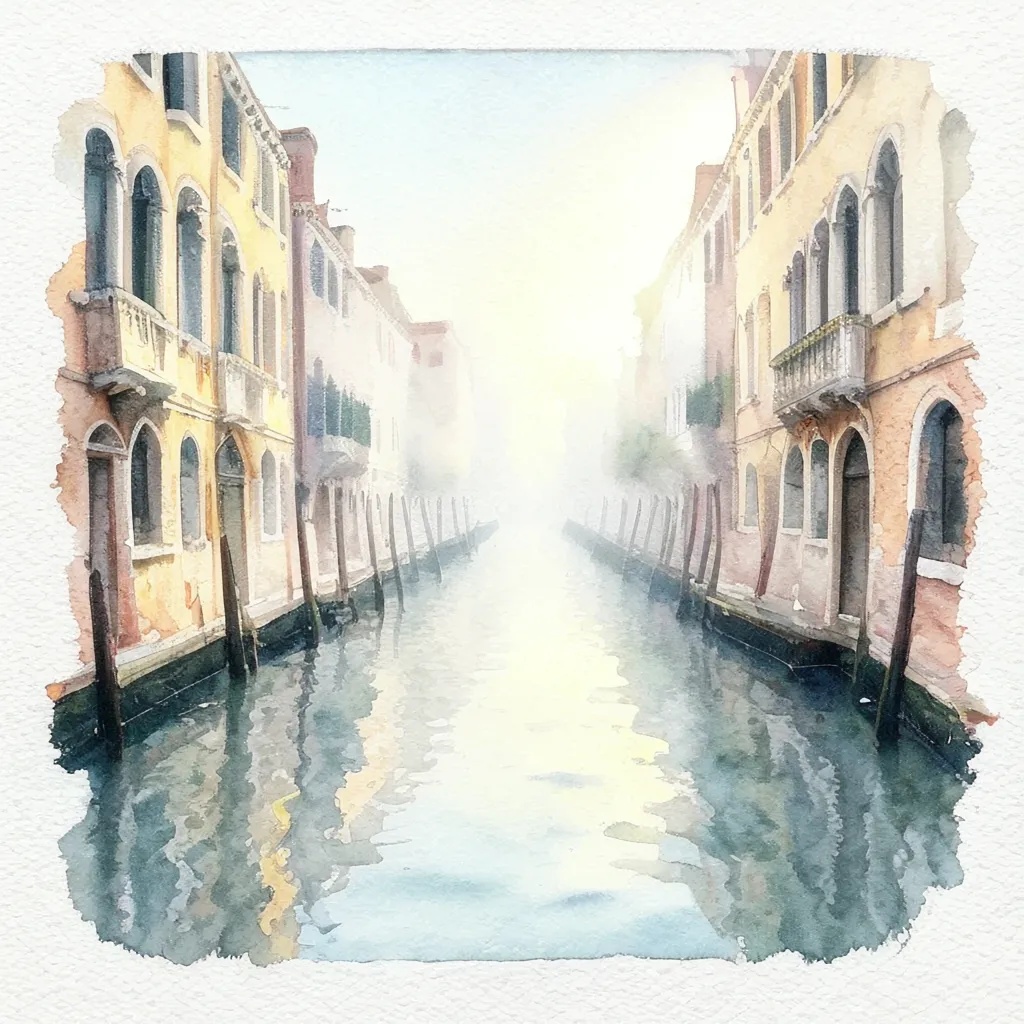}}
  \caption{Generation from our 4B model ($1024\times1024$) showcasing its ability to generate images with different styles thanks to the MONET Dataset.}
  \label{fig:gen_1024_3}
\end{figure*}

\begin{figure*}[p]
  \centering
  \captionsetup[subfigure]{position=below, labelformat = empty}
  \subfloat[\emph{An ultra-detailed nocturnal landscape of a hidden tropical lagoon. The water glows with intense neon blue bioluminescence where it laps against jet-black volcanic sand. Towering ancient banyan trees with glowing hanging vines frame the scene. In the background, a massive silver moon hangs low over a calm ocean, casting a shimmering path on the waves. Fireflies create bokeh light clusters in the dark jungle shadows. Intricate textures of wet sand and leaf veins. Surreal atmospheric lighting, high contrast, 16k masterwork, Unreal Engine 5 render style.}]{\includegraphics[width=2.5in]{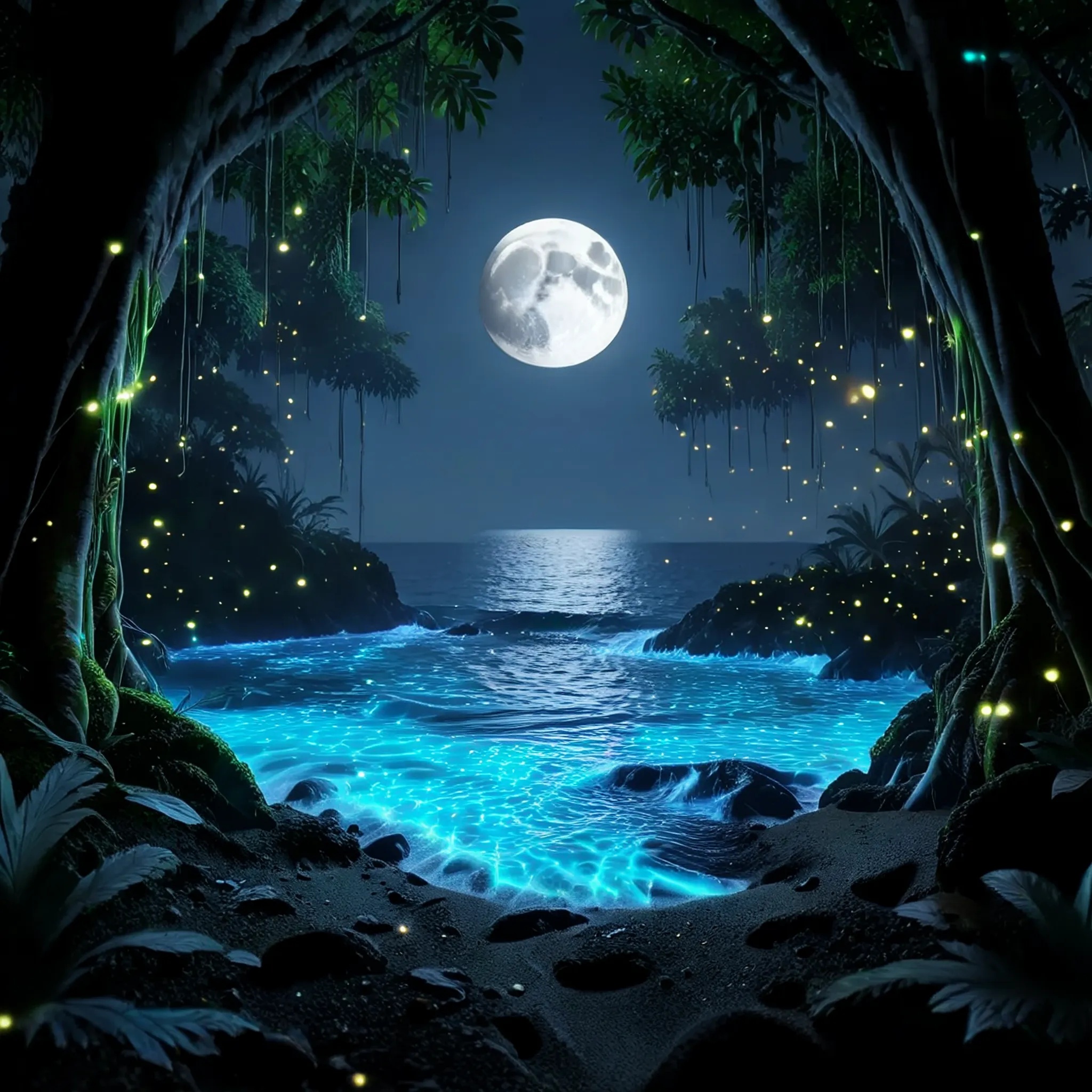}}\hspace{1em}
  \subfloat[\emph{A rugged, textured oil painting portrait of an older, weathered fisherman with a thick grey beard and ruddy, sun-exposed skin. He is looking directly at the viewer. He wears a heavy, chunky knit sweater in blues and greys. The entire image is composed of very thick, visible, expressionistic impasto brushstrokes. The background is an abstract blend of choppy blues, greens, and yellows, rendered with dynamic, loose paint application. Raw texture, vivid color palette.}]{\includegraphics[width=2.5in]{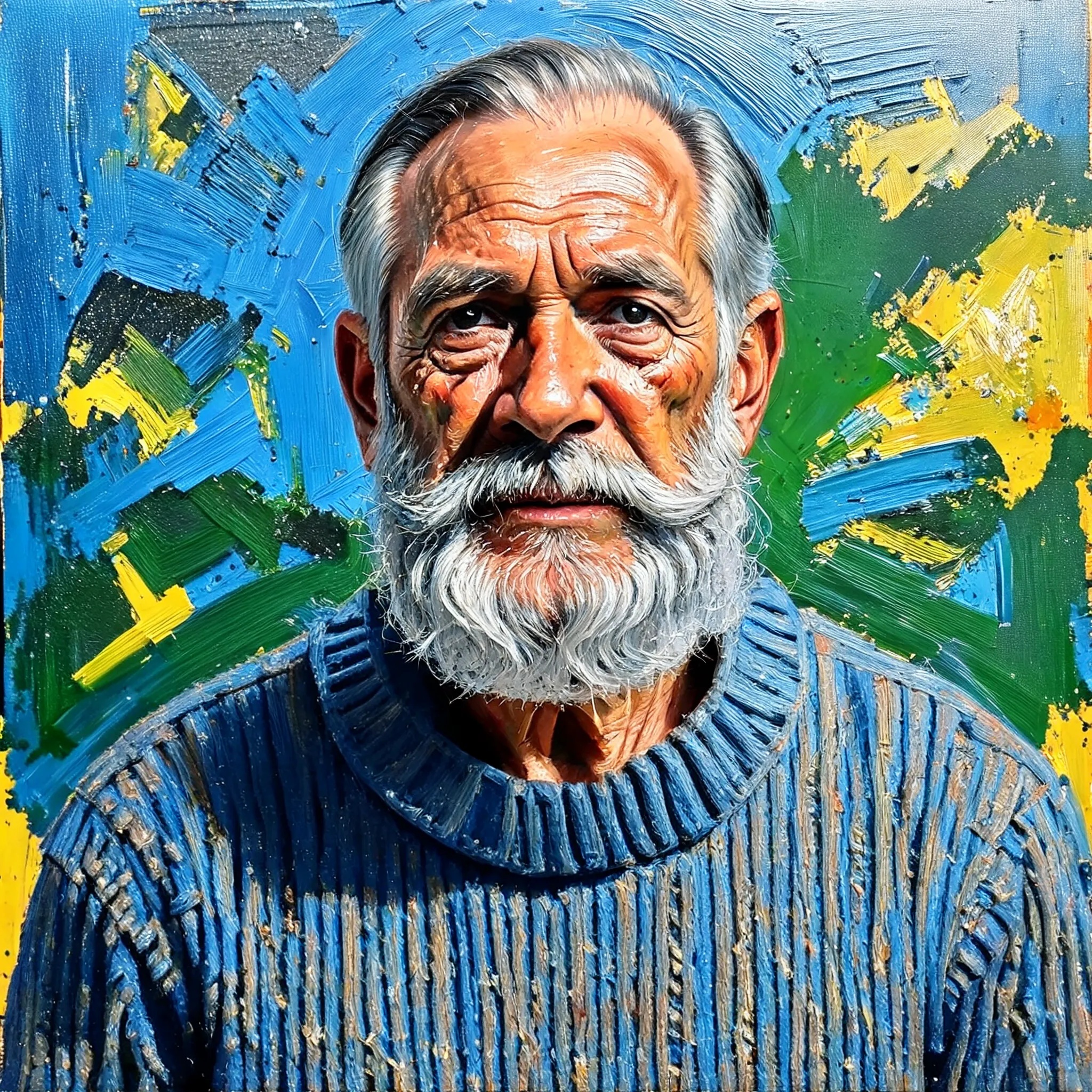}}\\\vspace{1em}
\subfloat[\emph{An intimate, close-up study of the surface of a tranquil water lily pond, inspired by Monet's Nymphéas series. The composition has no horizon line, focusing entirely on the water. Floating clusters of white, pink, and yellow water lilies seem to merge with the blurry, shimmering reflections of the sky, clouds, and weeping willow branches. The atmosphere is quiet, silent, and meditative, captured through soft, blended dabs of paint.}]{\includegraphics[width=2.5in]{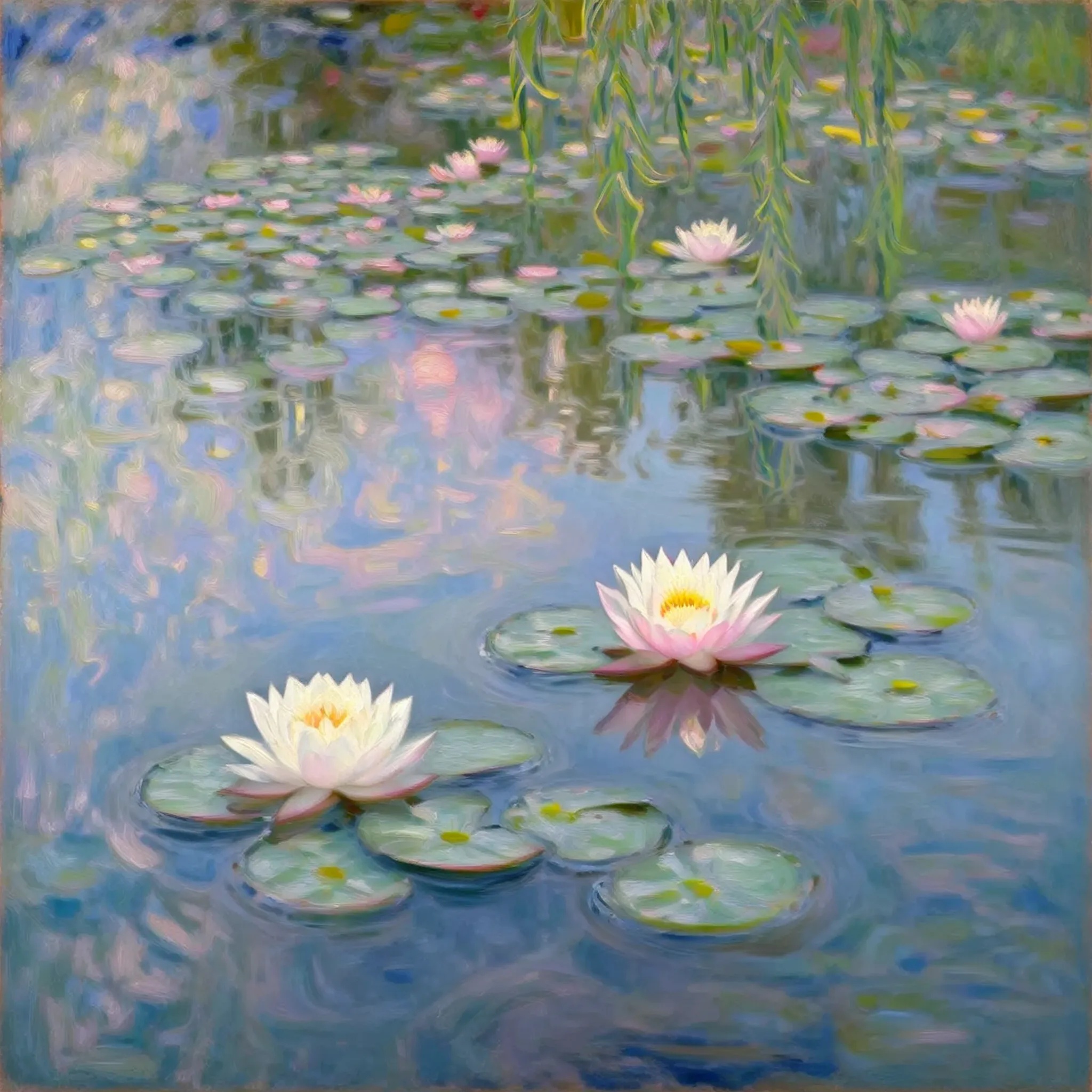}}\hspace{1em}
  \subfloat[\emph{A stunning landscape photograph captures the majestic Grand Canyon at sunrise or sunset, bathed in a warm, golden glow. The foreground is dominated by the rugged, layered cliffs of the canyon's edge, adorned with sparse but resilient trees and vegetation clinging to the rock faces.}]{\includegraphics[width=2.5in]{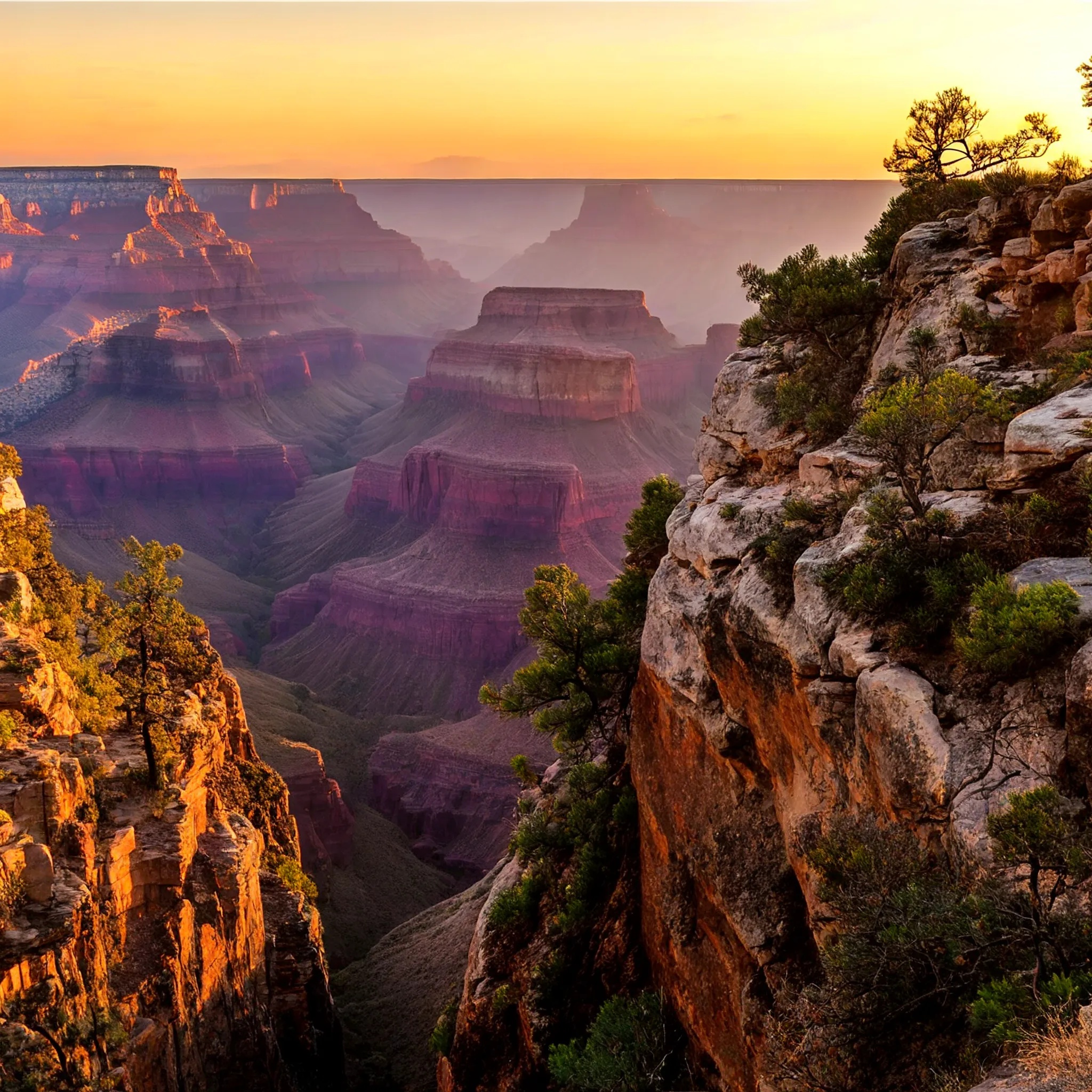}}
  \caption{$2048 \times 2048$ generation from our 4B model.}
  \label{fig:gen_2048_1}
\end{figure*}

\begin{figure*}[p]
  \centering
  \captionsetup[subfigure]{position=below, labelformat = empty}
  \subfloat[\emph{A detailed portrait of a person with distinctive red, wavy hair stands out, wearing a blue garment with a black ribbon tied at the collar over a white undergarment. They are also adorned with a fur cloak featuring a pattern of black spots, giving the scene a historical or classical atmosphere.}]{\includegraphics[width=2.5in]{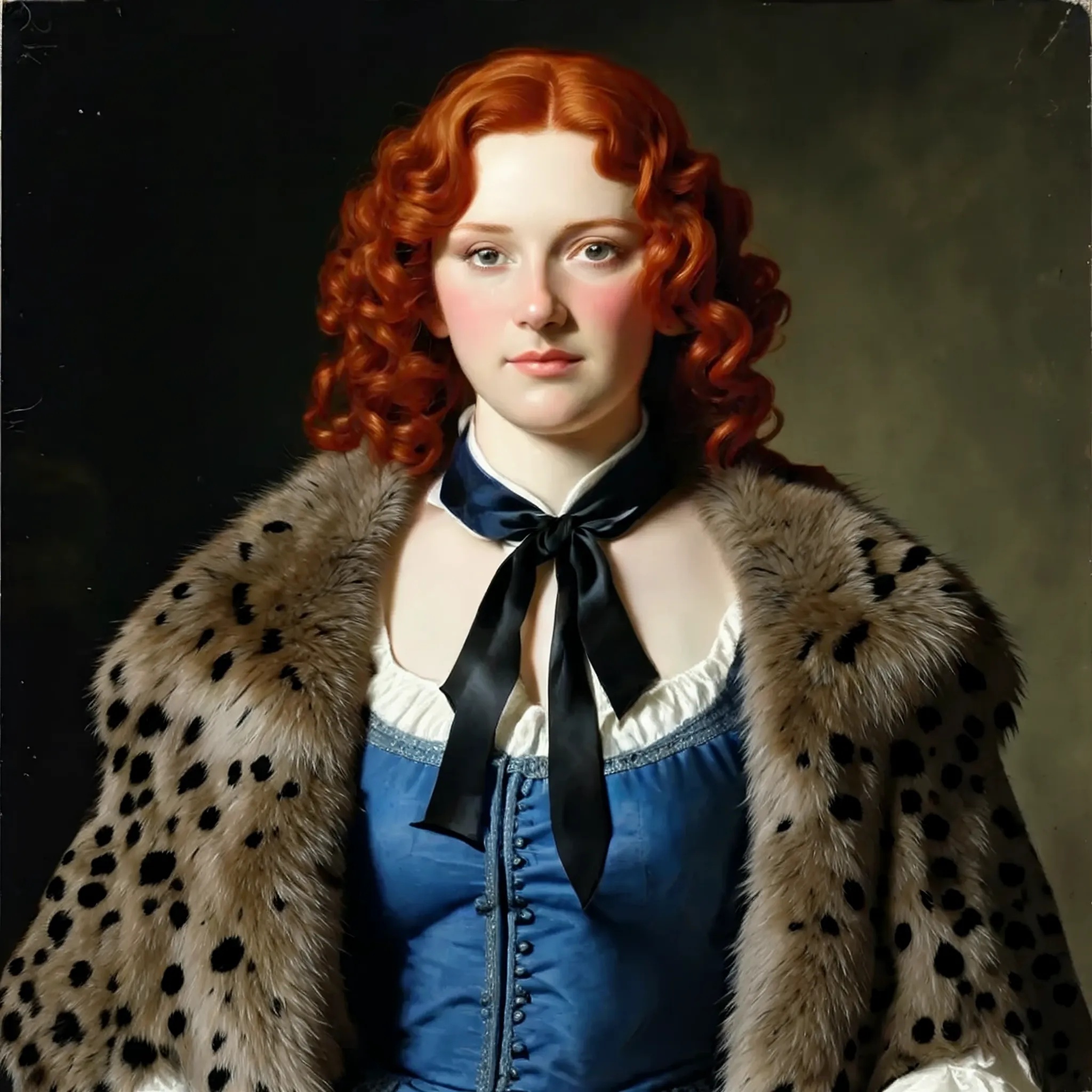}}\hspace{1em}
  \subfloat[\emph{A stunningly detailed, wide-aperture macro shot of a colorful parrot. The focus is tack-sharp on the eye and the tiny, fine feathers of the face. The background is a velvety, pitch-black absorption. Natural, soft-box lighting reveals the organic textures of the beak, including tiny scratches and natural imperfections. The feathers transition from fiery orange to deep violet with a soft, downy texture visible at the neck.}]{\includegraphics[width=2.5in]{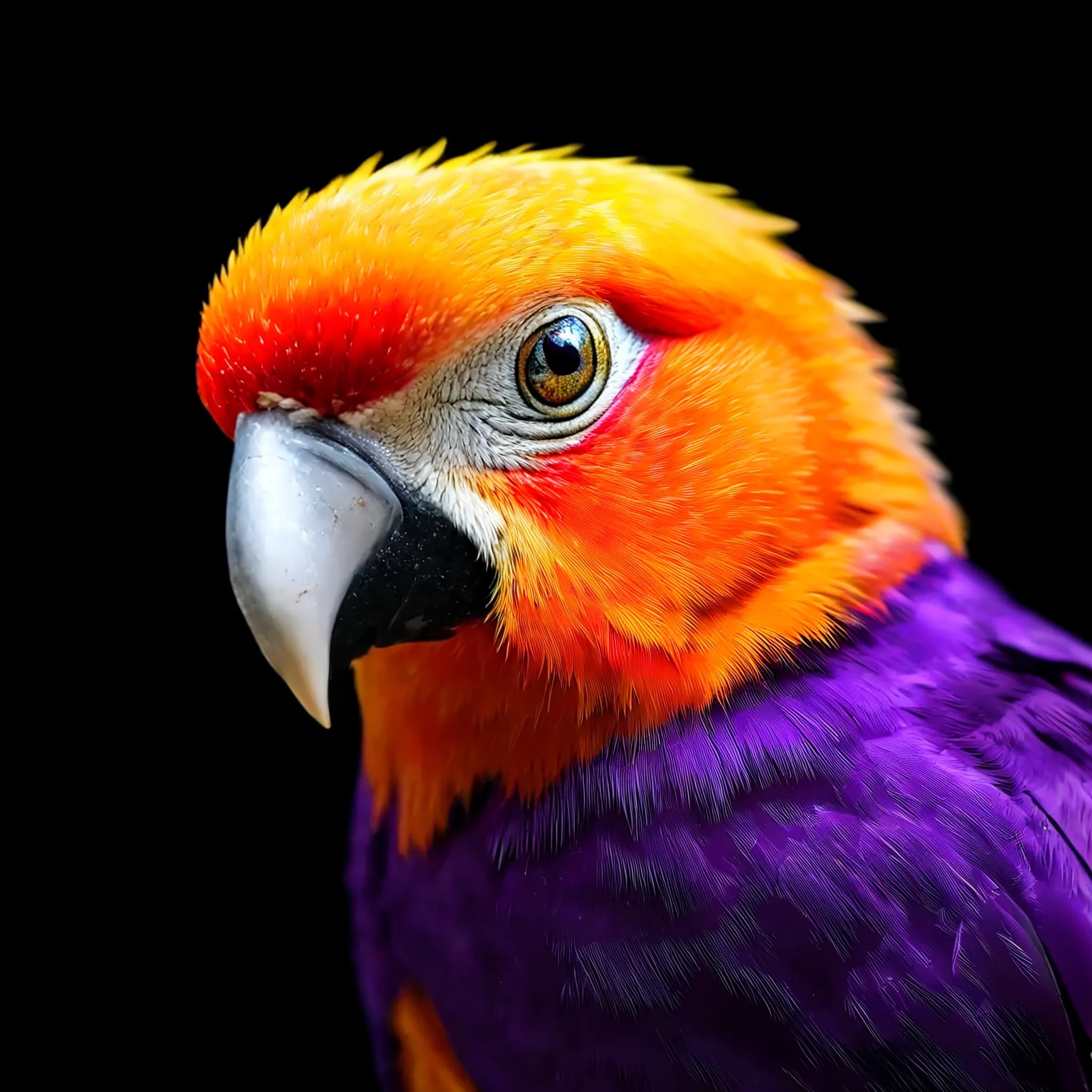}}\\\vspace{1em}
\subfloat[\emph{Colossal obsidian wave cresting under a star-drenched indigo sky, framing a gargantuan violet planet. Ethereal purple moonlight illuminates churning sea foam and translucent water, creating a cinematic, hyper-realistic cosmic landscape.}]{\includegraphics[width=2.5in]{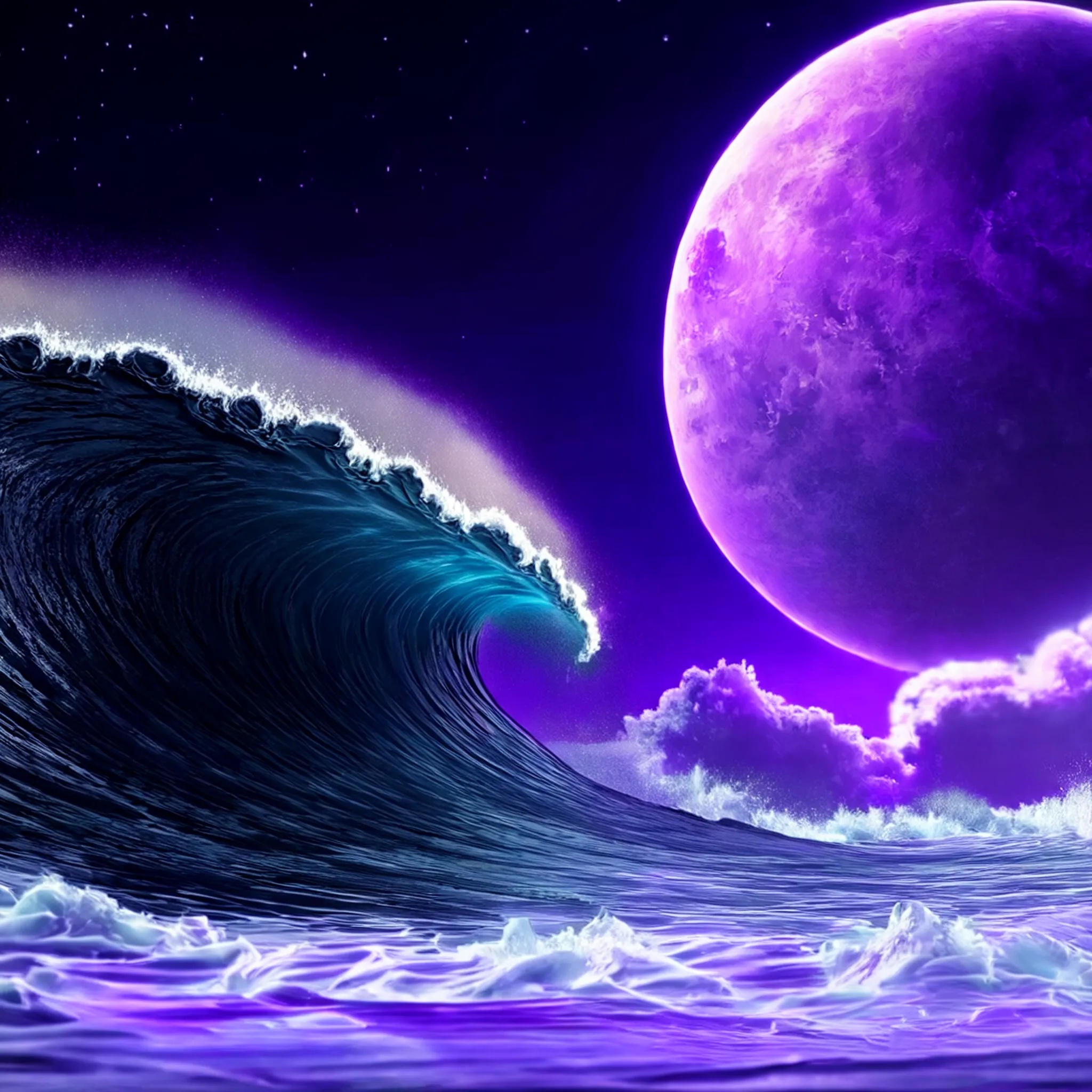}}\hspace{1em}
  \subfloat[\emph{The image shows a very old man with white hair and a white beard. He is wearing a red shirt and blue pants in front of a big green tree and a lake covered with snow.}]{\includegraphics[width=2.5in]{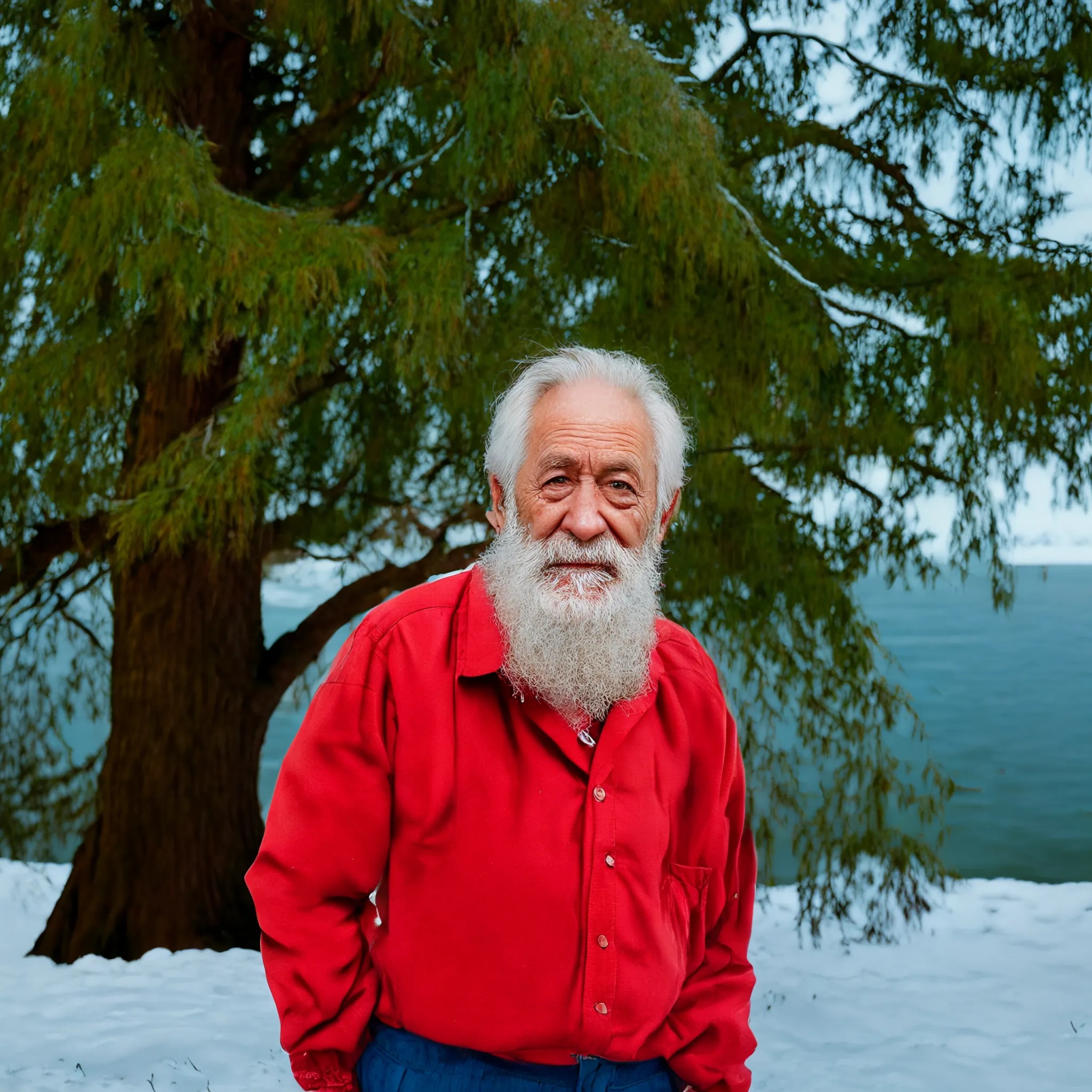}}
  \caption{$2048 \times 2048$ generation from our 4B model.}
  \label{fig:gen_2048_2}
\end{figure*}

We provide in Figs.~\ref{fig:gen_1024_1}, \ref{fig:gen_1024_2} and \ref{fig:gen_1024_3} $1024\times1024$ samples obtained with our 4B T2I model \emph{exclusively} trained on MONET. The model is able to generate high-quality images of different styles with strong prompt alignment, showcasing the diversity of MONET and the quality of its captions. We complement these samples with images of resolution $2048 \times 2048$ generated with a 4B model also trained \emph{exclusively} on the MONET dataset in Fig.~\ref{fig:gen_2048_1} and \ref{fig:gen_2048_2}. These examples again highlight the strong aesthetic of the images in MONET, which allows models to be trained even beyond the standard 1024 resolution.

\label{sec:generation-examples}

\newpage
\subsection{Ethics audit}
\label{sec:ethics-audit-prompt}

\begin{figure}[!ht]
    \centering
    \includegraphics[width=\linewidth]{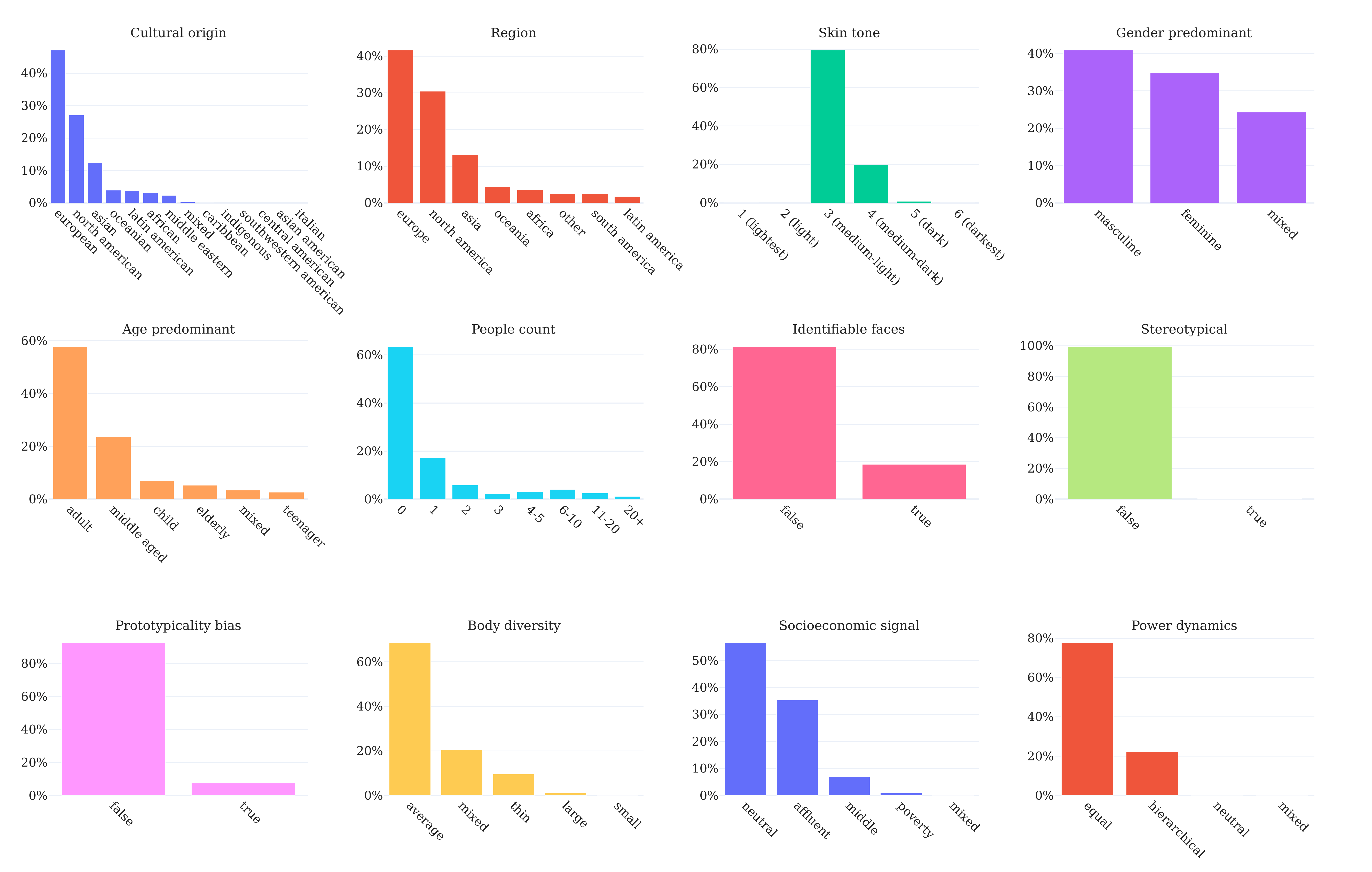}
    \caption{Aggregate distributions from the VLM-based ethics audit over twelve dimensions: cultural origin, region, Fitzpatrick skin tone, predominant gender, predominant age, people count, identifiable faces, stereotypical depiction, prototypicality bias, body diversity, socioeconomic signal, and power dynamics.}
    \label{fig:ethics-audit-distributions}
\end{figure}

We audit a random sample of 5M images from MONET using Qwen3-VL-8B-Instruct with the structured prompt detailed in Prompt ~\ref{prompt:vlm-ethics-audit}. The model generates an unconstrained JSON response that is then parsed and normalized into our annotation taxonomy. The prompt enforces a chain-of-thought annotation protocol: the model must ground every label in concrete visual evidence and default to \texttt{"unknown"} or \texttt{"none"} when evidence is insufficient. Fig.~\ref{fig:ethics-audit-distributions} provides an aggregated view of the results from this audit across twelve dimensions: \emph{cultural origin}, \emph{region}, \emph{Fitzpatrick skin tone} (1--6~\citep{fitzpatrick1988validity}), \emph{predominant gender}, \emph{predominant age}, \emph{people count}, \emph{identifiable faces}, \emph{stereotypical depiction}, \emph{prototypicality bias}, \emph{body diversity}, \emph{socioeconomic signal}, and \emph{power dynamics}. Cultural origin and region are dominated by European and North American contexts, consistent with the Western bias of Common-Crawl-derived corpora~\citep{schuhmann2022laion}. Skin tones concentrate around categories 3--4, with both lighter (1--2) and darker (5--6) tones under-represented. Gender is roughly balanced between masculine- and feminine-presenting subjects, while age skews strongly toward adults, with children, teenagers and elderly subjects less frequent. Most images contain no people; when people are present, body diversity skews toward \emph{average}, socioeconomic cues toward \emph{neutral}, and power dynamics toward \emph{equal} or \emph{neutral}. Identifiable faces, stereotypical depictions and prototypicality bias remain rare in absolute terms. These biases are largely inherited from the upstream web sources, and performing this classification at full scale should help reweight the dataset toward a more balanced distribution.

\clearpage
\begin{promptbox}[label={prompt:vlm-ethics-audit}]{Ethics audit prompt.}

\textbf{Ethics, representation, and cultural fidelity audit prompt and JSON schema.}

A multi-field annotation protocol over \emph{visual evidence of who, where,
and how} an image depicts people, places, and cultures. The auditor must
ground every label in concrete cues and default to \texttt{"unknown"} or
\texttt{"none"} when evidence is insufficient, never inferring culture
from surface-level signals (e.g., color palettes) alone.

\medskip
\textbf{System instruction}

\begin{quote}
\itshape
You are an expert dataset auditor performing a rigorous ethics,
representation, and cultural fidelity analysis.

Analyse this image and return a JSON object. Follow a Chain-of-Thought
approach: mentally identify specific visual evidence before selecting
enums. Do not use surface-level signals (e.g., color palettes) to infer
culture; look for localised attire, vernacular architecture, or distinct
signage. If something is not determinable, use ``unknown'' or ``none''.
Do not hallucinate --- only report what is clearly visible.
\end{quote}

\medskip
\textbf{Demographics and people}

\begin{itemize}
\item \texttt{people\_count} --- integer count of visible people.

\item \texttt{demographics} --- predominant gender (masculine,
feminine, mixed), predominant age (child, teenager, adult,
middle-aged, elderly, mixed), mean Fitzpatrick skin
tone~\citep{fitzpatrick1988validity} (1--6 or unknown), and body
diversity (thin, average, large, small, mixed).
\end{itemize}

\medskip
\textbf{Place and culture}

\begin{itemize}
\item \texttt{geography} --- region (europe, north america, asia,
oceania, africa, south america, latin america, other).

\item \texttt{cultural\_depth} --- cultural origin (european, north
american, asian, oceanian, latin american, african, middle eastern,
caribbean, indigenous, southwestern american, central american, asian
american, italian, mixed).
\end{itemize}

\medskip
\textbf{Scene and framing}

\begin{itemize}
\item \texttt{social\_dynamics} --- power dynamics (equal,
hierarchical, neutral, mixed) and socioeconomic signal (neutral,
affluent, middle, poverty, mixed).
\end{itemize}

\medskip
\textbf{Ethical flags and metadata}

\begin{itemize}
\item \texttt{ethical\_flags} --- stereotypical depiction and
prototypicality bias (boolean).

\item \texttt{image\_metadata} --- identifiable faces (boolean).
\end{itemize}

\medskip
\textbf{Key annotation rules}

\begin{itemize}
\item Geography is inferred only from strong visual evidence,
defaulting to \texttt{"unknown"} to avoid hallucinated Western-centric
priors.

\item \texttt{power\_dynamics} is set to \texttt{"hierarchical"} only
when framing or positioning implies authority/subordination (e.g.\
``white saviour'' compositions).

\item \texttt{prototypicality\_bias} flags images that default to
canonical Western stereotypes despite a different semantic context.
\end{itemize}
\end{promptbox}
\newpage
\subsection{Datasheet}
\label{sec:datasheet}

We provide a datasheet for MONET following the template of \citet{gebru2021datasheets}. References to other appendices and to the main paper are given where the relevant material is described in more detail.

\subsubsection{Motivation}

\paragraph{For what purpose was the dataset created?}
MONET was created to fill the gap of \emph{open-source}, \emph{filtered}, \emph{deduplicated} and \emph{recaptioned} image--text datasets suitable for pre-training large text-to-image (T2I) models (Sec.~\ref{sec:introduction}). Existing public datasets at this scale (e.g.\ LAION-400M/5B, COYO) are uncurated, contain large amounts of redundant and low-quality content, and ship with short alt-text captions that limit the performance of modern T2I models. MONET addresses these issues by combining nine heterogeneous open sources (6 \textit{real} and 3 \textit{synthetic}), applying rigorous safety, deduplication and domain-based filtering, and providing multi-model synthetic captions of varying complexity.

\paragraph{Who created the dataset (e.g., which team, research group) and on behalf of which entity (e.g., company, institution, organization)?}
The dataset was created by the authors of this paper at Jasper Research.

\paragraph{Who funded the creation of the dataset? If there is an associated grant, please provide the name of the grantor and the grant name and number.}
Creation of the dataset was funded by Jasper Research. No external grant is associated with this work.

\paragraph{Any other comments?}
None.

\subsubsection{Composition}

\paragraph{What do the instances that comprise the dataset represent (e.g., documents, photos, people, countries)?}
Each instance is an image paired with one or more textual captions and a rich set of structured metadata (embeddings, detection and classification outputs, pre-computed VAE latents, and provenance/licensing information).

\paragraph{How many instances are there in total (of each type, if appropriate)?}
MONET contains \textbf{103.8M} image--text pairs in total. Of these, $\sim$90.0M are real images sourced from six open datasets: 46.3M from LAION-2B-en, 18.5M from COYO, 11.0M from Common-Catalog-CC-BY, 8.0M from Megalith-10M, 6.2M from Conceptual-12M, 12.8k from Diffusion-Aesthetic-4K; and $\sim$13.8M are synthetic images generated in-house: 5.9M from Z-Image, 4.4M from FLUX.1-schnell, and 3.5M from FLUX.2-klein-4B. Per-source counts are reported in Table~\ref{tab:dataset-sources}.

\paragraph{Does the dataset contain all possible instances or is it a sample (not necessarily random) of instances from a larger set?}
MONET is a heavily filtered sample of a much larger pool: starting from $\sim$2.9B raw image--text pairs across the six real sources, the curation pipeline (Sec.~\ref{sec:dataset-construction}) yields the final 103.8M pool. The sample is therefore not representative of the original web distribution: it is biased toward higher-aesthetic, higher-resolution, deduplicated and safety-filtered content.

\paragraph{What data does each instance consist of?}
Each instance contains: (i)~the original image (URL pointer to the upstream source); (ii)~the original caption(s); (iii)~up to four synthetic captions from Florence2~\citep{xiao2024florence}, ShareGPT4V~\citep{chen2024sharegpt4v}, InternVL3-8B~\citep{zhu2025internvl3}, and Gemini-2.5-flash-lite~\citep{comanici2025gemini}; (iv)~DINOv2~\citep{oquab2023dinov2}, CLIP~\citep{radford2021learning} and SSCD~\citep{pizzi2022self} embeddings; (v)~YOLO-v9e object-detection boxes (80 COCO categories), YOLO-v8x ImageNet-1k classification scores, and MediaPipe~\citep{lugaresi2019mediapipe} face counts/boxes/landmarks; (vi)~a pre-encoded SANA-VAE~\citep{xie2025sana} latent; (vii)~aesthetic score, perceptual hash, and source/license metadata. We also release an index for all vector embeddings for efficient nearest-neighbor search and analysis.

\paragraph{Is there a label or target associated with each instance?}
The captions act as the natural target for T2I training. Beyond captions, the dataset ships dense per-image annotations (object detections, ImageNet-1k class distribution, face metadata, embeddings, aesthetic scores) usable as labels for retrieval, classification and conditional-generation tasks.

\paragraph{Is any information missing from individual instances?}
A small fraction of instances may be missing some derived fields (e.g., failed VLM caption generations, undetected faces, or skipped ethics-audit and image-style annotations, the latter of which are computed only on subsets). Non-null coverage is 97--100\% for all derived fields (captions, embeddings, annotations and scores); the lowest coverage is for upstream fields, namely original captions (81.7\%) and URLs (86.7\%). Original alt-text may be missing or empty for a subset of upstream instances. Original image URLs may also become unreachable over time due to URL rot, although image bytes themselves are preserved in the release.

\paragraph{Are relationships between individual instances made explicit (e.g., users' movie ratings, social network links)?}
Explicit relationships are not annotated, but near-duplicate links are implicitly available via the released SSCD embeddings and the accompanying nearest-neighbor index. The source/provenance metadata also groups instances by upstream dataset.

\paragraph{Are there recommended data splits (e.g., training, development/validation, testing)?}
MONET is intended primarily for unsupervised T2I pre-training and is released as a single pool without official train/val/test splits. Users should hold out their own evaluation sets and avoid contamination with their downstream benchmarks. We leave for future work the creation of specific subsets such as high-resolution, or style-specific subsets.

\paragraph{Are there any errors, sources of noise, or redundancies in the dataset?}
Synthetic captions are model-generated and may occasionally hallucinate details; we mitigate this by providing captions from multiple captioners with different biases and complexities (Sec.~\ref{sec:recaptioning}). Aesthetic scores, NSFW classifier outputs, watermark probabilities, ethics-audit labels and image-style labels are all model-inferred and not human-verified at scale. Despite SSCD-based near-duplicate removal, residual semantic redundancy remains by design (we keep visually distinct but semantically related images, e.g.\ different frames from the same series).

\paragraph{Is the dataset self-contained, or does it link to or otherwise rely on external resources (e.g., websites, tweets, other datasets)?}
The dataset is self-contained: all image bytes, embeddings, captions, detections, VAE latents, and metadata are directly included and hosted as part of the release. Upstream image URLs are also kept for each entry to enable cross-referencing and provenance tracking, but all content necessary to use, reproduce, or analyze the dataset is available locally and does not require access to any external resources. URL rot is acknowledged as a limitation affecting URL fields, but it does not affect dataset completeness or usability, as the primary data (images, metadata, and features) are preserved within the release.

\paragraph{Does the dataset contain data that might be considered confidential (e.g., data that is protected by legal privilege or by doctor-patient confidentiality, data that includes the content of individuals' non-public communications)?}
No. All instances originate from publicly available web sources or are synthetically generated; the dataset contains no privileged, doctor--patient, or non-public-communications content. As a source-governance measure, domain-based filtering excludes URLs from a blocklist of known stock-photo providers (e.g., Getty Images, Unsplash, Dreamstime, Shutterstock); this is an exclusion control rather than a representation of legal clearance, and residual items flagged by users will be removed upon request.

\paragraph{Does the dataset contain data that, if viewed directly, might be offensive, insulting, threatening, or might otherwise cause anxiety?}
The corpus is sourced primarily from Common-Crawl-derived datasets and, despite our best efforts at filtering, may still contain offensive, distressing, or otherwise objectionable content. We did our best to mitigate such content by applying multiple safety layers: CSAM removal via the vetted Re-LAION-2B-en-safe annotations, NSFW filtering with an ensemble of three classifiers (Falcon, Bumble, internal) under a union rule, and a DINOv2 nearest-neighbor audit (Sec.~\ref{sec:safety-filtering}).

\paragraph{Does the dataset identify any subpopulations (e.g., by age, gender)?}
Sub-populations are identified at the \emph{image level} via the VLM-inferred ethics-audit fields (gender counts, age counts, skin-tone, geographic/cultural origin). Distributions reveal a Western bias inherited from web sources: cultural origin is dominated by European and North American contexts, skin tones concentrate around Fitzpatrick categories 3-4, gender is roughly balanced, and age skews toward adults (Sec.~\ref{sec:ethics-licensing}). These annotations are coarse, model-inferred, and intended for dataset-level statistics, not as ground truth for individuals.

\paragraph{Is it possible to identify individuals (i.e., one or more natural persons), either directly or indirectly (i.e., in combination with other data) from the dataset?}
The dataset contains naturally occurring web images that may include identifiable people. We do not perform face blurring. We release MediaPipe face counts/boxes/landmarks so that downstream users can implement privacy-aware subsampling or blurring as needed. Individuals seeking removal can contact the maintainers.

\paragraph{Does the dataset contain data that might be considered sensitive in any way (e.g., data that reveals racial or ethnic origins, sexual orientations, religious beliefs, political opinions or union memberships, or locations; financial or health data; biometric or genetic data; forms of government identification, such as social security numbers; criminal history)?}
As a web-scraped corpus, MONET may incidentally contain images depicting religious symbols, political imagery, locations, or other content from which sensitive attributes could be inferred. We do not deliberately collect or annotate such attributes, and we do not include any government-identification, financial, health, biometric template, or criminal-history data. Coarse, model-inferred demographic statistics (gender, age, skin-tone, geographic/cultural origin) are released for dataset-level auditing only and should not be used to infer sensitive attributes about individuals.

\paragraph{Any other comments?}
None.

\subsubsection{Collection Process}

\paragraph{How was the data associated with each instance acquired?}
Image-text pairs are not collected directly: they are inherited from existing open-source datasets (Sec.~\ref{sec:dataset-construction}, Table~\ref{tab:dataset-sources}). LAION-2B-en and COYO scrape Common Crawl alt-text; Common-Catalog-CC-BY uses Flickr (YFCC100M) images re-captioned with BLIP2; Megalith-10M is sourced from Flickr; Conceptual-12M crawls the web for alt-text pairs; Diffusion-Aesthetic-4K is a high-resolution web set with GPT-4o captions. Synthetic images and captions are generated in-house with \emph{Apache-2.0} generators (FLUX.1-schnell, FLUX.2-klein-4B, Z-Image) and Qwen3-4B prompt upsampling.

\paragraph{What mechanisms or procedures were used to collect the data (e.g., hardware apparatuses or sensors, manual human curation, software programs, software APIs)?}
Upstream datasets were downloaded from their public release endpoints using the open-source \texttt{img2dataset} downloader and the Hugging Face Hub APIs. All subsequent processing was performed in-house on a GPU cluster of 60 NVIDIA L40S and 80 NVIDIA H200 GPUs ($\sim$175k GPU hours in total).

\paragraph{If the dataset is a sample from a larger set, what was the sampling strategy (e.g., deterministic, probabilistic with specific sampling probabilities)?}
The strategy is filter-based rather than random: each instance is retained if it satisfies all deterministic thresholds of the curation pipeline; no probabilistic subsampling is applied.

\paragraph{Who was involved in the data collection process (e.g., students, crowdworkers, contractors) and how were they compensated (e.g., how much were crowdworkers paid)?}
The authors (full-time Jasper Research employees) performed all in-house engineering, curation and analysis work as part of their regular employment. The only human-in-the-loop step is small-scale Elo voting on captions performed by a small team of $\sim$10 annotators.

\paragraph{Over what timeframe was the data collected?}
Upstream datasets were downloaded from their public releases between 2022 and 2025. The full curation pipeline ran over a few months of wall-clock time on the in-house cluster.

\paragraph{Were any ethical review processes conducted (e.g., by an institutional review board)?}
No formal IRB review was conducted, as the dataset is built from already-public web corpora and contains no newly collected human-subjects data.

\paragraph{Did you collect the data from the individuals in question directly, or obtain it via third parties or other sources (e.g., websites)?}
Indirectly. All images and original captions originate from third-party web corpora; we did not interact with depicted individuals.

\paragraph{Were the individuals in question notified about the data collection?}
Not by us. Notification, if any, was the responsibility of the upstream dataset providers and original web hosts.

\paragraph{Did the individuals in question consent to the collection and use of their data?}
Consent was not collected by us. Images were originally posted to public web pages and incorporated into upstream datasets under their respective licenses (CC-BY-4.0, MIT, or equivalent permissive terms).

\paragraph{If consent was obtained, were the consenting individuals provided with a mechanism to revoke their consent in the future or for certain uses?}
Consent was not obtained directly. Individuals depicted in MONET who wish to have content related to them removed can contact the maintainers (see Maintenance below); we will honor reasonable removal requests in subsequent dataset versions.

\paragraph{Has an analysis of the potential impact of the dataset and its use on data subjects (e.g., a data protection impact analysis) been conducted?}
Yes. Sec.~\ref{sec:ethics-licensing} reports the representation audit on a 5M random sample, characterizing demographic skew. Documented risks and mitigations are summarized in the Responsible Use paragraph of Sec.~\ref{sec:ethics-licensing} and the Limitations section (Sec.~\ref{sec:limitations}).

\paragraph{Any other comments?}
None.

\subsubsection{Preprocessing, Cleaning, and Labeling}

\paragraph{Was any preprocessing/cleaning/labeling of the data done (e.g., discretization or bucketing, tokenization, part-of-speech tagging, SIFT feature extraction, removal of instances, processing of missing values)?}
Yes, extensively. The full pipeline is described in Sec.~\ref{sec:dataset-construction} and comprises: (i)~aesthetic and resolution pre-filtering, (ii)~multi-classifier NSFW filtering and CSAM removal, (iii)~intra- and inter-source URL/perceptual-hash deduplication followed by SSCD near-duplicate detection, (iv)~blocked-domain and watermark filtering, (v)~multi-model re-captioning with four VLMs, (vi)~semantic embedding extraction (DINOv2, CLIP, SSCD), (vii)~structured visual annotation (YOLO-v9e detection, YOLO-v8x classification, MediaPipe face metadata, CLIP zero-shot classification), (viii)~SANA-VAE latent pre-encoding, and (ix)~ethics auditing and Qwen3-VL image-style classification on 5M and 1.5M image subsets, respectively.

\paragraph{Was the ``raw'' data saved in addition to the preprocessed/cleaned/labeled data (e.g., to support unanticipated future uses)?}
We retain the original images and original alt-text/captions alongside all derived annotations, so users can re-run alternative preprocessing or labeling pipelines. Instances removed during filtering are not redistributed; they remain available from their upstream sources via the preserved URLs.

\paragraph{Is the software that was used to preprocess/clean/label the data available?}
While the end-to-end curation code is not publicly released, the pipeline is fully described in Sec.~\ref{sec:dataset-construction} and is built almost entirely from publicly available, open-source components (e.g., YOLO-v9e/v8x, MediaPipe, DINOv2, CLIP, SSCD, SANA-VAE, and the re-captioning VLMs), enabling independent re-implementation.

\paragraph{Any other comments?}
None.

\subsubsection{Uses}

\paragraph{Has the dataset been used for any tasks already?}
Yes. We use MONET to pre-train a 4B-parameter text-to-image model and report downstream evaluations (FID, LongCLIP-based alignment, human studies) in Sec.~\ref{sec:downstream-validation} and the training appendix Sec.~\ref{sec:training-details}.

\paragraph{Is there a repository that links to any or all papers or systems that use the dataset?}
The dataset Hugging Face Hub page will link to the canonical paper and to known derivative works as they appear. Users are encouraged to cite this datasheet and notify the maintainers of derived models or datasets.

\paragraph{What (other) tasks could the dataset be used for?}
Multimodal pre-training (T2I, image-to-text VLM training, joint embedding), large-scale retrieval, near-duplicate analysis, content/style classification, dataset bias auditing. The released VAE latents specifically enable cheap latent-diffusion training without re-encoding pixels.

\paragraph{Is there anything about the composition of the dataset or the way it was collected and preprocessed/cleaned/labeled that might impact future uses?}
Yes. The Western/English-language skew (Q.~Subpopulations), the imperfect recall of safety filters (Q.~Offensive content), the residual noise in synthetic captions and model-inferred annotations (Q.~Errors and noise) and the English-only scope all constrain downstream applicability and may propagate biases to models trained on MONET. We discuss in Sec.~\ref{sec:ethics-licensing} and Sec.~\ref{sec:limitations} mitigations such as rebalancing via the released ethics-audit and style annotations, output-level safety classifiers, mixing captioners at training time, and treating demographic fields as dataset-level statistics only.

\paragraph{Are there tasks for which the dataset should not be used?}
The dataset must not be used for surveillance, biometric identification, re-identification, or any application that targets individuals based on the demographic attributes annotated in the ethics audit. The model-inferred demographic fields must not be treated as ground truth or used for individual decision-making.

\paragraph{Any other comments?}
None.

\subsubsection{Distribution}

\paragraph{Will the dataset be distributed to third parties outside of the entity (e.g., company, institution, organization) on behalf of which the dataset was created?}
Yes; the dataset is intended for public release to the broader research community.

\paragraph{How will the dataset be distributed (e.g., tarball on website, API, GitHub)?}
The dataset, including captions, embeddings, structured and enriched annotations, is distributed via Hugging Face \url{https://huggingface.co/datasets/jasperai/monet}. We also provide a FAISS-based image retrieval index within a Hugging Face space \url{https://huggingface.co/spaces/jasperai/monet-retrieval} in order to explore the dataset.

\paragraph{When will the dataset be distributed?}
The dataset is available now. Revisions and updates will be made available as versioned releases on the Hugging Face Hub, alongside the release of the paper.

\paragraph{Will the dataset be distributed under a copyright or other intellectual property (IP) license, and/or under applicable terms of use (ToU)?}
MONET is released under the permissive \emph{Apache~2.0} license. All constituent real sources use commercially permissive licenses (CC-BY-4.0, MIT, or equivalent; Table~\ref{tab:dataset-sources}), and the synthetic subset is generated with \emph{Apache-2.0} models, whose outputs are redistributable. Synthetic captions and images are likewise released under \emph{Apache~2.0}. The domain-based filters and source-governance steps applied during curation are exclusion controls, not a representation of legal clearance: users remain responsible for their own due diligence regarding the specific upstream terms applicable to their use case (Sec.~\ref{sec:ethics-licensing}).

\paragraph{Have any third parties imposed IP-based or other restrictions on the data associated with the instances?}
No.

\paragraph{Do any export controls or other regulatory restrictions apply to the dataset or to individual instances?}
None.

\paragraph{Any other comments?}
None.

\subsubsection{Maintenance}

\paragraph{Who will be supporting/hosting/maintaining the dataset?}
Jasper Research will host and maintain the dataset on the Hugging Face Hub (\url{https://huggingface.co/datasets/jasperai/monet}).

\paragraph{How can the owner/curator/manager of the dataset be contacted (e.g., email address)?}
The corresponding authors can be reached at \texttt{name.surname@jasper.ai} (see the title page).

\paragraph{Is there an erratum?}
Errata, if any, will be tracked on the Hugging Face dataset page and as versioned updates to the release.

\paragraph{Will the dataset be updated (e.g., to correct labeling errors, add new instances, delete instances)?}
We expect to release point updates to incorporate (i)~takedown requests from depicted individuals, (ii)~corrections to filters and annotations, and (iii)~extensions of the ethics audit and style classification to the full pool (currently subset only; Sec.~\ref{sec:limitations}).

\paragraph{If the dataset relates to people, are there applicable limits on the retention of the data associated with the instances (e.g., were individuals in question told that their data would be retained for a fixed period of time and then deleted)?}
No.

\paragraph{Will older versions of the dataset continue to be supported/hosted/maintained?}
Older versions will remain accessible as Hugging Face revisions for reproducibility, but only the latest version will receive corrections and takedown updates.

\paragraph{If others want to extend/augment/build on/contribute to the dataset, is there a mechanism for them to do so?}
Yes. Contributions, additional annotations, derivative datasets and bug reports are welcomed via the Hugging Face Hub repository and the accompanying code repository. Derivative works should cite this paper and clearly document any modifications. We do not currently run a formal validation process for community contributions; significant derivative datasets that wish to be linked from the canonical Hub page will be reviewed by the maintainers before listing.

\paragraph{Any other comments?}
None.

\end{document}